\documentclass[12pt]{article}
\usepackage{preamble_ARXIV_2026_09_23}

\newcommand{\conlab}[1]{\hspace{-2pt}--\hspace{5pt}{\emph{(#1)}}\,}%
\newcommand{\prjerr}{\tilde{\epsilon}_n}%
\newcommand{\anA}{a}%

\usepackage[capitalize]{cleveref} 

\usepackage{etoolbox}
\usepackage[user]{zref}

\usepackage{silence}
\makeatletter

\newcounter{ComMention}

\newcommand{\Com@addmention}[3]{%
  \ifcsundef{Com@mentions@#1@#2}
    {\csgdef{Com@mentions@#1@#2}{\zpageref{#3}}}
    {\csgappto{Com@mentions@#1@#2}{, \zpageref{#3}}}%
}

\newcommand{\ComNote}[4]{%
  \stepcounter{ComMention}%
  \zlabel{Com:mention:#1:#4:\arabic{ComMention}}%
  \note{#3\ref{#2:C#4}}%
  \protected@write\@auxout{}%
    {\string\Com@addmention{#1}{#4}{Com:mention:#1:#4:\arabic{ComMention}}}%
}

\newcommand{\AENote}[1]{ }
\newcommand{\RoneNote}[1]{ }
\newcommand{\RtwoNote}[1]{ }

\makeatother

\title{Statistical Properties of Deep Neural Networks with Dependent Data\thanks{This research did not receive any specific grant from funding agencies in the public, commercial, or not-for-profit sectors.}}
\date{ 
    \vspace{-.2cm} 
    \today
    \vspace{-.8cm}
}

\author{
Chad Brown%
    \thanks{Department of Economics, The University of Manchester. Email: chad.brown@manchester.ac.uk.}
}

\begin{document} 
    \setlength{\abovedisplayskip}{5pt}
    \setlength{\belowdisplayskip}{5pt}
\maketitle

\begin{abstract}
\vspace{-.1cm}
\renewcommand{\thefootnote}{\fnsymbol{footnote}}
\setstretch{1.2}
    \noindent 
    This paper develops theory for deep neural network (DNN) estimators under dependent data. 
    To provide theory applicable to a variety of DNN-based estimators,
    I first establish nonasymptotic probability bounds on the theoretical and empirical $\Lp{2}$-errors of nonparametric sieve estimators for a general class of estimation problems
    under possibly nonstationary $\beta$-mixing data taking values in unbounded sets. 
    I then apply the theory to fully connected and convolutional DNN estimators without bounds or sparsity restrictions on the DNN weights. 
    For both DNN classes, I derive general results when the function to be estimated is H\"older smooth
    and the data are nonstationary, subgaussian, and $\beta$-mixing with either exponential or polynomial decay.
    I then specialize these to nonparametric regression, logistic regression, and quantile regression settings.
    Under exponential $\beta$-mixing, the resulting estimators attain the nonparametric minimax rate of \citet{stone_optimal_1982} up to logarithmic factors.
    \\
    
    
\end{abstract}

\newpage
\section{Introduction}\label{sec:intro}

    Deep neural networks (DNNs) have 
    shown promising empirical performance
    for predicting
    time series in economics and finance,
    often improving on more traditional methods
        (e.g., 
        \citealp{gu_empirical_2020}; 
        \citealp{bucci_realized_2020};
        \citealp{lazcano_back_2024}%
        ).
    They are also increasingly used in a range of time-series forecasting applications
    including mean regression tasks such as predicting daily COVID-19 cases (\citealp{Chronopoulos_Panel_2023}), binary prediction problems such as loan default forecasting (\citealp{kvamme_predicting_2018,albanesi_predicting_2019,sadhwani_deep_2021}), and quantile regression models such as Value-at-Risk prediction (\citealp{petnehazi_quantile_2021,Chronopoulos_ForecastingDNNQuantile_2024}).
    However, existing theoretical results primarily focus on independent and identically distributed (i.i.d.\!)\! sampling 
    (e.g., \citealp{schmidt_hieber_nonparametric_2020,kohler_rate_2021,farrell_deep_2021,Padilla_Quantile_2022,bos_convergence_2022,Ohn_Nonconvex_2022,zhang_classification_2024,shen_nonparametric_2024,yara_nonparametric_2026}).
    This leaves an important gap between empirical applications and theoretical results for DNNs in time-series settings.

    For non-\iid\!data satisfying mixing or related dependence conditions, several recent papers study feed-forward DNNs
    under stationarity in general estimation problems (\citealp{Chen_Inference_2025,Kengne_Excess_2025,kengne_robust_2025,kengne_deep_2026}) and nonparametric mean regression (\citealp{ma_theoretical_2022,Men_Learning_2025}). 
    For nonstationary data, \citet{kurisu_adaptive_2025} provide results for nonparametric mean regression.
    This literature is important, but it does not accommodate several features common in modern DNN applications. 
    First, these papers impose bounds and sparsity restrictions on the 
\!\RtwoNote{1}\!%
    DNN weights.%
\footnote{Throughout, “DNN weights” refers to all trainable network parameters, including what are often called weights and biases in the DNN literature; see Subsection \ref{sec:DNN_sieve_spaces} for formal definitions.}
    Such constraints are often not used in practice (\citealp{yara_nonparametric_2026}), 
    sparsity restrictions have been shown to complicate training (\citealp{liu_rethinking_2018,evci_difficulty_2019}),
    and
    DNN weight bounds can lead to boundary-related optimization issues, limit approximation power, and increase computational costs
    (see \citealp{farrell_deep_2021} for more discussion).
    Second, the available theory is tied to particular network architectures and does not consider broader classes of DNN estimators.  
    Third, while these papers allow unbounded outcomes, they restrict the covariates to compact sets, which is limiting in autoregressive settings where the covariates contain lagged outcomes and the innovations may be unbounded. 
    Finally, the available results are usually stated as excess-risk or generalization-error bounds, often for conditional mean regression, rather than as probability bounds for estimation error in a framework that also covers classification and quantile regression. 
    %
    %


    This paper addresses these deficiencies by developing sieve-estimation results applicable to DNN-based estimators
    in a broad class 
    of nonparametric time-series estimation problems, with conditional mean, logistic, and quantile regression as leading examples.
    The theory is then applied to fully connected DNNs (FCDNNs) and convolutional DNNs (CDNNs) without bounds or sparsity restrictions on the DNN weights. The results yield finite-sample probability bounds for empirical and theoretical $\Lp{2}$-errors under possibly nonstationary $\beta$-mixing data 
    that may take values on unbounded sets.


    Mixing conditions provide a standard and flexible way to control dependence in time-series settings (\citealp{doukhan_mixing_1994,bosq_1998,fan_nonlinear_2003,gao_nonlinear_2007}). 
    Many models used in economics and finance are $\beta$-mixing under mild assumptions
    including many autoregressive (AR) specifications (e.g., \citealp{mokkadem_mixing_1988,chen_geometric_2000,carrasco_mixing_2002,bradley_basic_2005}; see also Subsection \ref{sec:Sieve_Est_Meas}). 
    Allowing for nonstationarity is also useful in forecasting applications, where stationarity assumptions may be too restrictive over the sample period.
    The theory developed here therefore applies to a wide range of regression and classification problems, including certain panel data settings with temporal dependence and fixed cross-sectional dimension.

    First, Theorem \ref{thrm:ROC_V2} provides nonasymptotic probability bounds on both
    the theoretical and empirical $\Lp{2}$-errors of nonparametric sieve estimators for a general class of statistical models under possibly nonstationary $\beta$-mixing data with unbounded outcomes and covariates. 
    %
    %
    This level of generality is useful
    because sieve estimation covers many DNN-based estimators rather than a single architecture.
    %
    The proof extends the localization and scale-insensitive complexity methods of \citet{farrell_deep_2021} beyond 
    settings with bounded \iid\!data.
    Unlike existing results under dependent data, this approach avoids conditions that are impractical for DNNs with unbounded weights, such as uniform metric entropy assumptions
    (e.g., \citealp{ma_theoretical_2022, kengne_robust_2025, kurisu_adaptive_2025, Kengne_Excess_2025, kengne_deep_2026})
\!\RoneNote{1}\!%
    or restrictions stated in terms of sieve basis functions 
    (\citealp{chen_optimal_2015,Chen_Inference_2025}).

    Next, I apply this general result to
    FCDNN estimators with architectures that reflect modern applications:
            fully connected feedforward networks
            with continuous piecewise-linear activation functions, 
            with 
            no constraints on the DNN weights. 
    I consider two architecture growth regimes: deep-and-wide networks with growing width and depth, or very-deep networks with fixed width and growing depth.
        %
    FCDNNs 
    are standard in practice
        (\citealp{almeida2020multilayer}; 
        \citealp{criado_ramon_electric_2022}),
    and are frequently used in time-series settings.%
\footnote{
Recurrent neural networks are also an important class of DNNs for time series applications not considered here. See Section \ref{sec:Conc} for brief discussion.
}
    For concreteness, I focus on the rectified linear unit (ReLU) activation function, 
    which is common in modern applications (\citealp{lecun_deep_2015}).%
\footnote{Compared to smooth activation functions, ReLU activation functions have been shown to offer improved properties both empirically (e.g., \citealp{orimoloye_comparing_2020,sadhwani_deep_2021}) 
and 
theoretically 
(\citealp{glorot_deep_2011}) in some settings.
}
    However, 
    Appendix \ref{sec:DNN_Alt_Arch} shows that the results extend to DNNs with any continuous piecewise-linear activation function, such as the leaky ReLU.


    Theorem \ref{thrm:FCDNN_ROC} applies Theorem \ref{thrm:ROC_V2} to FCDNN estimators in a broad class of estimation problems under possibly nonstationary, subgaussian data with either exponential or polynomial $\beta$-mixing decay, where the function to be estimated is H\"older smooth. 
    It is then specialized to nonparametric regression, logistic regression, and quantile regression settings
    where the theorem’s conditions are verified and concrete results are obtained.
    Under exponential $\beta$-mixing, these results imply the usual nonparametric minimax rate for H\"older classes up to a logarithmic factor.%
\footnote{Throughout, ``minimax" refers to the benchmark rate $n^{-\frac{2\smooth}{2\smooth+d}}$, from \citet{stone_optimal_1982}, where $\smooth$ is the H\"older smoothness index and $d$ is the dimension of the inputs. See the discussion following Theorem \ref{thrm:FCDNN_ROC}  for its interpretation in settings related to logistic and quantile regression.}
    In contrast to much of the existing FCDNN literature (e.g., \citealp{farrell_deep_2021,kurisu_adaptive_2025}), these results allow both deep-and-wide and very-deep fixed-width architectures.
    Further discussion of the contribution of each result is provided in Section \ref{sec:FCDNN_res}.

    Finally, I develop results for a class of CDNN architectures 
    consisting of blocks of expansive convolution layers separated by locally connected dimension-reduction layers.
    Although CDNNs are best known for image-recognition tasks, they have also shown promising empirical performance for time-series forecasting in finance and economics 
    (e.g., \citealp{cook_macroeconomic_2017,bai_empirical_2018,kvamme_predicting_2018,borovykh_conditional_2018,petnehazi_quantile_2021}).
    Relative to the FCDNN literature, theoretical work on CDNN estimators is less developed. 
    Under \iid\!data, existing work has obtained convergence rates for CDNN estimators in conditional mean estimation and classification settings (e.g., \citealp{liu_besov_2021,zhou_learning_2024,kohler_statistical_2025,yang_optimal_2025,Yang_CDNDconvergence2025}),
    often for architectures that also combine complexity-expanding convolution layers with down-sampling or dimension-reduction steps.
    However, theoretical work on CDNN estimators under dependent data remains limited.

    I first establish a complexity bound and a new approximation result for the CDNN class considered here.
    I then apply Theorem \ref{thrm:ROC_V2} under H\"older smoothness of the function to be estimated and possibly nonstationary $\beta$-mixing data to obtain results for general estimation problems and, in turn, for nonparametric regression, logistic regression, and quantile regression.
    Under exponential $\beta$-mixing, these results imply the usual nonparametric minimax rate of \citet{stone_optimal_1982} up to a logarithmic factor.
    Although there are many different CDNN architectures used in practice and the CDNNs considered here are just one specific class, these results illustrate how Theorem \ref{thrm:ROC_V2} accommodates convolutional architectures. 
    Similar results could be obtained for other CDNN specifications by employing the methods developed here.

    The rest of the paper is organized as follows. 
    Section \ref{sec:gen_res} 
        considers a general nonparametric estimation setting and gives the main results for sieve estimators.
    Section \ref{sec:FCDNN} 
        defines the FCDNN architectures and applies the results of Section \ref{sec:gen_res}.
    Section \ref{sec:CDNN} 
        defines the CDNN architectures and also applies the results of Section \ref{sec:gen_res}. 
    Section \ref{sec:Conc} 
        concludes and discusses avenues for future research.
    An appendix provides general measurability results for sieve estimation settings, proofs for all of the results presented in the paper, and 
    a discussion regarding extensions to alternative architectures.

\paragraph{Notation}

            $\R$ denotes the real numbers,
            $\N$ denotes the natural numbers, and $\N_0\coloneqq\N\cup\{0\}$.
            For a probability space $\probspace$, measurable space $(\mathbb{M}, \mathcal{M})$ and
            a random variable 
            $\X:\Omega \to \mathbb{M}$,
            define 
            $\Px(M) = \P(\X^{-1}(M))$
            for any $M\in\mathcal{M}$.
            For a measure space $(\mathbb{M}, \mathcal{M}, \mu)$, and $r\in\N$,
            let 
                $\Lp{r}(\mu)$
            denote the space of measurable functions 
                $
                    f:
                    \mathbb{M}
                    \to 
                    \R
                $
            such that 
                $$
                    \norm{f}_{\Lp{r}(\mu)}
                \coloneqq
                    \left(\int_{\mathbb{M}}|f|^rd\mu\right)^{1/r}
                < \infty
                .
                $$
            %
            %
            For 
                $\x\in\Xspace$,
            $\x^\tr$ denotes the transpose of $\x$,
            and for          
                $\x=(x_1,\ldots,x_d)^\tr$ 
            define 
                $$
                    \norm{\x}_{\lp{1}} 
                = 
                    \sum_{j=1}^d
                    |x_j|
                ,
                \quad
                \text{and}
                \quad
                    \norm{\x}_{\lp{2}} 
                = 
                    \sqrt{
                    \sum_{j=1}^d
                    (x_j)^2
                    }
                .
                $$
            For a function $f:\mathbbm{X}\to\R$ 
            and a subset 
            $A\subseteq \mathbbm{X}$ 
            define 
            $$
                \norm{f}_\infty
            \coloneqq
                \sup_{\x\in\mathbbm{X}}|f(\x)|,
            \quad
            \text{and}
            \quad
                \norm{f}_{\infty|A}
            \coloneqq
                \sup_{\x\in A}|f(\x)|
            .
            $$
            {%
            \renewcommand{\Xspace}{\mathbb{X}}%
            For $A\subseteq \Xspace$, $\mathbbm{1}_{A}$ is the indicator function.
            %
            }%
            For two non-negative sequences 
            $\{x_n\}_{n\in\mathbb{N}}$ and $\{y_n\}_{n\in\mathbb{N}}$, 
        the notation 
            $x_n\lesssim y_n$ 
        means there exists a constant $0<C<\infty$ such that    
            $x_n\leq Cy_n$
        for all $n$ sufficiently large. 
        The notation 
            $x_n \asymp y_n$
        means
            $x_n \lesssim y_n$ and $x_n \gtrsim y_n$.
        For $x\in\R$, $\lceil x \rceil$ denotes the smallest integer that is $\geq x$,
        and $\lfloor x \rfloor$ denotes the largest integer that is $\leq x$.

\section{General results for sieve extremum estimators}\label{sec:gen_res}

This section studies nonparametric sieve extremum estimation under dependent data. 
Starting from a general sieve-estimation framework serves two purposes.
First, it accommodates many time-series forecasting problems in which DNNs are commonly used, including the benchmark examples introduced below. 
Second, the resulting theory applies to a broad range of DNN-based estimators, rather than to a particular architecture.
Sections \ref{sec:FCDNN} and \ref{sec:CDNN} demonstrate how these general sieve results can be applied to FCDNN and CDNN architectures, respectively.

\subsection{Estimation Setting}\label{sec:Sieve_Est_Meas}

        Let $\Yt\in \R$ be an outcome variable and $\Xt\in \Xspace$ be the covariates, possibly containing lagged values of $\Yt$. 
        The data are collected together as 
            $\Zt = (\Yt, \Xt^{\tr})^\tr\in\Zspace$
        where 
            $\{\Zt\}_{t\in\mathbb{N}}$
        forms a, possibly nonstationary, stochastic sequence on a complete probability space $\probspace$.
        The parameter space, 
            $\measurableftns$, 
        is a space of functions with elements $f:\Xspace\to\R$ such that
            $(\measurableftns,\norm{\cdot}_{\Lp{2}(\P_{\Xt})})$ 
        is a metric space for each $t\in\N$.%
\footnote{Throughout, semi-norms and pseudo-metrics will be referred to simply as norms and metrics, respectively; and objects of the form $(\mathbb{M},\norm{\cdot})$ will be referred to as metric spaces, with the understanding that this refers to the metric induced by the norm $\norm{\cdot}$.}
        The goal is to estimate an unknown function 
            $
                \fO\in\measurableftns
            $
        defined by 
            $
                \E[\Crit(\fO)] \leq \E[\Crit(f)], 
            $
        for all $n\in\N$, $f\in\measurableftns$, where 
            $$
                \Crit(f)
            \coloneqq
                \sampavg \crit(\Zt,f)
            ,
            $$
        is the empirical criterion function, and $\crit: \Zspace \times \measurableftns \to \R$
        is the single observation criterion function, 
        which is such that
            $\crit\big(\cdot,f\big) \in 
            \Lp{1}(\P_{\Zt})
            $,
        for each $t\in\N$, $f\in\measurableftns$.

    This setup 
    accommodates a broad collection of nonlinear forecasting specifications where DNNs are commonly used. To illustrate its scope, three nonparametric benchmark examples are introduced below:
    a regression specification, 
    a classification problem in a logistic binomial autoregression model,
    and 
    a conditional quantile autoregression.
    These will be used later on to provide concrete settings for verifying the role of key regularity conditions, and obtaining properties of FCDNN and CDNN estimators. 
    All of the following examples may include auto-regressions with covariates where 
        $
            \Xt=(\V_{t-1},\Y_{t-1},$ $\ldots\Y_{t-r})
        $
    for some random sub-vector of covariates $\Vt\in\R^{d-r}$ for $d>r$.

    \begin{example}\label{ex:reg}
    \textbf{\textsc{(Nonparametric regression)}}
\!\AENote{2}\!%
        For all $t\in\N$, let
        $\Yt\in\Lp{2}(\P)$,
        $h\in\Lp{2}\big(\P_{\Xt}\big)$, 
        and
        $\upsilon_t\in\Lp{2}(\P)$,
        such that 
            \begin{equation}\label{eq:ex1}
            \begin{aligned}
                \Yt
                =
                \fO(\Xt) + h(\Xt)\upsilon_t
                ,
                \quad
                \E[\upsilon_t|\Xt]=0.
            \end{aligned}
            \end{equation}
        Then, 
            $
            \fO(\Xt)
            =
            \E[\Yt|\Xt]
            ,\;
            $
        and
            $
                \crit(\Zt,f)
            =
                \big(\Yt-f(\Xt)\big)^2
            .
            $

        Example \ref{ex:reg} is the nonparametric location-scale model considered by \citet{kurisu_adaptive_2025}. 
        It encompasses many common time-series specifications, such as nonlinear AR-ARCH models 
        obtained by letting
            $\Xt=(Y_{t-1},\ldots,Y_{t-d})^\tr$.%
\footnote{See \citet{kurisu_adaptive_2025} for additional special cases of Example \ref{ex:reg}.}
        These specifications include many settings in which DNNs have been used for empirical forecasting tasks.
        For example, FCDNNs have been used to predict
        daily COVID-19 cases (\citealp{Chronopoulos_Panel_2023})
        and
        asset returns (\citealp{gu_empirical_2020,fan_structural_2022}), 
        while both FCDNN and CDNN architectures have been used to forecast macroeconomic indicators such as unemployment (e.g., \citealp{cook_macroeconomic_2017}).
    \end{example}

    \begin{example}\label{ex:log}
\!\AENote{2}\!%
    \textbf{\textsc{(Nonparametric logistic regression)}}
    For all $t\in\N$, let 
        $\Yt\in\{0,1\}$.
        Suppose for any 
        $t\in\N$,
        $$
        \E[\Yt|\Xt]
            = 
            \frac{e^{\fO(\Xt)}}{1+e^{\fO(\Xt)}}
        $$
        where $|\fO(\x)|<\infty$ for all $\x\in\R^d$.
        Then,
            $$
                \fO(\Xt)
            =
                \logp{
                    \frac
                    {\E[\Yt|\Xt]}
                    {1-\E[\Yt|\Xt]}
                }
            ,
            \quad
            \text{and}
            \quad
                \crit(\Zt,f) 
            = 
                -\Yt f(\Xt)
                +
                \logp{1+e^{f(\Xt)}}
            .
            $$
        Lagged values of the covariates $\V_{t-2},\ldots,\V_{t-r}$ can trivially be included by replacing 
        $\V_{t-1}$
        with 
        $\V_{t-1}^*\coloneqq(\V_{t-1},\ldots,\V_{t-r})$.

    Example \ref{ex:log} captures binary classification problems, which are common in forecasting applications using DNNs.
    For instance, FCDNNs have been used for directional stock-index prediction (\citealp{orimoloye_comparing_2020})
    and
    loan default prediction (\citealp{albanesi_predicting_2019,sadhwani_deep_2021}),
    while CDNNs have been used for tasks such as mortgage-default forecasting (\citealp{kvamme_predicting_2018}).
    \end{example}

%
%
%
%
%
    \begin{example}\label{ex:quant}
\!\AENote{2}\!%
    \textbf{\textsc{(Nonparametric quantile regression)}}
        For all $t\in\N$, let
        $\Yt\in\Lp{1}(\P)$,
        and
        $
            \Xt=(\V_{t-1},\Y_{t-1},$ $\ldots\Y_{t-r})
        $
        for some random vector of covariates $\Vt\in\R^{d-r}$ for $d>r$.
        Fix a quantile index $\tau\in(0,1)$ and suppose $\fO$ is the $\tau$-quantile of $\Yt$ conditional on $\Xt$ given by 
            \begin{equation}\label{eq:ex3}
            \begin{aligned}
                \E\Big[
                    \mathbbm{1}_{\{\Yt \leq \fO(\Xt)\}}
                \mid 
                    \Xt
                \Big]
                =
                    \tau
                \qquad
                \text{(almost surely)}
                .
            \end{aligned}
            \end{equation}
        Then, 
            $
                \crit(\Zt,f)
            =
                \big(
                    \tau
                    -
                    \mathbbm{1}_{\{\Yt \leq f(\Xt)\}}
                \big)
                \big(
                    \Yt - f(\Xt)
                \big)
            .
            $
            
    Example \ref{ex:quant} captures forecasting problems in which the target is a conditional quantile, rather than a conditional mean or event probability.
    These settings are prevalent in financial forecasting applications, such as forecasting Value-at-Risk with FCDNNs (\citealp{Chronopoulos_ForecastingDNNQuantile_2024}), and CDNNs (\citealp{petnehazi_quantile_2021}).
    \end{example}

Throughout, the stochastic process $\{\Zt\}_{t\in\N}$ will be allowed to be nonstationary and the dependence of $\{\Zt\}_{t\in\N}$ will be controlled using $\beta$-mixing.
The following definition for the $\beta$-mixing coefficient is from 
\citet[Definition 3.1, p.19]{dehling_empirical_2002}, and is equivalent to many standard definitions.
\begin{definition}\label{def:beta_mixing}
\textbf{\textsc{($\boldsymbol{\beta}$-Mixing)}}
    The $\beta$-mixing coefficient is defined as, 
            $$
                \beta(j) 
            \coloneqq
                \E\Bigg[
                    \sup\bigg\{
                    \Big|
                        \P\big(
                            B|\,
                            \sigma\big(\{\Zt\}_{1}^k\big)
                        \big)
                        -
                        \P\big(
                            B
                        \big)
                    \Big|
                    :\,
                    {B\in \sigma\left(\{\Zt\}_{k+j}^\infty\right)}
                    ,\,
                    k\in\N
                    \bigg\}
                \Bigg]
            \quad
            \text{ for }
            j\in\N
            ,
            $$
        where 
            $\sigma\left(\{\Zt\}_{k+j}^\infty\right)$
        denotes the $\sigma$-algebra generated by 
            $\{\Zt\}_{t=k+j}^\infty$.
        $\{\Zt\}_{t\in\mathbb{N}}$ is $\beta$-mixing 
        (or absolutely regular) 
        if 
            $
            \lim_{j\to\infty}\beta(j)=0
            $.
    \end{definition}


    Many popular forecasting models used in economics and finance imply $\beta$-mixing under mild conditions 
    (e.g., \citealp{doukhan_mixing_1994, bosq_1998, bradley_basic_2005}).
    For instance, Examples \ref{ex:reg} and \ref{ex:quant} include many nonlinear auto-regressive models for which conditions ensuring that the $\beta$-mixing coefficient decays at an exponential rate are given by \citet{mokkadem_mixing_1988}, \citet{Lu_GEOMETRIC_ERGODICITY_1998}, \citet{chen_geometric_2000}, and \citet{carrasco_mixing_2002}; or at a subexponential rate by \citet{MEITZ_SAIKKONEN_subgeometric2021}.
    For categorical time-series models, such as Example \ref{ex:log}, see \citet{truquet_Local_2019}, \citet{Fokianos_categorical_2019}, and \citet{truquet_Coupling_2020}; see also \citet{Neumann_AbsoluteRegularity_2011} for Poisson count-process models.

\begin{remark}\label{rem:nonstationarity}\textbf{\textsc{(Nonstationarity)}}
\RoneNote{1}%
    %
    Allowing for nonstationary $\beta$-mixing is relevant not only for generality, but also because such settings arise naturally in many autoregressive forecasting settings.
    For instance, in the AR-ARCH case of Example \ref{ex:reg}, where 
    $\Xt=(Y_{t-1},\ldots,Y_{t-d})^\tr$,
    geometric ergodicity of the associated Markov chain implies $\beta$-mixing under mild conditions (e.g., \citealp{Liebscher_mixing_ergodicity2005,MEITZ_SAIKKONEN_subgeometric2021}).
    In that case, the process has a limiting distribution, but it need not be stationary if the initial state is not drawn from that distribution.
    The results herein therefore apply directly to forecasting applications where the observed sample is generated from historical initial conditions. 
\end{remark}

\subsubsection{Sieve estimators}

    Estimating $\fO$ by directly optimizing $\Crit$ over the possibly infinite-dimensional space $\measurableftns$ is typically infeasible.
    Instead, $\fO$ can be estimated using the method of sieves (\citealp{grenander_abstract_1981}; \citealp{wooldridge_white_1991}; \citealp{chen_sieve_1998}).
    Sieve estimation covers a wide variety of nonparametric methods,
    including series, B-spline, and neural network estimators.
    See \citet{chen_chapter_2007}
    for a standard reference on sieve estimation.

    Let $\FMLPg$ denote the sieve spaces such that, for each $n$, $\mathcal{F}_{n-1}\subseteq\FMLPg$ and
    $(\FMLPg,\norm{\cdot}_{\Lp{2}(\P_{\Xt})})$ 
        is a metric space for any $t\in\N$.
    Sieve estimation optimizes over lower-complexity sieve spaces, 
    $\FMLPg$,
    that grow with sample size to approximate $\measurableftns$ increasingly well, but grow slow enough to deliver feasible and consistent estimators.
        If $\theta_n$ is a non-negative random variable such that
            $\theta_n = \op(1)$,
        then
             $\f\in\FMLPg$ 
        is an \textit{approximate sieve estimator} if
            \begin{equation}\label{eq:fhat}
                \Crit(\f) 
            \leq 
                \inf_{f\in\FMLPg} \Crit(f) + \theta_n
            .
            \end{equation}
        $\theta_n$ captures finite-sample optimization error. In some cases, it may also absorb discrepancies due to regularization, including additive penalty terms, although regularization is not considered explicitly herein.

        The complexity of the sieve spaces, $\FMLPg$, will be controlled using pseudo-dimension.
    The following definition is from \citet[Definition 2]{bartlett_nearly_tight_2019}.
\begin{definition}\label{def:pseudodimension}%
\textbf{\textsc{(Pseudo-dimension)}}
    Let $\mathcal{S}$ be a class of functions from $\mathbb{X}\rightarrow \R$. 
    The pseudo-dimension of $\mathcal{S}$, denoted as $\mathrm{Pdim}(\mathcal{S})$, 
    is the largest $p\in\N$ for which there exists 
        $(x_1,\ldots,x_p,y_1,\ldots,y_p)\in \mathbb{X}^p \times \R^p$ 
    such that for any 
        $(b_1,\ldots,b_p)\in \{0,1\}^p$ 
    there exists $s\in\mathcal{S}$ such that 
    $
    \mathbbm{1}{\{s(x_i)- y_i >0\}}
    = b_i,
    $
    for all $i=1,\ldots,p.$
\end{definition}

    Pseudo-dimension provides a scale-free measure of complexity and is particularly well suited for DNN sieves with unbounded weights. \citet{bartlett_nearly_tight_2019} provides nearly tight pseudo-dimension bounds for common DNN architectures. These bounds can be translated into covering number bounds when needed 
    (\citealp[Theorem 12.2]{anthony_bartlett_neural_1999}).

        \begin{remark}\label{rem:meas}\textbf{\textsc{(Measurability)}}
\RtwoNote{3}%
\AENote{3}%
        When $\FMLPg$ is uncountable, the infimum in \eqref{eq:fhat}, and the mapping 
        $\omega\mapsto \f$ from $\Omega$ to $\FMLPg$ may not be measurable.
        Outer probability/integrals will not suffice for the proof of Theorem \ref{thrm:ROC_V2}, and standard compactness-based arguments commonly used for series and spline sieves (e.g., \citealp{wooldridge_white_1991}) do not apply directly when $\FMLPg$ is constructed using DNNs with unbounded weights.
        Appendix \ref{sec:meas} establishes 
        mild sufficient conditions to ensure existence and measurability of these objects and verifies that the DNN sieve spaces herein satisfy them.
        Throughout, all relevant objects are assumed measurable, and $\FMLPg$ is assumed to be finite-point separable (see Definition \ref{def:point_sep} in Appendix \ref{sec:meas}).
        \end{remark}

\subsection{Nonasymptotic error bounds}\label{sec:EB_SD}
This subsection presents Theorem \ref{thrm:ROC_V2}, which gives finite-sample $\Lp{2}$-error bounds for sieve estimators when $\{\Zt\}_{t\in\N}$ is a possibly nonstationary $\beta$-mixing process.
Below, I introduce generic \textit{conditions} for the general sieve estimation setting. 
In Sections \ref{sec:FCDNN} and \ref{sec:CDNN}, these are verified for DNN sieve estimators and the benchmark examples under model-specific \textit{assumptions}. This distinction separates the high-level sieve estimation framework from the more specialized assumptions used in the FCDNN and CDNN applications.



The first condition is a standard curvature requirement on the criterion function around the true parameter $\fO$.
This condition is standard in the nonparametric estimation literature (e.g., \citealp{chen_sieve_1998,chen_chapter_2007,farrell_deep_2021}). 

\begin{condition}\label{as:curv}\conlab{Curvature} 
    There exist constants $\Cpcq,\CpcqB>0$ with $\Cpcq\leq 1$, such that, for all $f\in\FMLPg$ and $t\in\N$, 
        $$
            \Cpcq 
            \E\Big[ 
                |f(\Xt)-\fO(\Xt)|^2
            \Big]
        \;\leq\;
            \E\big[\crit(\Zt,f)\big] - \E\big[\crit(\Zt,\fO)\big]
        \;\leq\;
            \CpcqB
            \E\Big[ 
                |f(\Xt)-\fO(\Xt)|^2
            \Big]
        .
        $$
\end{condition}
This curvature condition
holds for many common choices of $\crit$ under typical regularity conditions.
In Section \ref{sec:FCDNN}, Condition \ref{as:curv} will hold immediately for Example \ref{ex:reg} and holds for Examples \ref{ex:log} and \ref{ex:quant} under the regularity assumptions provided in Subsections 
\ref{sec:FCDNN_log} and \ref{sec:FCDNN_quant}, respectively.
See also \citet[Lemma 9]{farrell_deep_2021} for additional examples. 
Note that requiring  $\Cpcq\leq 1$ is without loss of generality, since if Condition \ref{as:curv} holds for some $\Cpcq'> 1$, then it also holds with $\Cpcq' = 1$.


The next condition introduces truncation sets $\Zbound\subseteq\Zspace$ to accommodate settings in which $\Zt\in\Zspace$ may take values in unbounded sets and the target function $\fO$ need not 
be bounded on $\Xspace$.
This is particularly relevant in autoregressive settings, such as the AR-ARCH case of Example \ref{ex:reg}, where 
    $\Xt=(Y_{t-1},\ldots,Y_{t-d})^\tr$
and $\Yt$ may be unbounded, for example under Gaussian innovations.


\begin{condition}\label{as:TruncLipEnv}
    Let 
    $\Zbound=\Ybound\times\Xbound\subseteq \R \times \Xspace$ be a measurable set. 
    \vspace{4pt}
    \begin{enumerate}
    [label=$\mathrm{\ref*{as:TruncLipEnv}.\arabic*}$]
    \item\label{as:Env}\conlab{Envelope}
            There exists 
            $\Fbound\in[1,\infty)$
            such that
            $
            \Fbound \geq \sup_{f\in\left\{\FMLPg\cup\{\fO\}\right\}}\snormDNNBound{f}
            $.
        \vspace{5pt}
        \item\label{as:lip}\conlab{Lipschitz}
        There exists 
        $\Mbound\in[1,\infty)$
            such that
        for any 
            $f,f'\in\left\{\FMLPg\cup\{\fO\}\right\}$ and $\z\in\Zspace$
            $$
                \big| \crit(\z,f) - \crit(\z,f') \big|
                \mathbbm{1}_{\{\z\in\Zbound\}}
            \leq 
                \Mbound
                \big|f(\x)-f'(\x)\big|
                \mathbbm{1}_{\{\x\in\Xbound\}}
            ,
            $$
    \end{enumerate}
\end{condition}

Condition \ref{as:Env} allows $\fO$ to be bounded only on subsets of $\Xspace$, such as when $\fO$ belongs to a H\"older class.
Condition \ref{as:lip} weakens standard Lipschitz assumptions used in nonparametric estimation (e.g., \citealp{farrell_deep_2021,kengne_deep_2026}) by requiring the condition only on the truncation sets $\Zbound\subseteq\Zspace$ and permitting the Lipschitz constant to grow with $n$.
In bounded-support settings, one can often take $\Fbound$ and $\Mbound$ to be fixed constants.
Section \ref{sec:FCDNN} gives model-specific assumptions verifying Condition \ref{as:TruncLipEnv} for Examples \ref{ex:reg}, \ref{ex:log}, and \ref{ex:quant}.


To quantify the effect of the truncation argument in Theorem \ref{thrm:ROC_V2}, define the tail event 
    $$
        \Ptrunc
    \coloneqq
        \bigcup_{t\in\{1,\ldots,n\}}
        \Big\{
        \Zt \notin \Zbound
        \Big\}
    ,
    $$
and define the truncation error terms $\truncf<\infty$ and $0<\trunc<\infty$ such that
    \begin{equation}\label{eq:trunc_term_def}
    \begin{gathered}
        \truncf
    \coloneqq
                \max_{t\in \{1,\ldots,n\}}
                \sup_{f\in\left\{\FMLPg\cup\{\fO\}\right\}}
                \E\Big[
                    |f(\Xt)|^2
                    \mathbbm{1}_{\{\Xt\notin\Xbound\}}
                \Big]
            ,
    \\
        \trunc
    \geq
            \max_{t\in \{1,\ldots,n\}}
            \sup_{f\in\left\{\FMLPg\cup\{\fO\}\right\}}
                    \E\Big[
                        \big|\crit\big(\Zt,f\big)\big|
                        \cdot
                        \mathbbm{1}_{\{\Zt\notin\Zbound\}}
                    \Big]
    .
    \end{gathered}
    \end{equation}
Here $\P(\Ptrunc)$ captures the probability that the sample takes values outside of the truncating sets, 
while $\trunc$ and $\truncf$ measure the contribution of these tail events to the 
error bounds in Theorem \ref{thrm:ROC_V2}.
With this, rather than requiring $\Zt$ and $\fO$ to be uniformly bounded, it suffices that 
$\P(\Ptrunc)$, $\truncf$, and $\trunc$ 
become small as $n$ grows.

    Define the approximation error term $\prjerr\geq 0$ such that there exists $\fpr\in\FMLPg$ satisfying 
    $$
    \snormDNNBound{\fpr-\fO} \leq \prjerr
    .
    $$
    $\prjerr$ is typically obtained using existing approximation results for the sieve spaces under consideration. 
    For instance, Subsection \ref{sec:FCDNN_res} uses \citet[Theorem 2]{kohler_rate_2021} to bound $\prjerr$ for FCDNN sieve spaces when $\fO$ belongs to a H\"older smoothness class.

    In practice, the truncation sets $\Zbound$ must be chosen to balance the tail terms $\P(\Ptrunc)$, $\truncf$, and $\trunc$ against the envelope, Lipschitz, and approximation terms $\Fbound$, $\Mbound$, and $\prjerr$. 
    Section \ref{sec:FCDNN_res} chooses $\Zbound$ to balance this tradeoff under a subgaussian tail assumption on $\{\Zt\}_{t\in\N}$, and a H\"older smoothness assumption on $\fO$.

    The following result gives finite-sample probability bounds on $\Lp{2}$-errors for general sieve estimators under dependent data. It serves as the basis for the DNN results in Sections \ref{sec:FCDNN} and \ref{sec:CDNN}.

\begin{theorem}\label{thrm:ROC_V2}
    There exists a constant $C>0$, depending only on 
        $\Cpcq$
        and
        $\CpcqB$,
    such that 
    if Conditions \ref{as:curv} and \ref{as:TruncLipEnv} hold,
    and 
        $n,\anA\in\N,\,$
        $\betaBound:\N\to(0,\infty),\,$ 
        $\delta>0$
    satisfy
\!\AENote{5}\!%
\!\RtwoNote{4}\!%
        \begin{equation}\label{eq:pbnd}
        \begin{gathered}
            n
            \geq
                \max\!\left\{
                    8
                ,\;
                    4\Pdim(\FMLPg)
                ,\;
                    \frac{16\Fbound^2}{\log(n)}
                \right\}\!
        ,
        \qquad
            \beta(\anA)\leq \betaBound(\anA) < \frac{\anA}{n}
        ,
        \qquad
            2
        <
            \anA
        \leq
                \frac{n}{2}
        ,
        \\[6pt]
        \text{ and }
        \quad
            \delta
        \geq
                \frac{4 \prjerr}{\Fbound}
                \!
                \left[
                \sqrt{
                    \frac{n}{\anA}
                    \log\!\left(
                        \frac{1}{\betaBound(\anA)}
                    \right)
                }
                +
                \frac{4}{3} 
                \log\!\left(
                        \frac{1}{\betaBound(\anA)}
                    \right)
                \right]\!
        \!,
        \end{gathered}
        \end{equation}
    then the following holds for any
    $\f$
    satisfying \eqref{eq:fhat}
        \begin{equation*}
        \begin{aligned}
            \Pr\!\left(
                \sampavg
                \int_{\Xspace}
                        \big[\f(\x)-\fO(\x)\big]^2
                \,
                d
                \P_{\Xt}\!(\x)
            >
                C\,
                \err(\delta,\anA)
            \right)
        &
        \,
        \leq
        \,
            e^{-\delta}
            +
            4\log(n)\!
            \left[\!
                \frac{n\betaBound(\anA)}{\anA}
                +
                \Pr(\Ptrunc)
            \right]\!
        , 
        \\ 
        \text{and} \quad
            \Pr\!\left(
                \sampavg
                        \big[\f(\Xt)-\fO(\Xt)\big]^2
            >
                C\,
                \err(\delta,\anA)
            \right)
        &
        \,
        \leq
        \,
            4e^{-\delta}
            +
            9\log(n)\!
            \left[\!
                \frac{n\,\betaBound(\anA)}{\anA}
                +
                \Pr(\Ptrunc)
            \right]\!
        , 
        \end{aligned}
        \end{equation*}
    where
    $
    \err(\delta,\anA)
        \coloneqq
            \big(
            \max\big\{\Mbound,\Fbound\big\}\big)^2\,
            \frac{\anA}{n}
            \Big[
                        \Pdim(\FMLPg)
                        \log(n)
                +
                    \delta
            \Big]
            +
                \prjerr^2
                +
                \trunc
                +
                \truncf
                +
                \theta_n
    $.
\end{theorem}

Although Theorem \ref{thrm:ROC_V2} is nonasymptotic, it yields convergence rates whenever sequences $\delta_n$ and $\anA_n$ can be chosen such that $\err(\delta_n,\anA_n)$ and the probability bound both converge to zero as $n$ grows. 
In that case, the bounds on $n$, $\betaBound$, and $\anA$ in \eqref{eq:pbnd} clearly hold for all sufficiently large $n$.
The lower bound on $\delta$ in \eqref{eq:pbnd} serves mainly to simplify $\err$ by absorbing terms arising from concentration inequalities used in the proof; it is typically not restrictive for large $n$ but could be weakened by incorporating those terms directly into $\err$.
For the DNN results in the following subsections, it will suffice to set $\delta_n\asymp \Pdim(\FMLPg)\log(n)$.

Theorem \ref{thrm:ROC_V2}%
\RoneNote{2}%
extends
\citet[Theorem 2]{farrell_deep_2021}, which considered feedforward DNNs with \iid data taking values in compact sets.
While blocking arguments for $\beta$-mixing processes are standard in broad terms, 
the present extension
is not immediate.
The proof must be reworked to accommodate dependence, truncation, and nonstationarity
throughout,
and requires 
extending the key maximal deviation bound of \citet{bartlett_local_2005} to the blocked nonstationary setting considered here.
The present approach is also distinct from that of 
\citet[Lemmas C.1, C.2]{kurisu_adaptive_2025}
which also consider general nonparametric estimators under nonstationary $\beta$-mixing data, but only  establish bounds on generalization error and expected empirical error (see Remark \ref{rem:risk}) for estimators of the mean function.


Theorem \ref{thrm:ROC_V2} controls sieve complexity through pseudo-dimension, which can be bounded for DNNs with unbounded weights using results such as \citet{bartlett_nearly_tight_2019}.
In contrast, existing DNN results for dependent data typically impose conditions based on entropy with bracketing or supremum norms (e.g., \citealp{kengne_robust_2025, kurisu_adaptive_2025, Kengne_Excess_2025, kengne_deep_2026,yara_nonparametric_2026}), which are not well suited to DNN classes with unbounded weights.
Theorem \ref{thrm:ROC_V2} also differs from this literature by yielding nonasymptotic probability bounds for $\Lp{2}$-errors. The following remark makes this distinction more precise by relating Theorem \ref{thrm:ROC_V2} to several risk criteria studied in the recent DNN literature.

\begin{remark}\label{rem:risk}
\textbf{\textsc{(Excess-risk and expected-risk bounds)}} 
Theorem \ref{thrm:ROC_V2} yields bounds for several related criteria studied in the recent DNN literature.
First, \citet{Kengne_Excess_2025} considers excess-risk bounds in probability. Theorem \ref{thrm:ROC_V2} implies the corresponding nonstationary version, since Condition \ref{as:curv} gives
    $$
        \sampavg
            \int_{\Zspace}
            \big[
                \crit(\z,\f)-\crit(\z,\fO)
            \big]
            d\P_{\Zt}(\z)
    \leq
        \CpcqB
        \sampavg
        \int_{\Xspace}
            \big[\f(\x)-\fO(\x)\big]^2
        \,d\P_{\Xt}(\x).
    $$
%
Second, \citet{kohler_rate_2021} study expected $\Lp{2}$-error under \iid\!data. Theorem \ref{thrm:ROC_V2} implies the analogous time-averaged expected $\Lp{2}(\P_{\Xt})$-error bound,
    \begin{equation*}
    \resizebox{\hsize}{!}{$
    \begin{aligned}
        \E\!\left[
            \sampavg
                \int_{\Xspace}\!
                        \big[\f(\x)-\fO(\x)\big]^2
                d
                \P_{\Xt}\!(\x)
        \right]\!
    &
    \leq    
        C\err 
        + 
        2(\Fbound^2+\truncf)
        \!\left(\!e^{-\delta}\!
            +
            4\log(n)\!
            \left[\!
                \frac{n\betaBound(\anA)}{\anA}
                +
                \Pr(\Ptrunc)\!
            \right]\!\right)\!.
    \end{aligned}
    $}
    \end{equation*}
Third, \citet{kengne_deep_2026} and \citet{kengne_robust_2025} consider expected excess-risk. Theorem \ref{thrm:ROC_V2} yields the corresponding nonstationary version by combining the expected $\Lp{2}(\P_{\Xt})$-error bound above with Condition \ref{as:curv}.
Similarly, in conditional mean regression settings, \citet{kurisu_adaptive_2025} evaluate prediction error on $\{\Xt^*\}_{t=1}^n$ that is independent of $\{\Xt\}_{t=1}^n$ and has the same distribution.
This will also follow from Theorem \ref{thrm:ROC_V2} using the time-averaged expected $\Lp{2}(\P_{\Xt})$-error bound above 
since $\P_{\Xt}=\P_{\Xt^*}$ for each $t$.
\end{remark}

\section{Feedforward fully connected deep neural networks}\label{sec:FCDNN}
    
    This section applies 
    Theorem \ref{thrm:ROC_V2}
    to obtain nonasymptotic error bounds for FCDNN estimators.
    Subsection \ref{sec:DNN_sieve_spaces} introduces the 
    FCDNN sieve spaces.
    Then, Subsection \ref{sec:FCDNN_res} presents the main results for FCDNN estimators.
    Theorem \ref{thrm:FCDNN_ROC} gives a result for general estimation settings, which is then applied to obtain concrete error bounds in the settings of Examples \ref{ex:reg}, \ref{ex:log}, and \ref{ex:quant}.

\newcommand{\Ctruncorder}{\const{Ctruncorder}}%

%

    \subsection{FCDNN architectures}%
    \label{sec:DNN_sieve_spaces}
    The architecture $\FMLP=\FMLPnon(\L,\Hb,\DNNBound)$ 
    is
    specified by 
    three
    user-chosen components: 
        (i) depth $\L\in\mathbb{N}$;
        (ii) width vector $\Hb=[\Hi{1},\ldots,\Hi{\L}]^{\tr} \in \N^{\L}$; 
        (iii) uniform output bound $\DNNBound\geq 2$. 
    The subscript $n$ indicates that $\L,\Hb,\DNNBound$ may grow with sample size.
    For concreteness, I focus on networks using the ReLU activation 
    $\relu(x) = \max\{0,x\}$,
    though all results herein apply similarly with any continuous piecewise linear activation (see Corollary \ref{prop:nonReLU}).
    Appendix \ref{sec:DNN_Alt_Arch} also discusses extensions to other architectures and classes of activation functions.

    The structure of $\FMLP$ is a directed acyclic graph 
    composed of sequentially connected layers where connections are restricted to adjacent layers only.
    The 
    $\Hi{0}\coloneqq d$
    inputs 
        $\x=[x_1,\ldots,x_d]^\tr$
    are passed 
    through $l=1,\ldots,\L$ hidden layers 
    comprised of
    $\Hi{l}$ nodes, denoted as $\nff^{(l)}_{h}$ for $h\in\{1,\ldots,\Hi{l}\}$. 
    Node $\nff^{(l)}_{h}$ receives inputs from all nodes in the previous layer,
    performs a computation, and passes its output to all nodes in the subsequent layer ($l+1$). 
    The final hidden layer $l=\L$ feeds into a single output node $\nff^{(\L+1)}_{1}$ which produces the network’s final output. See Figure \ref{fig:DNN_MLP} for an example of $\FMLP$.

{

\newcommand{\lin}{0}
\newcommand{\lI}{3}
\newcommand{\lII}{7}
\newcommand{\figlabLI}{\lII}
\newcommand{\lIII}{11}
\newcommand{\lout}{14}
\newcommand{\shft}{3pt}

\begin{figure}[t]    
    \centering

    \definecolor{input_node_fg}{HTML}{E90F0E}
    \definecolor{input_node_bg}{HTML}{F5C5C1}
    \definecolor{hidden_node_fg}{HTML}{3972FE}
    \definecolor{hidden_node_bg}{HTML}{D9E8FB}
    \definecolor{reduce_node_fg}{HTML}{8A6D3B}
    \definecolor{reduce_node_bg}{HTML}{FCF8E3}
    \definecolor{output_node_fg}{HTML}{9DBD88}
    \definecolor{output_node_bg}{HTML}{D6E7D4}

    
    \begin{tikzpicture}[
        node distance=0.55cm,
        input/.style={circle, draw=input_node_fg, thick, minimum size=0.8cm, inner sep=0pt, fill=input_node_bg},
        hidden/.style={rectangle, draw=hidden_node_fg, thick, minimum size=0.8cm, inner sep=0pt, fill=hidden_node_bg},
        output/.style={diamond, draw=output_node_fg, thick, minimum size=0.1cm, inner sep=1.0pt, fill=output_node_bg},
        conn/.style={->, shorten >= 1.8pt}
    ]
    

        \node[input] (x1) at (\lin, 0.75) {$x_{1}$};
        \node[input] (x2) at (\lin, -0.75) {$x_{2}$};

        \node[hidden] (u11) at (\lI, {0.6+1.2}) {$\!\nff^{(1)}_{1}\!$};
        \node[hidden] (u12) at (\lI, 0.6) {$\!\nff^{(1)}_{2}\!$};
        \node[hidden] (u13) at (\lI, -0.6) {$\!\nff^{(1)}_{3}\!$};
        \node[hidden] (u14) at (\lI, {-0.6-1.2}) {$\!\nff^{(1)}_{4}\!$};

        \node[hidden] (u21) at (\lII, 1.4) {$\!\nff^{(2)}_{1}\!$};
        \node[hidden] (u22) at (\lII, 0) {$\!\nff^{(2)}_{2}\!$};
        \node[hidden] (u23) at (\lII, -1.4) {$\!\nff^{(2)}_{3}\!$};

        \node[hidden] (u31) at (\lIII, 1.0) {$\!\nff^{(3)}_{1}\!$};
        \node[hidden] (u32) at (\lIII, -1.0) {$\!\nff^{(3)}_{2}\!$};

        \node[output] (y) at (\lout, 0.0) {$\!\nff^{(4)}_{1}\!$};

        \draw[conn] (x1.east) -- (u11);
        \draw[conn] (x1.east) -- (u12);
        \draw[conn] (x1.east) -- (u13);
        \draw[conn] (x1.east) -- (u14);
        \draw[conn] (x2.east) -- (u11);
        \draw[conn] (x2.east) -- (u12);
        \draw[conn] (x2.east) -- (u13);
        \draw[conn] (x2.east) -- (u14);

        \draw[conn] (u11.east) -- (u21);
        \draw[conn] (u11.east) -- (u22);
        \draw[conn] (u11.east) -- (u23);
        \draw[conn] (u12.east) -- (u21);
        \draw[conn] (u12.east) -- (u22);
        \draw[conn] (u12.east) -- (u23);
        \draw[conn] (u13.east) -- (u21);
        \draw[conn] (u13.east) -- (u22);
        \draw[conn] (u13.east) -- (u23);
        \draw[conn] (u14.east) -- (u21);
        \draw[conn] (u14.east) -- (u22);
        \draw[conn] (u14.east) -- (u23);

        \draw[conn] (u21.east) -- (u31);
        \draw[conn] (u21.east) -- (u32);
        \draw[conn] (u22.east) -- (u31);
        \draw[conn] (u22.east) -- (u32);
        \draw[conn] (u23.east) -- (u31);
        \draw[conn] (u23.east) -- (u32);

        \draw[conn] (u31.east) -- (y);
        \draw[conn] (u32.east) -- (y);

        \newcommand{\labfontsz}{10pt}
        \newcommand{\labheight}{2.1}
        \node at (\lin,\labheight) {\fontsize{\labfontsz}{2pt}\textbf{Inputs}};
        \node at (\figlabLI,{\labheight+0.4}) {\fontsize{\labfontsz}{2pt}\textbf{Hidden Layers}};
        \node at (\lout,{\labheight-0.0}) {\fontsize{\labfontsz}{2pt}\textbf{Output}};

    \end{tikzpicture}

    \caption{
    Example of $\FMLP$ architecture graph structure where 
        $\L=3$, $\Hi{1} =4$, $\Hi{2}=3$, $\Hi{3}=2$, $\W=38$, and $d=2$.
    }
    \label{fig:DNN_MLP}
\end{figure}
}

    Formally, each node $\nff^{(l)}_{h}$ is a function mapping $\R^{\Hi{l-1}}$ to $\R$ parameterized by a 
    vector of DNN weights,%
\footnote{
    In other literature the network parameters $\{\g_{l,h,k}: \forall\, l,h,k\}$ are often referred to as the weights and biases. Throughout these will be referred to collectively as the DNN weights.
}
\!\RtwoNote{1}\!%
    $$
    [\gamma^{(l)}_{h,0},\gamma^{(l)}_{h,1},\ldots,\gamma^{(l)}_{h,\Hi{l-1}}]^\tr \in \R^{\Hi{l-1}+1}.
    $$
    The weights for all nodes in $\FMLP$ are collected together as 
        $
                \gV
            \coloneqq 
                \{\g^{(l)}_{h,k}: \forall\, l,h,k\}
            \in 
                \R^{\W},
        $
    where $\W$ is the total number of weights in the network. Each node can be written as
        \begin{equation}\label{eq:nodes}
            \nff^{(l)}_{h}\left(\x \mid \gV \right) \coloneqq \left\{
        \begin{aligned}
        &
            \relu\bigg( 
            \sum_{i=1}^{\Hi{l-1}} \g^{(l)}_{h,i} \cdot \nff^{(l-1)}_{i}(\x \mid \gV) + \g^{(l)}_{h,0}
            \bigg),
            & 1\leq l \leq \L,
        \\& 
            \sum_{i=1}^{\Hi{\L}} \g^{(\L+1)}_{1,i} \cdot \nff^{(\L)}_{i}(\x \mid \gV) + \g^{(\L+1)}_{1,0},
            &  l = \L + 1,
        \end{aligned}
        \right.
        \end{equation}
    with the convention that $\nff^{(0)}_{i}(\x)=x_i$, $i=1\ldots,d$, for the inputs to layer $l=1$.

    The architecture $\FMLP = \FMLPnon(\L,\Hb,\DNNBound)$ is a class of functions
    defined as 
        \begin{equation}\label{eq:sievespace}
            \FMLPnon(\L,\Hb,\DNNBound)
        \coloneqq
            \bigg\{ 
                f:
                \;
                f(\x)
                =
                \max\Big\{\! -\!\DNNBound,\, \min\{\DNNBound, \nff^{(\L+1)}_{1}(\x \,| \gV)\}
                \Big\}
                ,
                \;  
                \gV \in \R^{\W}
            \bigg\},
        \end{equation}
    i.e., the truncation of $\nff^{(\L+1)}_{1}(\cdot \,|\, \cdot)$ to the range 
    $[-\DNNBound,\DNNBound]$.
    To simplify notation, let the widths 
    $\{\Hb=[\Hi{1},\ldots,\Hi{\L}]^{\tr}\}_{n\in\N}$ share common order $\H$, such that for some $\Cwidthmax\geq 1$
        \begin{equation}\label{eq:width_bound_order}
        \H \leq \Hi{l} \leq \Cwidthmax\H
        ,
        \;\; \forall l\in\{1,2,\ldots,\L\}
        ,\;\;
        \forall n\in\N
        .
        \end{equation}

    Note that the \textit{architecture} $\FMLP$ is a class of functions, and a \textit{network} $f\in \FMLP$ is a function
    $f : \R^d \rightarrow [-\DNNBound,\DNNBound]$ 
    determined by the DNN weights $\gV\in\R^{\W}$. 
    The estimator \eqref{eq:fhat} is a network that is typically `learned' using stochastic gradient descent and backpropagation algorithms which choose $\gV$ to minimize $\Crit$ evaluated on observations of $\data$.

\begin{remark}\label{rem:FCDNN_arch}
\textbf{\textsc{(Choosing the DNN architecture)}} 
\!\AENote{2}\!
The user-chosen components $\L$ and $\Hb$ are central to the performance of an FCDNN estimator. As with classical sieve estimators, there is a tradeoff between approximation and overfitting: increasing $\L$ or $\Hb$ improves approximation power but also raises the risk of overfitting.%
\footnote{See \citet{farrell_deep_2021} for discussion of parallels between DNNs and traditional nonparametric estimators.}
For DNNs, however, this tradeoff is less transparent because the interaction between depth and width is highly complex.

Empirical evidence often favors multilayer architectures over single-layer networks (see \citealp{szegedy_inception_v4_2016,orimoloye_comparing_2020,schmidt_hieber_nonparametric_2020} and citations therein), and theory suggests that depth should increase with sample size (\citealp[Theorem 6]{yarotsky_error_2017}). 
However, whether one should prefer depth $\L$ and widths $\Hb$ growing with $n$, or fixed widths and $\L$ growing with $n$ remains less clear.
Building on \citet{kohler_rate_2021}, the results below cover both regimes and show that near minimax rates are attainable for both, without imposing boundedness or sparsity restrictions on the DNN weights $\gV$.

A rough guideline sometimes used in practice is to choose $\Hb$ according to a geometric pyramid rule (\citealp{masters_practical_1993}). 
For example, in a setting similar to Example \ref{ex:reg}, \citet{gu_empirical_2020} forecast stock returns and find that, for their sample, an FCDNN estimator with $\L=3$ and $\Hb=[32,16,8]$ outperforms other architectures as well as linear and random forest models.

Similar to bandwidth selection for kernel estimators, cross-validation is also used to select FCDNN architectures. In a setting related to Example \ref{ex:reg}, \citet{Chronopoulos_Panel_2023} forecast daily new COVID-19 cases using cross-validation over candidate architectures with depths ranging from $1$ to $15$ layers and fixed widths ranging from $5$ to $30$ nodes per layer. For Example \ref{ex:log}, \citet{orimoloye_comparing_2020} predict stock-price direction using FCDNN architectures selected by cross-validation over networks with $2$ to $10$ hidden layers and $10$ to $100$ nodes per layer. For Example \ref{ex:quant}, \citet{Chronopoulos_ForecastingDNNQuantile_2024} forecast Value-at-Risk using cross-validation over architectures with depths between $1$ and $15$ layers and widths between $1$ and $10$ nodes per layer.

Many empirical DNN applications also employ regularization, including weight decay, dropout, and early stopping.
While useful in many cases, recent evidence suggests that the empirical gains from these techniques are not always robust across applications, and that model architecture often plays the more important role in performance (e.g., \citealp{zhang_understanding_2021,Balestriero_regularization_2022}, and citations therein). 
Given the wide range of regularization methods and the fact that their role is not yet well understood, 
the results here do not explicitly incorporate such techniques.
Nevertheless, the framework introduced here may be extended in future work to accommodate regularized estimators, for example through penalty terms absorbed into $\theta_n$ in \eqref{eq:fhat}.

\end{remark}

    \subsection{Nonasymptotic error bounds for FCDNNs}\label{sec:FCDNN_res}

    This subsection applies Theorem \ref{thrm:ROC_V2} to FCDNN sieve estimators where $\FMLPg$ is constructed using $\FMLP$, and then derives concrete results for the forecasting examples from Section \ref{sec:Sieve_Est_Meas}.
    To do so, model-specific assumptions are introduced to verify the conditions of Theorem \ref{thrm:ROC_V2}.


    In this setting, pseudo-dimension can be bounded using \citet{bartlett_nearly_tight_2019}, 
    and the approximation error $\prjerr$ can be controlled using \citet[Theorem 2]{kohler_rate_2021} \!\RtwoNote{2}\!under the following H\"older smoothness assumption on 
    $\fO$.%
\footnote{
I thank an anonymous referee for suggesting this approximation result.
}
    This type of smoothness assumption is standard in the nonparametric estimation literature (e.g. \citealp{stone_optimal_1982}; \citealp{chen_sieve_1998}; \citealp{chen_chapter_2007}; \citealp{chen_optimal_2015}; \citealp{farrell_deep_2021}).%
\footnote{
I focus on this H\"older smoothness class as a standard benchmark in nonparametric estimation.
Under stronger structural assumptions on $\fO$, results that mitigate the curse of dimensionality could in principle be obtained by combining the framework developed here with alternative approximation results, for example for hierarchical composition classes via \citet[Theorem 3]{kohler_rate_2021}. I leave such extensions for future work.
}

\begin{assumption}\label{as:data_holder} 
    \newcommand{\tempset}{\Xspace}%
        For 
        a smoothness parameter
            $\smooth=\holD+\holE$ 
        where
            $\holD \in \N_0$, 
            $\holE\in(0,1]$, 
        let $\fO\in\HolderCust{\smooth}$,
        where
        $\HolderCust{\smooth}(\tempset)$
        is defined 
        as the space of functions $f:\tempset\to\R$ such that, 
        for
        any multi-index  
            $\kb \in 
            \N_0^d
            $
        with
            $ 
            \norm{\kb}_{\lp{1}}
            \leq \holD$,
        the mixed partial derivative of order $\kb$, denoted by $D^{\kb}f$, exists and 
            $$
            \abs{
                D^{\kb}f(\x)
                -
                D^{\kb}f(\x')
            }
            \leq
                \norm{\x-\x'}^{\holE}_{\lp{2}}
            ,
            \quad
            \forall\,
            \x,\x'\in \tempset
            .
            $$
    \end{assumption}


    The following assumption requires the distributions of $\{\Zt\}_{t\in\N}$ to satisfy a uniform subgaussian tail bound.
    Together with Assumption \ref{as:data_holder}, it is used to choose suitable truncation sets $\Zbound$.
    Although this tail condition could likely be weakened, doing so would generally lead to less sharp probability and error bounds in the results below.

    \begin{assumption}\label{as:data_subg}
    \!\RoneNote{5}\!%
        There exists a constant $\Csubg>0$ such that, for any 
        sufficiently large
        $z>0$,
        the random variables 
        $\{\Zt\}_{t\in\N}$ satisfy the tail bound 
        $$
            \max_{t\in\N}
            \Pr\Big(
            \Zt\notin [-z,z]^{\dz}
            \Big)
        \leq
            2
            e^{-\Csubg z^2}    
        .
        $$
    \end{assumption}

    Assumption \ref{as:data_subg}
    allows for cases where
    $\{\Xt\}_{t\in\N}$, $\{\Yt\}_{t\in\N}$, or both, are not uniformly bounded, and holds trivially when they are bounded.
    To cover both cases, 
    choose 
        $\Zbound
    \coloneqq\Ybound\times\Xbound
    =
    [-\Yboundval,\Yboundval]\times[-\Xboundval,\Xboundval]^{d}
    $
    where
    \begin{equation}\label{eq:xbound_ybound_choice}
    \resizebox{0.91\hsize}{!}{$
        \Xboundval\!
    =
        \begin{cases}
        1, 
        &
        \mathrm{if}\,
        \max_{t\in\N}\P(\Xt\notin [-1,1]^d)\!=\!0,
        \\[-6pt]
        \log(n),
        \hspace{-6pt}
         &\mathrm{otherwise },
        \end{cases}
    \;\; 
    \mathrm{and} 
    \;\;
        \Yboundval\!
    =
        \begin{cases}
        1, 
        &
         \mathrm{if}\,
         \max_{t\in\N}\P(\Yt\notin [-1,1])\!=\!0,
        \\[-6pt]
        \log(n),
        \hspace{-6pt}
         &\mathrm{otherwise }.
        \end{cases}
    $}
    \end{equation}
    Then,
    Lemma \ref{lem:Ptrunc_truncf_env_verify} in Appendix \ref{sec:PROOF_SEC_DNN}
    shows 
        $
                \P(\Ptrunc)
            \lesssim
                n^{1-\Csubg\log(n)}
            .
        $
    For the results below, it always suffices to set $\Xboundval=\Yboundval=\log(n)$. However, \eqref{eq:xbound_ybound_choice} yields slightly tighter error-bounds when $\{\Xt\}_{t\in\N}$ or $\{\Yt\}_{t\in\N}$ are known to be bounded, such as Example \ref{ex:log} where $\Yt\in\{0,1\}$.

    For this choice of $\Xboundval$, 
    Assumption \ref{as:data_holder} implies,%
\footnote{This follows from \eqref{eq:fObound} in Appendix \ref{sec:PROOF_SEC_DNN}.}
        \begin{equation}\label{eq:CfObound}
                \snormDNNBoundval{\fO}\leq\CfObound\Xboundval
        ,\qquad
            \CfObound
        \coloneqq 
                |\fO(0)|
                +\sqrt{d}
        ,
        \end{equation}
    so
    the FCDNN output bound may be chosen to satisfy 
        \begin{equation}\label{eq:bound_order}
            \snormDNNBoundval{\fO} < \DNNBound \lesssim \Xboundval
            .
        \end{equation}
    Then,
    Condition \ref{as:Env} holds with $\Fbound=\DNNBound,$
    and 
        $
        \truncf
        \lesssim n^{-1}
        $
    by Lemma \ref{lem:Ptrunc_truncf_env_verify} in Appendix \ref{sec:PROOF_SEC_DNN}.
    It always suffices to set $\DNNBound=2\CfObound\Xboundval$. 
    However, given prior knowledge that $\snorm{\fO}<\infty$ or 
    $\P(\Xt\notin [-1,1]^d)\!=\!0$,
    then
    one may instead take $\DNNBound=\DNNBoundfix$ for any fixed 
    $\DNNBoundfix$ sufficiently large.
\!\RoneNote{6}\!%

    Under Assumptions \ref{as:data_holder} and \ref{as:data_subg}, Theorem \ref{thrm:FCDNN_ROC} applies Theorem \ref{thrm:ROC_V2} to obtain results for FCDNN sieve estimators under either exponential or polynomial decay of the $\beta$-mixing coefficient.
    To remain applicable across estimation settings,
    Theorem \ref{thrm:FCDNN_ROC}
    leaves Conditions \ref{as:curv} and \ref{as:lip}, 
    and bounds for $\trunc$ as assumptions,
    since they depend on 
    the particular choice of $\crit$.
    These will be verified  
    in the time-series forecasting settings of Examples \ref{ex:reg}, \ref{ex:log}, and \ref{ex:quant} in the following subsections.
    
    Theorem \ref{thrm:FCDNN_ROC} allows two types of growth regimes for $(\L,\H)$ indexed by a growth sequence $\{\growth\}_{n\in\N}$.
    Choose $\CLwide,\CHwide,\CLdeep,\CHdeep>0$ sufficiently large, 
    let
    the pair $(\L,\H)$ 
    correspond 
    to one of the following regimes:
    \vspace{-6pt}
        \begin{align}
        \label{eq:wide_arch}
            \text{{(deep and wide)}}
        \qquad
            &
                \L
            =
                \left\lceil 
                \CLwide
                    \log(\growth)
                \right\rceil
        ,
        \quad
        \text{and}
        \quad
                \H
            =
                \left\lceil 
                    \CHwide
                    \growth
                \right\rceil\!
        ;
        \quad
        \text{or}
        \\
        \label{eq:deep_arch}
        \text{{(fixed width very deep)}}
        \qquad
            &
                \L
            =
                \left\lceil 
                    \CLdeep
                    \growth
                \right\rceil
        ,
        \quad
        \text{and}
        \quad
                \H 
            =
                \left\lceil 
                    \CHdeep
                \right\rceil\!
        .
        \end{align}

\newcommand{\deltemp}{\delta}%
\newcommand{\betaBoundn}{\betaBound}%
\newcommand{\growthlog}{\zeta}
{
\renewcommand{\holD}{\smooth}
\begin{theorem}\label{thrm:FCDNN_ROC}
\newcommand{\temp}{\eta}%
\!\RoneNote{3}\!%
    Suppose Assumptions \ref{as:data_holder} and \ref{as:data_subg} hold,
    and
        $
        \Zbound
        $
    is chosen 
    according to \eqref{eq:xbound_ybound_choice}.
    Let
        $
            \FMLPg
        =
            \FMLPnon(\L,\Hb,\DNNBound)
        $
    be such that \eqref{eq:width_bound_order} and \eqref{eq:bound_order} hold,
    and the pair
    $(\L,\H)$ 
    satisfy
    either 
        \eqref{eq:wide_arch}
        or
        \eqref{eq:deep_arch}.
    Suppose 
    Condition \ref{as:curv} holds,
    and, for all $n$ sufficiently large,
    Condition \ref{as:lip} 
    and \eqref{eq:trunc_term_def} hold
    for some 
        $
        \Mbound
        $
        and 
        $
        \trunc
        .
        $
    Then,
    there exists 
    a constant $C>0$ such that, for any $n$ sufficiently large and $\f$ 
    satisfying \eqref{eq:fhat}, 
    \begin{equation}\label{eq:pbound_FCDNN_ROC}
        \begin{aligned}
            \Pr\!\left(\!
                \frac{1}{n}\!
                \sum_{t=1}^{n}
                \int_{\Xspace}\!
                        \big[\f(\x)-\fO(\x)\big]^2
                d
                \P_{\Xt}\!(\x)
            >
                C
                \err
            \!\right)
        &
        \leq
            4
            \, \Pbndterm
        ,
        \\
        \text{and} \quad
        \Pr\!\left(\!
                \sampavg\!
                        \big[\f(\Xt)-\fO(\Xt)\big]^2
            >
                C
                \err
            \!\right)
        &
        \leq
            9
            \, \Pbndterm
        ,
        \end{aligned}
        \end{equation}
    holds for the following two cases with $(\err,\Pbndterm)$ given therein.
    \begin{enumerate}
        \item 
        If 
            $\beta(j)\leq 
        e^{-\Cbeta\,j } $
        for some 
            $
            \Cbeta
            >0
            $,
        and 
            $
            \growth
            =
            n^{\frac{1}{2}\left(\frac{d}{2\smooth+d}\right)}
            \log(n)^{\growthlog}
            $
        for some $\growthlog\in\R$,
        then
        \eqref{eq:pbound_FCDNN_ROC} holds with 
            $
                \Pbndterm
            =
            e^{-\growth^2\log(n)^4}
            +
                    {n^{1-\Cbeta\log(n)}}/{\log(n)}
                    +
                    2
                    n^{1-\Csubg\!\log(n)}\!
                    \log(n)
            $,
        and
        \begin{equation*}
        \begin{aligned}
            \err
        &
        =
            n^{-\frac{2\smooth}{2\smooth+d}}
            \left[
            \big(\!\max\big\{\!\Mbound,\DNNBound\!\big\}\big)^2
            \log(n)^{6+2\growthlog}
            +
                \frac{\Xboundval^{8(\holD+1)}}{\log(n)^{\left(\frac{2\smooth}{d}\right)2\growthlog}}
            \right]
            +
            \frac{1}{n}
            +
            \trunc
            +
            \theta_n
        .
        \end{aligned}
        \end{equation*}
    \item 
        If $\beta(j)\leq j^{-\Cbeta} $ for some $\Cbeta>0$,
\RtwoNote{4}\!%
\AENote{3}\!%
        and 
            $
            \growth
            =
            n^{\frac{1}{2}(1-\temp)\left(\frac{d}{2\smooth+d}\right)}
            \log(n)^{\growthlog}
            $
        for some $\growthlog\in\R$ and 
        $\temp \in \big[(1+\Cbeta)^{-1}, 1 \big) $,
        then
        \eqref{eq:pbound_FCDNN_ROC} holds with 
        $
            \Pbndterm
        =
            e^{
                -\growth^2
                \log(n)^4
            }
            +
                n^{1-\temp(1+\Cbeta)}\!
                +
                2
                n^{1-\Csubg\!\log(n)}
                \!
                \log(n)
        $,
    and
        \begin{equation*}
        \begin{aligned}
            \err
        &
        =
            n^{-(1-\temp)\frac{2\smooth}{2\smooth+d}}
            \left[
            \big(\!\max\big\{\!\Mbound,\DNNBound\!\big\}\big)^2
            \log(n)^{5+2\growthlog}
            +
                \frac{\Xboundval^{8(\holD+1)}}{\log(n)^{\left(\frac{2\smooth}{d}\right)2\growthlog}}
            \right]
            +
            \frac{1}{n}
            +
            \trunc
            +
            \theta_n
        .
        \end{aligned}
        \end{equation*}
    \end{enumerate}
\end{theorem}
}%

    Theorem \ref{thrm:FCDNN_ROC} extends \citet[Theorem 1]{farrell_deep_2021} to settings with possibly nonstationary $\beta$-mixing data in which $\{\Zt\}_{t\in\N}$ and $\fO$ need not be uniformly bounded.%
\footnote{
Although 
 \citet[Theorem 1]{farrell_deep_2021} gives slower rates, similar rates should be attainable if \citet[Theorem 2]{farrell_deep_2021} were applied using the approximation result from \citet{kohler_rate_2021}.
}
    \citet{kurisu_adaptive_2025} also study DNN estimators under similar dependence assumptions. 
    However, Theorem \ref{thrm:FCDNN_ROC} differs from their results in three key respects: it yields probability bounds for the errors rather than only bounds for expected risk, it applies beyond nonparametric regression on compact sets, and it does not impose boundedness or sparsity restrictions on the DNN weights $\gV$.

    Note that $n^{-\frac{2\smooth}{2\smooth+d}}$ is the usual benchmark minimax rate for nonparametric estimation of $\fO\in\HolderCust{\smooth}(\Xspace)$
    under \iid processes. 
    For Example \ref{ex:reg}, \citet{stone_optimal_1982} shows that this is the minimax optimal rate of convergence in $\Lp{2}$-norms restricted to compact sets.
    Under standard regularity conditions, 
    this is also the relevant minimax benchmark for settings related to Example \ref{ex:log} (\citealp{Yang_minimaxCLASS_1999})
    and Example \ref{ex:quant} 
    (\citealp{Zhao_Minimax_2026}).
    
    Since $\DNNBound\lesssim\Xboundval\leq \log(n)$,
\!\AENote{1}\!%
\!\RtwoNote{2}\!%
    Theorem \ref{thrm:FCDNN_ROC}(i) implies this benchmark minimax rate up to a logarithmic factor 
    once Condition \ref{as:curv} and suitable bounds for $\Mbound$ and $\trunc$ are established.
    To focus on the statistical part of the problem, the following results assume the optimization error satisfies 
        $\theta_n\lesssim n^{-\frac{2\smooth}{2\smooth+d}}$.
    This is a mild requirement relative to the existing DNN literature, where optimization error is often implicitly taken to be zero (e.g., \citealp{farrell_deep_2021,kohler_rate_2021,kurisu_adaptive_2025}).

    The following subsections apply Theorem \ref{thrm:FCDNN_ROC}(i) in the setting of Examples \ref{ex:reg}, \ref{ex:log}, and \ref{ex:quant}.
    In each case, Condition \ref{as:curv} and suitable bounds for $\Mbound$ and $\trunc$ are verified, yielding the benchmark minimax rate 
    up to a logarithmic factor.
    In these settings, Theorem \ref{thrm:FCDNN_ROC}(ii) can give corresponding, but slower, rates under only polynomial $\beta$-mixing.

    As is standard in \!\AENote{4}\!nonparametric estimation, the tuning parameters for the sieve space, $\FMLPnon(\L,\Hb,\DNNBound)$, depend on quantities that are generally unknown in practice. 
    This is most consequential for $(\L,\Hb)$, which depend on the smoothness of $\fO$ through $\smooth$, whereas $\DNNBound$ only needs to be sufficiently large.
    Theorem \ref{thrm:FCDNN_ROC} therefore does not prescribe a feasible tuning rule, but instead identifies growth rates of depth and width sufficient for the stated bounds.
    However, if 
        $\fO\in\HolderCust{\smooth}(\Xspace)$
    for $\smooth=\holD+\holE$
    then 
        $\fO\in\HolderCust{\holD'+\holE}(\Xspace)$ 
    for any integer $0\leq \holD' \leq \holD$.
    Hence the theorem also applies under more conservative choices corresponding to $\smooth'\coloneqq\holD'+\holE\leq \smooth$, with error bounds determined by $\smooth'$.
    In practice, this may provide some guidance for the choice of candidate architectures (see Remark \ref{rem:FCDNN_arch}).

    \subsubsection{FCDNNs for nonparametric regression}\label{sec:FCDNN_reg}
    \AENote{2}\!%
    Consider the nonparametric time-series regression setting from Example \ref{ex:reg}.
    The goal is to estimate the function
        $\fO(\Xt)=\E[\Yt|\Xt]$,
    using the least squares criterion 
        $
            \crit(\Zt,f)
        \coloneqq
            (\Yt-f(\Xt))^2
        .
        $
    For nonparametric regression settings,
    Lemma \ref{lem:regression_conditions} in Appendix \ref{sec:PROOF_FCDNN_reg}
    shows that 
    Condition \ref{as:curv} holds with $\Cpcq=\CpcqB=1$,
    Condition \ref{as:lip} holds with
                $
                    \Mbound 
                = 2\big(
                        \Yboundval
                        +
                        \DNNBound
                    \big)
                ,
                $
    and if Assumptions \ref{as:data_holder} and \ref{as:data_subg} hold 
    then $\trunc\lesssim n^{-1}$.

    \begin{corollary}\label{cor:FCDNN_reg}
\!\AENote{1}\!%
\!\RtwoNote{2}\!%
    Consider the setting of Example \ref{ex:reg}.
    Suppose Assumptions
    \ref{as:data_holder}
    and
    \ref{as:data_subg}
    hold, 
    and $\beta(j)\leq 
        e^{-\Cbeta\,j } $
        for some 
            $
            \Cbeta
            >0
            $. 
    Let $\f$ 
    satisfy \eqref{eq:fhat}
    where
        $\theta_n\lesssim n^{-\frac{2\smooth}{2\smooth+d}}$
    and
        $
            \FMLPg
        =
            \FMLPnon(\L,\Hb,\DNNBound)
        $
    is such that 
    \eqref{eq:width_bound_order} holds,
    and
    the pair
    $(\L,\H)$ 
    satisfy
    either 
        \eqref{eq:wide_arch}
        or
        \eqref{eq:deep_arch}.
\begin{enumerate}
        \item 
\!\RoneNote{5}\!%
\!\RoneNote{1}\!%
        If 
            $
                \growth
            =
                \left[\!\sqrt{n}\log(n)^{4\smooth}\right]^{\!\frac{d}{2\smooth+d}}
            $
            and
            $\DNNBound=\CDNNBound\log(n)$ for some $\CDNNBound\geq 2\CfObound$,
        then \eqref{eq:pbound_FCDNN_ROC} holds with 
        $
        \err
        \asymp
        \big[n^{-1}\log(n)^{4d}\big]^{\!\frac{2\smooth}{2\smooth+d}}\!
        \log(n)^{8}
        .
        $
        \item 
            Suppose, in addition, 
            $\P(\Xt\notin \XspaceC)=0$ for all $t\in\N$.
            If 
                $
                    \growth
                =
                    n^{\frac{1}{2}\left(\frac{d}{2\smooth+d}\right)}
                ,
                $
            and $\DNNBound=\CDNNBound$ for some $\CDNNBound\geq 2\CfObound$,
            then \eqref{eq:pbound_FCDNN_ROC} holds with 
                $
                \err
                \asymp
                    n^{-\frac{2\smooth}{2\smooth+d}}\!
                    \log(n)^{8}\!
                .
                $
\end{enumerate}
    \end{corollary}

    Corollary \ref{cor:FCDNN_reg}
    implies near-minimax rates in relatively general settings. 
    Part (i) permits $\{\Zt\}_{t\in\N}$ to be nonstationary, exponentially $\beta$-mixing, and subgaussian, while also allowing $\fO$ to be unbounded on $\Xspace$.
    Allowing unbounded covariates
    is useful 
    in autoregressive settings where 
    $\Xt$ contains lags of $\Yt$, but it comes at the cost of a dimension-sensitive logarithmic factor. Part (ii) shows that when the covariates are almost surely bounded, the minimax rate is recovered up to a fixed logarithmic factor.

    The closest 
    comparisons are \citet{kohler_rate_2021} and \citet{kurisu_adaptive_2025}, who also study DNN estimators for nonparametric conditional mean regression.
    In \iid\!settings, \citet{kohler_rate_2021} study similar FCDNN architectures, allowing both deep-and-wide and fixed-width very-deep regimes without restrictions on the DNN weights $\gV$.
    For dependent data, \citet{kurisu_adaptive_2025} also allow nonstationary exponentially $\beta$-mixing observations,
        but impose boundedness and sparsity restrictions on the DNN weights $\gV$ and consider only a deep-and-wide growth regime.
    Unlike Corollary \ref{cor:FCDNN_reg}, both papers establish bounds on expected risk over compact sets for regression functions belonging to hierarchical composition classes.
    When $\fO$ has additional compositional structure these classes can yield faster rates, but without such structure they reduce to the H\"older case, where their results yield near-minimax rates similar to the expected-risk rate implied by Corollary \ref{cor:FCDNN_reg}(ii) together with Remark \ref{rem:risk}.
    Corollary \ref{cor:FCDNN_reg}(i), by contrast, additionally allows both $\Xt$ and $\Yt$ to be possibly unbounded.

    Other dependent-data DNN results also differ from the present setting in several respects (e.g., \citealp{ma_theoretical_2022,Chen_Inference_2025,Men_Learning_2025,kengne_robust_2025,Kengne_Excess_2025,kengne_deep_2026}). While several of these papers allow $\alpha$-mixing, they are typically developed for stationary data with uniformly bounded covariates and impose restrictions on the DNN weights, often together with sparsity constraints.

    Although Corollary \ref{cor:FCDNN_reg} is stated for the broader H\"older class, the same framework could in principle be applied to hierarchical composition classes, like those considered by much of the DNN literature (e.g., \citealp{schmidt_hieber_nonparametric_2020,kohler_rate_2021,kurisu_adaptive_2025}) by replacing the approximation result in Theorem \ref{thrm:FCDNN_ROC} with \citet[Theorem 3]{kohler_rate_2021}.

    \subsubsection{FCDNNs for nonparametric logistic autoregression}\label{sec:FCDNN_log}
    \AENote{2}\!%
    This section considers the logistic binomial autoregression model from  
    Example \ref{ex:log}.
    Such settings are of particular interest for DNNs since they are frequently employed for classification problems. 
    Although this binomial logistic setting is a simple case of the classification problem, 
    similar results for
    other settings could be attained with the methods employed here.
    See \citet[Lemma 9]{farrell_deep_2021} for an example of how Poisson, Gamma, and multinomial logistic models can be shown to satisfy Conditions \ref{as:curv} and \ref{as:lip}.


    In order to verify Condition \ref{as:curv} and obtain a bound for $\trunc$, the following additional assumption is used. 
    This ensures that $\fO$ is uniformly bounded. 
    Notably, if the covariates take values in a bounded set and Assumption \ref{as:data_holder} holds, then 
    only the values of $\fO$ on a compact subset of $\Xspace$ are relevant and
    Assumption \ref{as:data_logistic_fbound} mainly serves as a convenient global extension.
    
    

    \begin{assumption}\label{as:data_logistic_fbound}
        There exists $\Clog>0$ such that 
            $
                \snorm{\fO}
                \leq 
                \Clog
            .
            $
    \end{assumption}
    Assumption \ref{as:data_logistic_fbound} is equivalent to requiring
    $\P(\Yt=1|\Xt)$ be uniformly bounded away from $0$ and $1$, which is standard for binary classification problems (e.g., \citealp{Ohn_Nonconvex_2022}).
    %
    This follows
    since $\Yt\in\{0,1\}$, and
        $
        \P\big(
        \Yt=1
        \,
        \big|
        \,
            \Xt 
        \big)
        =
        \E[\Yt|\Xt]
        =
        e^{\fO(\Xt)}\big[1+e^{\fO(\Xt)}\big]^{-1}
        ,
        $
    so Assumption \ref{as:data_logistic_fbound} implies
        $$
        \frac{e^{-\Clog}}{1+e^{-\Clog}}
        \leq
        \E[\Yt|\Xt]
        \leq
        \frac{e^{\Clog}}{1+e^{\Clog}}
        .
        $$
    In the setting of Corollary \ref{cor:FCDNN_log},
    Lemma \ref{lem:logistic_conditions} in Appendix \ref{sec:PROOF_FCDNN_log}
    shows that Condition \ref{as:curv} holds, Condition \ref{as:lip} holds with $\Mbound=2$, 
    and $\trunc\lesssim n^{-\Csubg\log(n)}$.

    \begin{corollary}\label{cor:FCDNN_log}
\!\AENote{1}\!%
\!\RtwoNote{2}\!%
    Consider the setting of Example \ref{ex:log}.
    Suppose Assumptions
    \ref{as:data_holder},
    \ref{as:data_subg},
    and 
    \ref{as:data_logistic_fbound}
    hold,
    and $\beta(j)\leq 
        e^{-\Cbeta\,j } $
        for some 
            $
            \Cbeta
            >0
            $. 
    Let $\f$ 
    satisfy \eqref{eq:fhat}
    where
        $\theta_n\lesssim n^{-\frac{2\smooth}{2\smooth+d}}$
    and
        $
            \FMLPg
        =
            \FMLPnon(\L,\Hb,\DNNBound)
        $
    is such that 
    \eqref{eq:width_bound_order} holds,
        $\DNNBound
            =
            \CDNNBound\Clog
            $ 
            for some $\CDNNBound\geq2$,
    and
    the pair
    $(\L,\H)$ 
    satisfy
    either 
        \eqref{eq:wide_arch}
        or
        \eqref{eq:deep_arch}.
\begin{enumerate}
        \item 
\!\RoneNote{5}\!%
        If 
            $
                \growth
            =
                \!\left[\sqrt{n}\log(n)^{4\smooth+1}\right]^{\frac{d}{2\smooth+d}}\!
            ,
            $
        then \eqref{eq:pbound_FCDNN_ROC} holds with 
        $
        \err
        \asymp
        \big[
            n^{-1}
            \log(n)^{(4+{1}/{\smooth})d}
        \big]^{\!\frac{2\smooth}{2\smooth+d}}\!
        \log(n)^{6}
        .
        $
        \item 
            Suppose, in addition, that
            %
            $\P(\Xt\notin \XspaceC)=0$ for all $t\in\N$.
            If 
                $
                    \growth
                =
                    n^{\frac{1}{2}\left(\frac{d}{2\smooth+d}\right)}\!
                ,
                $
            then \eqref{eq:pbound_FCDNN_ROC} holds with 
            $
            \err
                \asymp
                    n^{-\frac{2\smooth}{2\smooth+d}}
                    \log(n)^{6}
            .
            $
\end{enumerate} 
    \end{corollary}

In the logistic regression setting, Corollary \ref{cor:FCDNN_log} shows that Theorem \ref{thrm:FCDNN_ROC} also yields $n^{-\frac{2\smooth}{2\smooth+d}}$ rates up to a logarithmic factor. Part (i) allows $\{\Zt\}_{t\in\N}$ to be nonstationary, exponentially $\beta$-mixing with subgaussian covariates and yields a rate with a dimension-sensitive logarithmic factor.
Part (ii) shows that when the covariates are almost surely bounded, the logarithmic factor no longer depends on $d$.

Existing theory for FCDNNs
in classification problems has predominantly assumed \iid data
(e.g., \citealp{farrell_deep_2021,yara_nonparametric_2026}). This work also often imposes boundedness or sparsity restrictions on the DNN weights $\gV$ (\citealp{bos_convergence_2022,Ohn_Nonconvex_2022,zhang_classification_2024}).

For DNN estimators in binary classification with logistic loss under dependent data, \citet{kengne_deep_2026} establish bounds on expected excess risk, whereas Corollary \ref{cor:FCDNN_log} gives nonasymptotic probability bounds for the $\Lp{2}$-errors which imply similar excess risk bounds with Remark \ref{rem:risk}. 
Their DNN class is also more restrictive, imposing boundedness and sparsity conditions on the network weights and allowing only deep-and-wide growth regimes.
Although their dependence framework permits strong mixing rather than $\beta$-mixing, it is restricted to stationary data with uniformly bounded covariates.

    \subsubsection{FCDNNs for nonparametric conditional quantile autoregression}\label{sec:FCDNN_quant}
    \AENote{2}\!%
    Consider the nonparametric conditional quantile autoregression setting of
                Example \ref{ex:quant}.
    For fixed $\tau\in(0,1)$ the goal 
    is to estimate $\fO$ satisfying
        $
        \E\big[
            \mathbbm{1}_{\{\Yt \leq \fO(\Xt)\}}
        \mid 
            \Xt
        \big]
        =
            \tau
        $,
    using the criterion 
        $
            \crit(\Zt,f)
        =
            \big(
                \tau
                -
                \mathbbm{1}_{\{\Yt \leq f(\Xt)\}}
            \big)
            \big(
                \Yt - f(\Xt)
            \big)
        .
        $
    To verify Condition \ref{as:curv} and obtain a bound for $\trunc$, the following additional assumption is imposed.

    \begin{assumption}\label{as:data_quantile}
        There exists $\Cquant>0$ such that 
            $
                \snorm{\fO}
                \leq 
                \Cquant
            ,
            $
        and the function
            $
                \FXY(y)
            \coloneqq
                \E\big[\mathbbm{1}_{\{\Yt \leq y\}}\mid\! \Xt\big]
            $
        admits a density, 
        $\densXY$,
        in the sense that
            $$
            F_{\Yt|\Xt(\omega)}
            (b)
            -
            F_{\Yt|\Xt(\omega)}
            (a)
            =
            \int_{a}^b
                \mu_{\Yt|\Xt(\omega)}
                (y)
                dy
            ,\quad
            \text{\emph{$\P$-almost every $\omega\in\Omega$, and all $a,b\in\R$}}
            .
            $$
        Moreover, 
        there exist constants 
            $
            \cdensi, \cdensii >0
            $
        such that 
            $$
                \cdensi 
            \leq 
                \mu_{\Yt|\Xt(\omega)}
                (y) 
            \leq 
                \cdensii
            ,\qquad
            \text{\emph{$(\P\otimes\lambda)$-almost every}}
            \;
            (\omega,y)\in \{\Omega\times\R: |y-\fO(\Xt(\omega))|\leq 3\Cquant \}
            ,
            $$
        where $\lambda$ denotes the Lebesgue measure on $\R$, and $\P\otimes\lambda$ is the product measure.
    \end{assumption}

    This is a standard regularity condition in nonparametric quantile regression: it requires the conditional density to exist and to be bounded away from zero and infinity in a neighborhood of the target quantile.
    It still allows $\Yt$ and $\Xt$ to be unbounded. For example, it covers nonlinear AR models of the form 
    $
        \Yt = \fO(\Xt) + \upsilon_t
        $
    where $\P(\upsilon_t\leq 0 | \Xt) = \tau,\,$
    $\upsilon_t|\Xt \sim N(0,1),\,$
    $\tau=1/2,$
    and
        $
            \Xt=(\V_{t-1},\Y_{t-1},$ $\ldots\Y_{t-r})
        $
    for some sub-vector of covariates $\Vt$.
    In the setting of Corollary \ref{cor:FCDNN_quant}, Lemma \ref{lem:quantile_conditions} in Appendix \ref{sec:PROOF_FCDNN_quant}
    shows that Condition \ref{as:curv} holds, Condition \ref{as:lip} holds with $\Mbound=1$; 
    and $\trunc\lesssim n^{-1}$.


\begin{corollary}\label{cor:FCDNN_quant}
\!\AENote{1}\!%
\!\RtwoNote{2}\!%
    Consider the setting of Example \ref{ex:quant}.
    Suppose Assumptions
    \ref{as:data_holder},
    \ref{as:data_subg},
    and 
    \ref{as:data_quantile}
    hold,
    and $\beta(j)\leq 
        e^{-\Cbeta\,j } $
        for some 
            $
            \Cbeta
            >0
            $. 
    Let $\f$ 
    satisfy \eqref{eq:fhat}
    where
        $\theta_n\lesssim n^{-\frac{2\smooth}{2\smooth+d}}$
    and
        $
            \FMLPg
        =
            \FMLPnon(\L,\Hb,\DNNBound)
        $
    is such that 
    \eqref{eq:width_bound_order} holds,
        $\DNNBound
        =
        2\Cquant
        $ 
    and the pair
    $(\L,\H)$ 
    satisfy
    either 
        \eqref{eq:wide_arch}
        or
        \eqref{eq:deep_arch}.
\begin{enumerate}
        \item 
\!\RoneNote{5}\!%
        If 
            $
                \growth
            =
                \!\left[\sqrt{n}\log(n)^{4\smooth+1}\right]^{\frac{d}{2\smooth+d}}\!
            ,
            $
        then \eqref{eq:pbound_FCDNN_ROC} holds with 
            $
            \err
            \asymp
            \big[
                n^{-1}
                \log(n)^{(4+{1}/{\smooth})d}
            \big]^{\!\frac{2\smooth}{2\smooth+d}}\!
            \log(n)^{6}
            .
            $
        \item 
            Suppose, in addition, that
            $\P(\Xt\notin \XspaceC)=0$ for all $t\in\N$.
            If 
                $
                    \growth
                =
                    n^{\frac{1}{2}\left(\frac{d}{2\smooth+d}\right)}\!
                ,
                $
            then \eqref{eq:pbound_FCDNN_ROC} holds with 
            $
                    \err
                \asymp
                    n^{-\frac{2\smooth}{2\smooth+d}}
                    \log(n)^{6}
            .
            $
\end{enumerate} 
\end{corollary}

    Corollary \ref{cor:FCDNN_quant} applies Theorem \ref{thrm:FCDNN_ROC} in the nonparametric quantile setting under possibly nonstationary exponentially $\beta$-mixing data, and yields nonasymptotic error bounds implying rates of order $n^{-\frac{2\smooth}{2\smooth+d}}$ up to logarithmic factors.
    Part (i) allows unbounded subgaussian covariates and outcomes, while part (ii) shows that when the covariates are almost surely bounded, the logarithmic factor is no longer dimension-sensitive.

    A classical antecedent is \citet{chen_improved_1999}, who obtain asymptotic results for conditional quantile estimation using single-hidden-layer neural networks under stationary $\beta$-mixing data and compact-support conditions.
    While important, that analysis is limited in several respects, most notably by its focus on shallow architectures.
    More recent quantile FCDNN results are otherwise largely confined to \iid\!settings (\citealp{shen2021deepquantileregressionmitigating}) and often impose restrictions on the DNN weights (\citealp{Padilla_Quantile_2022}) or bounds on DNN derivatives (\citealp{shen_nonparametric_2024}), which may be difficult to enforce in practice. 
    By contrast, Corollary \ref{cor:FCDNN_quant} studies multi-layer FCDNNs under non-\iid\!data, allows multiple growth regimes, and avoids restrictions on the DNN weights.
    In this way, it brings the theory closer to empirical work using deep FCDNNs for time-series quantile estimation, such as Value-at-Risk forecasting (e.g., \citealp{Chronopoulos_ForecastingDNNQuantile_2024}).

\section{Convolutional deep neural networks}\label{sec:CDNN}
\RoneNote{4}
\RoneNote{7}

This section develops results for a class of CDNNs to illustrate how Theorem \ref{thrm:ROC_V2} extends beyond fully connected architectures. The class considered here is only one example among the many convolutional architectures used in practice. Because CDNN architectures vary substantially across applications, the goal here is not to cover all specifications, but rather to show how the general time-series sieve-estimation framework applies to convolutional architectures.

    The CDNNs considered here are built from blocks of expansive convolution layers separated by locally connected dimension-reduction layers.
    Expansive convolutions with zero-padding admit approximation properties unavailable for many contracting convolutional architectures (see \citealp{lin_universal_2022}), but they increase network width with each convolution layer.
    The dimension-reduction layers are introduced to control this growth. 
    They are similar in spirit to downsampling or pooling, but are implemented here through locally connected sliding-window reduction matrices to preserve local temporal relationships in AR settings without introducing dense hidden-layer connections.
    In the theoretical \iid\!CDNN literature, complexity-expanding convolution layers are standard and are often combined with dimension-reduction steps (e.g., \citealp{Zhou_conv_2020,zhou_learning_2024,Yang_CDNDconvergence2025,kohler_statistical_2025}).%
\footnote{The CDNNs considered here use one-channel convolutions that increase layer width, which is natural for time-series problems in economics and finance. In applications such as image recognition, complexity-expanding convolutions more often increase the number of channels. In either case, the effect is to increase the number of DNN weights entering the subsequent convolution step.}
    Because convolution layers use localized filters with shared coefficients, and the reduction layers limit width growth, the resulting hidden layers contain far fewer DNN weights than comparably sized FCDNNs.

Although in practice CDNNs often include fully connected hidden layers, I do not consider such CDNN-FCDNN hybrids here. This keeps the CDNN results distinct from the FCDNN theory developed in the previous section. Closely related results for CDNN-FCDNN hybrids could be obtained by combining the CDNN and FCDNN methods developed in this paper. 
\citet{lin_universal_2022} also conduct experiments which find similar empirical performance of expansive CDNNs and the commonly used CDNN-FCDNN hybrid architectures.

    \subsection{CDNN architectures}%
    \label{sec:CDNN_sieve_spaces}

Throughout, let $d\geq 2$.
The CDNN architecture $\FCONV= \FCONVnon(\Lcnv,\filtlen,\DNNBoundcnv)$ is specified by the user-chosen tuning parameters:
(i) depth $\Lcnv \in \mathbb{N}$,
(ii) convolutional filter length $\filtlen \in \mathbb{N}$, $\filtlen\geq2$,
and
(iii) uniform output bound $\DNNBoundcnv \ge 2$.
The hidden layers consist of expansive convolution layers and locally connected dimension-reduction layers.

Let 
    $\gcv=\{\gc_k\}_{k=-\infty}^\infty$ 
be a filter of length $\filtlen$
where $\gc_k\in\R$ for each $k\in\{0,1,\ldots,\filtlen\}$ and $\gc_k=0$ whenever $k\notin\{0,1,\ldots,\filtlen\}$.
For a vector $\boldsymbol{a}\in\R^{d'}$ the convolution $\gcv*\boldsymbol{a}\in\R^{d'+\filtlen}$ is defined as
    $$
        \gcv*\boldsymbol{a}=
        [\gcv*\boldsymbol{a}]_{i=1}^{d'+\filtlen}
    ,\;\; \text{ where } \;\;
        [\gcv*\boldsymbol{a}]_i
    \coloneqq
        \sum_{k=1}^{d'}
            \gc_{i-k}a_{k}
    ,
    \quad
    i=1,\ldots,d'+s
    .
    $$
It will be convenient to express the convolution in matrix form $\gcv*\boldsymbol{a}= \Gcv\boldsymbol{a}$ using the associated Toeplitz convolution matrix $\Gcv\in \R^{(d'+s)\times d'}$ with elements $[\Gcv]_{ij}=\gc_{i-j}$.
The matrix $\Gcv$ admits the usual Toeplitz structure
with constant diagonals
{\renewcommand{\filtlen}{s}
$$
{\renewcommand{\arraystretch}{0.7}
\Gcv =
\begin{bmatrix}
\gc_0           & 0                 & 0\cdots   & \cdots            & \cdots            & \cdots0  & 0 \\
\gc_1           & \gc_0             & 0\cdots   & \cdots            & \cdots            & \cdots0  & 0 \\
\vdots          & \ddots            & \ddots    & \ddots            & \ddots            & \ddots  & \vdots \\
\gc_{\filtlen}& \gc_{\filtlen-1}    & \cdots    & \gc_0             & 0                 & \cdots0  & 0 \\
0               & \gc_{\filtlen}  & \cdots    & \gc_1             & \gc_0             & \cdots0  & 0 \\
\vdots          & \ddots            & \ddots    & \ddots            & \ddots            & \ddots  & \vdots \\
0               & \cdots            & \cdots0   & \gc_{\filtlen}  & \gc_{\filtlen-1}    & \cdots  & \gc_0 \\
0               & \cdots            & \cdots0   & 0                 & \gc_{\filtlen}  & \cdots  & \gc_1 \\
\vdots          & \ddots            & \ddots    & \ddots            & \ddots            & \ddots  & \vdots \\
0               & \cdots            & \cdots    & 0                 & 0                 & \cdots\gc_{\filtlen}  & \gc_{\filtlen-1} \\
0               & \cdots            & \cdots    & 0                 & 0                 & \cdots0  & \gc_{\filtlen}
\end{bmatrix}
}
\in \R^{(d'+s)\times d'}
.
$$
}%
\newcommand{\maxpool}[2]{\mathcal{M}_{#1,#2}}%
\newcommand{\HmaxA}{d_{1,\text{max}}}%
\newcommand{\HmaxB}{d_{\text{max}}}%
\newcommand{\LpoolA}{L_{1,\text{max}}}%
\newcommand{\LpoolI}{\overline{L}_1}%
\newcommand{\LpoolII}{\overline{L}_2}%
\newcommand{\HpoolI}{\overline{H}_1}%
\newcommand{\HpoolII}{\overline{H}_2}%
\newcommand{\Av}{\boldsymbol{A}}%
\newcommand{\Hreduc}{\overline{H}}%
\newcommand{\LCRM}{$\text{LCRM}$}%
\newcommand{\LCRMlayers}{\{\LpoolI + (j-1)\LpoolII:j\in\N\}}%
\newcommand{\LCRMlayersShort}{\{\LpoolI+(j-1)\LpoolII\}}%

The CDNNs considered here control the layer width using locally connected reduction matrices (\LCRM) $\Av_{r}$.
For $r\in\{d,\,2d+10\}$,
define
    $$
        \Hreduc \coloneqq 2d+10
    ,
    \qquad
        m_r \coloneqq r+\filtlen\left\lceil\frac{\Hreduc r}{\filtlen-1}\right\rceil
    ,
    \quad
    \text{ and }
    \quad
        w_r \coloneqq m_r-(\Hreduc-1)r
    .
    $$
An \LCRM\ 
$\Av_{r}\in \R^{\Hreduc \times m_r}$
is a sliding-window matrix with stride length $r$ and window size $w_r$ satisfying
     \begin{equation}\label{eq:LCRMdef}
        [\Av_r]_{ij}=0 \quad \text{whenever } j\notin\{(i-1)r+1,\ldots,(i-1)r+w_r\}
        .
    \end{equation}
Thus the possibly nonzero entries of each row are contained in the $w_r$ window,
which is shifted to the right by $r$ in each subsequent row, and every column falls inside of at least one window. See Figure \ref{fig:LCRM} for an example.
\begin{figure}[t]
\centering
\begin{tikzpicture}[scale=0.75]

    \def\H{14}      
    \def\N{58}     
    \def\rr{2}     
    \def\ww{32}     
    \def\cell{0.35}

    \draw[thick] (0,0) rectangle ({\N*\cell},{\H*\cell});
    \foreach \i in {1,...,\numexpr\N-1\relax} {
        \draw[gray!40] ({\i*\cell},0) -- ({\i*\cell},{\H*\cell});
    }
    \foreach \j in {1,...,\numexpr\H-1\relax} {
        \draw[gray!40] (0,{\j*\cell}) -- ({\N*\cell},{\j*\cell});
    }

    \foreach \row in {1,...,\H} {
        \pgfmathtruncatemacro{\start}{(\row-1)*\rr + 1}
        \pgfmathtruncatemacro{\finish}{(\row-1)*\rr + \ww}
        \foreach \col in {\start,...,\finish} {
            \fill[blue!25] ({(\col-1)*\cell},{(\H-\row)*\cell}) rectangle ({\col*\cell},{(\H-\row+1)*\cell});
        }
    }

    \foreach \row in {1,...,\H} {
        \pgfmathtruncatemacro{\start}{(\row-1)*\rr + 1}
        \pgfmathtruncatemacro{\finish}{(\row-1)*\rr + \ww}
        \draw[blue!70!black, thick]
        ({(\start-1)*\cell},{(\H-\row)*\cell})
        rectangle
        ({\finish*\cell},{(\H-\row+1)*\cell});
    }



    \node[left] at (0,{(\H-1+0.5)*\cell}) {\scriptsize$i=1$};
    \node[left] at (0,{0.5*\cell}) {\scriptsize $i=\Hreduc$};

    \node[below] at ({0.5*\cell},0) {\scriptsize $1$};
    \node[below] at ({(\N-0.5)*\cell},0) {\scriptsize $m_r$};


    \draw[decorate,decoration={brace,amplitude=5pt},yshift=-.1pt]
        (0,5) -- ({\ww*\cell},5)
        node[midway,above=4pt] {\scriptsize window length $w_r$};

    \draw[decorate,decoration={brace,amplitude=5pt},yshift=-.1pt]
        ({\ww*\cell},5) -- ({(\rr+\ww)*\cell},5)
        node[midway,above=4pt] {\scriptsize stride length $r$};

\end{tikzpicture}
\caption{
Structure of dimension reduction matrix $\Av_{r}\in \R^{\Hreduc \times m_r}$ with sliding-window structure for $d=2$, $s=2$, $r=d$. 
Entries with possible non-zero values in blue and always zero-valued entries unshaded.
Row \(i\) contains exactly \(w_r\) adjacent possibly nonzero entries, supported on columns
\(\{(i-1)r+1,\dots,(i-1)r+w_r\}\).
}
\label{fig:LCRM}
\end{figure}

The architecture $\FCONV = \FCONVnon(\L, \filtlen, \DNNBound)$ consists of hidden layers 
$l \in \{1,\ldots,\L\}$
with node vectors 
    $\nv^{(l)} : \R^d \to \R^{\Hicnv{l}}$, where $\Hicnv{0}=d$.
Layers are arranged in blocks of expansive convolution layers separated by reduction layers occurring at indices
    $$
    l\in\LCRMlayers
    ,
    $$
where 
    $$
        \LpoolI
    \coloneqq
        \left\lceil
            \frac%
            {\Hreduc d}%
            {(\filtlen-1)}%
        \right\rceil
        +1
    ,
    \quad
    \text{ and }
    \quad
        \LpoolII
    \coloneqq
        \left\lceil
            \frac%
            {\Hreduc^2}%
            {(\filtlen-1)}%
        \right\rceil
        +1
    .
    $$
For $l\notin\LCRMlayers$ layer $l$ is a convolution layer with width $\Hicnv{l}=\Hicnv{l-1}+\filtlen$. For $l=\LpoolI$, a reduction layer uses $\Av_d^{(l)}$, while for $l>\LpoolI$ with $l\in\{\LpoolI+(j-1)\LpoolII\}$, a reduction layer uses $\Av_{2d+10}^{(l)}$. In all cases, reduction layers map to width $\Hicnv{l}=\Hreduc$.

For each layer, $\nv^{(l)}$ is parameterized by DNN weights consisting of an intercept vector $\biasv^{(l)}\in\R ^{\Hicnv{l}}$, and either a convolution filter $\gcv^{(l)}$ or a reduction matrix $\Av_r^{(l)}$. The output node $\nv^{(\L +1)}$ is a linear combination of $\nv^{(\L )}$ with weights $\gv\in\R^{\Hicnv{\L}}$ and $\bias\in\R $. 
The full network depends on $\W$ DNN weights consisting of the filter coefficients, LCRM coefficients, intercept terms, and output node weights. These are 
collected together as $\gV\in\R^{\W}$ where
        $$
                \gV
            \coloneqq 
                \Big\{
                \big\{
                    \gcv^{(l)}
                \big\}_{1\leq l\leq \L,\ 
                    l \notin \LCRMlayersShort
                    }
                ,\;
                \big\{
                    \Av^{(l)}
                \big\}_{1\leq l\leq \L,\
                    l \in \LCRMlayersShort
                    }
                ,\;
                \{\biasv^{(l)}\}_{l=1}^{\L}
                ,\;
                \gv
                ,\;
                \bias
                \Big\}
            .
        $$
Then, 
adopting the convention that $\nv^{(0)}(\x)=\x$,
the 
layer computations are defined recursively as
    \begin{equation}\label{eq:nodes_conv_banded}
    \resizebox{\hsize}{!}{$
        \nv^{(l)}(\x \mid \gV) 
    \coloneqq 
        \begin{cases}
            \relu\!\left(\!
                \Gcv^{(l)} \nv^{(l-1)}(\x \mid \gV) + \biasv^{(l)}
            \!\right)\!,
        &
            l \leq \L, 
            \;
            l \notin 
            \LCRMlayers,
        \\[0.4em]
            \relu\!\left(\!
                \Av^{(l)}_{d} \nv^{(l-1)}(\x \mid \gV) + \biasv^{(l)}
            \!\right)\!,
        &
            l=\LpoolI,
        \\[0.4em]
            \relu\!\left(\!
                \Av^{(l)}_{2d+10} \nv^{(l-1)}(\x \mid \gV) + \biasv^{(l)}
            \!\right)\!,
        &
            \LpoolI < l \leq \L, 
            \;
            l \in 
            \LCRMlayers,
        \\[0.4em]
            \gv^{\tr}
            \nv^{(\L)}(\x \mid \gV) + \bias,
        &
            l = \L + 1
            ,
        \end{cases}
    $}
    \end{equation}
where $\relu$ is applied element-wise, 
and $\Gcv^{(l)}\in\R^{\Hicnv{l}\times\Hicnv{l-1}}$ is the 
convolution matrix 
$[\Gcv^{(l)}]_{ij}=\gc^{(l)}_{i-j}$.
The CDNN sieve spaces $\FCONV=\FCONVnon(\L,\filtlen,\DNNBoundfix)$ are defined as
    \begin{equation}\label{eq:sievespaceCNN}
            \FCONVnon(\L,\filtlen,\DNNBoundfix)
        \coloneqq
            \Big\{ 
            f:
                \;
                f(\x)
                =
            \max\big\{\! -\!\DNNBoundfix,\, \min\{\DNNBoundfix, \nv^{(\L+1)}(\x \,| \gV)\}
            \big\}
            ,
            \;  
            \gV \in \R^{\W}
            \Big\}
            .
    \end{equation}

{
\newcommand{\lin}{-0.1}
\newcommand{\lI}{1.7}
    \newcommand{\figlabLI}{2.5}
\newcommand{\lII}{3.6}
\newcommand{\pII}{5.6}
\newcommand{\lIII}{7.7}
    \newcommand{\figlabLII}{8.5}
\newcommand{\lIV}{9.6}
\newcommand{\pIV}{11.6}
\newcommand{\lV}{13.7}
\newcommand{\lout}{15.5}

\begin{figure}[t]    
    \centering

    \definecolor{input_node_fg}{HTML}{E90F0E}
    \definecolor{input_node_bg}{HTML}{F5C5C1}
    \definecolor{hidden_node_fg}{HTML}{3972FE}
    \definecolor{hidden_node_bg}{HTML}{D9E8FB}
    \definecolor{reduce_node_fg}{HTML}{8A6D3B}
    \definecolor{reduce_node_bg}{HTML}{FCF8E3}
    \definecolor{output_node_fg}{HTML}{9DBD88}
    \definecolor{output_node_bg}{HTML}{D6E7D4}
    
\resizebox{\hsize}{!}{
    \begin{tikzpicture}[
        node distance=0.55cm,
        input/.style={circle, draw=input_node_fg, thick, minimum size=0.55cm, fill=input_node_bg},
        hidden/.style={rectangle, draw=hidden_node_fg, thick, minimum size=0.70cm, fill=hidden_node_bg},
        reduce/.style={rectangle, rounded corners=2pt, draw=reduce_node_fg, thick, minimum size=0.70cm, fill=reduce_node_bg},
        output/.style={diamond, draw=output_node_fg, thick, minimum size=0.05cm, inner sep=2pt, fill=output_node_bg},
        conn/.style={->, shorten >= 1pt}
    ]

        \node[input] (x1) at (\lin, 1.1) {$x_{1}\!$};
        \node[input] (x2) at (\lin, 0.0) {$x_{2}\!$};
        \node[input] (x3) at (\lin,-1.1) {$x_{3}\!$};

        \node[hidden] (u11) at (\lI, 1.8) {$\!\n^{(1)}_{1}\!$};
        \node[hidden] (u12) at (\lI, 0.9) {$\!\n^{(1)}_{2}\!$};
        \node[hidden] (u13) at (\lI, 0.0) {$\!\n^{(1)}_{3}\!$};
        \node[hidden] (u14) at (\lI,-0.9) {$\!\n^{(1)}_{4}\!$};
        \node[hidden] (u15) at (\lI,-1.8) {$\!\n^{(1)}_{5}\!$};

        \node[hidden] (u21) at (\lII, 2.7) {$\!\n^{(2)}_{1}\!$};
        \node[hidden] (u22) at (\lII, 1.8) {$\!\n^{(2)}_{2}\!$};
        \node[hidden] (u23) at (\lII, 0.9) {$\!\n^{(2)}_{3}\!$};
        \node[hidden] (u24) at (\lII, 0.0) {$\!\n^{(2)}_{4}\!$};
        \node[hidden] (u25) at (\lII,-0.9) {$\!\n^{(2)}_{5}\!$};
        \node[hidden] (u26) at (\lII,-1.8) {$\!\n^{(2)}_{6}\!$};
        \node[hidden] (u27) at (\lII,-2.7) {$\!\n^{(2)}_{7}\!$};

        \node[reduce] (r31) at (\pII, 1.1)  {$\!\n^{(3)}_{1}\!$};
        \node[reduce] (r32) at (\pII, 0.0)  {$\!\n^{(3)}_{2}\!$};
        \node[reduce] (r33) at (\pII,-1.1)  {$\!\n^{(3)}_{3}\!$};

        \node[hidden] (u41) at (\lIII, 1.8) {$\!\n^{(4)}_{1}\!$};
        \node[hidden] (u42) at (\lIII, 0.9) {$\!\n^{(4)}_{2}\!$};
        \node[hidden] (u43) at (\lIII, 0.0) {$\!\n^{(4)}_{3}\!$};
        \node[hidden] (u44) at (\lIII,-0.9) {$\!\n^{(4)}_{4}\!$};
        \node[hidden] (u45) at (\lIII,-1.8) {$\!\n^{(4)}_{5}\!$};

        \node[hidden] (u51) at (\lIV, 2.7) {$\!\n^{(5)}_{1}\!$};
        \node[hidden] (u52) at (\lIV, 1.8) {$\!\n^{(5)}_{2}\!$};
        \node[hidden] (u53) at (\lIV, 0.9) {$\!\n^{(5)}_{3}\!$};
        \node[hidden] (u54) at (\lIV, 0.0) {$\!\n^{(5)}_{4}\!$};
        \node[hidden] (u55) at (\lIV,-0.9) {$\!\n^{(5)}_{5}\!$};
        \node[hidden] (u56) at (\lIV,-1.8) {$\!\n^{(5)}_{6}\!$};
        \node[hidden] (u57) at (\lIV,-2.7) {$\!\n^{(5)}_{7}\!$};

        \node[reduce] (r61) at (\pIV, 1.1) {$\!\n^{(6)}_{1}\!$};
        \node[reduce] (r62) at (\pIV, 0.0) {$\!\n^{(6)}_{2}\!$};
        \node[reduce] (r63) at (\pIV,-1.1) {$\!\n^{(6)}_{3}\!$};

        \node[hidden] (u71) at (\lV, 1.8) {$\!\n^{(7)}_{1}\!$};
        \node[hidden] (u72) at (\lV, 0.9) {$\!\n^{(7)}_{2}\!$};
        \node[hidden] (u73) at (\lV, 0.0) {$\!\n^{(7)}_{3}\!$};
        \node[hidden] (u74) at (\lV,-0.9) {$\!\n^{(7)}_{4}\!$};
        \node[hidden] (u75) at (\lV,-1.8) {$\!\n^{(7)}_{5}\!$};

        \node[output] (y) at (\lout, 0.0) {$\!\n^{\!(8)}_{1}\!$};

        \draw[conn] (x1.east) -- (u11);
        \draw[conn] (x1.east) -- (u12);
        \draw[conn] (x1.east) -- (u13);
        \draw[conn] (x2.east) -- (u12);
        \draw[conn] (x2.east) -- (u13);
        \draw[conn] (x2.east) -- (u14);
        \draw[conn] (x3.east) -- (u13);
        \draw[conn] (x3.east) -- (u14);
        \draw[conn] (x3.east) -- (u15);

        \draw[conn] (u11.east) -- (u21);
        \draw[conn] (u11.east) -- (u22);
        \draw[conn] (u11.east) -- (u23);

        \draw[conn] (u12.east) -- (u22);
        \draw[conn] (u12.east) -- (u23);
        \draw[conn] (u12.east) -- (u24);

        \draw[conn] (u13.east) -- (u23);
        \draw[conn] (u13.east) -- (u24);
        \draw[conn] (u13.east) -- (u25);

        \draw[conn] (u14.east) -- (u24);
        \draw[conn] (u14.east) -- (u25);
        \draw[conn] (u14.east) -- (u26);

        \draw[conn] (u15.east) -- (u25);
        \draw[conn] (u15.east) -- (u26);
        \draw[conn] (u15.east) -- (u27);

        \draw[conn] (u21.east) -- (r31);
        \draw[conn] (u22.east) -- (r31);
        \draw[conn] (u23.east) -- (r31);

        \draw[conn] (u23.east) -- (r32);
        \draw[conn] (u24.east) -- (r32);
        \draw[conn] (u25.east) -- (r32);

        \draw[conn] (u25.east) -- (r33);
        \draw[conn] (u26.east) -- (r33);
        \draw[conn] (u27.east) -- (r33);

        \draw[conn] (r31.east) -- (u41);
        \draw[conn] (r31.east) -- (u42);
        \draw[conn] (r31.east) -- (u43);
        \draw[conn] (r32.east) -- (u42);
        \draw[conn] (r32.east) -- (u43);
        \draw[conn] (r32.east) -- (u44);
        \draw[conn] (r33.east) -- (u43);
        \draw[conn] (r33.east) -- (u44);
        \draw[conn] (r33.east) -- (u45);

        \draw[conn] (u41.east) -- (u51);
        \draw[conn] (u41.east) -- (u52);
        \draw[conn] (u41.east) -- (u53);

        \draw[conn] (u42.east) -- (u52);
        \draw[conn] (u42.east) -- (u53);
        \draw[conn] (u42.east) -- (u54);

        \draw[conn] (u43.east) -- (u53);
        \draw[conn] (u43.east) -- (u54);
        \draw[conn] (u43.east) -- (u55);

        \draw[conn] (u44.east) -- (u54);
        \draw[conn] (u44.east) -- (u55);
        \draw[conn] (u44.east) -- (u56);

        \draw[conn] (u45.east) -- (u55);
        \draw[conn] (u45.east) -- (u56);
        \draw[conn] (u45.east) -- (u57);

        \draw[conn] (u51.east) -- (r61);
        \draw[conn] (u52.east) -- (r61);
        \draw[conn] (u53.east) -- (r61);

        \draw[conn] (u53.east) -- (r62);
        \draw[conn] (u54.east) -- (r62);
        \draw[conn] (u55.east) -- (r62);

        \draw[conn] (u55.east) -- (r63);
        \draw[conn] (u56.east) -- (r63);
        \draw[conn] (u57.east) -- (r63);

        \draw[conn] (r61.east) -- (u71);
        \draw[conn] (r61.east) -- (u72);
        \draw[conn] (r61.east) -- (u73);
        \draw[conn] (r62.east) -- (u72);
        \draw[conn] (r62.east) -- (u73);
        \draw[conn] (r62.east) -- (u74);
        \draw[conn] (r63.east) -- (u73);
        \draw[conn] (r63.east) -- (u74);
        \draw[conn] (r63.east) -- (u75);

        \draw[conn] (u71.east) -- (y);
        \draw[conn] (u72.east) -- (y);
        \draw[conn] (u73.east) -- (y);
        \draw[conn] (u74.east) -- (y);
        \draw[conn] (u75.east) -- (y);

        \newcommand{\labfontsz}{9pt}
        \node at (\lin,2.0) {\fontsize{\labfontsz}{2pt}\textbf{Inputs}};
        
        \node at (\figlabLI,3.5) {\fontsize{\labfontsz}{2pt}\textbf{Convolution}};
        \node at (\figlabLII,3.5) {\fontsize{\labfontsz}{2pt}\textbf{Convolution}};
        \node at (\lV,3.0) {\fontsize{\labfontsz}{2pt}\textbf{Convolution}};
        
        \node at (\lout,1.0) {\fontsize{\labfontsz}{2pt}\textbf{Output}};


        \node at (\pII,2.55) {\fontsize{\labfontsz}{2pt}\textbf{LCRM}};
        \node at (\pIV,2.55) {\fontsize{\labfontsz}{2pt}\textbf{LCRM}};
        \node[reduce_node_fg] at ({\pII},{2.55-0.35}) {\scriptsize $r=2$, $w=3$};
        \node[reduce_node_fg] at ({\pIV},{2.55-0.35}) {\scriptsize $r=2$, $w=3$};
        \node[reduce_node_fg] at ({\pII},{2.55-0.7}) {\scriptsize $\Hreduc=3$};
        \node[reduce_node_fg] at ({\pIV},{2.55-0.7}) {\scriptsize $\Hreduc=3$};

        \node[reduce_node_fg] at (\figlabLI,3.2) {\scriptsize $\filtlen=2$};
        \node[reduce_node_fg] at (\figlabLII,3.2) {\scriptsize $\filtlen=2$};
        \node[reduce_node_fg] at (\lV,2.7) {\scriptsize $\filtlen=2$};

    \end{tikzpicture}
}

    \caption{
    Example of a CDNN with local dimension-reduction layers, where $d=3$, $\filtlen=2$, and $L_n=7$.
    For readability, $\LpoolI=\LpoolII=3$, $\Hreduc=3$ and each reduction layer is illustrated 
    with window size $w=3$ and stride length $r=2$.
    }
    \label{fig:DNN_CNV}
\end{figure}
}

See Figure 
\ref{fig:DNN_CNV}
for a simplified example of the CDNN architecture $\FCONV$.
Writing $\nv^{(l)}$ using Toeplitz convolution matrices $\Gcv^{(l)}$ illustrates the structure of $\FCONV$:  
connections between convolution layers are local, with neighborhood size determined by $\filtlen$, and these connections have shared weights. 
Without \LCRM\ layers the widths would be increasing linearly as $\Hicnv{l}=d+l\filtlen$, $l\in\{1,\ldots,\L\}$.
However, with \LCRM\ layers, the layer widths never exceed $\Hreduc+(\LpoolII-1)\filtlen$.

The convolutional architecture substantially reduces the total number of free DNN weights relative to a fully connected network with the same layer widths. 
Each convolution layer $l$ contributes only $\filtlen+1$ filter coefficients and $\Hicnv{l}$ intercept terms, while each reduction layer contributes $\Hreduc \cdot w_r$ coefficients from the \LCRM\ $\Av_r^{(l)}$ and $\Hreduc$ intercept terms.
Thus, 
letting $w^{(\LpoolI)} = w_d$ and
$w^{(\LpoolI+(j-1)\LpoolII)} = w_{2d+10}$ for $j\geq 2$,
the total number of free CDNN weights is given by
    \begin{equation}\label{eq:WCDNN}
    \begin{aligned}
        \W^{(\text{CDNN})} 
    &
    =
        \hspace{-15pt}
        \underset{l\notin\LCRMlayersShort}{\sum_{1\leq l \leq \L,}}
        \hspace{-15pt}
        (\filtlen+1+\Hicnv{l}) 
        +
        \hspace{-15pt}
        \underset{l\in\LCRMlayersShort}{\sum_{\LpoolI \leq l \leq \L,}}
        \hspace{-15pt}
        \Hreduc (w^{(l)}+1)
        + \Hicnv{\L} + 1
    .
    \end{aligned}
    \end{equation}
By contrast, an FCDNN with the same widths has 
$\Hicnv{l} \Hicnv{l-1} +\Hicnv{l}$
free DNN weights per layer, yielding
    \begin{equation*}
    \begin{aligned}
        \W^{(\text{FCDNN})}
    &
    =
        \hspace{-15pt}
        \underset{l\notin\LCRMlayersShort}{\sum_{1\leq l \leq \L,}}
        \hspace{-15pt}
        (\Hicnv{l} \Hicnv{l-1} +\Hicnv{l}) 
        +
        \hspace{-15pt}
        \underset{l\in\LCRMlayersShort}{\sum_{\LpoolI \leq l \leq \L,}}
        \hspace{-15pt}
        \Hreduc (\Hicnv{l-1}+1)
        + \Hicnv{\L} + 1
    .
    \end{aligned}
    \end{equation*}
Thus, the CDNN architecture achieves a substantial reduction in complexity by replacing dense inter-layer connectivity with structured local interactions.

    \subsection{Nonasymptotic error bounds for CDNNs}\label{sec:CDNN_res}


 This subsection derives nonasymptotic error bounds where $\FMLPg$ is constructed using the CDNN sieve spaces $\FCONV=\FCONVnon(\L,\filtlen,\DNNBoundfix)$. In contrast to the FCDNN results of Section \ref{sec:FCDNN}, the architecture here has width controlled through the convolution and reduction structure, so the main tuning parameter is the depth $\L$ and filter length $\filtlen$.
 To apply  Theorem \ref{thrm:ROC_V2} it is enough to control two quantities: the approximation error of $\FCONV$ over H\"older classes and the complexity of the sieve spaces $\Pdim(\FCONV)$.
 Proposition \ref{prop:CDNN_approx} and Lemma \ref{lem:CDNN_pdim} provide these ingredients.

 Proposition \ref{prop:CDNN_approx} gives a sup-norm approximation bound for $\FCONV$ over Hölder-smooth functions on $\XspaceC$. 
 The result shows that the expansive-convolution architecture, together with the local reduction layers, retains sufficient approximation power despite the parameter sharing and local connection structure.
    
\newcommand{\Cconvapprox}{\const{Cconvapprox}}%
\begin{proposition}\label{prop:CDNN_approx}
\!\RoneNote{7}\!%
    Suppose $\DNNBoundfix> 0$, and $2\leq \filtlen\leq d$.
    Then there exists a constant $\Cconvapprox>0$, depending only on $\smooth,d$, such that for any 
    $\L\geq 2$,
    and any
    $
    \fO\in \HolderCust{\smooth}(\XspaceC)
    $
    where $\norm{\fO}_{\infty|\XspaceC} < \DNNBoundfix$,
    $$
            \inf_{f\in\FCONVnon(\L,\filtlen,\DNNBoundfix)}
            \norm{\fO-f}_{\infty|\XspaceC}
    \leq
        \Cconvapprox
        \L^{-\frac{2\smooth}{d}}
        \log(\L)^{\frac{2\smooth}{d}}
    .   
    $$
\end{proposition}


Proposition \ref{prop:CDNN_approx} 
is new to the literature and 
gives the approximation rate for the CDNN class studied here. See \citet[Table 1]{shen_approximation_2022} and \citet[Table 1]{Yang_CDNDconvergence2025} for summaries of existing CDNN approximation results. The approximation rate matches the fast approximation result of \citet{shen_approximation_2022} for CDNNs with down-sampling. This is strictly faster than approximation rates obtained for CDNNs that control complexity through constraints on the DNN weights rather than through dimension-reduction steps, such as \citet{Yang_CDNDconvergence2025} and the references cited therein.

Lemma \ref{lem:CDNN_pdim} applies \citet{bartlett_nearly_tight_2019} to bound $\Pdim(\FCONV)$. This result reflects the fact that the number of free DNN weights grows much more slowly than in comparably sized FCDNNs.

\newcommand{\Cconvpdim}{\const{Cconvpdim}}%
\begin{lemma}\label{lem:CDNN_pdim}
    There exists a constant $\Cconvpdim>0$ such that
    $
    \Pdim(\FCONVnon(\L,\filtlen,\DNNBoundfix))
    \leq
    \Cconvpdim
    \L^2\log(\L)
    .
    $
\end{lemma}

Combining Proposition \ref{prop:CDNN_approx} and Lemma \ref{lem:CDNN_pdim} with Theorem \ref{thrm:ROC_V2} yields the following CDNN result.
Since the approximation result is developed on $\XspaceC$, Theorem \ref{thrm:CDNN_ROC} is stated under bounded covariates. 

\begin{theorem}\label{thrm:CDNN_ROC}
\newcommand{\temp}{\eta}%
    Suppose Assumptions \ref{as:data_holder} and \ref{as:data_subg} hold,
    and $\P(\Xt\notin \XspaceC)=0$ for all $t\in\N$. 
    Let 
        $
        \Zbound
        $
    be chosen 
    according to \eqref{eq:xbound_ybound_choice}.
    Let 
        $
            \FMLPg
        =
            \FCONVnon(\L,\filtlen,\DNNBoundfix)
        $
    where
    $\DNNBoundfix> \norm{\fO}_{\infty|\XspaceC}$, and $2\leq \filtlen\leq d$.
    Suppose 
    Condition \ref{as:curv} holds,
    and that, for each $n$ sufficiently large,
        there exist 
        $
        \Mbound
        $
        and 
        $
        \trunc
        $
    satisfying 
    Condition \ref{as:lip} 
    and \eqref{eq:trunc_term_def}.
    \begin{enumerate}
        \item 
        If 
            $\beta(j)\leq 
        e^{-\Cbeta\,j } $
        for some 
            $
            \Cbeta
            >0
            $,
        and 
            $
            \L
            =
            n^{\frac{1}{2}\left(\frac{d}{2\smooth+d}\right)}
            \log(n)^{\growthlog}
            $
        for some $\growthlog\in\R$,
        then
        \eqref{eq:pbound_FCDNN_ROC} holds with 
            $
                \Pbndterm
            =
            e^{-\L^2\log(n)^2}
            +
                    {n^{1-\Cbeta\log(n)}}/{\log(n)}
                    +
                    2
                    n^{1-\Csubg\!\log(n)}\!
                    \log(n)
            $,
        and
        \begin{equation*}
        \begin{aligned}
            \err
        &
        =
            n^{-\frac{2\smooth}{2\smooth+d}}
            \left[
            \big(\!\max\big\{\!\Mbound,\DNNBoundfix\!\big\}\big)^2
            \log(n)^{4+2\growthlog}
            +
            \log(n)^{\left(\frac{4\smooth}{d}\right)(1-\growthlog)}
            \right]
            +
            \frac{1}{n}
            +
            \trunc
            +
            \theta_n
        .
        \end{aligned}
        \end{equation*}
    \item 
\!\RtwoNote{4}\!%
        If $\beta(j)\leq j^{-\Cbeta} $ for some $\Cbeta>0$,
        and 
            $
            \L
            =
            n^{\frac{1}{2}(1-\temp)\left(\frac{d}{2\smooth+d}\right)}
            \log(n)^{\growthlog}
            $
        for some $\growthlog\in\R$ and 
        $\temp \in \big[(1+\Cbeta)^{-1}, 1 \big) $,
        then
        \eqref{eq:pbound_FCDNN_ROC} holds with 
        $
            \Pbndterm
        =
            e^{-\L^2\log(n)^2}
            +
                n^{1-\temp(1+\Cbeta)}\!
                +
                2
                n^{1-\Csubg\!\log(n)}
                \!
                \log(n)
        $,
    and
        \begin{equation*}
        \begin{aligned}
            \err
        &
        =
            n^{-(1-\temp)\frac{2\smooth}{2\smooth+d}}
            \left[
            \big(\!\max\big\{\!\Mbound,\DNNBoundfix\!\big\}\big)^2
            \log(n)^{3+2\growthlog}
            +
            \log(n)^{\left(\frac{4\smooth}{d}\right)(1-\growthlog)}
            \right]
            +
            \frac{1}{n}
            +
            \trunc
            +
            \theta_n
        .
        \end{aligned}
        \end{equation*}
    \end{enumerate}
\end{theorem}

Theorem \ref{thrm:CDNN_ROC} provides the CDNN analogue of Theorem \ref{thrm:FCDNN_ROC}.
As in the FCDNN case, part (i) can yield a near minimax rate in the sense of \citet{stone_optimal_1982} under exponential $\beta$-mixing; part (ii) yields the corresponding slower rate under polynomial $\beta$-mixing.

Corollary \ref{cor:CDNN} specializes Theorem \ref{thrm:CDNN_ROC} to the benchmark regression, logistic regression, and quantile regression settings. 
In each case, the model-specific regularity conditions developed in Section \ref{sec:FCDNN} verify the curvature and tail requirements needed for Theorem  \ref{thrm:CDNN_ROC} to yield concrete results. 


\begin{corollary}\label{cor:CDNN}
\!\AENote{1}\!%
\!\RtwoNote{2}\!%
    For each of the following cases, 
    suppose Assumptions \ref{as:data_holder} and \ref{as:data_subg} hold, 
    $\P(\Xt\notin \XspaceC)=0$ for all $t\in\N$, and $\beta(j)\leq 
        e^{-\Cbeta\,j }$.
    Let 
        $\f$ 
    satisfy \eqref{eq:fhat}
    with
        $\theta_n\lesssim n^{-\frac{2\smooth}{2\smooth+d}}$
    and
        $
            \FMLPg
        =
            \FCONVnon(\L,\filtlen,\DNNBoundfix)
        $
    where
    $\DNNBoundfix> \norm{\fO}_{\infty|\XspaceC}$,
    $2\leq \filtlen\leq d$.
    \begin{enumerate}
    \item Consider the setting of Example \ref{ex:reg}.
    If 
        $
            \L
        \asymp
            n^{\frac{1}{2}\left(\frac{d}{2\smooth+d}\right)}
            \log(n)^{\frac{2\smooth-3d}{2\smooth+d}}
        $,
    then  \eqref{eq:pbound_FCDNN_ROC} holds with 
            $
            \err
            \asymp
            \left[n^{-1}\log(n)^{8}\right]^{\frac{2\smooth}{2\smooth+d}}
            .
            $
    \item
        Consider the setting of Example \ref{ex:log}.
        Suppose Assumption
        \ref{as:data_logistic_fbound}
        holds.
        If 
            $
            \L
        \asymp
            n^{\frac{1}{2}\left(\frac{d}{2\smooth+d}\right)}
            \log(n)^{\frac{2(\smooth-d)}{2\smooth+d}}
            $,  
        then \eqref{eq:pbound_FCDNN_ROC} holds with 
            $
            \err
            \asymp
            \left[n^{-1}\log(n)^{6}\right]^{\frac{2\smooth}{2\smooth+d}}
            .
            $
    \item
        Consider the setting of Example \ref{ex:quant}.
        Suppose Assumption
        \ref{as:data_quantile}
        holds.
        If
            $
            \L
        \asymp
            n^{\frac{1}{2}\left(\frac{d}{2\smooth+d}\right)}
            \log(n)^{\frac{2(\smooth-d)}{2\smooth+d}}
            $
            and
            $\DNNBoundfix
        =
        2\Cquant
        $,
        then \eqref{eq:pbound_FCDNN_ROC} holds with 
            $
            \err
            \asymp
            \left[n^{-1}\log(n)^{6}\right]^{\frac{2\smooth}{2\smooth+d}}
            .
            $
    \end{enumerate}
\end{corollary}

    In all three settings Corollary \ref{cor:CDNN} shows that, under possibly nonstationary exponential $\beta$-mixing data with subgaussian outcomes, CDNN estimators obtain the rate $n^{-\frac{2\smooth}{2\smooth+d}}$ up to logarithmic factors. 
    The logarithmic exponents differ across the regression, logistic, and quantile cases because the verification of the general conditions depends on the loss function and on the corresponding model-specific regularity conditions.

    Relative to the existing CDNN literature, these results are complementary in two respects. First, they are obtained under possibly nonstationary $\beta$-mixing time-series data, whereas existing nonparametric convergence-rate results for CDNNs are largely confined to \iid\!settings. Second, they cover conditional mean, logistic, and quantile regression. While the CDNN class considered here is only one specific example, Corollary \ref{cor:CDNN} shows that the sieve-estimation framework developed in this paper can be adapted to CDNNs in time-series settings where they have been used by applied work.

\section{Summary and extensions}
\label{sec:Conc}

    This paper addresses the lack of statistical foundation for empirical work using deep neural network (DNN) estimators under dependent data.
    I provide a general sieve-estimation framework that yields nonasymptotic control of both theoretical and empirical $\Lp{2}$-errors under non-\iid\! $\beta$-mixing observations, allowing for nonstationarity and unbounded data. This framework is sufficiently flexible to cover modern DNN architectures used in econometric applications. In particular, I derive results for fully connected ReLU networks with unrestricted weights and apply these results to nonparametric conditional mean, logistic, and quantile regression. I also establish results for convolutional architectures, illustrating that the general theory is not limited to standard FCDNN architectures. The practical relevance of my results beyond these direct estimation problems is demonstrated by \citet{brown_inference_2024}, which considers a partially linear regression model with dependent data and unbounded regressands,
    and shows that the estimator for the finite-dimensional parameter, constructed using DNN-estimated nuisance components, achieves $\sqrt{n}$-consistency and asymptotic normality.

    Perhaps the most important avenue for future research on DNN estimators under dependent data is recurrent neural networks (RNNs). 
    RNNs are a class of DNN architectures that are specifically designed for time series settings, due to a recursive feedback loop that gives the network a form of `memory' for past events.
    While the empirical results from \citet{lazcano_back_2024} indicate that the DNN architectures studied here outperform RNNs in simpler time-series models, RNNs have demonstrated superior empirical performance in more complex settings, such as in the work by \citet{bucci_realized_2020} on forecasting stock market volatility. 
    Thus far, very little work has been done on the theoretical properties of RNNs. 
    While \citet{kohler_recurrent_2023} provides an initial study of a specific recurrent architecture in time-series nonparametric regression, much remains to be understood, particularly for more general recurrent architectures, and dependence settings.
    Following the ideas of Appendix \ref{sec:DNN_Alt_Arch}, the sieve estimator framework introduced in this paper could facilitate future research for RNNs once their approximation power and complexity are better understood.

    Many other important aspects of DNNs are also not considered here,
    such as computational efficiency or potential gains from alternative architectures and regularization techniques.
    The results given here could also be adapted for classes of functions beyond the standard H\"older smoothness condition, using approximation results such as \citet{Imaizumi_DNN_nonsmooth_2019}. 
    These considerations are left for future research.

\pagebreak
\addappheadtotoc 
\appendix
\appendixpage

\section{Measurability of extrema of random functions}\label{sec:meas}
{
\renewcommand{\Zspace}{\boldsymbol{\mathcal{Z}}}%

\textbf{Notation:} 
For two $\sigma$-algebras $\mathcal{A},\mathcal{B}$ the product $\sigma$-algebra is 
            $
                \mathcal{A}\otimes\mathcal{B}
            \coloneqq 
                \sigma\big(\big\{
                    a\times b : a\in \mathcal{A}, b\in\mathcal{B}
                \big\}\big)
            .
            $
For a set $A$ in a topological space, let $\closure{A}$ denote the closure of $A$; i.e. the intersection of all closed sets containing $A$. If $A\subseteq\R^n$, then $\closure{A}$ is with respect to the standard topology on $\R^n$.
The extended real line is denoted as
            $\overline{\R}\coloneqq\R\cup\{-\infty,\infty\}$.
For a random sequence 
            $\{\Xt\}_{t\in\N}$
        let
            $\sigma\left(\{\Xt\}_{k}^n\right)$ 
        denote the $\sigma$-algebra generated by $\{\Xt\}_{t=k}^n$.
For a topological space $(\Xspace,\mathcal{O}_{\Xspace})$, let $\borel(\Xspace)\coloneqq\sigma(\mathcal{O}_{\Xspace})$ denote the Borel $\sigma$-algebra associated with $\Xspace$.

Let
    $\{\Zt\}_{t\in\mathbb{N}}$
be a stochastic sequence on
    $\probspace$,
with coordinates given by random vectors
    $\Zt:\Omega \to \Zspace\subseteq\R^{d_Z}$
for some $d_Z\in\N$ and each $t$.
The parameter space, $\measurableftns$, 
is a 
space
of functions 
with elements $f:\Zspace\to\R$
measurable-$\borel(\Zspace)/\borel(\R)$.
Let 
    $\crit: \Zspace \times \R \to \R$
be 
measurable-$\big(\borel(\Zspace)\otimes\borel(\R)\big)/\borel(\R)$,
and
    $\crit\big(\Zt(\cdot),f(\Zt(\cdot))\big) \in \Lp{1}\probspace$,
for each $t$ and $f\in\measurableftns$.%
\footnote{%
Requiring that $\crit$ be defined on 
    $\Zspace \times \R$ 
instead of the subset
    $\Zspace \times \measurableftns(\Zspace)$,
where
    $\measurableftns(\Zspace)
    \coloneqq 
    \cup_{\z\in\Zspace}
    \big\{
        f(\z): f\in\measurableftns
    \big\}\subseteq\R,
    $
is without loss of generality by
\citet[Lemma 2.14]{stinchcombe_measurability_1992}. 
}

Definition \ref{def:point_sep} and Proposition \ref{prop:meas_sieve} provide easily verifiable conditions that ensure measurability in many sieve estimation settings. 
    Different from existing measurability results
    these results are applicable in settings where $\FMLPg$ is constructed using DNNs with unbounded weights.
    Lemma \ref{lem:FMLP} demonstrates this for the FCDNN class $\FMLP$ defined as in \eqref{eq:sievespace}, and similar reasoning applies to the CDNN class $\FCONV$.

{
\renewcommand{\FMLPg}{\mathcal{H}}%
\newcommand{\ftn}{h}%
\renewcommand{\Xspace}{\R}%
\newcommand{\borelX}{\borel(\Xspace)}%
\newcommand{\CountSub}{\{\ftn_j\}_{j\in\N}}%
\begin{definition}\label{def:point_sep}
\textbf{\textsc{(Finite-Point Separable)}}
    Let $\FMLPg$ be a set of functions with elements 
        $\ftn:\Zspace \to \Xspace$
    that are measurable-$\borel(\Zspace)/\borelX$.
    The set $\FMLPg$ is finite-point separable if there is a countable subset 
        $\CountSub\subseteq\FMLPg$ 
    where for every 
        $\ftn\in\FMLPg$, 
        $(\z_1,\ldots,\z_n)^\tr\in\Zspace^n$, and 
        $\delta>0$,
    there exists
        $j=j(\z_1,\ldots,\z_n,\delta,\ftn) \in \N$
    such that 
    $\max_{1\leq k \leq n}|\ftn_j(\z_k)-\ftn(\z_k)|<\delta.$
\end{definition}
    \vx
}
{

\newcommand{\tempsub}{\mathcal{H}}%
\newcommand{\temp}{U_n}%
\newcommand{\supfun}{\upsilon_n}%
\newcommand{\selec}{s}%
\newcommand{\selecm}{\selec_m}%
\begin{proposition}\label{prop:meas_sieve}
    Fix $n\in\N$. Let $\probspace$ be a complete probability space,
    and $\tempsub$ be a nonempty finite-point separable class of
    functions $h:\Zspace\to\R$. 
    Let $\temp:\Omega\times\R^n\to{\R}$ be such that
    $\temp(\cdot,\x)$ is $\mathcal A/\mathcal B({\R})$-measurable
    for each $\x\in\R^n$, and $\temp(\omega,\cdot):\R^n\to\R$ is continuous for each
    $\omega\in\Omega$. 
    Define
        $$
            \supfun(\omega)
        \coloneqq
            \inf_{h\in \tempsub}
            \temp\left(
            \omega,\{h(\Z_t(\omega))\}_{t=1}^n
            \right)
        .
        $$
    \begin{enumerate}
        \item 
        Then, the mapping
                $
                \omega
                \mapsto
                \upsilon_n(\omega)
                $
        is $\Zsig/\borel({\R})$-measurable.
        \item 
        In addition, suppose $\supfun(\omega)>-\infty$ for each $\omega\in\Omega$, and 
        $\tempsub\subset\Lp{r}(\P_{\{\Zt\}_{t=1}^n})$ for some $1\leq r <\infty$.
        Then, for each $m>0$, there exists $\selecm:\Omega\to\tempsub$ 
        such that for each $\omega\in\Omega$
            $$
                \selecm(\omega)
            \in
                \bigg\{
                    h\in\tempsub
                :\;
                    \temp\big(
                        \omega\,
                        ,
                        \{h(\Z_t(\omega))\}_{t=1}^n
                    \big)
                <
                    \supfun(\omega) + \frac{1}{m}
                \bigg\}
            \neq
                \emptyset
            ,
            $$
        and $\selecm$ is measurable-$\Zsig/\borel(\tempsub)$, where $\borel(\tempsub)$ 
        uses 
        the topology 
        generated by 
            $\norm{\cdot}_{\Lp{r}(\P_{\{\Zt\}_{t=1}^n})}$.
    \end{enumerate}
\end{proposition}

{
\newcommand{\ftn}{f}%

Proposition \ref{prop:meas_sieve} provides sufficient conditions, which will apply to many common choices of $\FMLPg$ and $\crit$, to ensure the objects in the proof of Theorem \ref{thrm:ROC_V2} are well defined.
To see this, let $\tempsub$ be a nonempty finite-point separable class of
functions and define, for $\x=(\x_1,\ldots,\x_n)\in\R^n$,
$$
    \temp(\omega,\x)
    :=
    \sampavg \crit(\Zt(\omega),\x_t).
$$
If $\crit$ is measurable-
$\big(\borel(\Zspace)\otimes\borel(\R)\big)/\borel(\R)$
and $\crit(\z,\cdot):\R\to\R$ is continuous for each
$\z\in\Zspace$, then $\temp(\cdot,\x)$ is
$\Zsig/\borel(\R)$-measurable for each $\x\in\R^n$,
and $\temp(\omega,\cdot):\R^n\to\R$ is continuous for each
$\omega\in\Omega$. Hence Proposition \ref{prop:meas_sieve}(i) applies and 
ensures
$$
    \omega
    \mapsto
    \inf_{f\in\tempsub}
    \sampavg \crit(\Zt(\omega),f(\Zt(\omega)))
$$
is $\Zsig/\borel(\overline{\R})$-measurable.
Similarly, suprema are handled by applying the same result to $-\temp$.
In addition, Proposition \ref{prop:meas_sieve}(ii) ensures existence of a measurability mapping $\omega\mapsto\f$ from $\Omega$ to $\FMLPg$ satisfying \eqref{eq:fhat}.

This differs from measurable maximum theorems such as
\citet[Theorem 2.2]{wooldridge_white_1991}, which impose compactness-type
conditions to obtain exact measurable optimizers. Proposition
\ref{prop:meas_sieve} instead uses finite-point separability and continuity
of the finite-dimensional criterion to obtain measurable extrema and
measurable approximate optimizers. This is useful for DNN sieves with
unbounded weights, where compactness of the parameter space is not available.

The following lemma shows these results may apply with the FCDNNs with unbounded weights considered in Section \ref{sec:DNN_sieve_spaces}. Similar reasoning also applies to the CDNNs from Section \ref{sec:CDNN}.

}

\newcommand{\tempsubr}{\tempsub_{r}}%
\begin{lemma}\label{lem:FMLP}
$\FMLP$ as in \eqref{eq:sievespace} has the following useful properties:
    \begin{enumerate}
        \item 
            The mappings $(\x,\gV) \mapsto \nff^{(\L+1)}_1(\x\mid\gV)$ from $\Xspace \times \R^{\W}$ to $\R$ are continuous.
        \item 
            Let $r>0$ be fixed, $\mu$ be a finite measure on $(\Zspace,\borel(\Zspace))$, and $f_0\in \Lp{2}(\mu)$. Define
                $
                \tempsubr
                \coloneqq
                \left\{
                f\in\FMLP:\|f-f_0\|_{\Lp{2}(\mu)}\le r
                \right\}.
                $
            If $\tempsubr\neq \emptyset$, then
            $\tempsubr$ is a nonempty finite-point separable class.
    \end{enumerate}
\end{lemma}

\subsection{Proofs for Measurability Results}

\begin{lemma}\label{lem:closure_equivilant_form}
    Let $n\in\N$, and let $\tempsub$ be a nonempty finite-point separable class of
    functions $h:\Zspace\to\R$. 
    Then, there exists a
    countable subset $\{h_j\}_{j\in\N}\subset \tempsub$ such that, for every
    $(\z_1,\ldots,\z_n)^\tr\in\Zspace^n$,
    \begin{equation*}
        \closure{\left\{
    (h(\z_1),\ldots,h(\z_n)):h\in \tempsub
    \right\}}
    =
    \closure{\left\{
    (h_j(\z_1),\ldots,h_j(\z_n)):j\in\N
    \right\}}.
    \end{equation*}
\end{lemma}
\begin{proof}
    Take any
        $h\in\tempsub$.
    Then, for each $\delta>0$ there exists
        $j=j(\z_1,\ldots,\z_n,\delta,h) \in \N$
    such that 
        $\max_{1\leq k \leq n}|h_j(\z_k)-h(\z_k)|<\delta.$
    Since norms on finite dimensional spaces are equivalent, every $\delta$-neighborhood of $(h(\z_1),\ldots,h(\z_n))$ with respect to the Euclidean metric contains $(h_j(\z_1),\ldots,h_j(\z_n))$ for some $j\in\N$.
    Therefore, 
        $$
            \closure{\left\{
            (h(\z_1),\ldots,h(\z_n)):h\in \tempsub
            \right\}}
        \subseteq
            \closure{\left\{
            (h_j(\z_1),\ldots,h_j(\z_n)):j\in\N
            \right\}}
        .
        $$
    For the reverse inclusion, $\{h_j\}_{j\in\N}\subset \tempsub$ implies
        $$
            \closure{\left\{
            (h(\z_1),\ldots,h(\z_n)):h\in \tempsub
            \right\}}
        \supseteq
            \closure{\left\{
            (h_j(\z_1),\ldots,h_j(\z_n)):j\in\N
            \right\}}
        ,
        $$
    which completes the proof.
\end{proof}
    \vx

\begin{proof}[Proof of Proposition \ref{prop:meas_sieve}]
\newcommand{\Hvecj}{H_{n,j}}%
    By Lemma \ref{lem:closure_equivilant_form}, there exists 
    $\{h_j\}_{j\in\N}\subseteq \tempsub$ such that 
        \begin{equation}\label{eq:closure_equivilant_form}
            \closure{
            \left\{
            (h(\Z_1(\omega)),\ldots,h(\Z_n(\omega))):h\in \tempsub
            \right\}
            }
        =
            \closure{
            \left\{
            (h_j(\Z_1(\omega)),\ldots,h_j(\Z_n(\omega))):j\in\N
            \right\}}\!
        .
        \end{equation}
    Define
        \begin{gather*}
            \Psi_{\tempsub,n}(\omega)
        \coloneqq
            \left\{
            (h(\Z_1(\omega)),\ldots,h(\Z_n(\omega))):h\in \tempsub
            \right\}
        ,
        \quad \text{and}
        \\
            \Hvecj(\omega)
        \coloneqq
            (h_j(\Z_1(\omega)),\ldots,h_j(\Z_n(\omega))),\;\;j\in\N
        .
        \end{gather*}

    \textbf{Proposition \ref{prop:meas_sieve}(i)}.
    By \eqref{eq:closure_equivilant_form},  
    $
    \closure{\Psi_{\tempsub,n}(\omega)}
    =
    \closure{
        \big\{
        \Hvecj(\omega): j\in\N
        \big\}
    }.
    $ 
    Since $\temp(\omega,\cdot)$ is continuous
        \begin{equation}\label{eq:supfun_countable}
        \begin{aligned}
            \supfun(\omega)
        &=
            \inf_{\x\in \Psi_{\tempsub,n}(\omega) }\temp(\omega,\x)
        =
            \inf_{\x\in \operatorname{cl}(\Psi_{\tempsub,n}(\omega)) }\temp(\omega,\x)
        =
            \inf_{\x\in \operatorname{cl}(\{\Hvecj(\omega): j\in\N \}) }\temp(\omega,\x)
        \\&
        =
            \inf_{\x\in \{\Hvecj(\omega): j\in\N \} }\temp(\omega,\x)
        =
            \inf_{ j\in\N }\temp(\omega,\Hvecj(\omega))
        .
        \end{aligned}
        \end{equation}
    For each $j\in\N$, the function
        $
        \temp(\cdot,\Hvecj(\cdot)):\Omega\to \R 
        $
    is measurable-$\Zsig/\borel(\R)$,
    since 
    $\Hvecj$ 
        is measurable-$\Zsig/\borel({\R^n})$,
    and 
        $\temp$
    is measurable-$(\Zsig\otimes\borel(\R^n))/\borel(\R)$
        by 
        \citet[Lemma 4.51]{aliprantis_infinite_2006}.
    Thus, 
        $\inf_{ j\in\N }\temp(\cdot,\Hvecj(\cdot))$
    is measurable-$\Zsig/\borel(\R)$ since $\N$ is countable.
    With \eqref{eq:supfun_countable} this completes the proof of part (i).
        

\newcommand{\Isetm}{I_m}%
\newcommand{\Isetmmin}{j_m}%
\newcommand{\Hveck}{H_{n,k}}%
\newcommand{\Hvecl}{H_{n,\ell}}%
\newcommand{\Hvecjm}{H_{n,\Isetmmin(\omega)}}%
\newcommand{\hjm}{h_{\Isetmmin(\omega)}}%
    \textbf{Proposition \ref{prop:meas_sieve}(ii)}.
    Let $m>0$ be arbitrary. Define
        $$
        \resizebox{\hsize}{!}{$
            \Isetm(\omega)
        \coloneqq
            \left\{
                j\in\N:
                \temp(\omega,\Hvecj(\omega))
                <\supfun(\omega)+\frac{1}{m}
            \right\}
        ,
        \quad
            \Isetmmin(\omega) \coloneqq \min \Isetm(\omega)
        ,
        \;\;
        \text{and }
        \;\;
        \selecm(\omega) \coloneqq \hjm
        .
        $}
        $$
    
    First, we show there exists a mapping $\omega\mapsto\selecm(\omega)\in\tempsub$ for each $\omega\in\Omega$. 
    Note that
    $\Isetm(\omega)\neq \emptyset$ for all $\omega\in\Omega$, since  
        $\supfun(\omega)>-\infty$ 
        by assumption
    and
        $
            \supfun(\omega)
        =
            \inf_{ j\in\N }\temp(\omega,\Hvecj(\omega))
        $
        by \eqref{eq:supfun_countable}.
    Thus, $\selecm(\omega)\in \tempsub$ for each $\omega\in\Omega$.
        
    Next, we show $\omega\mapsto\selecm(\omega)$ from $\Omega$ to $\tempsub$ is measurable.
    For each $k\in\N$, 
    we have $\Isetmmin^{-1}(k)\in\Zsig$
    since $\Hvecj$ and $\supfun$ are measurable, and
        \begin{equation*}
        \resizebox{\hsize}{!}{$
        \begin{aligned}
            \Isetmmin^{-1}(k)
        &=
            \{ \omega\in\Omega: \Isetmmin(\omega)=k\}
        \\&
        =
                \left\{
                    \omega
                    \in\Omega
                    : 
                    \temp(\omega,\Hveck(\omega))
                    <\supfun(\omega)+\frac1m
                \right\}
            \cap
                \left\{
                    \bigcap_{\ell<k}
                    \left\{
                        \omega
                        \in\Omega
                        :
                        \temp(\omega,\Hvecl(\omega))
                        \geq \supfun(\omega)+\frac1m
                    \right\}
                \right\}\!
        .
        \end{aligned}
        $}
        \end{equation*}
    Thus,
    $\Isetmmin$ is measurable-$\Zsig/\borel(\N)$.
    Then, 
    $\selecm^{-1}(B)\in\Zsig$
    for any 
        $B\in\borel(\tempsub)$,
    since $\Isetmmin$ is measurable and
        $$
        \selecm^{-1}(B)
        =
        \Big\{
            \omega\in\Omega:
            \hjm\in B
        \Big\}
        =
        \bigcup_{\{k\in\N:\, h_k  \in B\}}
        \Big\{
            \omega\in\Omega:
            \Isetmmin(\omega) =k
        \Big\}
        .
        $$
    
    Finally, by construction,
        $$
        \temp\big(
        \omega,
        \{\selecm(\omega)\circ \Z_t(\omega)\}_{t=1}^n
        \big)
        =
         \temp(\omega,\Hvecjm(\omega) )
        < \supfun(\omega)+\frac{1}{m}
        .
        $$
    This completes the proof of part (ii).%
    \end{proof}
    \vx

\begin{proof}[Proof of Lemma \ref{lem:FMLP}]
    To simplify notation,
    for $\gv=\gV \in \R^{\W}$ by \eqref{eq:sievespace} we can write 
        $
        f_{\gv} \in \FMLP
        $
    to denote
        $$
            f_{\gv}(\x) 
        =
            \max\Big\{\! -\!\DNNBound,\, \min\{\DNNBound, \nff^{(\L+1)}_{1}(\x \,| \gv)\}
            \Big\}
        .
        $$
        
    \textbf{Lemma \ref{lem:FMLP}(i):}
    For $l\in \{1,\ldots,\L+1\}$, $h\in\{1,\ldots,\Hi{l}\}$ the function $\nff^{(l)}_{h}$ is composed of compositions of the ReLU activation function ($\relu$) and linear combinations of 
    $\{\g^{(l)}_{h}\}_{h=1}^{\Hi{l-1}}$ and $\{\nff_{l-1,h}\}_{h=1}^{\Hi{l-1}}$ (or $\x\in\Xspace$ when $l=1$).
    Thus, $\nff^{(l)}_{h}:\Xspace\times\R^{\W}\to \R$ is continuous for each $l,h$.
    Then, the proof of part (i) follows immediately since
        \begin{equation*}
            f_{\gv}(\x)
            =
            \begin{cases}
                \DNNBound
            ,
            \quad
            &\text{if}\;
                \nff^{(\L+1)}_{1}(\x \,| \gv)>\DNNBound,
            \\
            -\DNNBound
            ,
            \quad
            &\text{if}\;
                \nff^{(\L+1)}_{1}(\x \,| \gv)<-\DNNBound,
            \\
            \nff^{(\L+1)}_{1}(\x \,| \gv),
            \quad
            & \text{otherwise}.
            \end{cases}
        \end{equation*}

\newcommand{\gVspacer}{\Gamma_{\W,r}}
    \textbf{Lemma \ref{lem:FMLP}(ii):}
        Clearly,
        $f_{\gv}\in \Lp{2}(\mu)$ since $f_{\gv}$ is continuous and $\snorm{f_{\gv}}\leq \DNNBound$.
        Define the constrained DNN weight set
            $$
            \gVspacer
            \coloneqq
            \left\{
            \gv\in\R^{\W}
            :
            \norm{f_{\gv}-\fO}_{\Lp{2}(\mu)}\leq r
            \right\}
            .
            $$
        Then $\gVspacer$ is separable under the Euclidean topology since $\gVspacer \subseteq \R^{\W}$ and 
        $\R^{\W}$ is separable.
        Hence there exists a countable set
        $\{\gv_j\}_{j\in\mathbb N}\subseteq\gVspacer$ such that 
            for every $\gv\in\gVspacer$ and $\delta>0$ there exists $j\in\N$ such that
            $$
            \norm*{\gv_j-\gv}_{\lp{\infty}}<\delta.
            $$
        since norms on finite dimensional spaces are equivalent. 

        Now fix any $h\in\tempsubr$.
        Then, 
        $h=f_{\gv'}$
        for some $\gv'\in\gVspacer$.
        Fix $m\in\N$ and $\z_1,\ldots,\z_m\in\Zspace$.
        By Lemma \ref{lem:FMLP}(i) 
        $\gv\mapsto f_{\gv}(\z_k)$ is continuous for each $k=1,\ldots,m$.
        Hence, for any $\epsilon>0$
        and each $k\in\{1,\ldots,m\}$
        there exists 
        $\delta_{k,\epsilon}>0$
        such that 
            $$
            \abs*{f_{\gv}(\z_k)-f_{\gv'}(\z_k)}<\epsilon,
            \quad
            \forall \;
            \gv\in 
            \Big\{
                \gv\in 
                \gVspacer
                : \norm*{\gv-\gv'}_{\lp{\infty}}<\delta_{k,\epsilon}
            \Big\}
            .
            $$
        This implies
            $$
            \max_{k=1,\ldots,m}\abs*{f_{\gv}(\z_k)-f_{\gv'}(\z_k)}<\epsilon,
            \quad
            \forall \;
            \gv\in 
            \Big\{
                \gv\in 
                \gVspacer
                : \norm*{\gv-\gv'}_{\lp{\infty}}<
                \min_{k=1,\ldots,m}\delta_{k,\epsilon}
            \Big\}
            .
            $$
        Since $\{\gv_j\}_{j\in\N}$ is dense in $\gVspacer$, 
        there exists $j^*\in\N$ such that 
            $$
            \gv_{j^*}\in 
            \Big\{
                \gv\in 
                \gVspacer
                : \norm*{\gv-\gv'}_{\lp{\infty}}<
                \min_{k=1,\ldots,m}\delta_{k,\epsilon}
            \Big\}.
            $$
        For $j\in\N$ let $h_j\coloneqq f_{\gv_j}$, and recall $h=f_{\gv'}$.
        Then, 
        from the previous two displays,
            $$
            \max_{k=1,\ldots,m}\abs*{h_{j^*}(\z_k)-h(\z_k)}
            =
            \max_{k=1,\ldots,m}\abs*{f_{\gv_{j^*}}(\z_k)-f_{\gv'}(\z_k)}
            <\epsilon.
            $$
        This completes the proof since $
        \{f_{\gv}: \gv\in\{\gv_j\}_{j\in\N}\} \subseteq \tempsubr
        $
        is countable
        and
        $h$, $\z_1,\ldots,\z_m\in\Zspace$ and $\epsilon$ were arbitrary.%
\end{proof}
    \vx

}
}

\section{Proof of Theorem \texorpdfstring{\ref{thrm:ROC_V2}}{1}}\label{sec:PROOF_ROC_V2}
{
\renewcommand{\anA}{a}%
\newcommand{\anB}{b}%
\newcommand{\dif}[1]{g_{#1}}%
\newcommand{\difc}[1]{g_{#1}^{\comp}}%
\newcommand{\tempA}{\delta'}%
\newcommand{\tempB}{\delta}%
\newcommand{\critb}{{\crit}}%
\newcommand{\Critb}{{\Crit}}%
\newcommand{\one}{\mathbbm{1}_{nt}}%
\newcommand{\onec}{\one^{\comp}}%
The following proof for Theorem \ref{thrm:ROC_V2} generalizes the steps used by \citet{farrell_deep_2021}.
The ancillary results used throughout are listed after the proof, in Appendix \ref{sec:lem_ROC_V2}.


\subsection{Preliminaries}
Before proceeding, we first describe the independent blocking technique and provide some definitions which will be used throughout. 

\subsubsection{Independent blocks}\label{sec:IndBlock}
\newcommand{\Ione}[1]{T_{1,#1}}%
\newcommand{\Itwo}[1]{T_{2,#1}}%
\newcommand{\Ii}[2]{T_{#1,#2}}%
\newcommand{\Ik}[2]{T_{#1,#2}}%
\newcommand{\IR}{T_{R}}%
\newcommand{\IZ}{\Z^{*}}%
\newcommand{\IZt}{\IZ_t}%
\newcommand{\IX}{\X^{*}}%
\newcommand{\IXt}{\IX_t}%
\newcommand{\dataIZ}{\{\IZt\}_{t=1}^n}%
    This subsection constructs independent `blocks' commonly used when dealing with $\beta$-mixing processes (e.g. \citealp{chen_sieve_1998}; \citealp{kurisu_adaptive_2025}; \citealp{mirzaei_empirical_2025}) which will be used throughout the proof.
    By 
    assumption,  $2\leq\anA\leq n/2$, so 
        $\anB \coloneqq \big\lfloor n/(2\anA)\big\rfloor$
    is well defined, and $\anB\geq 1$.
    Then, we can divide 
        $\data$
    into $2\anB$ blocks of length $\anA$, and the remainder into a block of length $n-2\anB\anA$,
    using the 
    index sets
        \begin{equation}\label{eq:block_def}
        \begin{aligned}
            \Ik{1}{j} 
        &\coloneqq 
            \Big\{ t\in\N:\; 2(j-1)\anA +1 \leq t \leq (2j-1)\anA  \Big\},
        \quad j=1,\ldots,\anB;
        \\
            \Itwo{j} 
        &\coloneqq 
            \Big\{ t\in\N:\; (2j-1)\anA +1 \leq t \leq 2j\anA  \Big\},
        \quad j=1,\ldots,\anB;
        \\
            \IR 
        &\coloneqq
            \Big\{ t\in\N:\; 2\anB\anA +1 \leq t \leq n  \Big\}.
        \end{aligned}
        \end{equation}
    We use Berbee's Lemma to 
    redefine $\data$ on a richer probability space%
        \footnote{We will continue to refer to the richer probability space as $\probspace$ since the extension preserves the distribution of random variables defined on the original space. 
        }
    where there exists a random sequence $\dataIZ$ 
    such that, for $i\in\{1,2\}$ and
    $j,j'\in\{1,\ldots,\anB\}$ with $j\neq j'$,
    the blocks
        $
            \{\IZt\}_{t\in\Ii{i}{j}}
        $ 
    and 
        $
            \{\IZt\}_{t\in\Ii{i}{j'}}
        $ 
    are independent,
    and 
        $\P_{\{\IZt\}_{t\in\Ii{i}{j}}}=\P_{\{\Zt\}_{t\in\Ii{i}{j}}}$.
    %
    Together these two facts imply, for $i\in\{1,2\}$
        \begin{equation}\label{eq:IZT_ZT_blocks}
        \begin{aligned}
            \P_{
            \left\{\IZt: \;
            {
            t\,\in\, 
            \cup_{j=1}^{\anB} \Ik{i}{j}
            }
            \right\}} 
        &=
            \P_{\{\IZt\}_{t\in \Ik{i}{1}}} \times 
            \P_{\{\IZt\}_{t\in \Ik{i}{2}}} \times 
            \cdots \times 
            \P_{\{\IZt\}_{t\in \Ik{i}{\anB}}}
        \\&
        =
            \P_{\{\Zt\}_{t\in \Ik{i}{1}}} \times 
            \P_{\{\Zt\}_{t\in \Ik{i}{2}}} \times 
            \cdots \times 
            \P_{\{\Zt\}_{t\in \Ik{i}{\anB}}}
        ,
        \end{aligned}
        \end{equation}
    Next, 
    the usual $\beta$-mixing coefficient (e.g. \citealp[Definition 3.1, p.19]{dehling_empirical_2002}) can be equivalently written as (see \citealp{eberlein_weak_1984}) 
        $$
            \beta(m)
        = 
            \sup\left\{
                |\P(A \times B)- \P(A)\P(B)|
                :\,
                A\times B 
                \in 
                    \sigma\left(
                    \{\Zt\}_{t=1}^k
                    \right)
                \otimes
                    \sigma\left(
                    \{\Zt\}_{t=k+m+1}^\infty
                    \right)
                ,
                k\in\N
            \right\}
            .
        $$
    Hence, for each $k\in \{1,...,\anB-1\}$, for $i\in\{1,2\}$
        \begin{equation}\label{eq:beta_bound}
        \begin{aligned}
            \beta(\anA)
        \geq
            \sup\bigg\{
            \Big|&
                \P_{\left\{\Zt:\;t\,\in \cup_{j=1}^{\anB} \Ik{i}{j} \right\}}(A\times B)
                -
                \P_{\left\{\Zt:\;t\,\in \cup_{j=1}^k\Ik{i}{j}\right\}}(A)\;
                \P_{\left\{\Zt:\;t\,\in \cup_{j=k+1}^{\anB}\Ik{i}{j}\right\}}(B)
            \Big|
            : \;
        \\&
                A\times B\in 
                \sigma\big(
                \left\{\Zt:t\in \cup_{j=1}^k\Ik{i}{j}\right\}
                \big)
            \otimes
                \sigma\big(
                \left\{\Zt:t\in \cup_{j=k+1}^{\anB}\Ik{i}{j}\right\}
                \big)
            \bigg\}
        .
        \end{aligned}
        \end{equation}
    By \eqref{eq:beta_bound} the conditions for \citet{eberlein_weak_1984} Lemma 2 are satisfied. 
    So we apply this result, and use \eqref{eq:IZT_ZT_blocks},
    to obtain,
    for any measurable set $E$, for $i\in\{1,2\}$
        \begin{equation*}
        \begin{aligned}
            \Big|
                \Pr\Big(
                    \Big\{\Zt: 
                    t\in
                    \cup_{j=1}^{\anB} \Ik{i}{j}
                    \Big\}
                \in 
                    E
                \Big)
            -
                \Pr\Big(
                    \Big\{\IZt: 
                    t\in
                    \cup_{j=1}^{\anB} \Ik{i}{j}
                    \Big\}
                \in 
                    E
                \Big)
            \Big|
        &
        \; \leq \;
            (\anB-1)\beta(\anA)
        .
        \end{aligned}
        \end{equation*}
    Then, for $i\in\{1,2\}$, by the triangle inequality and
        $
            \anB 
        \coloneqq 
            \big\lfloor n/(2\anA)\big\rfloor
        <
        n/(2\anA)
        +1
        ,
        $
        \begin{equation}\label{eq:beta_bound2}
        \begin{aligned}
                \Pr\Big(
                    \Big\{\Zt: 
                    t\in
                    \cup_{j=1}^{\anB} \Ik{i}{j}
                    \Big\}
                \in 
                    E
                \Big)
        \leq
                \Pr\Big(
                    \Big\{\IZt: 
                    t\in
                    \cup_{j=1}^{\anB} \Ik{i}{j}
                    \Big\}
                \in 
                    E
                \Big)
        +
            \frac{n\,\beta(\anA)}{2\anA}
        .
        \end{aligned}
        \end{equation}
    Letting $\IZt = (\Yt^*,\IXt)$ the above also applies with $\Xt$ in place of $\Zt$.
\subsubsection{Preliminary Definitions}
%
\newcommand{\normbound}{\Xbound}
\newcommand{\normTj}[2]{\norm*{#1}_{T^{*}_{1,#2}|\normbound}}%
\newcommand{\normT}[1]{\norm*{#1}_{T^{*}_{1}|\normbound}}%
\newcommand{\normTij}[3]{\norm*{#2}_{T^{*}_{#1,#3}|\normbound}}%
\newcommand{\normTi}[2]{\norm*{#2}_{T^{*}_{#1}|\normbound}}%
\newcommand{\normLpTi}[2]{\norm*{#2}_{\mathcal{L}^{2}_{\!|\normbound}\!(\Pblocki{#1})}}%
\newcommand{\normLpITi}[2]{\norm*{#2}_{\mathcal{L}^{2}_{\!|\normbound}\!(\IPblocki{#1}\!)}}%
\newcommand{\IPblocki}[1]{\overline{\P}_{T^{*}_{\!#1}}}%
\newcommand{\Pblocki}[1]{\overline{\P}_{T_{\!#1}}}%
\newcommand{\normLpnS}[1]{ \norm*{#1}_{\mathcal{L}^{2}_{|\normbound\!}\!(\Pbarn\!) } }%
\newcommand{\Pbarn}{\overline{\P}_{\!\X_{1}^n}}%
\newcommand{\normLpn}[1]{\norm*{#1}_{\mathcal{L}^{2}(\Pbarn\!)}}%
    In light of Lemma \ref{lem:avgmeas}, we simplify notation by using the norm
    \begin{equation*}
        \normLpn{f}
        =
        \sqrt{
        \int_{\Xspace}
                [f(\x)]^2
        \,d
        \Pbarn
        (\x)
        }
    ,
    \quad
    \text{where}
    \quad
        \Pbarn
        \coloneqq \sampavg \P_{\Xt}
        .
    \end{equation*}
    Define $\normLpn{\cdot}$ restricted to 
    $\x\in\Xbound$ 
    as
        \begin{equation}\label{eq:normLpnS}
            \normLpnS{f}
            \coloneqq
            \sqrt{
            \int_{\x\in\Xbound}
                    [f(\x)]^2
            \,d
            \Pbarn
            (\x)
            }
            .
        \end{equation}
    Define the empirical block norms,
        \begin{equation}\label{eq:IndNorms}
        \begin{aligned}
            \normTij{i}{f}{j}
        &\coloneqq
                \bigg(\frac{1}{\anA}\sum_{t\in\Ik{i}{j}}
                \big|
                f\big(\IXt\big)
                \big|^2
                \mathbbm{1}_{\{\IXt \in \Xbound\}}
                \bigg)^{1/2},
            \quad
            j\in\{1,\ldots,\anB\}
            ,
            \; i\in\{1,2\},
        \\
            \normTi{i}{f}
        &\coloneqq
            \bigg(
                \frac{1}{\anB}
                \sum_{j=1}^{\anB}
                \normTij{i}{f}{j}^2
            \bigg)^{1/2}
        =
            \bigg(
                \frac{1}{\anB\anA}
                \sum_{j=1}^{\anB}
                \sum_{t\in \Ik{i}{j}}\big|f(\IXt)\big|^2
                \mathbbm{1}_{\{\IXt \in \Xbound\}}
            \bigg)^{1/2}
        ,
        \quad \; i\in\{1,2\}.
        \end{aligned}
        \end{equation}
    and the theoretical block norms 
        \begin{equation}\label{eq:normLpTi}
        \begin{aligned}
            \normLpTi{i}{f}
        &\coloneqq
            \bigg(
                \int_{\x\in\Xbound}
                |f(\x)|^2
                d
                \Pblocki{i}
                (\x)
            \bigg)^{1/2}
        \! , \; \text{ where } \;
            \Pblocki{i}
    \coloneqq 
        \frac{1}{\anB\anA}
        \sum_{j=1}^{\anB}
        \sum_{t\in \Ik{i}{j}} 
        \P_{\Xt}
        ,
        \;\; i\in\{1,2\}
        ,
        \end{aligned}
        \end{equation}
and similarly for $\normLpITi{i}{f}$ where 
    $\IPblocki{i}
    \coloneqq 
        \frac{1}{\anB\anA}
        \sum_{j=1}^{\anB}
        \sum_{t\in \Ik{i}{j}} 
        \P_{\IXt}$.
Note that by \eqref{eq:IZT_ZT_blocks}
        \begin{equation}\label{eq:exp_emp}
            \normLpTi{i}{f}^2
        =
            \normLpITi{i}{f}^2
        =
            \E\left[
            \normTi{i}{f}^2
            \right]
        .
        \end{equation}

    \newcommand{\RadX}[1]{\mathfrak{R}_{#1|\Xbound}}%
    \newcommand{\RadZ}[1]{\mathfrak{R}_{#1|\Zbound}}%
The following definition is the Rademacher complexity of a function class where the support is truncated. Rademacher complexity is commonly used in empirical process theory,
see, e.g.,
\citet{van_der_vaart_wellner_book_1996}; or \citet{bartlett_local_2005}.
{
\renewcommand{\Z}{\boldsymbol{W}}%
\renewcommand{\data}{\{\Zt\}_{t=1}^n}%
\newcommand{\tempspace}{\mathcal{H}}%
\newcommand{\fun}{h}%
\renewcommand{\Zspace}{\Xspace}%
\renewcommand{\Zbound}{\boldsymbol{\mathbb{W}}}%
\begin{definition}\label{def:Rad}
\textbf{\textsc{(Rademacher Complexity)}}
    For $n\in\N$, let $\data$ be random variables on $\probspace$ taking values in $\Zspace$.
    The Rademacher random variables, $\{\xi_t\}_{t=1}^{n}$, are i.i.d. random variables
    that are
    independent of 
        $\data$,
    and
        $\xi_t\in\{-1,1\}$ 
    where
        $\Pr(\xi_t=1)=\Pr(\xi_t=-1) = 1/2$. 
    For $\Zbound\subseteq\Zspace$ and a class of measurable functions $\tempspace$
    with elements
    $\fun:\Zspace\to\R$ 
    define 
        $$
            \RadZ{n}\tempspace
        \coloneqq
            \sup_{\fun\in\tempspace}
            \frac{1}{n}\sum_{t=1}^n
            \xi_t \,
            \fun(\Zt)
            \onezt
        .
        $$
    Define
    the Rademacher complexity as
        $\E[\RadZ{n}\tempspace],$ 
    and the empirical Rademacher complexity as
        $
            \E_{\xi}[\RadZ{n}\tempspace]
        $ $
        \coloneqq
            \E\big[
                \RadZ{n}\tempspace
                \;\big|
                \{\Zt\}_{t=1}^n
            \big]
        $.%
\footnote{
    Note that
        $\E[\RadZ{n}\tempspace]$  
    is well defined by letting $\{\xi_t\}_{t=1}^{n}$ be defined on $\probspace$ whenever $\probspace$ is rich enough, otherwise we can define $\{\xi_t\}_{t=1}^{n}$ on an auxiliary probability space 
        $(\Omega^{(\xi)},\Zsig^{(\xi)},\P^{(\xi)})$,
    and take the expectation over the product probability space
        $
            \probspace \times (\Omega^{(\xi)},\Zsig^{(\xi)},\P^{(\xi)})
        \coloneqq
            \big(
            \Omega \times \Omega^{(\xi)},
            \Zsig \otimes \Zsig^{(\xi)},
            \P \times \P^{(\xi)}\big)
        .
        $

}
\end{definition}
}

    We write
        $
            \RadX{\anA\anB}\FMLPg
        =
            \sup_{f\in\FMLPg}
            \frac{1}{\anA\anB}
            \sum_{j=1}^{\anB}
            \sum_{t\in\Ione{j}}
            \xi_{t}
            f(\Xt)
            \onext
        $
    since the double sum is of length $\anA\anB$.

\subsection{Main decomposition}
    Let 
        $
        \critt(f) \coloneqq \crit\big(\Zt,f\big)
        ,
        $
        $
            \one 
        \coloneqq 
                \mathbbm{1}_{\{
                \Zt \in \Zbound
                \}}
        ,
        $
        and
        $
            \onec
        \coloneqq 
            1-\one.
        $
    Define
        \begin{gather*}
            \dif{f}(\Zt)
        \coloneqq
            \big[\critt(f)-\critt(\fO)\big]
            \one
        ,
        \quad 
        \text{ and }
        \quad
            \difc{f}(\Zt)
        \coloneqq
            \big[\critt(f)-\critt(\fO)\big]
            \onec
        .
        \end{gather*}
    By Lemma \ref{lem:avgmeas} and Condition \ref{as:curv}, for any $f\in\FMLPg$,
        $$
                \Cpcq 
                \normLpn{f-\fO}^2  
            =
                \Cpcq
                \sampavg
                \E\Big[ 
                |f(\Xt)-\fO(\Xt)|^2
                \Big]
            \leq
                \E[\Critb(f)] - \E[\Critb(\fO)]
            .
        $$
    By \eqref{eq:fhat},
        $$
            0 
        \leq
            -\Critb(\f) + \Critb(\fpr) + \theta_n
        =
            -\big[\Critb(\f)- \Critb(\fO)\big]
            +
            \big[\Critb(\fpr)- \Critb(\fO)\big]
            + \theta_n
        .
        $$
\newcommand{\EPT}{(\mathrm{EmPr})_n}
\newcommand{\BiasT}{(\mathrm{Bias})_n}
\newcommand{\TruncT}{(\mathrm{Trunc})_n}
\newcommand{\fsub}{\mathcal{H}_n}
    Let 
        $$
        \fsub
        \coloneqq
        \left\{f\in\FMLPg: \normLpn{f-\fO}^2 \leq \normLpn{\f-\fO}^2\right\}
        $$
    With this, 
        \begin{equation}\label{eq:A1}
        \resizebox{0.915\hsize}{!}{$
        \begin{aligned}[b]
        &
            \Cpcq 
            \normLpn{\f-\fO}^2  
        \leq
            \Cpcq 
            \normLpn{\f-\fO}^2
            -\big[\Critb(\f)- \Critb(\fO)\big]
            +
            \big[\Critb(\fpr)- \Critb(\fO)\big]
            + \theta_n
        \\&
        \leq
            \sup_{f\in\fsub}
            \Big\{
            \Cpcq\normLpn{f-\fO}^2  
            -
            \big[\Critb(f)- \Critb(\fO)\big]
            \Big\}
            +
            \big[\Critb(\fpr)- \Critb(\fO)\big]
            + \theta_n
        \\&
        \leq
            \sup_{f\in\fsub}
            \Big\{
            \E\big[\Critb(f)- \Critb(\fO)\big]
            -
            \big[\Critb(f)- \Critb(\fO)\big]
            \Big\}
            +
            \big[\Critb(\fpr)- \Critb(\fO)\big]
            + \theta_n
        \\&
        =
            \sup_{f\in\fsub}
            \sampavg 
            \Big\{
            \E
            \big[ 
                \dif{f}(\Zt) + \difc{f}(\Zt)
            \big] 
            - 
            \big[ 
                \dif{f}(\Zt) + \difc{f}(\Zt)
            \big] 
            \Big\}
            +
                \big[\Critb(\fpr) - \Critb(\fO) \big]
            + \theta_n
        \\&
        \leq
            \EPT + \TruncT + \BiasT
            + \theta_n
        ,
        \end{aligned}
        $}
        \end{equation}
    where
        \begin{equation*}
        \begin{aligned}
            \EPT
        & \coloneqq
            \sup_{f\in\fsub}
            \sampavg 
            \Big\{
            \E
            \big[ 
                \dif{f}(\Zt)
            \big] 
            - 
                \dif{f}(\Zt)
            \Big\}
        ,
        \\
            \TruncT
        & \coloneqq
            \sup_{f\in\fsub}
            \sampavg 
            \Big\{
            \E
            \big[ 
                \difc{f}(\Zt)
            \big] 
            - 
                \difc{f}(\Zt)
            \Big\}
        ,
        \\
            \BiasT
        & \coloneqq
            \big[\Critb(\fpr) - \Critb(\fO) \big]
        ,
        \end{aligned}
        \end{equation*}
    are the empirical process term, truncation term, and bias term, respectively.

\subsubsection{Truncation term}
    First, 
        \begin{equation}\label{eq:difcprob}
        \begin{aligned}[b]
            \Pr\Bigg(
                \sup_{f\in\mathcal{H}_n}
                \sampavg
                \big|\difc{f}(\Zt)\big|
                >0
            \Bigg)
        &\leq
            \Pr\Bigg(
                \sup_{f\in\FMLPg}
                \sampavg
                \big|\critt(f)- \critt(\fO)\big|\onec
                >0
            \Bigg)
        \\&
        \leq
            \Pr\left(
                \sumin
                \onec
                >0
            \right)
        =
            \P(\Ptrunc)
        .
        \end{aligned}
        \end{equation}
    Second,
    by the triangle inequality, for any $f\in\FMLPg$
        \begin{equation}\label{eq:difctrunc}
        \begin{aligned}
            \E\big[ |\difc{f}(\Zt)|\big] 
        &
        =
            \E\Big[\big|\critt(f)-\critt(\fO)\big|\onec\Big]
        \leq
            2\sup_{f\in\left\{\FMLPg\cup\{\fO\}\right\}}
            \E\Big[
                \big|\critt(f)\big|
                \onec
            \Big]
        \leq
            2\trunc
        ,
        \end{aligned}
        \end{equation}
    which implies 
        $
            \sup_{f\in\mathcal{H}_n}
            \E\big[ |\difc{f}(\Zt)|\big]  
        <
            3\trunc
        $
    since $\trunc>0$ by definition.
    Then,
        \begin{equation}\label{eq:trunc}
        \begin{aligned}[b]
            \Pr\Big(
                    \TruncT
            \geq 3\trunc
            \Big)
        & =
            \Pr\Bigg(
                    \sup_{f\in\mathcal{H}_n}
                    \sampavg 
                    \Big\{
                    \E
                    \big[ 
                        \difc{f}(\Zt)
                    \big] 
                    - 
                        \difc{f}(\Zt)
                    \Big\}
            \geq 3\trunc
            \Bigg)
        \\
        &
        \leq
            \Pr\Bigg(
                3\trunc
                +
                \sup_{f\in\mathcal{H}_n}
                \sampavg
                \big|\difc{f}(\Zt)\big|
            > 3\trunc
            \Bigg)
        \\&
        =
            \Pr\Bigg(
                \sup_{f\in\mathcal{H}_n}
                \sampavg
                \big|\difc{f}(\Zt)\big|
                >0
            \Bigg)
        \\&
        \leq
            \P(\Ptrunc)
        .
        \end{aligned}
        \end{equation}

\subsubsection{Bias term}\label{sec:Bias}
    From the block definitions \eqref{eq:block_def}, we have 
        $
        \data
        =
        \big\{
        \{\Zt\}_{t=1}^{2\anA\anB}
        ,\,
        \{\Zt\}_{t\in\IR}
        \big\}
        .
        $
    With this
    and \eqref{eq:difcprob}, for any  $A_1, A_2 >0$, 
        \begin{equation}\label{eq:A2_init1}
        \resizebox{\hsize}{!}{$
        \begin{aligned}[b]
            \P\left(
                \Critb(\fpr) - \Critb(\fO)
            \geq
                A_1 + A_2 
            \right)
        &
        =
            \P\left(
                \sampavg \big\{\dif{\fpr}(\Zt)+\difc{\fpr}(\Zt)\big\}
            \geq
                A_1 + A_2 
            \right)
        \\&
        \leq
            \P\left(
                \sampavg \dif{\fpr}(\Zt)
            \geq
                A_1 + A_2 
            \right)
            +
            \Pr\Bigg(
                \sampavg
                \big|\difc{\fpr}(\Zt)\big|
                >0
            \Bigg)
        \\&
        \leq
            \P\left(
                \sampavg \dif{\fpr}(\Zt)
            \geq
                A_1 + A_2 
            \right)
            +
            \P(\Ptrunc)
        \\&
        \leq
            \P\left(
                \frac{1}{2\anA\anB}
                \sum_{t=1}^{2\anA\anB} 
                \dif{\fpr}(\Zt)
            \geq
                A_1 
            \!
            \right)
            +
            \P\left(
                \frac{1}{n} 
                \sum_{t\in\IR}
                \dif{\fpr}(\Zt)
            \geq
                A_2
            \!
            \right)
            +
            \P(\Ptrunc)
        .
        \end{aligned}
        $}
        \end{equation}
    where the first term in the last line uses 
        $n\geq 2\anA\anB$.
    
        Consider the first term of \eqref{eq:A2_init1}.
        By assumption 
        $
        n\betaBound(\anA)
        /\anA
        < 1
        ,
        $
        and 
        $1\leq\anA\leq n/2$.
        In addition, 
        by Condition \ref{as:lip} and the definition of $\prjerr$,
        \begin{equation}\label{eq:diffpr}
        \begin{aligned}
            \sup_{t\in\N}\abs{\dif{\fpr}(\Zt)}
        =
            \sup_{t\in\N}\big|\critt(\fpr)-\critt(\fO)\big|\one
        \leq
            \Mbound
            \snormDNNBound{\fpr-\fO}
        \leq
            \Mbound \prjerr
        .
        \end{aligned}
        \end{equation}
    Then, by Lemma \ref{lem:Bernstein_GenMix}
        \begin{equation*}
        \resizebox{\hsize}{!}{$
        \begin{aligned}
            \frac{n\betaBound(\anA)}{\anA}
        &
        \geq
            \P\left(
                \frac{1}{2\anA\anB}
                \sum_{t=1}^{2\anA\anB} 
                \Big\{
                \dif{\fpr}(\Zt) - \E\big[\dif{\fpr}(\Zt)\big]
                \Big\}
            \geq
                4\Mbound \prjerr
                \left[
                \sqrt{
                    \frac{\anA}{n}
                    \logp{
                        \frac{1}{\betaBound(\anA)}
                    }
                }
                +
                \frac{
                    4\anA
                }
                {3n} 
                \logp{
                    \frac{1}{\betaBound(\anA)}
                }
                \right]
            \right)
         \\&
        \geq
            \P\left(
                \frac{1}{2\anA\anB}
                \sum_{t=1}^{2\anA\anB} 
                \Big\{
                \dif{\fpr}(\Zt) - \E\big[\dif{\fpr}(\Zt)\big]
                \Big\}
            \geq
                \frac{\Mbound\Fbound\anA\tempB}{n}
            \right)
        ,
        \end{aligned}
        $}
        \end{equation*}
    where the last line follows because 
        \begin{equation*}
        \begin{aligned}
         &
                \frac{\Mbound\Fbound\anA\tempB}{n}
            \geq
                4\Mbound \prjerr
                \left[
                \sqrt{
                    \frac{\anA}{n}
                    \logp{
                        \frac{1}{\betaBound(\anA)}
                    }
                }
                +
                \frac{
                    4\anA
                }
                {3n} 
                \logp{
                    \frac{1}{\betaBound(\anA)}
                }
                \right]
        \\&
        \iff
        \;\;
                \tempB
            \geq
                \frac{4 \prjerr}{\Fbound}
                \left[
                \sqrt{
                    \frac{n}{\anA}
                    \logp{
                        \frac{1}{\betaBound(\anA)}
                    }
                }
                +
                \frac{4}{3} 
                \logp{
                    \frac{1}{\betaBound(\anA)}
                }
                \right]
        ,
        \end{aligned}
        \end{equation*}
    which holds by  the assumption on $\tempB$ in the statement of the theorem.
    Note that 
        $\E\big[|\difc{\fpr}(\Zt)|\big]\leq 2\trunc$
    by \eqref{eq:difctrunc},
    and
        $
            \E
            \big[
            \critt(\fpr)-\critt(\fO)
            \big]
        =
            \E\big[\dif{\fpr}(\Zt)\big]
            +
            \E\big[\difc{\fpr}(\Zt)\big]
        .
        $
    Then,
        \begin{equation*}
        \begin{aligned}
            \frac{1}{2\anA\anB}
            \sum_{t=1}^{2\anA\anB} 
            \E\big[\dif{\fpr}(\Zt)\big]
        &
        \leq
            \frac{1}{2\anA\anB}
            \sum_{t=1}^{2\anA\anB} 
            \left|
                \E
                \big[
                \critt(\fpr)-\critt(\fO)
                \big]
            -
                \E\big[\difc{\fpr}(\Zt)\big]
            \right|
        \\&
        \leq
                \frac{1}{2\anA\anB}
            \sum_{t=1}^{2\anA\anB} 
                \E
                \big[
                \critt(\fpr)-\critt(\fO)
                \big]
            + 2\trunc
        ,
        \end{aligned}
        \end{equation*}
    since 
        $
            \E
            \big[
            \critt(\fpr)-\critt(\fO)
            \big]
        \geq
            0
        $
    by Condition \ref{as:curv}.
    With this, by Condition \ref{as:curv},
        \begin{equation*}
        \begin{aligned}
        &
            \frac{1}{2\anA\anB}
            \sum_{t=1}^{2\anA\anB} 
            \E\big[\dif{\fpr}(\Zt)\big]
        \leq
            \frac{1}{2\anA\anB}
            \sum_{t=1}^{2\anA\anB} 
            \CpcqB
            \E\Big[ 
                |\fpr(\Xt)-\fO(\Xt)|^2
            \Big]
            + 2\trunc
        \\&
        \leq
            \CpcqB
            \snormDNNBound{\fpr-\fO}^2  
            +
            \frac{1}{2\anA\anB}
            \sum_{t=1}^{2\anA\anB} 
            \CpcqB
            \E\Big[ 
                |\fpr(\Xt)-\fO(\Xt)|^2
                \mathbbm{1}_{\{\Xt \notin \Xbound\}}
            \Big]
            +
            2\trunc
        \\&
        \leq
            \CpcqB (\prjerr^2 + 2\truncf)
            +
            2\trunc
        ,
        \end{aligned}
        \end{equation*}
    where 
    the last inequality follows from the definitions of $\prjerr$ and $\truncf$. 
    Thus, 
        \begin{equation}\label{eq:A2_init2}
        \begin{aligned}
            \frac{n\betaBound(\anA)}{\anA}
        \geq
            \P\left(
                \frac{1}{2\anA\anB}
                \sum_{t=1}^{2\anA\anB} 
                \dif{\fpr}(\Zt)
            \geq
                \frac{\Mbound\Fbound\anA\tempB}{n}
                +
                \CpcqB (\prjerr^2 + 2\truncf)
                +
                2\trunc
            \right)
        .
        \end{aligned}
        \end{equation}

    Consider the second term of \eqref{eq:A2_init1}. Denote the cardinality of $\IR$ as 
        $(\#\IR) =  n-2\anA\anB$,
    and note 
        $ (\#\IR)  
        <  2\anA $,
    since
        $\anB \coloneqq \big\lfloor n/(2\anA)\big\rfloor$
    implies
        $\anB>n/(2\anA)-1$.
    In addition, by Condition \ref{as:Env}
        $$
            \prjerr
        \coloneqq
            \snormDNNBound{\fpr-\fO}
        \leq 
            2 \Fbound
        $$
    With this, and \eqref{eq:diffpr}, for any $\tempA > \max\{\tempB, 1\}$
    \begin{equation*}
    \begin{aligned}
                \frac{1}{n} 
                \sum_{t\in\IR}
                \dif{\fpr}(\Zt)
            \leq
                \left( 
                    \frac{2 \anA}{n} 
                \right)
                \Mbound  \prjerr
            <
                \frac{4 \Mbound\Fbound\anA\tempA}{n}
            ,
    \end{aligned}
    \end{equation*}
    which implies
        \begin{equation*}
        \begin{aligned}
            \P\left(
                    \frac{1}{n} 
                    \sum_{t\in\IR}
                    \dif{\fpr}(\Zt)
                \geq
                    \frac{4 \Mbound\Fbound\anA\tempA}{n}
                \right)
                =0
                .
        \end{aligned}
        \end{equation*}
    Applying the previous display and \eqref{eq:A2_init2} to \eqref{eq:A2_init1}, for any $\tempA > \max\{\tempB, 1\}$
        \begin{equation}\label{eq:A2}
        \begin{aligned}[b]
            \P\!\left(\!
                \Critb(\fpr) - \Critb(\fO)
            \geq
                \frac{5 \Mbound\Fbound\anA\tempA}{n}
                +
                \CpcqB (\prjerr^2 + 2\truncf)
                +
                2\trunc
            \!\right)
        &
        \leq
            \frac{n\betaBound(\anA)}{\anA}
            +
            \P(\Ptrunc)
        .
        \end{aligned}
        \end{equation}


\subsection{Localization analysis}

    The remainder of the proof will employ the localization analysis of \citet{farrell_deep_2021}. 
    Step I shows that if a function $f$ is ``local" to $\fO$, i.e., within a neighborhood of $\fO$ w.r.t.\! an $\Lp{2}$-norm, then it is also within a neighborhood with respect to an empirical norm with high probability. 
    Step II applies this to obtain a local bound on the empirical process term. 
    Step III shows that if $\f$ is within a large neighborhood of $\fO$ then
    the radius of this neighborhood can be tightened to show $\f$ is within a smaller neighborhood of $\fO$ with high probability.
    Step IV finds a critical radius which is a lower bound for radii such that the Step III tightening can be applied repeatedly to obtain.
    Step V obtains a global bound for $\norm*{\f-\fO}_{\Lp{2}}$ by applying Step III iteratively down to the critical radius from Step IV.
    Step VI uses steps I and V to obtain a bound on the empirical error.

\subsubsection{Step I: Quadratic process bound}
    Suppose for some $r>0$ and $\tempA > 0$ we have
        \begin{equation}\label{eq:A4}
        \begin{aligned}
            r^2
        \geq
            12
             \Fbound
            \sqrt{2\anA}\,
            \max_{i=1,2}
            \left(\!
            \E\left[
                \RadX{\anA\anB}
                \left\{
                    f-\fO:
                    f\in\FMLPg,
                    \normLpTi{i}{f-\fO}
                    \leq r
                \right\}
            \right] 
            \!\right)
        ,
        \end{aligned}
        \end{equation}
    and
        \begin{equation}\label{eq:A5}
        \begin{aligned}
            r^2
        \geq
            \frac{55\,\anA \Fbound^2\tempA}{n}
        .
        \end{aligned}
        \end{equation}
    Such an $r$ exists since 
        $
            \RadX{\anA\anB}
            \left\{
                f-\fO:
                f\in\FMLPg
            \right\}
        \leq
            2 \Fbound
        ,
        $
    follows from 
        $
        \sup_{f\in\FMLPg}\snormDNNBound{f-\fO}
            \leq
                2 \Fbound
        ,
        $
    by Condition \ref{as:Env}.
    Thus, 
    any 
        $
        r^2\!
        \geq 
        \max\!\Big\{\!
            24 \Fbound^2\sqrt{2\anA}
        ,
            55\anA \Fbound^2\tempA/n
        \Big\}
        $
    will always suffice 
    as an initial radius satisfying \eqref{eq:A4} and \eqref{eq:A5}.
    
    For 
    arbitrary fixed 
    $i\in\{1,2\}$, this step will show that if \eqref{eq:A4} and \eqref{eq:A5} hold, then 
    $
            \normTi{i}{f-\fO}
        \leq 
            2r
        $
    with probability greater than $1-e^{-\tempA}$
    for any $f$ such that
        \begin{equation}\label{eq:localf}
            f\in
        \left\{f\in\FMLPg:
        \normLpTi{i}{f-\fO}
        \leq r\right\}
        .
        \end{equation}
    First, we verify the conditions for  Proposition \ref{prop:bartsym1}. 
    By \eqref{eq:exp_emp},
        \begin{equation}\label{eq:step1ev0}
            \normTi{i}{f-\fO}^2
            -
            \normLpTi{i}{f-\fO}^2
        =
            \frac{1}{\anB}
            \sum_{j=1}^{\anB}
            \left\{
                \normTij{i}{f-\fO}{j}^2
                -
                \E\big[\normTij{i}{f-\fO}{j}^2\big]
            \right\}
        .
        \end{equation}
    For all $j\in\{1,\ldots,\anB\}$, and any $f\in\FMLPg$ by Condition \ref{as:Env}
        $
            \normTij{i}{f-\fO}{j}^2
        \leq
            4 \Fbound^2
        .$
    In addition, for $f$ satisfying \eqref{eq:localf}, by \eqref{eq:exp_emp}
        $$
            \E
            \Big[\normTi{i}{f-\fO}^4\Big]
        \leq
            (2\Fbound)^2
            \normLpTi{i}{f-\fO}^2
        \leq
            4 \Fbound^2 r^2
        .
        $$
    Recall from Subsection \ref{sec:IndBlock} that 
        $
            \{\IZt\}_{t\in\Ii{i}{1}},
            \{\IZt\}_{t\in\Ii{i}{2}},
            \ldots,
            \{\IZt\}_{t\in\Ii{i}{\anB}}
        $
    is a sequence of independent blocks.
    Consequently, so is $\big\{\normTij{i}{f-\fO}{j}\big\}_{j=1}^{\anB}$.
    By Remark \ref{rem:meas}, $\FMLPg$ is pointwise-separable.
    Then, by Proposition \ref{prop:bartsym1}
    for any $\tempA>0$ 
        \begin{equation}\label{eq:A3prelim}
        \begin{aligned}
        e^{-\tempA} 
        \geq
        \P\Bigg(
        &
            \sup_{\{f\in\FMLPg:\normLpTi{i}{f-\fO}\leq r\}}
            \frac{1}{\anB}
            \sum_{j=1}^{\anB}
            \left\{
                \normTij{i}{f-\fO}{j}^2
                -
                \E\big[\normTij{i}{f-\fO}{j}^2\big]
            \right\}
        \\&
        \quad
        \geq
            3
            \E\left[
                \RadX{\anB}
                \left\{
                    \normTij{i}{f-\fO}{j}^2:
                    f\in\FMLPg,
                    \normLpTi{i}{f-\fO}\leq r
                \right\}
            \right] 
        \\&
        \quad
            +
            2 \Fbound r
            \sqrt{
                \frac{2\tempA}{\anB}
            }
            +
            \frac{20 \Fbound^2\tempA}{\anB}
            \Bigg)
        .
        \end{aligned}
        \end{equation}
    By $a^2-b^2=(a+b)(a-b)$
    and the reverse triangle inequality 
        \begin{equation*}
        \begin{aligned}
                (f-\fO)^2
                -
                (f'-\fO)^2
        &=
            \Big(
                (f-\fO)
                +
                (f'-\fO)
            \Big)
            \cdot
            \Big(
                (f-\fO)
                -
                (f'-\fO)
            \Big)
        \\&
        \leq
            4 \Fbound
            \Big|
                (f-\fO)
                -
                (f'-\fO)
            \Big|
        .
        \end{aligned}
        \end{equation*}
    Then, by Lemma \ref{lem:Rad_Contraction}
        \begin{equation*}
        \resizebox{\hsize}{!}{$
        \begin{aligned}
            \E\bigg[
        &
                \RadX{\anB}
                \left\{
                    \normTij{i}{f-\fO}{j}^2:
                    f\in\FMLPg,
                    \normLpTi{i}{f-\fO}\leq r
                \right\}
            \bigg] 
        \\&
        =
            \frac{1}{\anB}
            \E
            \left[
                \sup\left\{
                    \sum_{j=1}^{\anB}
                    \xi_j\cdot
                    \bigg(
                        \frac{1}{\anA}
                        \sum_{t\in\Ii{i}{j}}
                        \big(f(\Xt)-\fO(\Xt)\big)^2
                        \onext
                    \bigg)
                :f\in\FMLPg, \normLpTi{i}{f-\fO}\leq r
                \right\}
            \right]
        \\&
        \leq
            \frac{4 \Fbound\sqrt{2/\anA}}{\anB}
            \E
            \left[
                \sup\left\{
                    \sum_{j=1}^{\anB}
                    \sum_{t\in\Ii{i}{j}}
                    \xi_t
                    \big(f(\Xt)-\fO(\Xt)\big)
                    \onext
                :f\in\FMLPg, \normLpTi{i}{f-\fO}\leq r
                \right\}
            \right]
        \\&
        =
            4 \Fbound\sqrt{2\anA}\,
            \E\bigg[
                \RadX{\anA\anB}
                \left\{
                    f-\fO:
                    f\in\FMLPg,
                    \normLpTi{i}{f-\fO}\leq r
                \right\}
            \bigg] 
        \end{aligned}
        $}
        \end{equation*}
    Applying this Rademacher complexity bound, 
    \eqref{eq:step1ev0},
    and 
        $\anB \coloneqq \big\lfloor n/(2\anA)\big\rfloor>n/(4\anA)$,
    to 
        \eqref{eq:A3prelim} 
        \begin{equation}\label{eq:A3}
        \begin{aligned}
        \hspace{-.4cm}e^{-\tempA} 
        \geq
        \P\Bigg(&
            \sup_{\{f\in\FMLPg:\normLpTi{i}{f-\fO}\leq r\}}
            \Big\{
                \normTi{i}{f-\fO}^2
                -
                \normLpTi{i}{f-\fO}^2
            \Big\}
        \\&
        \quad
        \geq
            12 \Fbound\sqrt{2\anA}\,
            \E\bigg[
                \RadX{\anA\anB}
                \left\{
                    f-\fO:
                    f\in\FMLPg,
                    \normLpTi{i}{f-\fO}\leq r
                \right\}
            \bigg] 
        \\&
        \qquad
            +
            r
            \sqrt{
                \frac{16\,\anA \Fbound^2\tempA}{n}
            }
            +
            \frac{80\,\anA \Fbound^2\tempA}{n}
            \Bigg)
        .
        \end{aligned}
        \end{equation}
    Note that if \eqref{eq:A5} holds then
        $
            2r^2
        \geq 
            r
            \sqrt{
                \tfrac
                {16\,\anA \Fbound^2\tempA}{n}
            }
            +
            \tfrac
            {80\,\anA \Fbound^2\tempA}{n}
        .
        $
    Therefore,
    \eqref{eq:A3}
    implies that
    for all $r, \tempA >0$ such that \eqref{eq:A4} and \eqref{eq:A5} hold
        \begin{equation}\label{eq:A6}
        \begin{aligned}[b]
            e^{-\tempA}
        &\geq
            \Pr\Bigg(
                \sup_{\{f\in\FMLPg:\normLpTi{i}{f-\fO}\leq r\}}
                \Big\{
                    \normTi{i}{f-\fO}^2 
                    -
                    \normLpTi{i}{f-\fO}^2
                \Big\}
            \geq
                3r^2
            \Bigg)
        \\&
        \geq
            \Pr\Bigg(
                \sup_{\{f\in\FMLPg:\normLpTi{i}{f-\fO}\leq r\}}
                \normTi{i}{f-\fO}^2 
            \geq
                4r^2
            \Bigg)
        \\&
        =
            \Pr\Bigg(
                \sup_{\{f\in\FMLPg:\normLpTi{i}{f-\fO}\leq r\}}
                \normTi{i}{f-\fO}
            \geq
                2r
            \Bigg)
        .
        \end{aligned}
        \end{equation}

\subsubsection{Step II: Local empirical process term bound}\label{sec:step2}
\newcommand{\rO}{r}%
\newcommand{\eventfr}{E_{f,\rO}}%
\newcommand{\BlockIZ}[3]{{G}^{*(#1)}_{#2,#3}}%
\newcommand{\BlockZ}[3]{G^{(#1)}_{#2,#3}}%
\newcommand{\supsubset}{\sup_{\!\{\!f\in\FMLPg:\normLpTi{i}{f-\fO}\leq 2\rO\!\}\!}}%
\newcommand{\Diam}{D_{n}}%
\newcommand{\Ptermi}{\P_i}
\newcommand{\PtermR}{\P_R}
    Suppose $\rO>0$, $\tempA\geq 1$ are
    such that \eqref{eq:A4} and \eqref{eq:A5} hold for all $r'\geq \rO$. 
    This step will show that 
    if
        $
        \Pr\big(
        \normLpnS{\f-\fO}
        \leq 
        \rO
        \big)>0
        ,
        $
    then $\rO$ can be used to obtain probability bound on the empirical process term from \eqref{eq:A1}.

    For 
        $i\in\{1,2\}$, 
    and
        $j\in\{1,...,\anB\}$,
    define
        $$
            \BlockZ{i}{j}{f}
        \coloneqq 
            \frac{1}{\anA}
            \sum_{t\in \Ii{i}{j}}
            \dif{f}(\Zt)
        ,
        \quad \text{ and } \quad
            \BlockIZ{i}{j}{f}
        \coloneqq 
            \frac{1}{\anA}
            \sum_{t\in \Ii{i}{j} }
            \dif{f}(\IZt)
        .
        $$
    Then, the empirical process term from \eqref{eq:A1} can be decomposed using
        \begin{equation*}
        \begin{aligned}
        &
            \sampavg 
            \left\{
                \E
                \big[ 
                    \dif{f}(\Zt)
                \big] 
                - 
                \dif{f}(\Zt)
            \right\}
        \\&
        \quad
        =
            \left(\frac{\anA\anB}{n}\right)
            \frac{1}{\anB}
            \sum_{j=1}^{\anB}
            \left\{
                \E\big[\BlockZ{1}{j}{f}\big]
                -
                \BlockZ{1}{j}{f}
                +
                \E\big[\BlockZ{2}{j}{f}\big]
                -
                \BlockZ{2}{j}{f}
            \right\}
        +
            \frac{1}{n}
            \sum_{t\in\IR}
            \left\{
                \E
                \big[ 
                    \dif{f}(\Zt)
                \big] 
                - 
                \dif{f}(\Zt)
            \right\}
        .
        \end{aligned}   
        \end{equation*}
    Thus, for 
        $A_1,A_2>0$ 
    to be specified later,
        \begin{equation}\label{eq:emp_proc_prob}
        \begin{aligned}[b]
            &\Pr\!\left(
                \left\{
                    \EPT
                \geq
                    2A_1 + A_2
                \right\}
            \cap
                \left\{
                    \normLpnS{\f-\fO}
                \leq 
                    \rO
                \right\}
            \right)
        \leq
            2\max_{i\in\{1,2\}}\Ptermi + \PtermR
        ,
        \end{aligned}
        \end{equation}
    where 
        \begin{equation*}
        \begin{aligned}
            \Ptermi
        &
        \coloneqq
            \Pr\!\left(\!
                \left\{
                    \sup_{f\in\fsub}
                    \left(\!\frac{\anA\anB}{n}\!\right)\!
                    \frac{1}{\anB}
                    \sum_{j=1}^{\anB}\!
                    \left\{
                        \E\big[\BlockZ{i}{j}{f}\big]
                        -
                        \BlockZ{i}{j}{f}
                    \right\}\!
                \geq
                    A_1\!
                \right\}
            \cap
                \left\{
                    \normLpnS{\f-\fO}
                \leq 
                    \rO
                \right\}
            \right)
        ,
        \\
            \PtermR
        &
        \coloneqq
            \Pr\!\left(
                \sup_{f\in\fsub}
                \frac{1}{n}
                \sum_{t\in\IR}\!
                \left\{
                    \E
                    \big[ 
                        \dif{f}(\Zt)
                    \big] 
                    - 
                    \dif{f}(\Zt)\!
                \right\}\!
            \geq
                A_2
            \right)
        \end{aligned}
        \end{equation*}

    Consider $\PtermR$. 
    By Conditions \ref{as:lip} and \ref{as:Env}, for any $f\in\FMLPg$
        \begin{equation}\label{eq:difbnd}
        \begin{aligned}
            \norm*{\dif{f}}_\infty
        =
            \norm*{(\critt(f)-\critt(\fO))\one}_\infty
        \leq
            \Mbound\snormDNNBound{f-\fO}
        \leq
            2 \Mbound\Fbound
        ,
        \end{aligned}
        \end{equation}
    hence
        $
            \big\lVert
            \E\big[\dif{f}(\Zt)\big] - \dif{f}(\Zt)
            \big\lVert_{\infty}
        \leq
            4 \Mbound\Fbound
        .
        $ 
    Denote the cardinality of $\IR$ as 
        $(\#\IR) =  n-2\anA\anB$,
    and note 
        $ (\#\IR)  
        <  2\anA $,
    since
        $\anB \coloneqq \big\lfloor n/(2\anA)\big\rfloor$
    implies
        $\anB>n/(2\anA)-1$.
    With this,
        \begin{equation*}
        \begin{aligned}
            \sup_{f\in\FMLPg}
                \frac{1}{n}
                \sum_{t\in\IR}
                \left\{
                    \E
                    \big[ 
                        \dif{f}(\Zt)
                    \big] 
                    - 
                    \dif{f}(\Zt)
                \right\}
        &
        \leq
                \sup_{f\in\FMLPg}
                \frac{2\anA}{n}\,
                \big\lVert
                    \E\big[\dif{f}(\Zt)\big] - \dif{f}(\Zt)
                \big\lVert_{\infty}
        \leq
                \frac{8 \Mbound\Fbound\anA}{n}
        .
        \end{aligned}
        \end{equation*}
    Thus,
        \begin{equation}\label{eq:A7_P2}
        \begin{aligned}
            A_2 \coloneqq \frac{9 \Mbound\Fbound\anA}{n}
        \quad 
        \implies
        \quad
            \PtermR
        \leq
            \Pr\!\left(
                \sup_{f\in\FMLPg}
                \frac{1}{n}
                \sum_{t\in\IR}\!
                \left\{
                    \E
                    \big[ 
                        \dif{f}(\Zt)
                    \big] 
                    - 
                    \dif{f}(\Zt)\!
                \right\}\!
            \geq
                A_2
            \right)
        =0
            .
        \end{aligned}
        \end{equation}

    Consider $\Ptermi$ for arbitrary $i\in\{1,2\}$.
    Note that
        $$
            \left\{
                \normLpnS{\f-\fO}
            \leq 
                \rO
            \right\}
        \;\;\implies\;\;
            \fsub
            \subseteq
            \left\{
                f\in\FMLPg:
                \normLpnS{f-\fO}
            \leq 
                \rO
            \right\}
        ,
        $$
    which implies
        \begin{equation*}
        \begin{aligned}
            \Ptermi
        &
        \leq
            \Pr\!\left(\!
                    \sup_{\{
                        f\in\FMLPg:
                            \normLpnS{f-\fO}
                        \leq 
                            \rO
                        \}}
                    \left(\!\frac{\anA\anB}{n}\!\right)\!
                    \frac{1}{\anB}
                    \sum_{j=1}^{\anB}\!
                    \left\{
                        \E\big[\BlockZ{i}{j}{f}\big]
                        -
                        \BlockZ{i}{j}{f}
                    \right\}\!
                \geq
                    A_1\!
            \right)
        .
        \end{aligned}
        \end{equation*}
    From the norm definitions \eqref{eq:normLpTi}, \eqref{eq:normLpnS}, and Lemma \ref{lem:avgmeas}
        \begin{equation}\label{eq:normLpTi_v_normLpnS}
        \begin{aligned}[b]
        \max_{i\in\{1,2\}}
        \normLpTi{i}{f}^2
        &
        \leq
            \int_{\x\in\Xbound}
                |f(\x)|^2
                d\Pblocki{1} (\x)
            +
            \int_{\x\in\Xbound}
                |f(\x)|^2
                d\Pblocki{2} (\x)
        \\&
        =
            \frac{1}{\anB\anA}
            \sum_{j=1}^{\anB}
            \left(
                \sum_{t\in \Ik{1}{j}} 
                \int_{\x\in\Xbound}
                    |f(\Xt)|^2
                    d\P_{\Xt}
            +
                \sum_{t\in \Ik{2}{j}} 
                \int_{\x\in\Xbound}
                    |f(\Xt)|^2
                    d\P_{\Xt}
            \right)
        \\&
        \leq
            \left(\frac{n}{\anB\anA}\right) 
            \cdot
            \sampavg
                \int_{\x\in\Xbound}
                    |f(\Xt)|^2
                    d\P_{\Xt}
        \leq
            4\normLpnS{f}^2
        ,
        \end{aligned}
        \end{equation}
    since 
        $n/(4\anA)
        <
            \anB$.
    Then,
        $
            \max_{i\in\{1,2\}}
            \normLpTi{i}{f-\fO}
        \leq
            2
            \normLpnS{f-\fO}
        $
    for any $f\in\FMLPg$,
    and 
        $$
            \left\{
                f\in\FMLPg:
                \normLpnS{f-\fO}
            \leq 
                \rO
            \right\}
        \subseteq
            \left\{
                f\in\FMLPg:
                \normLpTi{i}{f-\fO}
            \leq 
                2\rO
            \right\}
        .
        $$
    Thus,
        \begin{equation*}
        \begin{aligned}
            \Ptermi
        &
        \leq
            \Pr\!\left(\!
                \supsubset
                \left(\frac{\anA\anB}{n}\right)
                \frac{1}{\anB}
                \sum_{j=1}^{\anB}
                \left\{
                    \E\big[\BlockZ{i}{j}{f}\big]
                    -
                    \BlockZ{i}{j}{f}
                \right\}
            \geq
                A_1
            \right)
        .
        \end{aligned}
        \end{equation*}
    The set 
        $
        \big\{f\in\FMLPg:\normLpTi{i}{f-\fO}\leq 2\rO \big\}
        $
    is non-stochastic, so by 
    \eqref{eq:beta_bound2}
        \begin{equation}\label{eq:Poneinit}
        \begin{aligned}[b]
            \Ptermi
        &\leq
            \P
            \Bigg(
                \supsubset
                \left(\frac{\anA\anB}{n}\right)
                \frac{1}{\anB}
                \sum_{j=1}^{\anB}
                \left\{
                    \E\big[\BlockIZ{i}{j}{f}\big]
                    -
                    \BlockIZ{i}{j}{f}
                \right\}
            \geq
                A_1
            \Bigg)
        +
            \frac{n\,\beta(\anA)}{2\anA}
        \\&
        \leq
            \P
            \Bigg(
                \supsubset
                \frac{1}{\anB}
                \sum_{j=1}^{\anB}
                \left\{
                    \E\big[\BlockIZ{i}{j}{f}\big]
                    -
                    \BlockIZ{i}{j}{f}
                \right\}
            \geq
                2 A_1
            \Bigg)
        +
            \frac{n\,\beta(\anA)}{2\anA}
        ,        
        \end{aligned}
        \end{equation}
    since
        $\anB
        \leq
            n/(2\anA)
        .
        $
    We bound the probability term above
    with
    Proposition \ref{prop:bartsym1}.
    By \eqref{eq:difbnd}
        $
            \norm*{\dif{f}}_\infty
        \leq
            2 \Mbound\Fbound
        $
    for any $f\in\FMLPg$,
    so for any $f\in\big\{f\in\FMLPg:\normLpTi{i}{f-\fO}\leq \rO \big\}$
        \begin{equation*}
        \begin{aligned}
            \E
                \left[
                    \frac{1}{\anB}
                    \sum_{j=1}^{\anB}
                    \big(\BlockIZ{i}{j}{f}\big)^2
                \right]
            &
            \leq
                \E
                \left[
                    \frac{1}{\anB}
                    \sum_{j=1}^{\anB}
                    \bigg(
                    \frac{1}{\anA}
                    \sum_{t\in \Ii{i}{j}}
                    \Mbound
                    \big|f(\IXt)-\fO(\IXt)\big|
                    \one
                    \bigg)^2
                \right]
            \\&
            \leq
                \Mbound^2\,
                \E
                \left[
                    \frac{1}{\anA\anB}
                    \sum_{j=1}^{\anB}
                    \sum_{t\in \Ii{i}{j}}
                    \big|f(\Xt)-\fO(\Xt)\big|^2
                    \one
                \right]
            \\&
            =
                \Mbound^2\,
                \normLpTi{i}{f-\fO}^2
            \leq
                4
                \Mbound^2
                \,\rO^2
        \end{aligned}
        \end{equation*}
    where the 
    first inequality uses Condition \ref{as:lip}, and the
    second inequality uses 
    the Cauchy-Schwarz inequality and 
        $
            \P_{\{\IZt\}_{t\in\Ii{i}{j}}}
        =
            \P_{\{\Zt\}_{t\in\Ii{i}{j}}}
        $
    for any $j\in\{1,\ldots,\anB\}$.
    Then, since 
        $
            \{\IZt\}_{t\in\Ii{i}{1}},
        $ 
        $
            \{\IZt\}_{t\in\Ii{i}{2}},
        $ 
        $
            \ldots,
            \{\IZt\}_{t\in\Ii{i}{\anB}}
        $
    are independent blocks, we can apply Proposition \ref{prop:bartsym1}
    to obtain, 
        \begin{equation*}
        \begin{aligned}[b]
        1-2e^{-\tempA}
        &\leq
            \P\Bigg(
                \supsubset
                \frac{1}{\anB}
                    \sum_{j=1}^{\anB}
                    \Big\{
                        \E\big[\BlockIZ{i}{j}{f}\big]
                        -
                        \BlockIZ{i}{j}{f}
                    \Big\}
            \leq
        \\& \qquad\quad
                6
                \E_{\xi}
                \Big[
                    \RadZ{\anB}
                    \Big\{
                        \BlockIZ{i}{j}{f}:
                        f\in\FMLPg,
                        \normLpTi{i}{f-\fO}\leq 2\rO
                    \Big\}
                \Big] 
                +
                14
                \Mbound
                \rO
                \sqrt{
                    \frac{2\tempA}{\anB}
                }
                +
                \frac{58 \Mbound\Fbound\tempA}{\anB}
            \Bigg)
        \\&
        \leq
            \P\Bigg(
                \supsubset
                \frac{1}{\anB}
                    \sum_{j=1}^{\anB}
                    \Big\{
                        \E\big[\BlockIZ{i}{j}{f}\big]
                        -
                        \BlockIZ{i}{j}{f}
                    \Big\}
            \leq
        \\&
        \qquad \quad
                6
                \E_{\xi}
                \Big[
                    \RadZ{\anB}
                    \Big\{
                        \BlockIZ{i}{j}{f}:
                        f\in\FMLPg,
                        \normTi{i}{f-\fO} \leq 4\rO
                    \Big\}
                \Big] 
                +
                14
                \Mbound
                \rO
                \sqrt{
                    \frac{2\tempA}{\anB}
                }
                +
                \frac{58 \Mbound\Fbound\tempA}{\anB}
            \Bigg)
        \\&
        \quad
        +
            \Pr\Bigg(
                \supsubset
                \normTi{i}{f-\fO}
            >
                4\rO
            \Bigg)
        .
        \end{aligned}
        \end{equation*}
    By the assumptions on $\rO$ in this step, 
    \eqref{eq:A6}
    can be applied 
        \begin{equation}\label{eq:A7_P1_pre}
        \resizebox{0.9\hsize}{!}{$
        \begin{aligned}[b]
            1-
            3e^{-\tempA}
        &\leq
            \P\Bigg(
                \supsubset
                \frac{1}{\anB}
                    \sum_{j=1}^{\anB}
                    \Big\{
                        \E\big[\BlockIZ{i}{j}{f}\big]
                        -
                        \BlockIZ{i}{j}{f}
                    \Big\}
            \leq
        \\&
        \qquad \quad
                6
                \E_{\xi}
                \Big[
                    \RadZ{\anB}
                    \Big\{
                        \BlockIZ{i}{j}{f}:
                        f\in\FMLPg,
                        \normTi{i}{f-\fO} \leq 4\rO
                    \Big\}
                \Big] 
                +
                14
                \Mbound
                \rO
                \sqrt{
                    \frac{2\tempA}{\anB}
                }
                +
                \frac{58 \Mbound\Fbound\tempA}{\anB}
            \Bigg)
        .
        \end{aligned}
        $}        
        \end{equation}
    Now we address the Rademacher complexity term.
    For any $f,f'\in\FMLPg$, by 
    Condition \ref{as:lip}
        \begin{equation*}
        \begin{aligned}
                \big|
                \dif{f}(\IZt)
                -
                \dif{f'}(\IZt)
                \big|
        &=
                \big|
                \critb(\IZt,f)
                -
                \critb(\IZt,f')
                \big|\mathbbm{1}_{\{\IZt \in \Zbound\}}
        \leq
            \Mbound
            \big|f(\IXt)-f'(\IXt)\big|\mathbbm{1}_{\{\IXt \in \Xbound\}}
        \\&
        =
            \Mbound
            \Big|\big(f(\IXt)-\fO(\IXt)\big)-\big(f'(\IXt)-\fO(\IXt)\big)\Big|\mathbbm{1}_{\{\IXt \in \Xbound\}}
        ,
        \end{aligned}
        \end{equation*}
    so we can apply Lemma \ref{lem:Rad_Contraction} to obtain
        \begin{equation}\label{eq:RadBound_init}
        \resizebox{.9\hsize}{!}{$
        \begin{aligned}[b]
        &
            6\E_{\xi}
            \bigg[
                \RadZ{\anB}
                \bigg\{
                    \BlockIZ{i}{j}{f}:
                    f\in\FMLPg,
                    \normTi{i}{f-\fO} \leq 4\rO
                \bigg\}
            \bigg]
        \\&
        \leq
            \frac{6\Mbound\sqrt{2}}{\anB\sqrt{\anA}}
            \E_{\xi}
            \left[
                \sup\left\{
                    \sum_{j=1}^\anB
                    \sum_{t\in\Ii{i}{j}}
                    \xi_{j,t}\,
                    \big(f(\IXt)-\fO(\IXt)\big)
                    \onext
                :f\in\FMLPg, 
                \normTi{i}{f-\fO} \leq 4\rO
                \right\}
            \right]
        \\&
        =
            6\Mbound\,\sqrt{2\anA}\,
            \E_{\xi}
            \Big[
                \RadX{\anA\anB}
                \Big\{
                    f-\fO:
                    f\in\FMLPg,
                    \normTi{i}{f-\fO} \leq 4\rO
                \Big\}
            \Big] 
        .
        \end{aligned}
        $}
        \end{equation}
    We bound \eqref{eq:RadBound_init} 
    with Proposition \ref{prop:rad_dudley_bound}. 
    To apply this, first note that
        $\normTi{i}{\cdot}$ 
    is a sum of $\anA\anB$ terms,
    and 
        $
        n/4 < \anA\anB
        $
    since 
        $
            n/(4\anA)
        <
            \anB
        $
    by construction.
    Then, 
        $\anA\anB\geq\Pdim(\FMLPg)$
    since 
        $
        n/4
        >\Pdim(\FMLPg)
        $
    by assumption.
    Let
        $
            \Diam
        \coloneqq
            \min
            \left\{
                2 \Fbound
                ,\,
                4\rO
            \right\}
        ,
        $
    and recall 
        $
        \sup_{f\in\FMLPg}\snormDNNBound{f-\fO} \leq 2 \Fbound
        $,
    so 
        $$
            \big\{
                f\in\FMLPg:
                \normTi{i}{f-\fO} \leq 2\rO
            \big\}
        =
            \big\{
                f\in\FMLPg:
                \normTi{i}{f-\fO} \leq \Diam
            \big\}
        .
        $$
    Clearly, $\Diam\leq 2 \Fbound < 4 \Fbound$, so Proposition \ref{prop:rad_dudley_bound} can be applied 
        \begin{equation}\label{eq:RadBound0}
        \begin{aligned}[b]
            \E_{\xi}
            \Big[
                \RadX{\anA\anB}
                \Big\{
                    f-\fO:
            &\,
                    f\in\FMLPg,
                    \normTi{i}{f-\fO} \leq 4\rO
                \Big\}
            \Big] 
        \\&
        =
            \E_{\xi}
            \Big[
                \RadX{\anA\anB}
                \Big\{
                    f-\fO:
                    f\in\FMLPg,
                    \normTi{i}{f-\fO} \leq \Diam
                \Big\}
            \Big] 
        \\&
        \leq
            12\Diam
                \sqrt{
                    \frac{ 2\,\Pdim(\FMLPg)}{\anA\anB}
                      \cdot
                      \logp{
                            \frac%
                            {4e^2 \Fbound \anA\anB}%
                            {\Diam \cdot \Pdim(\FMLPg)}
                        }
                }
        \\&
        \leq
            24\Diam
                \sqrt{
                    \frac{ 2\,\Pdim(\FMLPg)}{n}
                      \cdot
                      \logp{
                            \frac%
                            {2e^2 \Fbound n}%
                            {\Diam \cdot \Pdim(\FMLPg)}
                        }
                }
        ,
        \end{aligned}
        \end{equation}
    where the last inequality uses
        $
        n/4 < \anA\anB \leq n/2
        $
    since 
        $
            n/(4\anA)
        <
            \anB
        \leq
            n/(2\anA)
        .
        $
    Note that
        $$
            4\rO
        \geq
            4\sqrt{\frac{55\,\anA \Fbound^2\tempA}{n}}
        >
            \frac{2e^2 \Fbound}{n}
        ,
        \quad
        \text{ and }
        \quad
            2 \Fbound
        >
            \frac{2e^2 \Fbound}{n}
        ,
        $$
    where the $4\rO$ bound follows from 
        \eqref{eq:A5}, with $\anA, \tempA\geq 1$ and $4\sqrt{55}>2e^2$;
    and the $2 \Fbound$ bound follows from 
        $e^2/n<1$,
    because
        $n\geq 8 $ 
    by assumption.
    Thus,
        $
                \Diam 
            > 
                2e^2 \Fbound/n
            ,
        $
    which implies
        $$
        \logp{
                \frac%
                {2e^2 \Fbound n}%
                {\Diam \Pdim(\FMLPg)}
            }
        <
            \log(n)
        ,
        $$
    since 
        $\Pdim(\FMLPg)\geq 1$
    by definition.
    With this, \eqref{eq:RadBound0},
    and since $\Diam \leq 4\rO$ by definition,
        \begin{equation}\label{eq:RadBound1}
        \begin{aligned}[b]
        \E_{\xi}&
            \Big[
                \RadX{\anA\anB}
                \Big\{
                    f-\fO:
                    f\in\FMLPg,
                    \normTi{i}{f-\fO} \leq 4\rO
                \Big\}
            \Big] 
        \leq
                96\rO
                \sqrt{
                    \frac{ 2\Pdim(\FMLPg)}{n}
                      \logp{n}
                }
        .
        \end{aligned}
        \end{equation}
    Combining \eqref{eq:RadBound_init} and \eqref{eq:RadBound1},
        \begin{equation*}
        \begin{aligned}
        &
            \E_{\xi}
            \bigg[
                \RadX{\anB}
                \bigg\{
                    \BlockIZ{i}{j}{f}:
                    f\in\FMLPg,
                    \normTi{i}{f-\fO} \leq 2\rO
                \bigg\}
            \bigg]
        \leq
            \rO\,
            (6\cdot 96\cdot 2)
            \Mbound
            \sqrt{
                    \frac{ \anA\Pdim(\FMLPg)}{n}
                    \logp{ n}
                }
        .
        \end{aligned}
        \end{equation*}
    With this, \eqref{eq:A7_P1_pre}, 
    and
        $
        \textstyle
            \anB
        >
            n/(4\anA),
        $
        \begin{equation*}
        \begin{aligned}[b]
            3e^{-\tempA}
        &\geq
            \P\Bigg(
                \supsubset
                \frac{1}{\anB}
                    \sum_{j=1}^{\anB}
                    \Big\{
                        \E\big[\BlockIZ{i}{j}{f}\big]
                        -
                        \BlockIZ{i}{j}{f}
                    \Big\}
            \geq
        \\&
        \qquad \quad
                \rO\,
                1152\Mbound\,
                    \sqrt{
                        \frac{\anA \Pdim(\FMLPg)}{n}
                        \log(n)
                    }
                +
                14\rO\,\Mbound
                \sqrt{
                    \frac{2\tempA}{\anB}
                }
                +
                \frac{58 \Mbound\Fbound\tempA}{\anB}
            \Bigg)
        \\&
        \geq
            \P\Bigg(
                \supsubset
                \frac{1}{\anB}
                    \sum_{j=1}^{\anB}
                    \Big\{
                        \E\big[\BlockIZ{i}{j}{f}\big]
                        -
                        \BlockIZ{i}{j}{f}
                    \Big\}
            \geq
        \\&
        \qquad \quad
                \rO\,
                1152\Mbound\,
                    \sqrt{
                        \frac{\anA \Pdim(\FMLPg)}{n}
                        \log(n)
                    }
                +
                14\rO\, \Mbound
                \sqrt{
                    \frac{8\anA\tempA}{n}
                }
                +
                \frac{232 \Mbound\Fbound\anA\tempA}{n}
            \Bigg)
        .
        \end{aligned}
        \end{equation*}
    Since $i\in\{1,2\}$ is arbitrary, applying the above display to 
    \eqref{eq:Poneinit} gives
        \begin{equation}\label{eq:A7_P1}
        \begin{aligned}
        2\max_{i\in\{1,2\}}\Ptermi 
        &
        \leq
            2\left[
            \max_{i\in\{1,2\}}
            \P
            \Bigg(
                \supsubset
                \frac{1}{\anB}
                \sum_{j=1}^{\anB}
                \left\{
                    \E\big[\BlockIZ{i}{j}{f}\big]
                    -
                    \BlockIZ{i}{j}{f}
                \right\}
            \geq
                2 A_1
            \Bigg)
        +
            \frac{n\,\beta(\anA)}{2\anA}
        \right]
        \\&
        \leq
            6e^{-\tempA} + \frac{n\,\beta(\anA)}{\anA},
        \\
        \text{ where }
        &
        \quad
            2A_1 
        = 
            \rO\,
            1152\Mbound\,
                \sqrt{
                    \frac{\anA \Pdim(\FMLPg)}{n}
                    \log(n)
                }
            +
            14\rO\, \Mbound
            \sqrt{
                \frac{8\anA\tempA}{n}
            }
            +
            \frac{232 \Mbound\Fbound\anA\tempA}{n}
        .        
        \end{aligned}
        \end{equation}
    Thus, 
    we can update \eqref{eq:emp_proc_prob}
    using 
    \eqref{eq:A7_P2} and \eqref{eq:A7_P1}, 
    to obtain the following whenever \eqref{eq:A4}, and \eqref{eq:A5} hold for all $r'\geq \rO$
        \begin{equation}\label{eq:A8}
        \begin{aligned}
        &
            \Pr\Bigg(
            \left\{
                    \EPT
                \geq
                    \rO\cdot
                    \left(
                        \frac
                            {1152 \Mbound\, \sqrt{\anA}}
                            {\sqrt{n}}
                    \right)\!
                    \Big[
                            \sqrt{
                                \Pdim(\FMLPg)
                                \log(n)
                            }
                        +
                            \sqrt{\tempA}
                    \Big]
                    +
                    \frac{241 \Mbound\Fbound\anA\tempA}{n}
                \right\}
            \\&
            \qquad
                \cap
                \left\{
                    \normLpnS{\f-\fO}
                \leq 
                    \rO
                \right\}
                \Bigg)
        \\&
        \leq
        6e^{-\tempA}+\frac{n\,\beta(\anA)}{\anA}
        ,
        \end{aligned}
        \end{equation}
    since 
        $1152>14\sqrt{8}$,
    and
    $\tempA\geq 1$ 
    implies
    $232\tempA + 9 \leq 241\tempA$.

\subsubsection{Step III: Radius one-step tightening for \texorpdfstring{$\mathcal{L}^{2}_{|\Zbound\!}\!(\Pbarn\!)$}{L2}-error }\label{sec:step3}
{
\renewcommand{\rO}{r}
\newcommand{\rone}{r'}
    From the norm definition \eqref{eq:normLpnS},
    $
    \normLpnS{\f-\fO}^2
    \leq
    \normLpn{\f-\fO}^2
    ,
    $
    so 
    \eqref{eq:A1} also applies to 
    $\normLpnS{\f-\fO}^2$. 
    Therefore, if 
    $\tempA>\max\{\tempB,1\}$
    and $\rO$ are such that \eqref{eq:A4} and \eqref{eq:A5} hold for all $r''\geq r$, 
    then
    we may apply \eqref{eq:A1}, \eqref{eq:trunc}, \eqref{eq:A2}, 
    and \eqref{eq:A8} to obtain%
\footnote{
To see this, let $X$, $Y$, $Z$ be random variables, and let $y$, $z$, $c$ be constants. Suppose $\Pr(Y\geq y) \leq c$, 
$\Pr(Z\geq z) \leq c$,
and
$\Pr(X\leq Y + Z) = 1$.
Then, 
$
\Pr(X > y + z) 
\leq 
\Pr(Y + Z > y + z) 
\leq 
\Pr(Y > y)
+
\Pr(Z > z)
\leq 2c
$.
}
        \begin{equation}\label{eq:A9}
        \begin{aligned}[b]
        &
            \Pr
            \Bigg(
            \Bigg\{
                \Cpcq 
                \normLpnS{\f-\fO}^2
            >
                \rO\cdot
                \left(
                    \frac
                        {1152 \Mbound\, \sqrt{\anA}}
                        {\sqrt{n}}
                \right)
                \Big[
                        \sqrt{
                            \Pdim(\FMLPg)
                            \log(n)
                        }
                    +
                        \sqrt{\tempA}
                \Big]
        \\&
        \qquad
                +
                \frac{246 \Mbound\Fbound\anA\tempA}{n}
                +
                    \CpcqB (\prjerr^2 + 2\truncf)
                +
                5\trunc
                +
                \theta_n
            \Bigg\}
            \cap
                \left\{
                    \normLpnS{\f-\fO}
                \leq 
                    \rO
                \right\}\!
            \!\Bigg)
        \\&
        \leq 
            6e^{-\tempA}
            +
            \frac{2n\,\betaBound(\anA)}{\anA}
            +
            2\P(\Ptrunc)
        ,
        \end{aligned}
        \end{equation}
    since $\beta(\anA) \leq \betaBound(\anA)$.
    Thus, 
        if 
    $\rO$ is sufficiently larger than
        \begin{equation*}
        \begin{aligned}
            \max
            \left\{
                \left(
                    \frac
                        {\Mbound\, \sqrt{\anA}}
                        {\sqrt{n}}
                \right)
                \Bigg[
                        \sqrt{
                            \Pdim(\FMLPg)
                            \log(n)
                        }
                    +
                        \sqrt{\tempA}
                \Bigg]
            \, , \,
                \frac{ \Mbound\Fbound\anA\tempA}{n}
            \, , \,
                \prjerr^2
            \, , \,
                \trunc
            \, , \, 
                \truncf
            \, , \,
                \theta_n
            \right\}
        \end{aligned}
        \end{equation*}
    then
    \eqref{eq:A9} implies that
    there exists
        $
            \rone 
        <
            \rO
        $
    such that 
        \begin{equation*}
        \begin{aligned}
            \Pr\big(
                \normLpnS{\f-\fO} \leq \rone 
            \big)
        \geq
            \Pr\big(
            \normLpnS{\f-\fO} \leq \rO 
            \big)
            - 6e^{-\tempA}
            - \frac{2n\,\betaBound(\anA)}{\anA}
            - 2\Pr(\Ptrunc)
        \end{aligned}
        \end{equation*}
    since $\Pr(A^{\comp} \cap B) = \Pr(B)-\Pr(A \cap B)$ for any events $A,B$.
    This can be done repeatedly as long as the new bound $\rone$ is such that  
    \eqref{eq:A4} and \eqref{eq:A5} hold for all $r''\geq \rone$.
}

\subsubsection{Step IV: Radius tightening lower bound}
\newcommand{\rcrit}{\overline{r}}%
\newcommand{\event}{E}%
    This step obtains a critical radius, $\rcrit$, that is a lower-bound for 
    radii 
    such that the one-step tightening  of the previous step can be applied.
    Define
        \begin{equation*}
        \resizebox{\hsize}{!}{$
        \begin{aligned}
            &\Xi
        \coloneqq
        \\&
            \inf\!
            \bigg\{\!
                    r\!>\!0
                :
                    12
                     \Fbound
                    \sqrt{2\anA}\,
                    \max_{i=1,2}
                    \left(\!
                        \E\big[
                            \RadX{\anA\anB}\!
                            \big\{
                                f-\fO:
                                f\!\in\FMLPg,
                                \normLpTi{i}{f-\fO}
                                \leq s
                            \big\}
                        \big] 
                    \right)\!
                <
                s^2,
                \,
                \forall s\geq r
            \bigg\}
            ,
        \end{aligned}
        $}
        \end{equation*}
    and 
        \begin{equation*}
        \begin{aligned}
        &
            \rcrit
        \coloneqq
            \max\left\{
                \Xi
            ,\;
                \frac{1}{2}
                \sqrt{\frac{55\,\anA \Fbound^2\log(n)}{n}}
            \right\}
        .
        \end{aligned}
        \end{equation*}
    The definition of $\rcrit$ implies
        $$
            2\rcrit
        \geq
            \sqrt{\frac{55\,\anA \Fbound^2\log(n)}{n}},
        $$
    so $2\rcrit$ satisfies \eqref{eq:A5} for $\tempA=\log(n)$, and
    by construction $2\rcrit$ satisfies \eqref{eq:A4}.
    Hence, for the event 
        \begin{gather*}
            \event
        \coloneqq
            \event_1 \cap \event_2
        ,\quad \text{where} 
        \quad
            \event_i
        \coloneqq
            \Bigg\{
                \sup_{\{f\in\FMLPg:\normLpTi{i}{f-\fO}\leq 2\rcrit\}}
                \normTi{i}{f-\fO}
            \leq
                4\rcrit
            \Bigg\}
        ,
        \;\; i \in \{1,2\}
        .
        \end{gather*}
    we can apply \eqref{eq:A6}, with $\tempA=\log(n)$, to obtain
        \begin{equation}\label{eq:Peventbnd}
        \begin{aligned}
            \Pr(\event)
        &
        =
            1-\Pr(\event^{\comp})
        =
            1-\Pr\Big(\event_1^{\comp} \cup \event_2^{\comp} \Big)
        \geq
            1-\Pr\big(\event_1^{\comp}\big) - \Pr\big(\event_2^{\comp} \big)
        \geq
            1 - 2e^{-\tempA}
        = 
            1 - \frac{2}{n}
        .
        \end{aligned}
        \end{equation}

    First, consider the case where
        $
        \rcrit = \Xi
        ,
        $
    or equivalently,
        \begin{equation*}
        \begin{aligned}
        &
                    \frac{1}{2}
                    \sqrt{\frac{55\,\anA \Fbound^2\log(n)}{n}}
       \;\leq\;
            \Xi
        .
        \end{aligned}
        \end{equation*}
    In this case,
    note that%
\footnote{To see this, suppose 
    $
        \rcrit^2
    >
        12
         \Fbound\sqrt{2\anA}\,
        \max_{i=1,2}
            \left(\!
                \E\left[
                    \RadX{\anA\anB}
                    \left\{
                        f-\fO:
                        f\in\FMLPg,
                        \normLpTi{i}{f-\fO}
                        \leq \rcrit
                    \right\}
                \right] 
            \right)
    .
    $
Then, there exists $\upsilon>0$ such that 
    $$
        (\rcrit-\upsilon)^2
    >
        12
         \Fbound\sqrt{2\anA}\,
        \max_{i=1,2}
            \left(\!
                \E\left[
                    \RadX{\anA\anB}
                    \left\{
                        f-\fO:
                        f\in\FMLPg,
                        \normLpTi{i}{f-\fO}
                        \leq \rcrit
                    \right\}
                \right] 
            \right)
        .
    $$
However, 
    $
    \big\{
    f\in\FMLPg:
    \normLpTi{i}{f-\fO} \leq \rcrit-\upsilon
    \big\}
    \subseteq
    \big\{
        f\in\FMLPg:
        \normLpTi{i}{f-\fO} \leq \rcrit
    \big\},
    $
so we have
    $$
        (\rcrit-\upsilon)^2
    >
        12
         \Fbound\sqrt{2\anA}\,
        \max_{i=1,2}
            \left(\!
                \E\left[
                    \RadX{\anA\anB}
                    \left\{
                        f-\fO:
                        f\in\FMLPg,
                        \normLpTi{i}{f-\fO}
                        \leq \rcrit-\upsilon
                    \right\}
                \right] 
            \right),
    $$
which contradicts 
$\rcrit=\Xi$.
}
    \begin{equation}\label{eq:rcritcase1}
        \rcrit^2
        \leq
            12
             \Fbound\sqrt{2\anA}\,
            \max_{i=1,2}
            \left(\!
                \E\left[
                    \RadX{\anA\anB}
                    \left\{
                        f-\fO:
                        f\in\FMLPg,
                        \normLpTi{i}{f-\fO}
                        \leq \rcrit
                    \right\}
                \right] 
            \right)
            .
    \end{equation}
    Recall, 
        $
        \textstyle
            \sup_{f\in\FMLPg}\snorm{f-\fO}
        \leq
            2 \Fbound
        ,
        $
    which implies
        \begin{equation}\label{eq:unifradbound}
        \begin{aligned}
            2 \Fbound
        &
        \geq 
            \RadX{\anA\anB}
            \left\{
                f-\fO:
                f\in\FMLPg
            \right\}
        \\&
        \geq
            \RadX{\anA\anB}\!
                \left\{\!
                    f-\fO:
                    f\in\FMLPg,
                    \normLpTi{i}{f-\fO}\!
                    \leq 2\rcrit
                \right\}\!
            ,\quad
            \forall i\in\{1,2\}\!
        .
        \end{aligned}
        \end{equation}
    Using \eqref{eq:rcritcase1} with \eqref{eq:unifradbound} 
    and 
    \eqref{eq:Peventbnd}
        \begin{equation*}
        \begin{aligned}
            \rcrit^2
        &
        \leq
            12 \Fbound\sqrt{2\anA}\,
            \max_{i=1,2}
            \left(\!
                \E\left[
                    \RadX{\anA\anB}\!
                    \left\{\!
                        f-\fO:
                        f\in\FMLPg,
                        \normLpTi{i}{f-\fO}\!
                        \leq 2\rcrit
                    \right\}\!
                \right] 
            \right)
        \\&
        \leq
            12 \Fbound\sqrt{2\anA}\,
                \max_{i=1,2}
                \left(\!
                \E\bigg[
                    \E_{\P^{\xi}}\Big[
                        \RadX{\anA\anB}
                        \left\{
                            f-\fO:
                            f\in\FMLPg,
                            \normLpTi{i}{f-\fO}\!
                            \leq 2\rcrit
                        \right\}
                    \Big]
                \mathbbm{1}_{\event}
            \bigg] 
            \right)
            \\&
            \qquad 
                +
                12 \Fbound\sqrt{2\anA}\,
                \E\Big[
                2 \Fbound(1-\mathbbm{1}_{\event})
            \Big] 
        \\&
        \leq
            12 \Fbound\sqrt{2\anA}\,
                \max_{i=1,2}
                \left(\!
                \E\bigg[
                    \E_{\P^{\xi}}\Big[
                        \RadX{\anA\anB}
                        \left\{
                            f-\fO:
                            f\in\FMLPg,
                            \normTi{i}{f-\fO}\leq 4\rcrit
                        \right\}
                    \Big]
                \mathbbm{1}_{\event}
            \bigg] 
            \right)
            \\&
            \qquad 
                +
                12 \Fbound\sqrt{2\anA}\,
                \E\Big[
                2 \Fbound(1-\mathbbm{1}_{\event})
            \Big] 
        \\&
        \leq
            12 \Fbound\sqrt{2\anA}\,
            \max_{i=1,2}
                \left(\!
            \E\bigg[
                \E_{\P^{\xi}}\Big[
                    \RadX{\anA\anB}
                    \left\{
                        f-\fO:
                        f\in\FMLPg,
                        \normTi{i}{f-\fO}\leq 4\rcrit
                    \right\}
                \Big]
            \bigg] 
            \right)
            +
            \frac{48 \Fbound^2}{n}
        .
        \end{aligned}
        \end{equation*} 
    Letting
        $
            \Diam
        \coloneqq
            \min
            \left\{
                2 \Fbound
                ,\,
                4\rcrit
            \right\}
        ,
        $
    then Proposition \ref{prop:rad_dudley_bound} can be used, similar to how \eqref{eq:RadBound0} and \eqref{eq:RadBound1} were obtained, to show
        \begin{equation*}
        \begin{aligned}[b]
            \E_{\xi}
            \Big[
                \RadX{\anA\anB}\!
                \Big\{
                    f-\fO:
                    f\in\FMLPg,
                    \normTi{i}{f-\fO}\! \leq 4\rcrit
                \Big\}
            \Big] 
        &
        \leq
            \rcrit\cdot
            96
                \sqrt{
                    \frac{ 2\,\Pdim(\FMLPg)}{n}
                      \logp{n}
                }
        ,\;\;
            \forall i\in\{1,2\}
        .
        \end{aligned}
        \end{equation*}
    Combining the previous two displays
        \begin{equation*}
        \begin{aligned}[b]
            \rcrit^2
        &
        \leq
            \rcrit\cdot
            2304
            \cdot
             \Fbound
                \sqrt{
                    \frac{ \anA\,\Pdim(\FMLPg)}{n}
                      \logp{n}
                }
            +
            \frac{48 \Fbound^2}{n}
        \\&
        \leq
            \rcrit\cdot
            2352
            \cdot
             \Fbound
                \sqrt{
                    \frac{ \anA\,\Pdim(\FMLPg)}{n}
                      \logp{n}
                }
        ,
        \end{aligned}
        \end{equation*}
    where the last inequality uses 
        $
            \rcrit
        \geq
            \sqrt{{55\,\anA \Fbound^2\log(n)}/{n}}
        >
             \Fbound/\sqrt{n}
        .
        $
    Thus,
        $$
            \frac{1}{2}
            \sqrt{\frac{55\,\anA \Fbound^2\log(n)}{n}}
        \;\leq\;
            \Xi
        \quad 
        \implies 
        \quad
            \rcrit
            \;\leq\;
            2352
             \Fbound
                \sqrt{
                    \frac{ \anA\,\Pdim(\FMLPg)}{n}
                      \logp{n}
                }
        .
        $$

    For the alternative case, 
        $$
            \frac{1}{2}
            \sqrt{\frac{55\,\anA \Fbound^2\log(n)}{n}}
        >
            \Xi
        \;
        \implies 
        \;
            \rcrit
            =
            \frac{1}{2}
                    \sqrt{\frac{55\,\anA \Fbound^2\log(n)}{n}}
            \leq
            2352 
             \Fbound
            \sqrt{
                \frac{ \anA\Pdim(\FMLPg)}{n}
                \log(n)
            }
        .
        $$
    since $\Pdim(\FMLPg)\geq 1$ by definition.
    Combining the previous two displays shows,
    in either case,
        \begin{equation}\label{eq:rstar}
        \begin{aligned}
            \rcrit
        \leq
            2352
             \Fbound
            \sqrt{
                \frac{ \anA\Pdim(\FMLPg)}{n}
                \log(n)
            }
        .
        \end{aligned}
        \end{equation}

\subsubsection{Step V: Final \texorpdfstring{$\Lp{2}$}{L2}-error bound}\label{sec:step4}
\newcommand{\rate}{r_*}%
\newcommand{\ffOdist}{\normLpnS{\f-\fO}}%
\newcommand{\MFB}{(\Mbound\vee\Fbound)}%
    For $\delta$ from the statement of this theorem,
    choose 
        $$
            \tempA=\tempB+\log(6J)
        ,
        \quad
        \text{ where }
        \quad
            J
        \coloneqq
            \left\lfloor
                \log_{2}\left(
                \frac{4 \Fbound\sqrt{n}}{\sqrt{\log(n)}}
                \right)
            \right\rfloor
        .
        $$
    Then $\tempA>\max\{\tempB,1\}$ holds for any $\tempB>0$, 
        since 
        $\Fbound\geq 1$ 
        by Condition \ref{as:Env},
    implies
        $J\geq \log_2(4) \geq 2$, so $\tempA \geq \tempB + \log(14)>1$.

    Let $x\vee y = \max\{x,y\}$, and define 
        \begin{equation}\label{eq:A14}
        \begin{aligned}
            \rate
        \;\coloneqq \;\;
        & 
            \rcrit
            +
                4\cdot
                \frac{ 1152 }{\Cpcq}
                \left(
                    \frac
                        {\MFB\, \sqrt{\anA}}
                        {\sqrt{n}}
                \right)
                \Bigg[
                        \sqrt{
                            \Pdim(\FMLPg)
                            \log(n)
                        }
                    +
                        \sqrt{\tempA}
                \Bigg]
        \\&
            +
            \sqrt{\frac{2}{\Cpcq}}
            \Bigg[
                    \frac{246 \Mbound\Fbound\anA\tempA}{n}
                +
                    \CpcqB (\prjerr^2 + 2\truncf)
                +
                    5\trunc
                +
                    \theta_n
            \Bigg]^{1/2}
        .
        \end{aligned}
        \end{equation}
    Now we will use \eqref{eq:A9} to show, for any $j\in\{1,\ldots,J\}$, 
        \begin{equation}\label{eq:A9_localization}
        \begin{aligned}
            \P\Big( 
                \big\{
                    \ffOdist
                >
                    2^{j-1}\rate
                \big\}
                \cap
                \big\{
                    \ffOdist
                \leq
                    2^{j}\rate
                \big\}
            \Big)
        \leq
            6e^{-\tempA}
            +
            \frac{2n\,\beta(\anA)}{\anA}
            +
            2\P(\Ptrunc)
        .
        \end{aligned}
        \end{equation}
    To apply \eqref{eq:A9}, note that
    \eqref{eq:A4} and \eqref{eq:A5}
    hold with 
    $2^{j}\rate$ 
    because:
    $\rate$ satisfies \eqref{eq:A4} since $\rate>\rcrit$; and
    $\rate$ satisfies \eqref{eq:A5} since 
        $
            {(4\cdot 1152)   }
            /{\Cpcq}
            \geq
            \sqrt{55}
        $
    follows from $\Cpcq\leq 1$ by Condition \ref{as:curv}.
    Next, $\ffOdist \leq 2^{j}\rate$ can be tightened to $2^{j-1}\rate$ using \eqref{eq:A9}, since 
    for any $j\in\{1,\ldots,J\}$
        \begin{equation*}
        \resizebox{\linewidth}{!}{$
        \begin{aligned}
        &
            2^j\rate \cdot
                \frac{1}{\Cpcq}
                \left(
                    \frac
                        {1152 \Mbound \sqrt{\anA}}
                        {\sqrt{n}}
                \right)
                \Bigg[
                        \sqrt{
                            \Pdim(\FMLPg)
                            \log(n)
                        }
                    +
                        \sqrt{\tempA}
                \Bigg]
                +
                \frac{1}{\Cpcq}
                \Bigg[
                    \frac{246 \Mbound\Fbound\anA\tempA}{n}
                +
                    \CpcqB (\prjerr^2 + 2\truncf)
                +
                    5\trunc
                +
                    \theta_n
            \Bigg]
        \\&
        \leq
            \frac{(2^j\rate)^2}{8}
            +
            \frac{(2^j\rate)^2}{8}
        \;=\; 
            \frac{(2^j\rate)^2}{4}
        \;=\;
            (2^{j-1}\rate)^2
        ,
        \end{aligned}
        $}
        \end{equation*}
    where the second inequality follows because, 
    by
    \eqref{eq:A14}
        $$
            2^j\rate \cdot
            \frac{1}{\Cpcq}
                \left(
                    \frac
                        {1152 \Mbound\, \sqrt{\anA}}
                        {\sqrt{n}}
                \right)
                \Bigg[
                        \sqrt{
                            \Pdim(\FMLPg)
                            \log(n)
                        }
                    +
                        \sqrt{\tempA}
                \Bigg]
            \;\leq\;
                2^j\rate \cdot
                \frac{\rate}{4}
            \;\leq\;
                \frac{(2^{j}\rate)^2}{8},
        \quad \text{ and }
        $$
        $$
            \frac{1}{\Cpcq}
            \Bigg[
                    \frac{246 \Mbound\Fbound\anA\tempA}{n}
                +
                    \CpcqB (\prjerr^2 + 2\truncf)
                +
                    5\trunc
                +
                    \theta_n
            \Bigg]
            \;\leq\;
                \frac{\rate^2}{2}
            \;\leq\;
                \frac{(2^j\rate)^2}{8}
            .
        $$
    Therefore, \eqref{eq:A9_localization} follows from \eqref{eq:A9}.

    Now the local bound from \eqref{eq:A9_localization} can be used to obtain a global bound for $\ffOdist$.
    First, by \eqref{eq:A9_localization}
        \begin{equation}\label{eq:prob_bound_j}
        \resizebox{\hsize}{!}{$
        \begin{aligned}[b]
            \P\Big(
                \ffOdist
            >
                2^{j-1}\rate
            \Big)
        &
        \leq
            \P\Big(
                \big\{\ffOdist > 2^{j-1}\rate \big\} 
                \cap 
                \big\{\ffOdist \leq 2^{j}\rate \big\}
            \Big)
        \\&
        \qquad
            +
            \P\Big(
                \big\{\ffOdist > 2^{j}\rate \big\}
            \Big)
        \\&
        \leq
            6e^{-\tempA}
            +
            \frac{2n\,\beta(\anA)}{\anA}
            +
            2\Pr(\Ptrunc)
            +
            \P\Big(
                \ffOdist > 2^{j}\rate 
            \Big)
        .
        \end{aligned}
        $}
        \end{equation}
    Note that 
        $\rate > \sqrt{\log(n)/n}$,
    and 
        \begin{equation*}
        \begin{aligned}
            J
        &=
            \left\lfloor
                \log_{2}\left(
                \frac{4 \Fbound\sqrt{n}}{\sqrt{\log(n)}}
                \right)
            \right\rfloor
        =
            \left\lfloor
                \log_{2}\left(
                    \frac{2 \Fbound\sqrt{n}}{\sqrt{\log(n)}}
                \right)
                +1
            \right\rfloor
        \geq
                \log_{2}\left(
                    \frac{2 \Fbound\sqrt{n}}{\sqrt{\log(n)}}
                \right)
        ,
        \end{aligned}
        \end{equation*}
    which, with Condition \ref{as:Env}, implies
        $$
            2^J\rate
        \geq
            \left(
                \frac{2 \Fbound\sqrt{n}}{\sqrt{\log(n)}}
            \right)
            \rate
        >
            2 \Fbound
        \geq
            \sup_{f\in\FMLPg}
            \snormDNNBound{f-\fO}
        \geq
            \ffOdist
        .
        $$
    Then, 
        $
            \P\big(
                \ffOdist > 2^{J}\rate 
            \big)
            =0
        $,
    and with \eqref{eq:prob_bound_j}
        \begin{equation*}
        \begin{aligned}
            \P\Big(
                \ffOdist
            >
                2^{J-1}\rate
            \Big)
        &\leq
            6e^{-\tempA}
            +
            \frac{2n\,\beta(\anA)}{\anA}
            +
            2\Pr(\Ptrunc)
        .
        \end{aligned}
        \end{equation*}
    Hence, it follows via induction that
        \begin{equation}\label{eq:prob_bound_j2}
        \begin{aligned}[b]
            \P\Big(
                \ffOdist
            >
                \rate
            \Big)
        &\leq
            J\cdot
            \bigg(
            6e^{-\tempA}
            +
            \frac{2n\,\beta(\anA)}{\anA}
            +
            2\Pr(\Ptrunc)
            \bigg)
        \\&
        \leq
            e^{-\tempB}
            +
            4\log(n)
            \left[
                \frac{n\,\beta(\anA)}{\anA}
                +
                \Pr(\Ptrunc)
            \right]
        ,
        \end{aligned}
        \end{equation}
    where the first term used 
        $6Je^{-\tempA}= 6Je^{-\delta-\log(6J)} =e^{-\tempB}$, 
    and the second term follows since
    by assumption 
    $n\geq16 \Fbound^2/\log(n)$, which implies $4 \Fbound/\sqrt{\log(n)}\leq \sqrt{n}$,
    so we have
        \begin{equation*}
        \begin{aligned}
            J
        &
        \leq
                \log_{2}\left(
                \frac{4 \Fbound\sqrt{n}}{\sqrt{\log(n)}}
                \right)
        \leq 
            \log_{2}(n)
        \leq
            2\log(n)
        .
        \end{aligned}
        \end{equation*}

    To obtain the bound for 
    $\normLpn{\f-\fO}$,
    using the definition of $\truncf$.
        \begin{equation*}
        \begin{aligned}
            &
                \normLpn{\f-\fO}^2
            = 
                \normLpnS{\f-\fO}^2
                +
                \int_{\x\notin\Xbound}
                |\f(\x) - \fO(\x)|^2
                d\Pbarn(\x)
            \\&
            \leq
                \normLpnS{\f-\fO}^2
                +
                \sampavg
                \int_{\x\notin\Xbound}\!
                \big(|\f(\x)|^2 + |\fO(\x)|^2 + 2|\f(\x)\fO(\x)|\big)
                d\P_{\Xt}(\x)
            \\&
            \leq
                \normLpnS{\f-\fO}^2
                +
                \frac{4}{n}
                \sumin
                \sup_{f\in\left\{\FMLPg\cup\{\fO\}\right\}}
                \E\Big[
                    |f(\Xt)|^2
                    \mathbbm{1}_{\{\Xt\notin\Xbound\}}
                \Big]
            \\&
            \leq
                \normLpnS{\f-\fO}^2
                +
                4\truncf
            .
        \end{aligned}
        \end{equation*}
    With this and \eqref{eq:prob_bound_j2}
        \begin{equation}\label{eq:prob_bound_j3}
        \begin{aligned}[b]
            \P\left(
                \normLpn{\f-\fO}
            >
                \rate
                +
                2\sqrt{\truncf}
            \right)
        &\leq
            e^{-\tempB}
            +
            4\log(n)
            \left[
                \frac{n\,\beta(\anA)}{\anA}
                +
                \Pr(\Ptrunc)
            \right]
        .
        \end{aligned}
        \end{equation}

    With \eqref{eq:prob_bound_j3}, the proof will be complete by showing
        $\rate + 2\sqrt{\truncf} \leq C\, \sqrt{\err(\tempB,\anA)}$,
    for a suitable constant $C>0$, and $\err$ defined as in the statement of this theorem.
    Note that
        \begin{equation}\label{eq:delta_term2}
        \begin{aligned}
            \sqrt{\tempA}
        \leq
            \sqrt{\log\big(10\log(n)\big) + \tempB}
        <
            \sqrt{9\log\log(n) + \tempB}
        \leq
            3\sqrt{\log\log(n) + \tempB}
        \end{aligned}
        \end{equation}
    since $n\geq 4$ by assumption, and  $J\leq 2\log(n)$ was shown previously.
    Then we have
        \begin{equation}\label{eq:A17}
        \resizebox{.9\hsize}{!}{$
        \begin{aligned}[b]
        &
            \rate + 2\sqrt{\truncf}
        \\
        &
        \leq
            \rcrit
            +
             \frac{4\cdot 1152 }{\Cpcq}
                \left(\MFB\sqrt{\frac{\anA}{n}}\right)
                \Big[
                        \sqrt{
                            \Pdim(\FMLPg)
                            \log(n)
                        }
                    +
                        \sqrt{\tempA}
                \Big]
            +
                \sqrt{\frac{492 }{\Cpcq}}
                \left(\sqrt{\Mbound\Fbound}\sqrt{\frac{\anA}{n}}\right)
                \sqrt{\tempA}
        \\&
        \qquad
            +
                \sqrt{
                    \frac{2\CpcqB + 5}{\Cpcq}
                }
                \sqrt{
                    \prjerr^2
                    +
                    \trunc
                    +
                    \truncf
                    +
                    \theta_n
                }
                + 2\sqrt{\truncf}
        \\& 
        \leq
                \left(\!
                2352
                +
                \frac{4\cdot1152 }{\Cpcq}
                +
                 \sqrt{\frac{492 }{\Cpcq}}
                \right)\!
                \sqrt{\frac{\anA}{n}}
                \Big[
                        \MFB
                        \sqrt{
                            \Pdim(\FMLPg)
                            \log(n)
                        }
                    +
                        2\MFB
                        \sqrt{\tempA}
                \Big]
        \\&
        \qquad
            +
                \sqrt{
                    \frac{2\CpcqB + 5}{\Cpcq}
                }
                \sqrt{
                    \prjerr^2
                    +
                    \trunc
                    +
                    \truncf
                    +
                    \theta_n
                }
                + 2\sqrt{\truncf}
        \\& 
        \leq
                C'\,\Bigg(
                \MFB
                \sqrt{\frac{\anA}{n}}
                \Big[
                        \sqrt{
                            \Pdim(\FMLPg)
                            \log(n)
                        }
                    +
                        {\sqrt{\log\log(n)+\tempB}}
                \Big]
            +
                \sqrt{
                    \prjerr^2
                    +
                    \trunc
                    +
                    \truncf
                    +
                    \theta_n
                }
            \Bigg)
        \\& 
        \leq
             C'\, \sqrt{2\err(\tempB,\anA)}
        ,
        \end{aligned}
        $}
        \end{equation}
    where
        the second inequality used \eqref{eq:rstar} with $\sqrt{\Mbound\Fbound}\leq\MFB$,
    the third inequality used \eqref{eq:delta_term2} with 
        $$
            C' 
        =
            2
            \sqrt{
                    \frac{2\CpcqB + 5}{\Cpcq}
                }
        \;\vee\;
                \left(\!
                2352
                +
                \frac{4\cdot1152 }{\Cpcq}
                +
                 \sqrt{\frac{492 }{\Cpcq}}
                \right)\!
        ,
        $$
    and the last inequality used the definition of $\err(\delta,\anA)$
    with 
    $(x+y)^2 \leq 2(x^2+y^2)$ for all $x,y\geq0$,
    and
    $\Pdim(\FMLPg)\log(n) \geq \log\log(n)$ since $\Pdim(\FMLPg)\geq 1$ and $n\geq8$.
    Therefore,
        $C'$
    only depends on 
        $\Cpcq$
    and
        $\CpcqB$.
    Thus, \eqref{eq:prob_bound_j3} and \eqref{eq:A17} complete the proof of the first result.

\subsubsection{Step VI: Final empirical error bound}\label{sec:EmpErrBound}
    Let
        \begin{gather*}
            \norm*{\f-\fO}_{2,n}^2 
        \coloneqq
            \sampavg
            \big[\f(\Xt)-\fO(\Xt)\big]^2
        ,
        \quad \text{and}
        \\
            \norm*{\f-\fO}_{2,n|\Xbound}^2 
        \coloneqq
            \sampavg
            \big[\f(\Xt)-\fO(\Xt)\big]^2\onext
        .
        \end{gather*}
    Note that 
        $\big[\f(\Xt)-\fO(\Xt)\big]^2\onezt \leq \big[\f(\Xt)-\fO(\Xt)\big]^2\onext$,
    and
    by De Morgan's laws
        $$
            \{\Ptrunc\}^\comp
        =
            \bigcap_{t\in\{1,\ldots,n\}}
                \Big\{
                \Zt \in \Zbound
                \Big\}
        .
        $$
    Thus,
    for any $A>0$,
        \begin{equation}\label{eq:emperr0}
        \begin{aligned}
            \Pr\Big(
                \norm*{\f-\fO}_{2,n}
            \geq
                \sqrt{A}
            \Big)
        &=
            \Pr\Big(
                \left\{
                    \norm*{\f-\fO}_{2,n}^2
                \geq
                    A
                \right\}
                \cap \{\Ptrunc\}^{\comp}
            \Big)
            +
            \Pr\Big(
                \left\{
                    \norm*{\f-\fO}_{2,n}^2
                \geq
                    A
                \right\}
                \cap \{\Ptrunc\}
            \Big)
        \\&
        \leq
            \Pr\Big(
                    \norm*{\f-\fO}_{2,n|\Xbound}^2
                \geq
                    A
            \Big)
            +
            \Pr(\Ptrunc)
        .
        \end{aligned}
        \end{equation}
    Note that
        \begin{equation*}
        \resizebox{\hsize}{!}{$
        \begin{aligned}
            \norm*{\f-\fO}_{2,n|\Xbound}^2
        &=
            \left(
                \frac{1}{n}
                \sum_{j=1}^{\anB}
                \sum_{t\in \Ione{j}}
                \big[\f(\Xt)-\fO(\Xt)\big]^2
                \onext
            \right)
            +
            \left(
                \frac{1}{n}
                \sum_{j=1}^{\anB}
                \sum_{t\in \Itwo{j}}
                \big[\f(\Xt)-\fO(\Xt)\big]^2
                \onext
            \right)
        \\&
        \qquad
            + 
            \left(
            \frac{1}{n}
            \sum_{t\in\IR}
            \big[\f(\Xt)-\fO(\Xt)\big]^2
            \onext
            \right)
        .
        \end{aligned} 
        $}
        \end{equation*}
    Then, by \eqref{eq:A17}
        \begin{equation}\label{eq:emperr1}
        \begin{aligned}
            \Pr\Big(
        &
                \norm*{\f-\fO}_{2,n|\Xbound}
            \geq
                \sqrt{17}\, C'\sqrt{2\err(\tempB,\anA)}
            \Big)
        \leq
            \Pr\Big(
                \norm*{\f-\fO}_{2,n|\Xbound}^2
            \geq
                17\rate^2
            \Big)
        \\&
        \leq
            2
            \max_{i\in\{1,2\}}
            \Pr\left(
                \frac{1}{n}
                \sum_{j=1}^{\anB}
                \sum_{t\in \Ik{i}{j}}
                \big[\f(\Xt)-\fO(\Xt)\big]^2
                \onext
            \geq    
                8
                \rate^2
            \right) 
        \\&
        \qquad
        +
            \Pr\left(
                \frac{1}{n}
                \sum_{t\in\IR}
                \big[\f(\Xt)-\fO(\Xt)\big]^2
                \onext
            \geq 
                \rate^2
            \right)
        \\&
        =:
            2\max_{i\in\{1,2\}}\P_3^{(i)} + \P_4
        \end{aligned}
        \end{equation}

    Consider $\P_3^{(i)}$ for arbitrary $i\in\{1,2\}$,
        \begin{equation*}
        \resizebox{\hsize}{!}{$
        \begin{aligned}
            \P_3^{(i)}
        &
        \leq
            \Pr\left(
                \Bigg\{
                    \frac{1}{n}
                    \sum_{j=1}^{\anB}
                    \sum_{t\in \Ik{i}{j}}
                    \big[\f(\Xt)-\fO(\Xt)\big]^2
                    \onext
                \geq    
                    8
                    \rate^2
                \Bigg\}
            \cap
                \Big\{
                        \ffOdist
                    \leq
                        \rate
                \Big\}
            \right)     
        \\&
        \qquad
             +
            \P\Big(
                \ffOdist
            >
                \rate
            \Big)
        \\&
        \leq
            \Pr\left(
                \sup_{\{f\in\FMLPg:\normLpTi{i}{f-\fO}\leq 2\rate\!\}}
                    \frac{1}{n}
                    \sum_{j=1}^{\anB}
                    \sum_{t\in \Ik{i}{j}}
                    \big[f(\Xt)-\fO(\Xt)\big]^2
                    \onext
                \geq    
                    8
                    \rate^2
            \right)     
             +
            \P\Big(
                \ffOdist
            >
                \rate
            \Big)
        \\&
        \leq
            \Pr\left(
                \sup_{\{f\in\FMLPg:\normLpTi{i}{f-\fO}\leq 2\rate\!\}}
                (\anA\anB/n)
                \normTi{i}{f-\fO}^2
            \geq
                8
                \rate^2
            \right)
            +
            \frac{n\,\beta(\anA)}{2\anA}
             +
            \P\Big(
                \ffOdist
            >
                \rate
            \Big)
        \\&
        \leq
            \Pr\left(
                \sup_{\{f\in\FMLPg:\normLpTi{i}{f-\fO}\leq 2\rate\!\}}
                \normTi{i}{f-\fO}^2
            \geq
                16
                \rate^2
            \right)
            +
            \frac{n\,\beta(\anA)}{2\anA}
            +
            e^{-\tempB}
            +
            4
            \log(n)
            \left[
                \frac{n\,\beta(\anA)}{\anA}
                +
                \Pr(\Ptrunc)
            \right]
        ,
        \end{aligned}
        $}
        \end{equation*}
    where 
        the second inequality uses \eqref{eq:normLpTi_v_normLpnS},
        the third inequality uses  \eqref{eq:beta_bound2}, 
        and the last inequality uses 
        $\anB\leq n/(2\anA)$
        and
        \eqref{eq:prob_bound_j2}.
    By the same reasoning used in Subsection \ref{sec:step4} 
    we can apply \eqref{eq:A6} to the previous display, and since $i\in\{1,2\}$ was arbitrary,
        \begin{equation}\label{eq:P3}
        \begin{aligned}
            \max_{i\in\{1,2\}}\P_3^{(i)}
        &\leq
            2e^{-\tempB}
            +
            4
            \log(n)
            \left[
                \frac{n\,\beta(\anA)}{\anA}
                +
                \Pr(\Ptrunc)
            \right]
            +\frac{n\,\beta(\anA)}{2\anA}
        ,
        \end{aligned}
        \end{equation}

    Consider $\P_4$.
    Recall 
        $ (\#\IR)  
        \coloneqq  n-2\anA\anB  
        <  2\anA $,
    from Section \ref{sec:step2}.
    Then, 
    by 
    Condition \ref{as:Env}
        $$
            \frac{1}{n}
            \sum_{t\in\IR}
            \big[\f(\Xt)-\fO(\Xt)\big]^2
            \onext
        \leq
            \frac{2\anA}{n}
            \snormDNNBound{\f-\fO}^2
        <
            \frac{8 \Fbound^2\anA}{n}
        <
            \rate^2
        $$
    where the last inequality has used the definition of $\rate$ from \eqref{eq:A14}.
    Therefore,
        \begin{equation}\label{eq:P4}
        \begin{aligned}
            \P_4
        =
            \Pr\left(
                \frac{1}{n}
                \sum_{t\in\IR}
                \big[\f(\Xt)-\fO(\Xt)\big]^2
                \onext
            \geq 
                \rate^2
            \right)
        =
            0
        .
        \end{aligned}
        \end{equation}

    Applying \eqref{eq:P4}, \eqref{eq:P3} and \eqref{eq:emperr1} to \eqref{eq:emperr0}
        \begin{equation*}
        \begin{aligned}
            \Pr\Big(
                \norm*{\f-\fO}_{2,n}
            \geq
                C' 
                \sqrt{34\err(\tempB,\anA)}
            \Big)
        &
        \leq
            2
            \left(
                2e^{-\tempB}
                +
                4
                \log(n)
                \left[
                    \frac{n\,\beta(\anA)}{\anA}
                    +
                    \Pr(\Ptrunc)
                \right]
                +\frac{n\,\beta(\anA)}{2\anA}
            \right)
            +
            \Pr(\Ptrunc)
        \\&
        \leq
            4e^{-\tempB}
            +
            9
            \log(n)
            \left[
                \frac{n\,\beta(\anA)}{\anA}
                +
                \Pr(\Ptrunc)
            \right]
        .
        \end{aligned}
        \end{equation*}
    This completes the proof of Theorem \ref{thrm:ROC_V2} with $C= 34 (C')^2$.

\subsection{Supporting Results}\label{sec:lem_ROC_V2}

{
\begin{lemma}\label{lem:avgmeas}
Let $\probspace$ be a probability space and let
$\X_1,\ldots,\X_n$ be random vectors where $\Xt:\Omega\to\R^d$ and
$\E\big[\|\X_t\|\big]<\infty$ for each $t\le n$.\footnote{Any norm on $\R^d$
may be used.}
Let $\P_{\X_t}:=\P\circ \X_t^{-1}$ denote the law of $\X_t$ on
$(\R^d,\mathcal B(\R^d))$.
Then, for any 
$g:\R^d\to\R$ satisfying
$g\in\Lp{1}(\P_{\Xt})$
for all $t\le n$,
\[
    \frac{1}{n}\sum_{t=1}^n \E[g(\X_t)]
    =
    \int_{\R^d} g(x)\, d\overline{\P}_{\X_1^n}(x)
    ,
    \quad
    \text{ where }
    \quad
    \overline{\P}_{\X_1^n}:=\frac{1}{n}\sum_{t=1}^n \P_{\Xt}
    .
\]
\end{lemma}
\begin{proof}
Fix $t\le n$. By definition of 
$\P_{\X_t}=\P\circ \X_t^{-1}$, 
for any 
$g\in\Lp{1}(\P_{\Xt})$
\begin{equation}\label{eq:pushforward_d}
\E[g(\X_t)]
=
\int_{\R^d} g(x)\, d\P_{\X_t}(x).
\end{equation}
Therefore,
\[
\frac{1}{n}\sum_{t=1}^n \E[g(\X_t)]
=
\frac{1}{n}\sum_{t=1}^n \int_{\R^d} g(x)\, d\P_{\X_t}(x).
\]
It remains to show that, for any integrable Borel-measurable $g$,
\begin{equation}\label{eq:mixture_d}
\int_{\R^d} g(x)\, d\overline{\P}_{\X_1^n}(x)
=
\frac{1}{n}\sum_{t=1}^n \int_{\R^d} g(x)\, d\P_{\X_t}(x).
\end{equation}

We first verify \eqref{eq:mixture_d} for nonnegative simple functions.
Let
\[
s(x)=\sum_{k=1}^K a_k \mathbbm{1}_{A_k}(x),
\qquad
a_k\ge 0,\; A_k\in\mathcal B(\R^d).
\]
By the definition of $\overline{\P}_{\X_1^n}$ and linearity of integration,
\[
\int_{\R^d} s(x)\, d\overline{\P}_{\X_1^n}(x)
=
\sum_{k=1}^K a_k\,\overline{\P}_{\X_1^n}(A_k)
=
\frac{1}{n}\sum_{t=1}^n \sum_{k=1}^K a_k\,\P_{\X_t}(A_k)
=
\frac{1}{n}\sum_{t=1}^n \int_{\R^d} s(x)\, d\P_{\X_t}(x).
\]

Next, let $g\ge 0$ be Borel-measurable. There exists a sequence of nonnegative
simple functions $\{s_m\}_{m\in\N}$, $s_m:\R^d\to\R$, such that $s_m\leq s_{m+1} \leq g$ and $s_m\to g$ pointwise.
By the previous step
\[
\int_{\R^d} s_m(x)\, d\overline{\P}_{\X_1^n}(x)
=
\frac{1}{n}\sum_{t=1}^n \int_{\R^d} s_m(x)\, d\P_{\X_t}(x)
,
\qquad\text{for each }m\in\N
.
\]
Applying the monotone convergence theorem to both sides gives
\[
\int_{\R^d} g(x)\, d\overline{\P}_{\X_1^n}(x)
=
\frac{1}{n}\sum_{t=1}^n \int_{\R^d} g(x)\, d\P_{\X_t}(x).
\]

Finally, for a general integrable function $g$, write $g=g^+-g^-$, where
$g^+,g^-\ge0$.
Applying the previous argument to $g^+-g^-$ 
yields \eqref{eq:mixture_d}. 
With \eqref{eq:pushforward_d} this
completes the proof.
\end{proof}
    \vx
}

{
\newcommand{\Wt}{W_t}%
\renewcommand{\X}{X}%
Lemma \ref{lem:mixing_mapping} is a simplified version of \citet[Theorem 15.1]{davidson_stochastic_2022} that will be useful throughout.
For a random sequence $\{\Wt\}_{t\in\N}$ the mixing coefficients associated with $\{\Wt\}_{t\in\N}$ are denoted as $\beta_{\X}$.
    \begin{lemma}\label{lem:mixing_mapping}
        Let $U:\Zspace\to\R$ be measurable-$\borel(\Zspace)/\borel(\R)$ and define $\Wt\coloneqq U(\Zt)$.
        Then
            $\beta_{W}(j)\leq \beta_{Z}(j)$ for any $j\in\N$.
    \end{lemma}
    \begin{proof}
        Note that $\Wt$ is measurable-$\sigma(\Zt)/\borel(\R)$ for each $t\in\N$. 
        Consequently, 
            $
                \sigma\big(\{\Wt\}_{1}^k\big)
            \subseteq
                \sigma\big(\{\Zt\}_{1}^k\big)
            $
        and
            $
                \sigma\left(\{\Wt\}_{k+j}^\infty\right)
            \subseteq
                \sigma\left(\{\Zt\}_{k+j}^\infty\right)
            ,
            $
        for any $k,j\in\N$.
        With this, the desired follows immediately from Definition
            \ref{def:beta_mixing}.
    \end{proof}
    \vx
}

{
\newcommand{\ftn}{f}%
\newcommand{\Boundtemp}{B}%
\newcommand{\tempn}{n}%
\renewcommand{\Zt}{\boldsymbol{X}_t}%
\renewcommand{\Zspace}{\boldsymbol{\mathcal{X}}}%
\newcommand{\Varbound}{\Boundtemp^2}%
    Lemma \ref{lem:Bernstein_GenMix}
    follows from the Bernstein-type inequality for general $\beta$-mixing processes in 
    \citet[Theorem 1]{mirzaei_empirical_2025}.
\begin{lemma}\label{lem:Bernstein_GenMix}
    Let 
        $\{\Zt\}_{t\in\N}$ be a $\beta$-mixing process
    where each $\Zt$ takes values in $\Zspace\subseteq\R^d$,
    and there exists a non-increasing function $\betaBound:\N\to[0,\infty)$ such that $\beta(j)\leq \betaBound(j)$.
    Let $\ftn:\Zspace\to\R$ be measurable-$\borel(\Zspace)/\borel(\R)$, such that 
        $\sup_{t\in\N}\abs{\ftn(\Zt)}<\Boundtemp$.
    If 
    $\anA\in\N$, $n\in\N$ are such that 
        $
        n\betaBound(\anA)
        /\anA
        < 1
        $
    and 
        $1\leq\anA\leq n/2$,
    and
        $\anB \coloneqq \big\lfloor n/(2\anA)\big\rfloor$,
    then 
        \begin{equation*}
            \P\!\left(
                \bigg|
                \frac{1}{2\anA\anB} \sum_{t=1}^{2\anA\anB} 
                \Big\{\ftn(\Zt)- \E[\ftn(\Zt)]\Big\}
                \bigg|
            >
                4\Boundtemp\!
                \left[
                \sqrt{
                    \frac{\anA}{n}
                    \logp{
                        \frac{1}{\betaBound(\anA)}
                    }
                }
                +
                \frac{
                    4\anA
                }
                {3n} 
                \logp{
                    \frac{1}{\betaBound(\anA)}
                }
                \right]
            \right)
        \leq
            \frac{n\betaBound(\anA)}{\anA}
        .
        \end{equation*}
\end{lemma}
\begin{proof}
\newcommand{\Ptemp}{\Xi}
    By Lemma \ref{lem:mixing_mapping},
    $\{\ftn(\Zt)\}_{t=1}^{\anA}$ has a $\beta$-mixing coefficient that is less than or equal to $\beta(j)$ from the statement of this lemma.
    %
    By \citet[Theorem 1]{mirzaei_empirical_2025}, if
        $
        2(\anB-1)\beta(\anA)
        <
        \tempB
        <
        1
        ,
        $
    then, 
        \begin{gather*}
            \P\left(
                \bigg|
                \frac{1}{2\anA\anB} \sum_{t=1}^{2\anA\anB} 
                \Big\{\ftn(\Zt)- \E[\ftn(\Zt)]\Big\}
                \bigg|
            >
                \Ptemp(\tempB)
            \right)
        \leq
            \tempB
        , 
        \qquad
        \text{where}
        \\
            \Ptemp(\tempB)
        \coloneqq
            \sqrt{
                \frac{\Boundtemp^2}{\anB}
                \left[
                1+2
                \logp{
                    \frac{2}{\tempB-2(\anB-1)\beta(\anA)}
                }
                \right]
            }
            +
            \frac{
                4\Boundtemp
            }
            {3\anB} 
            \logp{
                \frac{2}{\tempB-2(\anB-1)\beta(\anA)}
            }
            .
        \end{gather*}
    Choose 
        $\tempB = 2\anB\betaBound(\anA)$
    which satisfies the requirements on $\tempB$ since 
        $
        2(\anB-1)\beta(\anA)
        <
            2\anB\betaBound(\anA)
            =
            \tempB
        $
    and
    $b\coloneqq \big\lfloor n/(2\anA)\big\rfloor \leq {n}/({2\anA})$ implies
        $$
            \tempB 
            = 
            2\anB\betaBound(\anA)
        \leq
            \frac{n\betaBound(\anA)}{\anA}
        <
            1
        ,
        $$
    because 
        $
        n\betaBound(\anA)/\anA
        < 1
        $ 
    by assumption. 
    Note that
        $$
            \frac{2}{\tempB-2(\anB-1)\beta(\anA)}
        \leq
            \frac{2}{\tempB-2(\anB-1)\betaBound(\anA)}
        =
            \frac{2}{2\anB\betaBound(\anA)-2(\anB-1)\betaBound(\anA)}
        =
            \frac{1}{\betaBound(\anA)}
        .
        $$
    With this, and 
        $n/(4\anA)
        <
            \anB$,
        \begin{equation*}
        \begin{aligned}
        \Ptemp(\tempB)
        &
        \leq
            2\Boundtemp
            \sqrt{
                \frac{\anA}{n}
                \left[
                1+2
                \logp{
                    \frac{1}{\betaBound(\anA)}
                }
                \right]
            }
            +
            \frac{
                16\Boundtemp\anA
            }
            {3n} 
            \logp{
                \frac{1}{\betaBound(\anA)}
            }
        \\&
        \leq
            4\Boundtemp
            \sqrt{
                \frac{\anA}{n}
                \logp{
                    \frac{1}{\betaBound(\anA)}
                }
            }
            +
            \frac{
                16\Boundtemp\anA
            }
            {3n} 
            \logp{
                \frac{1}{\betaBound(\anA)}
            }
        ,
        \end{aligned}
        \end{equation*}
    where the last inequality uses
        $1+2\log(x)<4\log(x)$ for all $x\geq2$,
    and
        $$
        \betaBound(\anA)
        < 
        \frac{\anA}{n}
        \leq
        \frac{1}{2}  
        ,
        $$
    since 
        $
        n\betaBound(\anA)/\anA
        < 1$
    and 
        $\anA\leq n/2$
    by assumption.
\end{proof}
\vx
}

    The following proposition, preceded by an ancillary lemma, bounds the maximal difference between empirical means and true means of function classes using their Rademacher averages.
    First, Lemma \ref{lem:bennett} uses \citet[Theorem 1.1c]{klein_concentration_2005} to obtain a Bennett type concentration inequality that generalizes similar \iid results, such as \citet[Theorem 2.3]{bousquet_bennett_2002} or \citet[Theorem A.1]{bartlett_local_2005}, to settings with only independence. 
    Then, Proposition \ref{prop:bartsym1} uses Lemma \ref{lem:bennett} to generalize \citet[Theorem 2.1]{bartlett_local_2005} from settings with independent random variables sharing a common distribution, to settings with only independence. Notably, the constants could likely be improved using similar arguments as the proof of \citet[Theorem 2.1]{bartlett_local_2005}.

{
    \renewcommand{\Zt}{\boldsymbol{X}_t}
    \renewcommand{\FMLPg}{\mathcal{F}}
    \newcommand{\func}{h_{tf}}
    \newcommand{\funcB}{\overline{\func}}
    \renewcommand{\Zspace}{\boldsymbol{\mathbb{X}}}
    \renewcommand{\Zbound}{\mathsf{H}_n}
    \newcommand{\Vbound}{\mathsf{V}_n}
    \newcommand{\Vsbound}{\textcolor{blue}{\mathsf{V}}}
    \newcommand{\emp}{\mu_{\funcB}}
    \newcommand{\countsub}{\breve{\FMLPg}}
    \newcommand{\supf}{\sup_{f\in\FMLPg}}
\begin{lemma}\label{lem:bennett}
    Fix $n\in\N$.
    Let $\{\Zt\}_{t\in\N}$ be independent random variables taking values in $\Zspace\subseteq\R^d$. 
    Let $\FMLPg$ be a finite-point separable class of measurable functions $f:\Xspace\to \R$.
    For all $t\in\{1,\ldots,n\}$, $f\in\FMLPg$, $\x\in\Xspace$, let the mapping $(f,t,\x)\mapsto\func(\x)\in\R$ be such that $\func:\Xspace\to \R$ is continuous and $\E[\func(\Zt)]=0$.
    Suppose there exist
    constants $\Zbound,\Vbound>0$, such that 
        $$
            \max_{t\in\{1,\ldots,n\}}
            \supf
            \snorm{\func}\leq \Zbound
        ,
        \quad
        \text{and}
        \quad
            \supf
            \E\left[\sampavg\func(\Zt)^2\right]\leq \Vbound 
        .
        $$
    Then, for any $\delta>0$,
        \begin{equation*}
        \begin{aligned}
            e^{-\delta}
        &
        \geq
            \P\left(
            \!
                \supf
                \sampavg 
                \func(\Zt)
            \geq
                \frac{3}{2}
                \E\left[
                    \supf
                    \sampavg 
                    \func(\Zt)
                \right]\!
                +
                \sqrt{
                    \frac{2\delta \Vbound}{n}
                    }
                + \frac{5\Zbound\delta}{n}
            \!
            \right)\!
        .
        \end{aligned}
        \end{equation*}
\end{lemma}
\begin{proof}
    Define $\funcB \coloneqq \func/\Zbound$, so $\funcB:\Zspace\to [-1,1]$.
    Since
    $\{\Zt\}_{t\in\N}$ are independent
    and
    $\E[\func(\Zt)]=0$
        $$
            \sup_{f\in\FMLPg}\mathrm{Var}\left[\sumin\funcB(\Zt)\right]
        =
            \sup_{f\in\FMLPg}\sumin\mathrm{Var}\left[\,\funcB(\Zt)\right]
        =
            \sup_{f\in\FMLPg}
            \E
            \left[\sumin \frac{\func(\Zt)^2}{\Zbound^2}\right]
        \leq
            \frac{n\Vbound}{\Zbound^2}
        .
        $$
    By assumption $\FMLPg$ is a finite-point separable class so let $\countsub\subseteq \FMLPg$ be a countable dense subset as in Definition \ref{def:point_sep}. 
    Then, by \citet[Theorem 1.1c]{klein_concentration_2005}, for any $\kappa>0$
        \begin{equation*}
        \begin{aligned}
            &
            \P\left(
            \!
                \sup_{f\in\countsub}
                \sumin 
                \funcB(\Zt)
            \geq
                \E\left[
                    \sup_{f\in\countsub}
                    \sumin 
                    \funcB(\Zt)
                    \!
                \right]\!
                +
                \kappa
                \!
            \right)
        \leq
            \exp\!\left(\frac{-\kappa^2}{2\upsilon+3\kappa}\right)\!
        ,
        \\&
        \quad
        \text{where}
        \quad
                \upsilon
            \coloneqq
                2
                \E\left[
                    \sup_{f\in\countsub}
                    \sumin 
                    \funcB(\Zt)
                \right]\!
                +
                \frac{n\Vbound}{\Zbound^2}
        .
        \end{aligned}
        \end{equation*}
    For any $\delta>0$, choose 
    $\kappa = \big(3\delta + \sqrt{9\delta^2+8\delta\upsilon}\big)/2$,
    so $\delta = \kappa^2/(2\upsilon+3\kappa)$ by the quadratic formula, and
        $\kappa \leq 
        \sqrt{2\delta\upsilon}
                +
                3\delta$.
    With this, 
        \begin{equation*}
        \begin{aligned}
            e^{-\delta}
        &
        \geq
            \P\left(
            \!
                \sup_{f\in\countsub}
                \sumin 
                \funcB(\Zt)
            \geq
                \E\left[
                    \sup_{f\in\countsub}
                    \sumin 
                    \funcB(\Zt)
                \right]
                +
                \sqrt{2\delta\upsilon}
                +
                3\delta    
            \!
            \right)\!
        .
        \end{aligned}
        \end{equation*}
    For any $a,b\geq0$, by \citet[Lemma A.3]{bartlett_local_2005}, 
        $\sqrt{a+b}\leq \sqrt{a}+\sqrt{b}$, 
    and for any $c>0$,
        $2\sqrt{ab}\leq ca+b/c$,
    which implies
        \begin{equation*}
        \begin{aligned}
            \E
            &
            \left[
                \sup_{f\in\countsub}
                \sumin 
                \funcB(\Zt)
            \right]
            +
            \sqrt{2\delta\upsilon}
            +
            3\delta    
        \\&
        =
                \E
                \left[
                    \sup_{f\in\countsub}
                    \sumin 
                    \funcB(\Zt)
                \right]
            +
                \sqrt{2\delta
                \left(2
                \E\left[
                    \sup_{f\in\countsub}
                    \sumin 
                    \funcB(\Zt)
                \right]\!
                +
                \frac{n\Vbound}{\Zbound^2}
                \right)}
            + 3\delta    
        \\&
        \leq
            \E
            \left[
                \sup_{f\in\countsub}
                \sumin 
                \funcB(\Zt)
            \right]
            +
            \sqrt{
                4\delta
                \E\left[
                    \sup_{f\in\countsub}
                    \sumin 
                    \funcB(\Zt)
                \right]\!
            }
            +
            \sqrt{
                2\delta
                \frac{n\Vbound}{\Zbound^2}
                }
            + 3\delta   
        \\&
        \leq
            \frac{3}{2}
                \E\left[
                    \sup_{f\in\countsub}
                    \sumin 
                    \funcB(\Zt)
                \right]\!
            +
            \sqrt{
                \frac{2\delta n\Vbound}{\Zbound^2}
                }
            + 5\delta
        .
        \end{aligned}
        \end{equation*}
    Consequently,
        \begin{equation*}
        \begin{aligned}
            e^{-\delta}
        &
        \geq
            \P\left(
            \!
                \sup_{f\in\countsub}
                \sumin 
                \funcB(\Zt)
            \geq
                \frac{3}{2}
                \E\left[
                    \sup_{f\in\countsub}
                    \sumin 
                    \funcB(\Zt)
                \right]\!
                +
                \sqrt{
                    \frac{2\delta n\Vbound}{\Zbound^2}
                    }
                + 5\delta
            \!
            \right)\!
        \\&
        =
            \P\left(
            \!
                \sup_{f\in\countsub}
                \sampavg 
                \frac{\func(\Zt)}{\Zbound}
            \geq
                \frac{3}{2}
                \E\left[
                    \sup_{f\in\countsub}
                    \sampavg 
                    \frac{\func(\Zt)}{\Zbound}
                \right]\!
                +
                \sqrt{
                    \frac{2\delta \Vbound}{\Zbound^2n}
                    }
                + \frac{5\delta}{n}
            \!
            \right)\!
        \\&
        =
            \P\left(
            \!
                \sup_{f\in\countsub}
                \sampavg 
                \func(\Zt)
            \geq
                \frac{3}{2}
                \E\left[
                    \sup_{f\in\countsub}
                    \sampavg 
                    \func(\Zt)
                \right]\!
                +
                \sqrt{
                    \frac{2\delta \Vbound}{n}
                    }
                + \frac{5\Zbound\delta}{n}
            \!
            \right)\!
        .
        \end{aligned}
        \end{equation*}
    This completes the proof because the suprema are unchanged when $\countsub$ is replaced by $\FMLPg$.
\end{proof}
    \vx
}

{
    \renewcommand{\Zt}{\boldsymbol{X}_t}
    \renewcommand{\FMLPg}{\mathcal{F}}
    \newcommand{\func}{f}
    \newcommand{\funcB}{\overline{\func}}
    \renewcommand{\Zspace}{\boldsymbol{\mathbb{X}}}
    \renewcommand{\Zbound}{\mathsf{F}}
    \newcommand{\Vbound}{\mathsf{V}}
    \newcommand{\emp}{\mu_{\funcB}}
    \newcommand{\countsub}{\breve{\FMLPg}}
\begin{proposition}\label{prop:bartsym1}
    Let $\{\Zt\}_{t\in\N}$ be independent random variables taking values in $\Zspace\subseteq\R^d$. 
    Let $\FMLPg$ be a finite-point separable class of measurable functions on $\Zspace$ taking values in $\R$ such that for constants $\Zbound,\Vbound>0$,
        $$
            \sup_{\func\in\FMLPg}
            \snorm{\func}\leq \Zbound
        ,
        \quad
        \text{and}
        \quad
            \sup_{\func\in\FMLPg}
            \E\left[\sampavg\func(\Zt)^2\right]\leq \Vbound 
        .
        $$
    Then, for any $\delta>0$,
        \begin{equation*}
        \begin{aligned}
            e^{-\delta}
        &
        \geq
            \P\left(
            \!
                \sup_{\func\in\FMLPg}
                \sampavg 
                \left\{
                \func(\Zt)-\E[\func(\Zt)]\right\}
            \geq
                3
                \E\left[\mathfrak{R}_n\FMLPg\right]
                +
                \sqrt{
                    \frac{2\delta \Vbound}{n}
                    }
                + \frac{10\Zbound\delta}{n}
            \!
            \right)\!
        ,
        \end{aligned}
        \end{equation*}
    and
        \begin{equation*}
        \begin{aligned}
            2e^{-\delta}
        &
        \geq
            \P\left(
            \!
                \sup_{\func\in\FMLPg}
                \sampavg 
                \left\{
                \func(\Zt)-\E[\func(\Zt)]\right\}
            \geq
                6
                \E_{\xi}\left[
                \mathfrak{R}_n\FMLPg
                \right]
                +
                7
                \sqrt{
                    \frac{2\delta \Vbound}{n}
                    }
                + \frac{34\Zbound\delta}{n}
            \!
            \right)\!
        .
        \end{aligned}
        \end{equation*}
    Similar results also hold for 
                $\sup_{\func\in\FMLPg}
                \sampavg 
                \left\{
                \E[\func(\Zt)]-\func(\Zt)\right\}
                .
                $
\end{proposition}
\begin{proof}
    Let  $h_{tf}\coloneqq f-\E_{\P_{\Xt}}[f]$.
    Note that
        $$
        \E\left[\sampavg h_{tf}(\Zt)^2\right]
        =
        \sampavg \E\left[ \big(\func(\Zt)-\E[\func(\Zt)]\big)^2\right]
        =
        \sampavg\mathrm{Var}[\func(\Zt)]
        \leq \Vbound 
        .
        $$
    Then, applying Lemma \ref{lem:bennett} to $h_{tf}$
        \begin{equation*}
        \resizebox{\hsize}{!}{$
        \begin{aligned}
            e^{-\delta}
        &
        \geq
            \P\left(
            \!
                \sup_{\func\in\FMLPg}
                \sampavg 
                \left\{\func(\Zt)-\E[\func(\Zt)]\right\}
            \geq
                \frac{3}{2}
                \E\left[
                    \sup_{\func\in\FMLPg}
                    \sampavg 
                    \left\{\func(\Zt)-\E[\func(\Zt)]\right\}
                \right]\!
                +
                \sqrt{
                    \frac{2\delta \Vbound}{n}
                    }
                + \frac{10\Fbound\delta}{n}
            \!
            \right)\!
        .
        \end{aligned}
        $}
        \end{equation*}
    By standard symmetrization results%
\footnote{
Note that results such as \citet[Lemma 2.3.1]{van_der_vaart_wellner_book_1996} and \citet{bartlett_local_2005} assume $\{\Zt\}_{t\in\N}$ are independent random variables where each $\Zt$ is distributed according a common $\P$ and are therefore identically distributed. However, the proofs of these results are unchanged if $\{\Zt\}_{t\in\N}$ are only assumed to be independent and the quantity $(\P-\Pn)f$ is replaced by $\textstyle\sampavg \{f(\Xt)-\E[f(\Xt)]\} $.
}
        \begin{equation*}
        \begin{aligned}
            \E
                \left[
                    \sup_{\func\in\FMLPg}
                    \sampavg 
                    \left\{\func(\Zt)-\E[\func(\Zt)]\right\}
                \right]
        \leq
            2\E\left[\mathfrak{R}_n\FMLPg\right]
        \end{aligned}
        \end{equation*}
    The previous two displays imply the first result.

Now consider the second result.
    Define $\funcB\coloneqq{\func}/{\Zbound}$
    By assumption $\FMLPg$ is a finite-point separable class so let $\countsub\subseteq \FMLPg$ be a countable dense subset as in Definition \ref{def:point_sep}. 
    Note that 
        $
        \E\left[
            \xi_t 
            \right]
        =0
        $
    implies
        $$
            \E\left[
                \sumin 
                \xi_t 
                \funcB(\Zt)
                \!
            \right]
        =
            \sumin 
            \E\left[
            \xi_t 
            \right]
            \E\left[
            \funcB(\Zt)
            \!
            \right]
        =
            0
        .
        $$
    Since $\{\Zt\}_{t\in\N}$ are independent
        $$
            \mathrm{Var}
            \left[
                \sumin 
                \xi_t 
                \funcB(\Zt)
                \!
            \right]
        =
            \sumin 
            \mathrm{Var}
            \left[
                \xi_t 
                \funcB(\Zt)
                \!
            \right]
        =
            \sumin 
            \E\left[
                \left(\xi_t 
                \funcB(\Zt)
                \right)^2
                \!
            \right]
        \leq
            \frac{n\Vbound}{\Zbound^2}
        .
        $$
\newcommand{\temp}{\Phi_n}%
    Define 
    $$
            \temp 
        \coloneqq 
            \sup_{\func\in\countsub}
            \sumin 
            \xi_t 
            \funcB(\Zt)
    ,
    \quad
    \text{and}
    \quad        
            \upsilon
        \coloneqq
            2
            \E\left[
                \sup_{\func\in\countsub}
                \sumin 
                \xi_t 
                \funcB(\Zt)
            \right]\!
            +
            \frac{n\Vbound}{\Zbound^2}
    .
    $$
    Then, applying Jensen's inequality followed by \citet[Theorem 1.2a]{klein_concentration_2005} yields, for any $\tau>0$,
        \begin{equation*}
        \begin{aligned}
            \E\left[
            e^{
                -\tau\left(
                    \E_{\xi}[\temp]
                    -
                    \E[\temp]
                \right)
            }
            \right]
        =
            \E\left[
            e^{
                -\tau\E_{\xi}\left[
                    \temp
                    -
                    \E[\temp]
                \right]
            }
            \right]
        \leq
            \E
            \left[
            e^{
                -\tau\left(
                    \temp
                    -
                    \E[\temp]
                \right)
            }
            \right]
        \leq
            \exp\left[
                \frac{\upsilon}{9}
                \left(
                    e^{3\tau}
                    -
                    3\tau
                    -1
                \right)
            \right]
        .
        \end{aligned}
        \end{equation*}
    From the usual Chernoff bound and optimization of constants, this implies,
    for any $\kappa>0$
        \begin{equation*}
        \begin{aligned}
            \exp\!\left(\frac{-\kappa^2}{2\upsilon+2\kappa}\right)
        & \geq
            \P\left(
                \E_{\xi}\left[\temp
                \right]
            \leq
                \E\left[
                    \temp
                \right]
                -
                \kappa
            \right)
        =
            \P\left(
                \E_{\xi}\left[
                \sup_{\func\in\countsub}
                \sumin 
                \xi_t 
                \funcB(\Zt)
                \right]
            \leq
                \E\left[
                    \sup_{\func\in\countsub}
                    \sumin 
                    \xi_t 
                    \funcB(\Zt)
                    \!
                \right]\!
                -
                \kappa
                \!
            \right)
        .
        \end{aligned}
        \end{equation*}
    For any $\delta>0$, choose 
    $\kappa = \big(2\delta + \sqrt{4\delta^2+8\delta\upsilon}\big)/2$,
    so $\delta = \kappa^2/(2\upsilon+2\kappa)$ by the quadratic formula, and
        $\kappa \leq 
        \sqrt{2\delta\upsilon}
                +
                2\delta$.
    With this, 
        \begin{equation*}
        \begin{aligned}
            e^{-\delta}
        &
        \geq
            \P\left(
            \!
                \E_{\xi}\left[
                \sup_{\func\in\countsub}
                \sumin 
                \xi_t 
                \funcB(\Zt)
                \right]
            \leq
                \E\left[
                    \sup_{\func\in\countsub}
                    \sumin 
                    \xi_t 
                    \funcB(\Zt)
                \right]
                -
                \sqrt{2\delta\upsilon}
                -
                2\delta    
            \!
            \right)\!
        .
        \end{aligned}
        \end{equation*}
    For any $a,b\geq0$, by \citet[Lemma A.3]{bartlett_local_2005}, 
        $\sqrt{a+b}\leq \sqrt{a}+\sqrt{b}$, 
    and for any $c>0$,
        $2\sqrt{ab}\leq ca+b/c$,
    which implies   
        \begin{equation*}
        \begin{aligned}
            \E
            &
            \left[
                \sup_{\func\in\countsub}
                \sumin 
                \xi_t 
                \funcB(\Zt)
            \right]
            -
            \sqrt{2\delta\upsilon}
            -
            2\delta    
        \\&
        =
                \E
                \left[
                    \sup_{\func\in\countsub}
                    \sumin 
                    \xi_t 
                    \funcB(\Zt)
                \right]
            -
                \sqrt{2\delta
                \left(2
                \E\left[
                    \sup_{\func\in\countsub}
                    \sumin 
                    \xi_t 
                    \funcB(\Zt)
                \right]\!
                +
                \frac{n\Vbound}{\Zbound^2}
                \right)}
            - 2\delta    
        \\&
        \geq
            \E
            \left[
                \sup_{\func\in\countsub}
                \sumin 
                \xi_t 
                \funcB(\Zt)
            \right]
            -
            \sqrt{
                4\delta
                \E\left[
                    \sup_{\func\in\countsub}
                    \sumin 
                    \xi_t 
                    \funcB(\Zt)
                \right]\!
            }
            -
            \sqrt{
                2\delta
                \frac{n\Vbound}{\Zbound^2}
                }
            - 2\delta   
        \\&
        \geq
            \frac{1}{2}
                \E\left[
                    \sup_{\func\in\countsub}
                    \sumin 
                    \xi_t 
                    \funcB(\Zt)
                \right]\!
            -
            \sqrt{
                \frac{2\delta n\Vbound}{\Zbound^2}
                }
            - 4\delta
        .
        \end{aligned}
        \end{equation*}
    Combining the previous two displays
        \begin{equation*}
        \begin{aligned}
            e^{-\delta}
        &
        \geq
            \P\left(
            \!
                \E_{\xi}\left[
                \sup_{\func\in\countsub}
                \sumin 
                \xi_t 
                \funcB(\Zt)
                \right]
            \leq
                \frac{1}{2}
                \E\left[
                    \sup_{\func\in\countsub}
                    \sumin 
                    \xi_t 
                    \funcB(\Zt)
                \right]\!
            -
            \sqrt{
                \frac{2\delta n\Vbound}{\Zbound^2}
                }
            - 4\delta
            \!
            \right)\!
        \\&
        =
            \P\left(
            \!
                \E_{\xi}\left[
                \sup_{\func\in\countsub}
                \sampavg 
                \xi_t 
                \func(\Zt)
                \right]
            \leq
                \frac{1}{2}
                \E\left[
                    \sup_{\func\in\countsub}
                    \sampavg
                    \xi_t 
                    \func(\Zt)
                \right]\!
            -
            \sqrt{
                \frac{2\delta \Vbound}{n}
                }
            - 
                \frac{4\delta\Zbound}{n}
            \!
            \right)\!
        .
        \end{aligned}
        \end{equation*}
    Thus, since the suprema are unchanged when $\countsub$ is replaced by $\FMLPg$, 
        \begin{equation*}
        \begin{aligned}
            e^{-\delta}
        &
        \geq
            \P\left(
            \!
                3\E\left[\mathfrak{R}_n\FMLPg\right]
            \geq
                6
                \E_{\xi}\left[
                \mathfrak{R}_n\FMLPg
                \right]
            +
            6
            \sqrt{
                \frac{2\delta \Vbound}{n}
                }
            + 
                \frac{24\delta\Zbound}{n}
            \!
            \right)\!
        .
        \end{aligned}
        \end{equation*}
    Combining this with the first result of this lemma,
        \begin{equation*}
        \begin{aligned}
            1-e^{-\delta}
        &
        \leq
            \P\left(
            \!
                \sup_{\func\in\FMLPg}
                \sampavg 
                \left\{
                \func(\Zt)-\E[\func(\Zt)]\right\}
            \leq
                3
                \E\left[\mathfrak{R}_n\FMLPg\right]
                +
                \sqrt{
                    \frac{2\delta \Vbound}{n}
                    }
                + \frac{5\Zbound\delta}{n}
            \!
            \right)\!
        \\&
        \leq
            \P\left(
            \!
                \sup_{\func\in\FMLPg}
                \sampavg 
                \left\{
                \func(\Zt)-\E[\func(\Zt)]\right\}
            \leq
                6
                \E_{\xi}\left[
                \mathfrak{R}_n\FMLPg
                \right]
                +
                7
                \sqrt{
                    \frac{2\delta \Vbound}{n}
                    }
                + \frac{29\Zbound\delta}{n}
            \!
            \right)\!
            +
            e^{-\delta}
        ,
        \end{aligned}
        \end{equation*}
    which proves the second result. 

    To see that the same results hold for 
        $\sup_{\func\in\FMLPg}
        \sampavg 
        \left\{
        \E[\func(\Zt)]-\func(\Zt)\right\}
        ,
        $
    simply apply these results to $-\FMLPg=\{-\func\in\FMLPg\}$ and note that the Rademacher complexities of $\FMLPg$ and $-\FMLPg$ are the same. 
\end{proof}
    \vx
}

    
Lemma \ref{lem:Rad_Contraction} is a contraction inequality for Rademacher complexities of sums that follows from
    \citet[Theorem 3]{maurer_vector_contraction_2016}.
{
\newcommand{\zt}{\z_t}%
\newcommand{\xt}{\x_t}%
\renewcommand{\Ione}[1]{T_{#1}}%
\renewcommand{\Xspace}{\mathbb{X}}%
\renewcommand{\Xspace}{\Zspace}%
\renewcommand{\FMLPg}{\mathcal{S}}
\newcommand{\difb}[2]{h_{#2}(#1)}%
\renewcommand{\dif}[2]{g_{#2}(#1)}%
    \begin{lemma}\label{lem:Rad_Contraction}
        For $\anA,\anB\in\N$, and let $\{\Ione{j}\}_{j=1}^{\anB}$ form a partition of 
        $\{1,\ldots,\anA\anB\}$ 
        such that  $\#\Ione{j}=\anA$ for each $j$.
        Let $\FMLPg$ be a finite-point separable space.
        For $t\in\{1,\ldots,\anA\anB\}$, let $g_t:\FMLPg\to\R$ 
        and $h_t:\FMLPg\to\R$
        be 
        such that
        there exists a constant $L>0$, where
            $\dif{f}{t}-\dif{f'}{t}\leq L|\difb{f}{t}-\difb{f'}{t}|$
        for each $t$ and $f,f'\in\FMLPg$.
        Then,
            \begin{equation*}
            \begin{aligned}
                \E_{\xi}
                \left[
                    \sup_{f\in\FMLPg}
                    \sum_{j=1}^{\anB}
                    \xi_{j}\cdot
                    \bigg(
                        \frac{1}{\anA}
                        \sum_{t\in \Ione{j}}
                        \dif{f}{t}
                    \bigg)
                \right]
            \leq
                \sqrt{2/\anA}\,L\,
                \E_{\xi}
                \left[
                    \sup_{f\in\FMLPg}
                    \sum_{j=1}^{\anB}
                    \sum_{t\in\Ione{j}}
                    \xi_{t}\,
                    \difb{f}{t}
                \right]
            ,
            \end{aligned}
            \end{equation*}
        where 
            $\{\xi_{j}\}_{j=1}^\anB$ 
        and 
            $\{\xi_{t}\}_{t=1}^{\anA\anB}$
        are sequences of \iid Rademacher random variables.
    \end{lemma}
    \begin{proof}
\newcommand{\normE}[1]{\norm{#1}_{E}}%
        For $\x\in\R^{\anA}$ 
        let $\normE{\x}$ denote the Euclidean norm on $\R^\anA$.
        By \citet[Theorem 3]{maurer_vector_contraction_2016}%
\footnote{
There appears to be a typo in the statement of Theorem 3 in \citet{maurer_vector_contraction_2016}, the term $\psi(s_i')$ in the contraction condition should be $\psi_i(s')$,
as in Lemma 7 therein. 
}
        for countable set $\mathcal{H}$ and functions 
            $\psi_j:\mathcal{H}\to \R$,
            $\boldsymbol{\phi}_j:\mathcal{H}\to \R^{\anA}$,
            $j\in\{1,\ldots,\anB\}$
        such that
            $$
                \psi_j(f)-\psi_j(f') 
            \leq
                \normE{\boldsymbol\phi_j(f)-\boldsymbol\phi_j(f')}
            ,
            \quad
            \forall f,f'\in\mathcal{H}, \; j\in\{1,\ldots,\anB\},
            $$
        we have, 
        for 
            $   
                \boldsymbol{\phi}_j
            = 
                 \{\phi_{t}\}_{t\in\Ione{j}}
            ,
            $
            $$
                \E_{\xi}
                \left[
                    \sup_{f\in\mathcal{H}}
                    \sum_{j=1}^{\anB}
                    \xi_j
                    \psi_j(f)
                \right]
            \leq
                \sqrt{2}\,
                \E_{\xi}
                \left[
                    \sup_{f\in\mathcal{H}}
                    \sum_{j=1}^{\anB}
                    \sum_{t\in\Ione{j}}
                    \xi_{t}\,
                    \phi_{t}(f)
                \right].
            $$
        By assumption $\FMLPg$ is finite-point separable so
        choose 
            $\mathcal{H}$
        to be the countable dense subset of $\FMLPg$.
        Let
            \begin{gather*}
                \psi_j(f)
            = 
                \frac{1}{\anA}
                \sum_{t\in \Ione{j}}
                \dif{f}{t}
            ,
            \quad
            \text{ and }
            \quad
                \boldsymbol{\phi}_j(f)
            =
                    \big\{
                    L
                    \difb{f}{t}
                    /\sqrt{\anA}
                    \big\}_{t\in\Ione{j}}
            .
            \end{gather*}
        With this, 
            \begin{equation*}
            \begin{aligned}
                \psi_j(f)-\psi_j(f') 
            &=
                \frac{1}{\anA}
                \sum_{t\in \Ione{j}}
                \big\{
                \dif{f}{t}
                -
                \dif{f'}{t}
                \big\}
            \leq
                \frac{1}{\anA}
                \sum_{t\in \Ione{j}}
                L|\difb{f}{t}-\difb{f'}{t}|
            \\&
            \leq
                \Bigg(
                \frac{1}{\anA}
                \sum_{t\in \Ione{j}}
                \big|L\difb{f}{t}-L\difb{f'}{t}\big|^2
                \Bigg)^{1/2}
            =
                \normE{\boldsymbol\phi_j(f)-\boldsymbol\phi_j(f')}
            .
            \end{aligned}
            \end{equation*}
        Hence, 
        \citet[Theorem 3]{maurer_vector_contraction_2016} implies
            \begin{equation*}
            \begin{aligned}
                    \E_{\xi}
                    \left[
                        \sup_{f\in\mathcal{H}}
                        \sum_{j=1}^{\anB}
                        \xi_j
                        \bigg(
                            \frac{1}{\anA}
                            \sum_{t\in \Ione{j}}
                            \dif{f}{t}
                        \bigg)
                    \right]
                \leq
                    \sqrt{2/\anA}\,L\,
                    \E_{\xi}
                    \left[
                        \sup_{f\in\mathcal{H}}
                        \sum_{j=1}^{\anB}
                        \sum_{t\in\Ione{j}}
                        \xi_{j,t}\,
                        \difb{f}{t}
                    \right].
            \end{aligned}
            \end{equation*}
        This completes the proof because the supremum is unchanged when $\mathcal{H}$ is replaced by $\FMLPg$.
    \end{proof} 
}
    \vx

    Now we give Proposition \ref{prop:rad_dudley_bound}, preceded by a definition and two ancillary lemmas, which provide bounds for the Rademacher complexities using the pseudo-dimension. 
    The first ancillary lemma uses standard chaining arguments to bound the Rademacher complexity with a metric entropy integral; the second lemma is a consequence of \citet[Theorem 12.2]{anthony_bartlett_neural_1999} which bounds the covering number with pseudo-dimension.

    \begin{definition}\label{def:CoveringNumber}%
\newcommand{\temp}{\mathbb{M}}%
\newcommand{\tempsub}{\mathcal{G}}%
\renewcommand{\Z}{\X}%
\renewcommand{\Zspace}{\Xspace}%
    \textbf{\textsc{(Covering Number)}}
    Let 
        $\delta>0$, 
    and let 
        $(\temp,\norm{\cdot})$
    be a 
    metric space.
    \begin{enumerate}
        \item 
            For $\tempsub\subset \temp$, 
            the $\boldsymbol{\delta}$\textbf{-covering number} of $\tempsub$, 
                denoted as $\cover(\delta,\tempsub,\norm{\cdot})$,
            is the smallest $J\in\N$ such that there exists a collection
                $
                \{m_j\}_{j=1}^J
                \subseteq
                \temp,
                $
            where
                $$
                    \tempsub
                \;\subseteq\;
                    \bigcup_{j=1}^J 
                    \Big\{
                        g\in\tempsub
                        :\,
                        \norm{g-m_j}<\delta
                    \Big\}
                ,
                $$
            and if no such $J\in\N$ exists let 
                $\cover(\delta,\tempsub,\norm{\cdot})=\infty$.
        \item 
            When 
                $\temp$
            is a space of
            functions with elements $f:\Zspace\to \R$, 
            then, for any 
                $a\in\N$,
            define
                $  
                    \temp|_{\{\Zt\}_{t=1}^a}
                \coloneqq
                    \big\{
                        \big(f(\Z_1), f(\Z_2), \ldots, f(\Z_a)\big)
                        :
                        f\in\temp
                    \big\}
                .
                $
            For any 
            $r\geq 1$
            let
                $
                    \cover^{(\infty)}_{r}(\delta,\temp,a)
                \coloneqq
                    \sup
                    \Big\{
                    \cover
                    \big(
                        \delta,\temp|_{\{\Z_t(\omega)\}_{t=1}^a},\norm{\cdot}_{r,a}
                    \big)
                    :
                    \omega\in\Omega
                    \Big\}
                .
                $
    \end{enumerate}
    \end{definition}
    For 
    $\mathcal{G}\subset\measurableftns$
    and
    a sample $\{\Zt\}_{t=1}^a$, note that 
        $
        \cover
        \big(
            \delta,\mathcal{G}|_{\{\Z_t\}_{t=1}^a},\norm{\cdot}_{r,a}
        \big)
        $
    depends on $\omega\in\Omega$, but 
        $\cover^{(\infty)}_{r}(\delta,\mathcal{G},a)$
    does not.
    Following the usual convention, if $\mathcal{G}=\emptyset$ then
    $\cover(\delta,\mathcal{G},\norm{\cdot})=1$, for any $\delta>0$.

{
\renewcommand{\Diam}{r}%
\newcommand{\Rad}[1]{\mathfrak{R}_{#1}}%
\renewcommand{\anA}{n}
\renewcommand{\normT}[1]{\norm*{#1}}%
\newcommand{\ft}{f_t}%
\renewcommand{\rO}{r}%
\renewcommand{\FMLPg}{\mathcal{G}}%
\newcommand{\ftn}{f}%
\renewcommand{\sumin}{\sum_{t=1}^{\anA}}%
\renewcommand{\Zspace}{\R^d}%
\newcommand{\EPspace}{\R^{d\times\anA}}%
\renewcommand{\z}{\x}%
\renewcommand{\Z}{\X}%
    \begin{lemma}\label{lem:Dudley_entropy}%
        Let $\FMLPg$ be a finite-point separable set of functions with elements $\ftn:\Zspace\to\R$ such that $\norm{\ftn}_{\infty}< \infty$ for each $\ftn\in\FMLPg$. 
        Then, for any $\anA\in\N$ 
        we have
        \begin{equation*}
        \begin{aligned}
            \E_{\xi}
            \Big[
                \Rad{\anA}
                \Big\{
                    \ftn:
                    \ftn\in\FMLPg,
                    \,
                    \normT{\ftn} \leq \Diam
                \Big\}
            \Big] 
        &
        \leq
            \inf_{0<\alpha<\Diam}
            \Bigg\{
                4\alpha
                +
                8\sqrt{\frac{2}{\anA}}
                \int_{\alpha}^{\Diam}
                \sqrt{
                     \log 
                    \cover(
                        \upsilon,\FMLPg 
                        ,\normT{\cdot} 
                    )
                    }
            \,d\upsilon
            \Bigg\}
        ,
        \end{aligned}
        \end{equation*}
    where 
        $
        \textstyle
            \normT{f} = \left(\sampavg f(\Zt)^2\right)^{1/2}
        $.
    \end{lemma}
    \begin{proof}
        Consider the case where 
        $\{\Zt\}_{t=1}^{\anA} =\{\z_t\}_{t=1}^{\anA}$ is an arbitrary fixed element of 
        $\R^{d\times \anA}$.
        Then 
            $
            \textstyle
                \Big\{
                    \frac{1}{\sqrt{\anA}}
                    \sumin
                    \xi_t
                    \ftn(\z_t)
                    :
                    \ftn\in\FMLPg,\;
                    \normT{\ftn} 
                    \leq 
                        \Diam
                \Big\}
            $
        is a zero mean sub-Gaussian empirical process; since
        Hoeffding's inequality for Rademacher random variables implies,%
\footnote{
This result is a simple modification of \citet[Theorem 2]{hoeffding_probability_1963}, and can be found in \citet[Lemma 2.2.7]{van_der_vaart_wellner_book_1996}.
} 
        \begin{equation*}
        \begin{aligned}
            \Pr\left(
                \frac{1}{\sqrt{\anA}}
                    \sumin
                    \xi_t
                    \ftn(\z_t)
                \geq
                    \upsilon
            \right)
        &\leq
            2\exp\left[
                \frac
                    {-\upsilon^2 \anA}
                    {2\sumin
                    \ftn(\z_t)^2
                    }
            \right]
        \leq
            2\exp\left[
                \frac
                    {-\upsilon^2}
                    {
                    2\normT{\ftn}^2
                    }
            \right]
        ,
        \end{aligned}
        \end{equation*}
    for any 
        $\ftn\in\big\{\ftn\in\FMLPg,\normT{\ftn} \leq \Diam\big\}$
    and $\upsilon>0$.
    Hence, Dudley's entropy integral can be applied%
\footnote{
This is a well-known result with many formulations. This version, and it's proof can be found in 
    \citet[p. 11]{bartlett_theoretical_2013},
or in the proof of \citet[Theorem 2.2.4]{van_der_vaart_wellner_book_1996}.
This result is sometimes referred to as Dudley's chaining (e.g. \citealp[Lemma 3]{farrell_deep_2021}). 
}
    to obtain, for each $\alpha\in(0,\Diam)$
        \begin{equation*}
        \begin{aligned}
        \E_{\xi}
            \Big[
        &
                \Rad{\anA}
                \Big\{
                    \ftn:
                    \ftn\in\FMLPg|_{\{\Zt\}_{t=1}^\anA},
                    \,
                    \normT{\ftn} \leq \Diam
                \Big\}
            \Big] 
        =
            \frac{1}{\sqrt{\anA}}\,
            \E_{\xi}
            \left[
            \sup_{\{\ftn\in\FMLPg,\normT{\ftn} \leq \Diam\}}
                \frac{1}{\sqrt{\anA}}
                    \sumin
                    \xi_t
                    \ftn(\z_t)
            \right]
        \\&
        \leq
            \frac{1}{\sqrt{\anA}}
            \left(
            2\E_{\xi}
            \left[
            \sup_{\{\ftn\in\FMLPg,\normT{\ftn} \leq 2\alpha\}}
                \frac{1}{\sqrt{\anA}}
                    \sumin
                    \xi_t
                    \ftn(\z_t)
            \right]
            +
            8\sqrt{2}
            \int_{\alpha}^{\Diam}
                \sqrt{
                     \log 
                    \cover(
                        \upsilon,\FMLPg 
                        ,\normT{\cdot} 
                    )
                    }
                \,d\upsilon
            \right)
        \\&
        \leq
            2\sup_{\{\ftn\in\FMLPg,\normT{\ftn} \leq 2\alpha\}}
                \left\{
                \frac{1}{{\anA}}
                    \sumin
                    \big|\ftn(\z_t) 
                    \big|
            \right\}
            +
            8\sqrt{\frac{2}{\anA}}
            \int_{\alpha}^{\Diam}
                \sqrt{
                     \log 
                    \cover(
                        \upsilon,\FMLPg 
                        ,\normT{\cdot} 
                    )
                    }
                \,d\upsilon
        \\&
        \leq
            4\alpha
            +
            8\sqrt{\frac{2}{\anA}}
            \int_{\alpha}^{\Diam}
                \sqrt{
                     \log 
                    \cover(
                        \upsilon,\FMLPg 
                        ,\normT{\cdot} 
                    )
                    }
                \,d\upsilon.
        \end{aligned}
        \end{equation*}
    The desired result follows since this holds for any 
        $\{\z_t\}_{t=1}^\anA\in\R^{d\times \anA}$ 
    and 
        $\alpha\in(0,\Diam)$.
    \end{proof}
    \vx
}

{
    \renewcommand{\FMLPg}{\mathcal{G}}%
    \newcommand{\Boundtemp}{B}%
    \newcommand{\an}{n}%
    \newcommand{\ftn}{f}%
    \begin{lemma}\label{lem:entropy_bound}%
        Let $\FMLPg$ be a set of real-valued functions such that 
        $
        \sup_{\ftn\in\FMLPg}\norm{\ftn}_{\infty}\leq \Boundtemp
        $. 
        Then,
        for any $r\in[1,\infty]$, $\delta\in(0,2\Boundtemp]$, and $\an\in\N$ such that $\an\geq\Pdim(\FMLPg)$,
            $$
                \cover^{(\infty)}_{r}
                \hspace{-3pt}
                \left(
                    \delta
                    ,\,
                    \FMLPg
                    ,\,
                    \an
                \right)
            \leq
                \left( 
                    \frac%
                    {2e\Boundtemp \an}%
                    {\delta \cdot \Pdim(\FMLPg)}
                \right)^{\Pdim(\FMLPg)}
            .
            $$
    \end{lemma}
    \begin{proof}
        By \citet[Theorem 12.2]{anthony_bartlett_neural_1999}
            $$
                \cover^{(\infty)}_{\infty}
                \hspace{-3pt}
                \left(
                    \delta
                    ,\,
                    \FMLPg
                    ,\,
                    \an
                \right)
            \leq
                \left( 
                    \frac%
                    {2e\Boundtemp \an}%
                    {\delta \cdot \Pdim(\FMLPg)}
                \right)^{\Pdim(\FMLPg)}
            .
            $$
    For any $r\in[1,\infty]$, $\an\in\N$, and $\x\in\R^{\an}$, we have
        $\norm{\x}_{r,\an}\leq\norm{\x}_{\infty,\an}$;
    so any $\delta$-cover with respect to $\norm{\cdot}_{\infty,\an}$ is also a cover for $\norm{\cdot}_{r,\an}$.
    Thus, 
        $
            \cover^{(\infty)}_{r}
            \big(
                \delta
                ,\,
                \FMLPg
                ,\,
                \an
            \big)
        \leq
            \cover^{(\infty)}_{\infty}
            \big(
                \delta
                ,\,
                \FMLPg
                ,\,
                \an
            \big)
        .
        $
    \end{proof}
    \vx
}

{
\renewcommand{\Diam}{\delta}%
\newcommand{\Boundtemp}{B}%
\newcommand{\Rad}[1]{\mathfrak{R}_{#1}}%
\renewcommand{\rO}{\delta}%
\renewcommand{\FMLPg}{\mathcal{G}}%
\newcommand{\ftn}{f}%
\renewcommand{\Zspace}{\boldsymbol{\mathbb{X}}}%
\newcommand{\an}{a}%
\renewcommand{\normTi}[2]{\norm*{#2}_{2,\an}}%
\renewcommand{\z}{\x}%
\renewcommand{\Z}{\X}%
    \begin{proposition}\label{prop:rad_dudley_bound}
        Let $\Zspace \subseteq \R^d$, and let $\FMLPg$ be a finite-point separable set of functions with elements $\ftn:\Zspace\to\R$ such that        
            $\sup_{\ftn\in\FMLPg}\norm{\ftn}_{\infty}< \Boundtemp$ for some $\Boundtemp>0$. 
        Then,
        for any 
        $\Diam\in(0,2\Boundtemp]$, and $\an\in\N$ such that 
        $\an\geq\Pdim(\FMLPg)$,
            \begin{equation*}
            \begin{aligned}[b]
            \E_{\xi}
                \Big[
                    \Rad{\an}
                    \Big\{
                        \ftn\in\FMLPg:
            &
                        \normTi{i}{\ftn} \leq \Diam
                    \Big\}
                \Big] 
            \leq
                12\Diam
                \sqrt{
                    \frac{ 2\,\Pdim(\FMLPg)}{\an}
                      \cdot
                      \logp{
                            \frac%
                            {2e^2\Boundtemp \an}%
                            {\Diam \cdot \Pdim(\FMLPg)}
                        }
                }
            .
            \end{aligned}
            \end{equation*}
    \end{proposition}
    \begin{proof}
    Using Lemma \ref{lem:Dudley_entropy} then Lemma \ref{lem:entropy_bound}
        \begin{equation*}
        \begin{aligned}
            \E_{\xi}
            \Big[
                \Rad{\an}
                \Big\{
                    \ftn\in\FMLPg:
                    \normTi{i}{\ftn} \leq \Diam
                \Big\}
            \Big] 
        &
        \leq
            \inf_{0<\alpha<\Diam}
            \Bigg\{
                4\alpha
                +
                8\sqrt{\frac{2}{\an}}
                \int_{\alpha}^{\Diam}
                \sqrt{
                    \log 
                    \cover(
                        \upsilon,\FMLPg
                        ,\normTi{i}{\cdot} 
                    )
                    }
            \,d\upsilon
            \Bigg\}
        \\&
        \leq
            \inf_{0<\alpha<\Diam}
            \Bigg\{
                4\alpha
                +
                8\sqrt{\frac{2}{\an}}
                \int_{\alpha}^{\Diam}
                \sqrt{
                    \log 
                    \cover^{(\infty)}_{2}(
                        \upsilon,\FMLPg,
                        \an
                    )
                    }
            \,d\upsilon
            \Bigg\}
        \\&
        \leq
            \inf_{0<\alpha<\Diam}
            \Bigg\{
                4\alpha
                +
                8\sqrt{\frac{2}{\an}}
                \int_{\alpha}^{\Diam}
                \sqrt{
                    \Pdim(\FMLPg)
                    \cdot
                    \log\!
                    \left( 
                        \frac%
                        {2e\Boundtemp \an}%
                        {\upsilon \cdot \Pdim(\FMLPg)}
                    \right)
                    }
            \,d\upsilon
            \Bigg\}
        .
        \end{aligned}
        \end{equation*}
    Clearly, 
        $
        \log\!
        \left( 
            \frac%
            {2e\Boundtemp \an}%
            {\upsilon \cdot \Pdim(\FMLPg)}
        \right)
        >
        0
        $
    for any $\upsilon \in (0, \Diam]$,
    since 
    $\Diam\in(0,2\Boundtemp]$ 
    and
    $\an\geq\Pdim(\FMLPg)$
    by assumption.
    Then, by the Cauchy-Schwarz inequality,%
\footnote{
For any $f:\R\to\R$ and $a,b\in\R$ such that $a<b$ and $f(x)\geq0$ for all $x\in[a,b]$, by the Cauchy-Schwarz inequality 
$$ 
\textstyle
\int_{a}^{b}\sqrt{f(x)}\;dx
\leq 
\Big(\int_{a}^{b}f(x)\;dx\Big)^{\frac{1}{2}}\Big(\int_{a}^{b}1\;dx\Big)^{\frac{1}{2}}
=
\sqrt{b-a}\Big(\int_{a}^{b}f(x)\;dx\Big)^{\frac{1}{2}}
.
$$
}
        \begin{equation*}
        \begin{aligned}[b]
        \E_{\xi}
            \Big[
                \Rad{\an}
                \Big\{
                    \ftn\in\FMLPg:
        &
                    \normTi{i}{\ftn} \leq \Diam
                \Big\}
            \Big] 
        \leq
            \inf_{0<\alpha<\Diam}
            \Bigg\{
                4\alpha
                +
                8\sqrt{
                    \frac{2\Diam \cdot \Pdim(\FMLPg)}{\an}
                    \int_{\alpha}^{\Diam}
                        \logp{
                            \frac%
                            {2e\Boundtemp \an}%
                            {\upsilon \cdot \Pdim(\FMLPg)}
                        }
                    \,d\upsilon
                }\,
            \Bigg\}
        \\&
        =
            \inf_{0<\alpha<\Diam}
            \Bigg\{
                4\alpha
                +
                8
                \sqrt{
                    \frac{ 2\Diam \cdot \Pdim(\FMLPg)}{\an}
                      \Bigg[  
                      \upsilon
                      \cdot
                      \logp{
                            \frac%
                            {2e^2\Boundtemp \an}%
                            {\upsilon \cdot \Pdim(\FMLPg)}
                        }
                \Bigg]_{\upsilon=\alpha}^{\Diam}
                }\,
                \Bigg\}
        \\&
        \leq
            4\Diam \sqrt{\frac{ \Pdim(\FMLPg)}{\an}}
            +
            8
            \sqrt{
                \frac{ 2\Diam \cdot \Pdim(\FMLPg)}{\an}
                \Bigg[  
                \upsilon
                \cdot
                \logp{
                    \frac%
                    {2e^2\Boundtemp \an}%
                    {\upsilon \cdot \Pdim(\FMLPg)}
                }
            \Bigg]_{\upsilon=\Diam \sqrt{\frac{ \Pdim(\FMLPg)}{\an}}}^{\Diam}
            }
        \\&
        \leq
            4\Diam \sqrt{\frac{ \Pdim(\FMLPg)}{\an}}
            +
            8\Diam
            \sqrt{
                \frac{ 2\,\Pdim(\FMLPg)}{\an}
                  \cdot
                  \logp{
                        \frac%
                        {2e^2\Boundtemp \an}%
                        {\Diam \cdot \Pdim(\FMLPg)}
                    }
            }
        \\&
        \leq
            12\Diam
            \sqrt{
                \frac{ 2\,\Pdim(\FMLPg)}{\an}
                  \cdot
                  \logp{
                        \frac%
                        {2e^2\Boundtemp \an}%
                        {\Diam \cdot \Pdim(\FMLPg)}
                    }
            }
        ,
        \end{aligned}
        \end{equation*}
    where the second inequality chooses 
        $
        \alpha =
        \Diam \sqrt{{ \Pdim(\FMLPg)}/{\an}}
        \in (0,\Diam)
        $,
    and the last inequality uses 
        $$
            \logp{
                \frac%
                {2e^2\Boundtemp \an}%
                {\Diam \cdot \Pdim(\FMLPg)}
            }
        \geq
            \logp{
                \frac%
                {e^2\an}%
                { \Pdim(\FMLPg)}
            }
        \geq
            \logp{
                e^2
            }
        > 1
        ,
        $$
    since 
    $\Diam\in(0,2\Boundtemp]$ 
    and
    $\an\geq\Pdim(\FMLPg)$
    by assumption.
    \end{proof}
    \vx
}

}

\section{Proofs for Section 
\texorpdfstring{\ref{sec:FCDNN}}{3}}\label{sec:PROOF_SEC_DNN}

Before proceeding with the 
Proofs for Section \ref{sec:FCDNN},
we provide Lemma \ref{lem:Ptrunc_truncf_env_verify} which provides bounds on $\snormDNNBound{\fO}$, $\P(\Ptrunc)$, and $\truncf$; as well as
Theorem \ref{thrm:FCDNN_ROC_GEN} a more general FCDNN result which will be used to prove Theorem \ref{thrm:FCDNN_ROC}. All other ancillary results can be found in Appendix \ref{sec:FCDNN_sup_res}.

    \begin{lemma}\label{lem:Ptrunc_truncf_env_verify}
    Let $n\geq 3$,
    and
    $\Zbound
    \coloneqq\Ybound\times\Xbound
    =
    [-\Yboundval,\Yboundval]\times[-\Xboundval,\Xboundval]^{d}
    $
    where 
    $\Xboundval$ and $\Yboundval$ are chosen as in \eqref{eq:xbound_ybound_choice}.
    Let $\CfObound$ be defined as in \eqref{eq:CfObound}.
    \begin{enumerate}
        \item 
            If Assumption \ref{as:data_holder} holds, then
            for any $0\leq \norm{\kb}_{1}\leq \smooth-\holE$ and $\Xboundval\geq 1$
            \begin{equation}\label{eq:fObound}
                \sup_{\x\in[-\Xboundval,\Xboundval]^d}
                |
                D^{\kb}
                \fO(\x)
                |
            \leq
                |
                D^{\kb}
                \fO(0)
                |
                +
                \sqrt{d}\,
                \Xboundval
            .
            \end{equation}
        \item 
    If Assumption \ref{as:data_subg} holds,
    then 
            $
                \P(\Ptrunc)
            \leq
                2n^{1-\Csubg\log(n)}
            $
    for all $n$ sufficiently large. 
        \item 
            If 
            $\FMLPg=\FMLP$
            and 
            $\DNNBound\geq \snormDNNBoundval{\fO}$
            then 
            \begin{equation*}
            \truncf
            \leq
                \max_{t\in\{1,\ldots,n\}}\Big(
                \DNNBound^2
                \Pr\left(\Xt\notin\Xbound\right)
                +
                \E\Big[
                    |\fO(\Xt)|^2
                    \mathbbm{1}_{\{\Xt\notin\Xbound\}}\!
                \Big]
                \Big)
            .
            \end{equation*}
        If
            Assumptions \ref{as:data_holder} and \ref{as:data_subg} hold, 
        and $\DNNBound=\CDNNBound\Xboundval$ for any $\CDNNBound\geq \CfObound$, 
            then
        $\DNNBound\geq \snormDNNBoundval{\fO}$
        and
        $
            \truncf
        \lesssim n^{-1}
        .
        $
    \end{enumerate}
    \end{lemma}

\begin{proof}[Proof of Lemma \ref{lem:Ptrunc_truncf_env_verify}(i)]
    For any $\fO$ satisfying Assumption \ref{as:data_holder}, and $\x\in\Xspace$
        $$
            \max\{
                |D^{\kb} \fO(\x)-D^{\kb} \fO(0)|
                :
                \norm{\kb}_{1}\leq \smooth-\holE
            \}
        \leq
            \norm{\x}_{\lp{2}}^\holE
        ,
        $$
    for all $\x\in\Xspace$.
    Then, 
    using the triangle inequality and $\norm{\x}_{\lp{2}}\leq\sqrt{d}\norm{\x}_{\lp{\infty}}$ for all $\x\in\R^d$,
    for any $0\leq \norm{\kb}_{1}\leq \smooth-\holE$
        \begin{equation*}
            |
            D^{\kb}
            \fO(\x)
            |
        \leq
            |
            D^{\kb}
            \fO(0)
            |
            +
            (\sqrt{d}\,\norm{\x}_{\lp{\infty}})^\holE
        .
        \end{equation*}
    Thus, 
        $
        \snormDNNBoundval{\fO}
        \leq
            |
            \fO(0)
            |
            +
            (\sqrt{d}\,\Xboundval)^\holE
        \lesssim
        \Xboundval
        $
    for all $\Xboundval\geq 1$,
    since 
        $
            |
            D^{\kb}
            \fO(0)
            |
        $
    is constant, 
    and $\holE\in(0,1]$.
\end{proof}
\vx

\begin{proof}[Proof of Lemma \ref{lem:Ptrunc_truncf_env_verify}(ii)]
%
    If $\Xboundval=\Yboundval=1$, then $\P(\Zt \notin \Zbound)=0$ follows immediately.
    
    Consider the case where $\Xboundval=\Yboundval=\log(n)$, and let $\Zboundval=\log(n)$.
    By 
    Assumption \ref{as:data_subg},
        $
        \Pr\big(
            \Zt \notin \Zbound
        \big)
        \leq
            2e^{-\Csubg\log(n)^2}
        =
            2n^{-\Csubg\log(n)}
        $
    for each $t=1,\ldots,n$, and all $n$ sufficiently large.
    Thus, 
    by subadditivity of measures
        \begin{equation*}
        \begin{aligned}
            \P(\Ptrunc)
        =
            \P\left(
                \bigcup_{t\in\{1,\ldots,n\}}
                \Big\{
                \Zt \notin \Zbound
                \Big\}
            \right)
        \leq
            \sumin
            \P\big(
                \Zt \notin \Zbound
            \big)
        \leq
            2n^{1-\Csubg\log(n)}
        .
        \end{aligned}
        \end{equation*}

    Consider the case where $\Xboundval=\log(n)$ and $\Yboundval=1$.
    By 
    Assumption \ref{as:data_subg}, for each $t=1,\ldots,n$, 
        \begin{equation*}
            \Pr\big(
                \Zt \notin \Zbound
            \big)
        \leq
            \Pr\Big(
                \Xt \notin[-\Xboundval,\Xboundval]^{d}
            \Big)
            +
            \Pr\Big(
                \Yt \notin[-\Yboundval,\Yboundval]
            \Big)
        =
            \Pr\Big(
                \Xt \notin[-\Xboundval,\Xboundval]^{d}
            \Big)
        ,
        \end{equation*}
    and the result follows by the same argument used in the previous case. 
    Finally, the case where $\Xboundval=1$ and $\Yboundval=\log(n)$ follows similarly.
    \end{proof}
\vx

\begin{proof}[Proof of Lemma \ref{lem:Ptrunc_truncf_env_verify}(iii)]
    If $\Xboundval=1$, then $\truncf=0$ follows immediately.

    Consider the case where $\Xboundval=\log(n)$.
    Let $t\in \{1,\ldots,n\}$ be arbitrary. Note that
        \begin{equation}\label{eq:unifintf1}
        \begin{aligned}[b]
            \sup_{f\in\left\{\FMLP\cup\{\fO\}\right\}}\!
            \E\Big[
                |f(\Xt)|^2
                \mathbbm{1}_{\{\Xt\notin\Xbound\}}\!
            \Big]
        &\leq
            \E\Big[
                    |\fO(\Xt)|^2
                    \mathbbm{1}_{\{\Xt\notin\Xbound\}}\!
                \Big]
            +
            \sup_{f\in\FMLP}\!
                \E\Big[
                    |f(\Xt)|^2
                    \mathbbm{1}_{\{\Xt\notin\Xbound\}}\!
                \Big]
        \\&
        \leq
                \E\Big[
                    |\fO(\Xt)|^2
                    \mathbbm{1}_{\{\Xt\notin\Xbound\}}\!
                \Big]
            +
            \DNNBound^2
            \Pr\left(\Xt\notin\Xbound\right)
        ,
        \end{aligned}
        \end{equation}
    which implies the first result.
    For the second result, first we have
        \begin{equation}\label{eq:unifintf2}
        \begin{aligned}[b]
            \E\Big[
        &
                |\fO(\Xt)|^2
                \mathbbm{1}_{\{\Xt\notin\Xbound\}}
            \Big]
        \\
        &
        =
            \E\Big[
                |\fO(\Xt)|^2
                \mathbbm{1}_{\{\Xt\notin\Xbound\}}
                \mathbbm{1}_{\{|\fO(\Xt)|\leq \DNNBound\}}
            \Big]
            +
            \E\Big[
                |\fO(\Xt)|^2
                \mathbbm{1}_{\{\Xt\notin\Xbound\}}
                \mathbbm{1}_{\{|\fO(\Xt)| > \DNNBound\}}
            \Big]
        \\&
        \leq
            \DNNBound^2
            \Pr\left(\Xt\notin\Xbound\right)
            +
            \E\Big[
                |\fO(\Xt)|^2
                \mathbbm{1}_{\{\Xt\notin\Xbound\}}
                \mathbbm{1}_{\{|\fO(\Xt)| > \DNNBound\}}
            \Big]
        .
        \end{aligned}
        \end{equation}
    By \eqref{eq:fObound},
    for any $\Xboundval\geq1$ and $\z = (y,\x^{\tr})^\tr\in\Zspace$ 
    such that 
    $\x\notin\Xbound=[-\Xboundval,\Xboundval]^d$,
    we have
        $$
            |\fO(\x)|
        \leq
            \sup_{
                \x'\in [-\norm{\x}_{\lp{\infty}},\,\norm{\x}_{\lp{\infty}}]^d
            }
            |\fO(\x')|
        \leq
            |
            \fO(0)
            |
            +
            (\sqrt{d}\,\norm{\x}_{\lp{\infty}})^\holE
        \leq
            \CfObound \norm{\x}_{\lp{\infty}}
        \leq
            \CfObound \norm{\z}_{\lp{\infty}}
        .
        $$
    Furthermore, for any $z>0$
    note that 
        $
         \Zt\notin [-z,z]^{\dz}
        $
    if and only if 
        $
        \norm{\Zt}_{\lp{\infty}}>z
        $,
    so Assumption \ref{as:data_subg} implies
        $
        \Pr
        \big(
            \Zt\notin [-z,z]^{\dz}
        \big)
        =
        \Pr
        \big(
            \norm{\Zt}_{\lp{\infty}}
            >z
        \big)
        $.
    With this and the previous display, Lemma \ref{lem:subgausexp} can be applied to obtain
        $$
            \E\left[
                e^{2|\fO(\Xt)|}
                \mathbbm{1}_{\{\Xt\notin\Xbound\}}
            \right]
        \,\leq\,
            \E\left[
                e^{2\CfObound\norm{\Zt}_{\lp{\infty}}}
            \right]
        \,\leq\,
            1+4\CfObound\sqrt{\pi/\Csubg}\,e^{\CfObound^2/\Csubg}
        ,
        $$
    and
        \begin{equation*}
        \begin{aligned}
            \E\Big[
            e^{2|\fO(\Xt)|}
            \mathbbm{1}_{\{\Xt\notin\Xbound\}}
            \Big]
        &
        \geq
            \E\Big[
            e^{2|\fO(\Xt)|}
            \mathbbm{1}_{\{\Xt\notin\Xbound\}}
            \mathbbm{1}_{\{|\fO(\Xt)| > \DNNBound\}}
            \Big]
        \\&
        \geq
            e^{\DNNBound}
            \E\Big[
            e^{|\fO(\Xt)|}
            \mathbbm{1}_{\{\Xt\notin\Xbound\}}
            \mathbbm{1}_{\{|\fO(\Xt)| > \DNNBound\}}
            \Big]
        \\&
        \geq
            e^{\DNNBound}
            \E\Big[
                |\fO(\Xt)|^2
                \mathbbm{1}_{\{\Xt\notin\Xbound\}}
                \mathbbm{1}_{\{|\fO(\Xt)| > \DNNBound\}}
            \Big]
        .
        \end{aligned}
        \end{equation*}
    The previous two displays imply
        $$
            \E\Big[
                |\fO(\Xt)|^2
                \mathbbm{1}_{\{\Xt\notin\Xbound\}}
                \mathbbm{1}_{\{|\fO(\Xt)| > \DNNBound\}}
            \Big]
        \leq
            \frac{
                \E\Big[
                e^{2|\fO(\Xt)|}
                \mathbbm{1}_{\{\Xt\notin\Xbound\}}
                \Big]
            }{
                e^{\DNNBound}
            }
        \leq
            \frac{
                1+4\CfObound\sqrt{\pi/\Csubg}\,e^{\CfObound^2/\Csubg}
            }{
                e^{\DNNBound}
            }
        .
        $$
    Thus, with this, \eqref{eq:unifintf2}, and \eqref{eq:unifintf1}
        \begin{equation*}
        \begin{aligned}[b]
            \sup_{f\in\left\{\FMLP\cup\{\fO\}\right\}}\!
            \E\Big[
                |f(\Xt)|^2
                \mathbbm{1}_{\{\Xt\notin\Xbound\}}\!
            \Big]
        &\leq
            2\DNNBound^2
            \Pr\left(\Xt\notin\Xbound\right)
            +
            \frac{
                1+4\CfObound\sqrt{\pi/\Csubg}\,e^{\CfObound^2/\Csubg}
            }{
                e^{\DNNBound}
            }
        ,
        \end{aligned}
        \end{equation*}
    The result follows since 
        $
            \snormDNNBound{\fO} 
        \leq
            (
                |\fO(0)|
                +\sqrt{d}
            )\Xboundval
        \leq
        \CfObound \Xboundval
        $
    by \eqref{eq:fObound},
    and $\DNNBound=\CDNNBound\Xboundval$ where $\CDNNBound\geq \CfObound$,
    so
    $\DNNBound\geq \snormDNNBound{\fO}$.
\end{proof}
\vx

{
\begin{theorem}\label{thrm:FCDNN_ROC_GEN}
    Suppose Assumption \ref{as:data_holder}  holds.
    Let 
        $
        \Zbound
        =
        [-\Yboundval,\Yboundval]\times[-\Xboundval,\Xboundval]^{d}
        $
    for some 
        $\{\Yboundval\}_{n\in\N}$, $\{\Xboundval\}_{n\in\N}$ 
    where 
        $\Yboundval\geq 0$ 
        and 
        $1 \leq \Xboundval \leq \sqrt{n}$
    for all $n$.
    Let
        $
            \FMLPg
        =
            \FMLPnon(\L,\Hb,\DNNBound)
        $
    be such that \eqref{eq:width_bound_order} and \eqref{eq:bound_order} hold,
    and the pair
    $(\L,\H)$ 
    satisfy
    either 
        \eqref{eq:wide_arch}
        or
        \eqref{eq:deep_arch}
    for $\growth$ satisfying
        \begin{equation}\label{eq:growth_cond}
            \growth>2
        ,
        \qquad    
            \lim_{n\to\infty}\left\{\frac{\growth^2\log(\growth)^3}{n}\right\}=0
        ,
        \qquad
        \lim_{n\to\infty}
        \left\{
            \frac%
                {\Xboundval^{4(\holD+1)}}%
                {\growth^{\left(\frac{2\smooth}{d}\right)}}
        \right\}
        =0,
        \end{equation}
    Suppose 
    Condition \ref{as:curv} holds,
    and the sequences
        $
        \{
        \Mbound
        \}_{n\in\N}
        ,
        \{
        \trunc
        \}_{n\in\N}
        $
    are such that
    Condition \ref{as:lip} 
    and \eqref{eq:trunc_term_def} hold
    for all $n$ sufficiently large.
    Then, there exists a constant $C>0$ such that, for each $n$ sufficiently large,
    any $\f$ 
    satisfying \eqref{eq:fhat},
    and any
        $\anA$,
        $\betaBoundn(\anA)$,
        and
        $\deltemp$
    satisfying
        \begin{equation*}
        \begin{gathered}
                2
            <
                \anA
            \leq
                \frac{n}{2}
        ,
        \quad
            \beta(\anA)\leq \betaBoundn(\anA)
        ,
        \quad
            e^{-\frac{n}{2\anA}}
            \leq 
            \betaBoundn(\anA)
            <
            \frac{\anA}{n}
        ,
        \quad
        \deltemp
        \geq
                \frac{\Xboundval^{4(\holD+1)}}{\growth^{\left(\frac{2\smooth}{d}\right)}}
                \!
                \sqrt{
                    \frac{n}{\anA}
                    \!
                    \log\!\left(
                        \frac{1}{\!\betaBoundn(\anA)\!}
                    \right)
                }\!
        ,
        \end{gathered}
        \end{equation*}
    \eqref{eq:pbound_FCDNN_ROC} holds for any 
        \begin{equation*}
        \begin{aligned}
            \err
        \geq
            \big(\!\max\big\{\!\Mbound,\DNNBound\!\big\}\big)^2
            \,\frac{\anA}{n}
            \Big[
            \growth^2\log(\growth)^3\log(n) + \deltemp
            \Big]
            +
            \left(
                \frac{\Xboundval^{4(\holD+1)}}{\growth^{\left(\frac{2\smooth}{d}\right)}}
            \right)^{\!2}
                +
                \trunc
                +
                \truncf
                +
                \theta_n
        ,
        \end{aligned}
        \end{equation*}
    and
        $
         \Pbndterm
        \geq
            e^{
                -
                \deltemp
            }
            +
            \log(n)\!
                \left[
                    n\betaBoundn(\anA)/\anA
                    +
                    \Pr(\Ptrunc)
                \right]
        .
        $
\end{theorem}
}%
\begin{proof}
    Condition \ref{as:Env} holds for $\Fbound=\DNNBound$ by Assumption \ref{as:data_holder} and \eqref{eq:bound_order}.
    By \eqref{eq:fObound}, \eqref{eq:growth_cond}, and $\smooth>\holD$
        $$
        \frac{
                \Big[
                \max\left\{
                    \Xboundval,\, \norm*{\fO}_{\HolderCust{\holD}{ }([-\Xboundval,\Xboundval]^d)}
                \right\}\Big]^{4(\holD+1)}
            }{\growth^{\frac{2\smooth}{d}}}
        \lesssim
            \frac{ \Xboundval^{4(\holD+1)} }{ \growth^{\frac{2\smooth}{d}} }
        \to 0
        \text{ as }
        n\to \infty.
        $$
    Then, since \eqref{eq:width_bound_order} holds, for $n$ sufficiently large, and any architecture satisfying either \eqref{eq:wide_arch} or \eqref{eq:deep_arch}, Lemma \ref{lem:approximationNEW} can be applied.
    This implies 
    there exists $\fpr\in \FMLP$ such that 
        \begin{equation}\label{eq:prjerrFCDNN_ROC}
            \prjerr
        \coloneqq
            \snormDNNBoundval{\fpr-\fO} 
        \lesssim
            \frac{
                \Big[
                \max\left\{
                    \Xboundval,\, \norm*{\fO}_{\HolderCust{\holD}{ }([-\Xboundval,\Xboundval]^d)}
                \right\}\Big]^{4(\holD+1)}
            }{\growth^{\frac{2\smooth}{d}}}
        \lesssim
            \frac{ \Xboundval^{4(\holD+1)} }{ \growth^{\frac{2\smooth}{d}} }
        .
        \end{equation}
    For $(\L,\H)$
    satisfying \eqref{eq:wide_arch} or \eqref{eq:deep_arch}, we have
        ${ \log(\H^2\L)}
        \lesssim
        \log(\growth)
        .
        $
    Thus, by Lemma \ref{lem:Pdim_boundNEW} 
        \begin{equation}\label{eq:pdimFCDNN_ROC}
            \Pdim(\FMLP)
        \lesssim
            \H^2\L^2\log(\H^2\L)
        \lesssim
            \growth^2\log(\growth)^3
        .
        \end{equation}

    To apply Theorem \ref{thrm:ROC_V2} we first verify \eqref{eq:pbnd}.
    Note that
        $
            n
        \geq
            \max\!\left\{
                2\Pdim(\FMLPg)
            ,\;
                {16\Fbound^2}/{\log(n)}
            \right\}
        $
    for all large $n$,
    by 
    \eqref{eq:growth_cond}
    and
    \eqref{eq:bound_order}
        $$
        \Pdim(\FMLP)
        \lesssim
            \growth^2\log(\growth)^3
        = \littleo(n)
        ,
        \quad
        \text{ and }
        \quad
            \frac{16\Fbound^2}{\log(n)}
        \lesssim
            \frac{\DNNBound^2}{\log(n)}
        \lesssim
            \frac{\Xboundval^2}{\log(n)}
        \leq
            \frac{n}{\log(n)}
        = \littleo(n)
        .
        $$
    From \eqref{eq:prjerrFCDNN_ROC} and the conditions on $\anA$,
        $\betaBoundn(\anA)$,
        and
        $\deltemp$, the constant $C$ can be chosen large enough such that $C\deltemp$ satisfies the lower bound on $\delta$ from \eqref{eq:pbnd}, and the probability bound still holds since $e^{-C\deltemp}\leq e^{-\deltemp}$.
    All remaining requirements from \eqref{eq:pbnd} follow from 
    the bounds on 
        $
            \anA,
            \betaBoundn,
        $ and $
            \deltemp
        $
    so Theorem \ref{thrm:ROC_V2} can be applied.
    Finally, $\err$ 
    follows from \eqref{eq:prjerrFCDNN_ROC}, \eqref{eq:pdimFCDNN_ROC}.
    \end{proof}
\vx

    \begin{proof}[Proof of Theorem \ref{thrm:FCDNN_ROC}(i)]
    The proof will follow by applying Theorem \ref{thrm:FCDNN_ROC_GEN}.
        Using $\Xboundval\leq \log(n)$ by \eqref{eq:xbound_ybound_choice} and
    $\smooth=\holD+\holE>\holD$, 
        \begin{equation}\label{eq:prjerrFCDNN}
        \begin{aligned}
            \frac{\Xboundval^{4(\holD+1)}}{\growth^{\left(\frac{2\smooth}{d}\right)}}
        &\leq
            n^{
                    - \frac{\smooth}{2\smooth+d}
            }
            \log(n)^{4(\smooth+1)-\left(\frac{2\smooth}{d}\right)\growthlog}
        .
        \end{aligned}
        \end{equation}
    so \eqref{eq:growth_cond} holds.
    Let
    \begin{gather*}
        \betaBoundn(j) =  e^{-\Cbeta\,j}
    ,
    \quad
        \anA = \log(n)^{2}
    ,
    \quad
        \deltemp
        =
        \growth^2\log(n)^4
    .
    \end{gather*}
    Now we verify the conditions on $\anA$,
        $\betaBoundn(\anA)$,
        and
        $\deltemp$ from Theorem \ref{thrm:FCDNN_ROC_GEN}.
    Clearly, the conditions on $\anA$ hold. 
    The condition on $\betaBoundn(\anA)$ follows since
    $
    \betaBoundn(\anA)
    =
     n^{-\Cbeta\,\log(n)}
    ,
    $
and for all $n$ sufficiently large
    \begin{equation*}
        \frac{n\betaBoundn(\anA)}{\anA}
    =
    \frac{ n^{1-\Cbeta\,\log(n)}}{\log(n)^{2}}
            < 1
    .
    \end{equation*}
All that remains is the lower bound on $\deltemp$. 
Note that
    $$
        \logp{
                \frac{1}{\betaBoundn(\anA)}
            }
    = 
        \logp{
            n^{\Cbeta\,\log(n)}
        }
    = 
        \Cbeta\,\log(n)
        \logp{n}
    \lesssim
        \log(n)^{2}
        =
        \anA
    ,
    $$
so by \eqref{eq:prjerrFCDNN}, 
    \begin{equation*}
    \begin{aligned}
        \frac{\Xboundval^{4(\holD+1)}}{\growth^{\left(\frac{2\smooth}{d}\right)}}
                \!
                \sqrt{
                    \frac{n}{\anA}
                    \log\!\left(
                        \frac{1}{\betaBoundn(\anA)}
                    \right)
                }
    &\lesssim
        \frac{\Xboundval^{4(\holD+1)}}{\growth^{\left(\frac{2\smooth}{d}\right)}}
        \sqrt{n}
    \leq
        n^{
            \frac{1}{2}\left(
                \frac{d}{2\smooth+d}
            \right)
        }
        \log(n)^{4(\smooth+1)-\left(\frac{2\smooth}{d}\right)\growthlog}
    \lesssim
        \deltemp
    .
    \end{aligned}
    \end{equation*}
Thus, Theorem \ref{thrm:FCDNN_ROC_GEN} can be applied. 
Finally, 
$\log(\growth)\asymp \log(n)$, so
    \begin{equation*}
    \begin{aligned}
        \growth^2\log(\growth)^3\log(n) 
    &
    \asymp
        n^{\frac{d}{2\smooth+d}}
                \log(n)^{4+2\growthlog}
    =
        \deltemp
    .
    \end{aligned}
    \end{equation*}
Then, $\err$ and $\Pbndterm$ follow from Theorem \ref{thrm:FCDNN_ROC_GEN} with the above display, \eqref{eq:prjerrFCDNN}, and Lemma \ref{lem:Ptrunc_truncf_env_verify}.
    \end{proof}
\vx

\begin{proof}[Proof of Theorem \ref{thrm:FCDNN_ROC}(ii)]
\newcommand{\temp}{\eta}%
Let
    $
        \betaBoundn(j) =  j^{-\Cbeta}\log(n)^{\Cbeta}
    ,$
    $
        \anA = n^{\temp}
        \log(n)
    ,$ 
    $
        \deltemp
        =
        \growth^2\log(n)^4
    .
    $
Note that   
    $(1+\Cbeta)^{-1}\leq \temp < 1$
implies 
    $1-\temp(1+\Cbeta)\leq 0$, so
for all $n$ sufficiently large
    \begin{equation*}
        \frac{n\betaBoundn(\anA)}{\anA}
    =
        \frac{n\log(n)^{\Cbeta}}{\anA^{1+\Cbeta}}
    =
        \frac{n^{1-\temp(1+\Cbeta)}}{\log(n)}
    < 1
    .
    \end{equation*}
In addition, $\log(1/\betaBoundn(\anA))\lesssim\log(n)$ and
    \begin{equation*}
    \begin{aligned}
        \frac{\Xboundval^{4(\holD+1)}}{\growth^{\left(\frac{2\smooth}{d}\right)}}
                \!
                \sqrt{
                    \frac{n}{\anA}
                    \log\!\left(
                        \frac{1}{\betaBoundn(\anA)}
                    \right)
                }
    &\lesssim
        \frac{\Xboundval^{4(\holD+1)}}{\growth^{\left(\frac{2\smooth}{d}\right)}}
        n^{\frac{1}{2}(1-\temp)}
    \leq
        n^{
            \frac{1}{2}(1-\temp)\left(
                \frac{d}{2\smooth+d}
            \right)
        }
        \log(n)^{4(\smooth+1)-\left(\frac{2\smooth}{d}\right)\growthlog}
    \lesssim
        \deltemp
    ,
    \end{aligned}
    \end{equation*}
    since $\temp<1$.
The desired result follows by Theorem \ref{thrm:FCDNN_ROC_GEN} using similar reasoning as the proof of Theorem \ref{thrm:FCDNN_ROC}(i).
\end{proof}
\vx


\subsection{Proofs for FCDNN results in nonparametric regression settings}\label{sec:PROOF_FCDNN_reg}

    \begin{lemma}\label{lem:regression_conditions}
    \renewcommand{\Ctruncorder}{C}%
    Consider the setting of Example \ref{ex:reg}. 
    Let $\Zbound=\Ybound\times\Xbound=[-\Yboundval,\Yboundval]\times[-\Xboundval,\Xboundval]^d$ 
    for some $\Xboundval,\Yboundval\geq 1$.
    Let $\FMLPg=\FMLP$ and 
    $\DNNBound=\CDNNBound\Xboundval$ for any $\CDNNBound\geq \CfObound$.
    \begin{enumerate}
        \item 
            Condition \ref{as:curv} holds with $\Cpcq=\CpcqB=1$.
        \item 
            Condition \ref{as:lip} holds with
                $
                    \Mbound 
                = 2\big(
                        \Yboundval
                        +
                        \DNNBound
                    \big)
                .
                $
        \item 
        The requirement on $\trunc$ in \eqref{eq:trunc_term_def} is satisfied by
            \begin{equation*}
                \trunc 
            =
            \max\left\{
            \frac{1}{n}
            ,\;\,
                3\truncf
                +
                3\max_{t\in\{1,\ldots,n\}}
                \Big(
                    \E\big[
                        \Yt^2\,
                        \mathbbm{1}_{\{|\Yt|\notin\Ybound\}}
                    \big]
                    +
                    \P\big(
                        \Zt\notin\Zbound
                    \big)
                    \big[
                    \DNNBound^2
                    +
                    \Yboundval^2
                    \big]
                \Big)
            \right\}
            .
            \end{equation*}
            If 
            Assumptions \ref{as:data_holder} and \ref{as:data_subg} hold,
            and
            $\Yboundval=\log(n)$ and $\Xboundval$ is chosen as 
            \eqref{eq:xbound_ybound_choice},
            then 
            $\trunc\lesssim n^{-1}$.
    \end{enumerate}
    \end{lemma}
    \begin{proof}[Proof of Lemma \ref{lem:regression_conditions}(i)]
    Recall $\fO(\Xt)=\E[\Yt|\Xt]$. By iterated expectations, for any $f\in\Lp{2}$ 
        \begin{equation*}
        \begin{aligned}[b]
            \E\big[\crit(\Zt,f)\big] - \E\big[\crit(\Zt,\fO)\big]
        &= 
            \E
            \Big[ f(\Xt)^2 - \fO(\Xt)^2 - 2f(\Xt) \Yt + 2\fO(\Xt) \Yt \Big] 
        \\& 
        =
            \E
            \Big[ f(\Xt)^2 - \fO(\Xt)^2 - 2f(\Xt)\fO(\Xt) + 2 \fO(\Xt)^2 \Big] 
        \\& 
        =
            \E\Big[ 
                |f(\Xt)-\fO(\Xt)|^2
            \Big]
        .
        \end{aligned}
        \end{equation*}
    Thus, Condition \ref{as:curv} holds with $\Cpcq=\CpcqB=1$.
    \end{proof}
    \vx
    
    \begin{proof}[Proof of Lemma \ref{lem:regression_conditions}(ii)]
    For any $f,g\in\left\{\FMLP\cup\{\fO\}\right\}$
    and $\z\in\Zspace$
    we have
        \begin{equation}\label{eq:crit_lip}
        \begin{aligned}[b]
            \big| \crit\big(\z,f\big) - \crit\big(\z,g\big) \big|
            &
            \mathbbm{1}_{\{\z\in\Zbound\}}
        =
            \Big|
                f(\x)^2 
                - 2yf(\x)
                - 
                g(\x)^2
                + 2yg(\x)
            \Big|
            \mathbbm{1}_{\{\z\in\Zbound\}}
        \\&
        =
            \Big| 
            \big[f(\x)+g(\x)\big]  \big[f(\x)-g(\x)\big]
            -
            2y\big[f(\x)-g(\x)\big]
            \Big|
            \mathbbm{1}_{\{\z\in\Zbound\}}
        \\&=
            \Big|
            \big( 
                f(\x) + g(\x) - 2y
            \big)
            \big(f(\x)-g(\x)\big)
            \Big|
            \mathbbm{1}_{\{\z\in\Zbound\}}
        \\&\leq
            \Big(
            2\sup_{f'\in\left\{\FMLP\cup\{\fO\}\right\}} \big|y - f'(\x)\big|
            \Big)
            \big|f(\x)-g(\x)\big|
            \mathbbm{1}_{\{\z\in\Zbound\}}
        .
        \end{aligned}
        \end{equation}
\end{proof}
    \vx

\begin{proof}[Proof of Lemma \ref{lem:regression_conditions}(iii)]
    Consider arbitrary $t\in \{1,\ldots,n\}$, and $f\in\left\{\FMLP\cup\{\fO\}\right\}$. 
    Note that 
        \begin{equation*}
        \resizebox{\hsize}{!}{$
        \begin{aligned}
            \E\big[
                \big| f(\Xt)\Yt \big|
                \mathbbm{1}_{\{\Zt\notin\Zbound\}}
            \big]
        &=
            \E\big[
                \big| f(\Xt)\Yt \big|
                \mathbbm{1}_{\{\Zt\notin\Zbound\}}\!
                \mathbbm{1}_{\{|\Yt|\geq|f(\Xt)|\}}
            \big]
            +
            \E\big[
                \big| f(\Xt)\Yt \big|
                \mathbbm{1}_{\{\Zt\notin\Zbound\}}\!
                \mathbbm{1}_{\{|\Yt|<|f(\Xt)|\}}
            \big]
        \\&
        \leq
            \E\big[
                    \Yt^2
                \mathbbm{1}_{\{\Zt\notin\Zbound\}}
            \big]
            +
            \E\big[
                f(\Xt)^2 
                \mathbbm{1}_{\{\Zt\notin\Zbound\}}
            \big]
        .
        \end{aligned}
        $}
        \end{equation*}
    With this and the triangle inequality
        \begin{equation}\label{eq:trunc_0}
        \begin{aligned}[b]
            \E\Big[
                \big|\critt\big(f\big)\big|
                \mathbbm{1}_{\{\Zt\notin\Zbound\}}\!
            \Big]
        &
        \leq
            \E\big[
                    \Yt^2 \,
                \mathbbm{1}_{\{\Zt\notin\Zbound\}}
            \big]
            +
            \E\big[
                f(\Xt)^2 
                \mathbbm{1}_{\{\Zt\notin\Zbound\}}
            \big]
            +
            2
            \E\big[
                \big| f(\Xt)\Yt \big|
                \mathbbm{1}_{\{\Zt\notin\Zbound\}}
            \big]
        \\&
        \leq
            3\E\big[
                    \Yt^2 \,
                \mathbbm{1}_{\{\Zt\notin\Zbound\}}
            \big]
            +
            3\E\big[
                f(\Xt)^2 
                \mathbbm{1}_{\{\Zt\notin\Zbound\}}
            \big]
        .
        \end{aligned}
        \end{equation}
    For
    the second term 
    in the last line
    of \eqref{eq:trunc_0}, since $\DNNBound\geq\snormDNNBoundval{\fO}$
        \begin{equation*}
        \begin{aligned}
            \E\big[
                f(\Xt)^2 
                \mathbbm{1}_{\{\Zt\notin\Zbound\}}
            \big]
        &\leq
            \E\big[
                f(\Xt)^2 
                \mathbbm{1}_{\{\Zt\notin\Zbound\}}
                \mathbbm{1}_{\{\Xt\notin\Xbound\}}
            \big]
            +
            \E\big[
                f(\Xt)^2 
                \mathbbm{1}_{\{\Zt\notin\Zbound\}}
                \mathbbm{1}_{\{\Xt\in\Xbound\}}
            \big]
        \\&
        \leq
            \truncf
            +
            \DNNBound^2
            \P\big(\{\Zt\notin\Zbound\}\big)
        .
        \end{aligned}
        \end{equation*}
    For the first term in the last line of \eqref{eq:trunc_0} 
        \begin{equation*}
        \begin{aligned}
            \E\big[
                \Yt^2\,
                \mathbbm{1}_{\{\Zt\notin\Zbound\}}
            \big]
        &=
            \E\big[
                \Yt^2\,
                \mathbbm{1}_{\{|\Yt|\leq \Yboundval\}}
                \mathbbm{1}_{\{\Zt\notin\Zbound\}}
            \big]
            +
            \E\big[
                \Yt^2\,
                \mathbbm{1}_{\{|\Yt|> \Yboundval\}}
                \mathbbm{1}_{\{\Zt\notin\Zbound\}}
            \big]
        \\&
        \leq
            \Yboundval^2
            \P\big(\{\Zt\notin\Zbound\}\big)
            +
            \E\big[
                \Yt^2\,
                \mathbbm{1}_{\{|\Yt|> \Yboundval\}}
            \big]
        .
        \end{aligned}
        \end{equation*}
    Thus, 
        \begin{equation}\label{eq:trunc_1}
        \begin{aligned}[b]
            \E\Big[
                \big|\critt\big(f\big)\big|
                \mathbbm{1}_{\{\Zt\notin\Zbound\}}\!
            \Big]
        &
        \leq
             3
            \Big(
                \truncf
                +
                \E\big[
                    \Yt^2\,
                    \mathbbm{1}_{\{|\Yt|\notin\Ybound\}}
                \big]
                +
                \P\big(
                    \Zt\notin\Zbound
                \big)
                \big[
                \DNNBound^2
                +
                \Yboundval^2
                \big]
            \Big)
        .
        \end{aligned}
        \end{equation}
    Next, for any $z>0$,
        $
        \P\big(\{\Zt\notin[-z,z]^{\dz}\}\big)
        =
        \P\big(\norm{\Zt}_{\lp{\infty}}>z\big)
        .
        $
    Then, 
    under Assumption \ref{as:data_subg},
    by Lemma \ref{lem:subgausexp} 
        $$
            \E\left[
                e^{2|\Yt|}
            \right]
        \leq
            \E\left[
                e^{2\norm{\Zt}_{\lp{\infty}}}
            \right]
        \leq
            1+4\sqrt{\pi/\Csubg}\,e^{1/\Csubg}
        .
        $$
    In addition,
        $$
        \E\Big[
            e^{2|\Yt|}
            \Big]
        \geq
            \E\Big[
            e^{2|\Yt|}
            \mathbbm{1}_{\{|\Yt|> \Yboundval\}}
            \Big]
        \geq
            e^{\Yboundval}
            \E\Big[
            e^{|\Yt|}
            \mathbbm{1}_{\{|\Yt|> \Yboundval\}}
            \Big]
        \geq
            e^{\Yboundval}
            \E\Big[
            \Yt^2
            \mathbbm{1}_{\{|\Yt|> \Yboundval\}}
            \Big]
        .
        $$
    Thus, from the previous two displays
        \begin{gather}\label{eq:Yunif_int}
            \E\Big[
            \Yt^2
            \mathbbm{1}_{\{|\Yt|> \Yboundval\}}
            \Big]
        \leq
            \frac{
                \E\Big[
                    e^{2|\Yt|}
                \Big]
            }{
                e^{\Yboundval}
            }
        \leq    
            \frac{
                1+4\sqrt{\pi/\Csubg}\,e^{1/\Csubg}
            }{
                e^{\Yboundval}
            }
        .
        \end{gather}
    With this, Assumption \ref{as:data_subg}, and \eqref{eq:trunc_1}
        \begin{equation*}
        \begin{aligned}[b]
            \E\Big[
                \big|\critt\big(f\big)\big|
                \mathbbm{1}_{\{\Zt\notin\Zbound\}}\!
            \Big]
        &
        \leq
            3\truncf
            +
                \frac{
                    3+12\sqrt{\pi/\Csubg}\,e^{1/\Csubg}
                }{
                    e^{\Yboundval}
                }
                +
                3
                \P\big(
                    \Zt\notin\Zbound
                \big)
                \big[
                \DNNBound^2
                +
                \Yboundval^2
                \big]
        \\&
        \leq
            3\truncf
            +
            \frac{
                    3+12\sqrt{\pi/\Csubg}\,e^{1/\Csubg}
                }{
                    n
                }
            +
            12\CDNNBound \log(n)^2 n^{-\Csubg\log(n)}
        .
        \end{aligned}
        \end{equation*}
    The desired result follows from Lemma \ref{lem:Ptrunc_truncf_env_verify}(ii).
\end{proof}
\vx

    \begin{proof}[Proof of Corollary \ref{cor:FCDNN_reg}]
For part (i),
    $
    \Xboundval\leq \log(n)
    $,
    $
        \Yboundval\leq \log(n),
    $
    and $\DNNBound\lesssim\log(n)$.
Then $\Mbound\lesssim\log(n)$ by Lemma \ref{lem:regression_conditions},
and proof will follow by applying Theorem \ref{thrm:FCDNN_ROC}(i). 

For part (ii),
    $
    \Xboundval=1
    $,
    $
        \Yboundval\leq \log(n),
    $
    and $\DNNBound\lesssim 1$.
Then, the  proof will follow by applying Theorem \ref{thrm:FCDNN_ROC}(i) using Lemma \ref{lem:regression_conditions}.
\end{proof}
\vx

    

\subsection{Proofs for FCDNN results in nonparametric logistic autoregression settings}\label{sec:PROOF_FCDNN_log}

    \begin{lemma}\label{lem:logistic_conditions}
    Consider the setting of Example \ref{ex:log}. 
    Let $\FMLPg=\FMLP$ and $\DNNBound=\DNNBoundfix\coloneqq\CDNNBound\Clog$ for $\CDNNBound\geq1$.
    \begin{enumerate}
        \item 
            If Assumption \ref{as:data_logistic_fbound} holds, 
            then
            Condition \ref{as:curv} holds with 
                $\Cpcq= \big[2(e^{\DNNBoundfix} +e^{-\DNNBoundfix}+2)\big]^{-1}$,
            and
                $\CpcqB=1/8$.
        \item 
            Condition \ref{as:lip} holds with
                $\Mbound = 2.$
        \item 
            Let $\Zbound=\Ybound\times\Xbound=[-\Yboundval,\Yboundval]\times[-\Xboundval,\Xboundval]^d$ 
            where
            $\Yboundval=1$
            and
            $\Xboundval$ is chosen as in 
            \eqref{eq:xbound_ybound_choice}.
            If 
            Assumptions \ref{as:data_holder}, \ref{as:data_subg} and \ref{as:data_logistic_fbound} hold,
            then there exists 
            $\trunc$
            satisfying \eqref{eq:trunc_term_def} such that 
            $\trunc\lesssim n^{-\Csubg\log(n)}$.
    \end{enumerate}
    \end{lemma}
\begin{proof}[Proof of Lemma \ref{lem:logistic_conditions}(i)]
   \newcommand{\ftn}{g_{a}}%
   The proof follows similarly to \citet[Lemma 8]{farrell_deep_2021}.
    Let $t\in\N$ be arbitrary.
    By definition 
        $
        \E[\Yt|\Xt]
        =
        e^{\fO(\Xt)}\big[1+e^{\fO(\Xt)}\big]^{-1}
        .
        $
    Then, for any $f\in\FMLP$, by iterated expectations
            \begin{equation*}
            \begin{aligned}[b]
                \E[\critt(f)]-\E[\critt(\fO)]
            &= 
                \E\left[
                    \frac{e^{\fO(\Xt)}}{1+e^{\fO(\Xt)}}
                    \Big(
                        \fO(\Xt)
                        -
                        f(\Xt)
                    \Big)
                    +
                    \logp{
                        \frac{1+e^{ f(\Xt)}}{1+e^{ \fO(\Xt)}}
                        }
                \right]
            .
            \end{aligned}
            \end{equation*}
        Let 
            $
                \ftn(b)
            \coloneqq
                -
                \frac{e^{a}}{1+e^{a}}
                \big(
                    b
                    -
                    a
                \big)
                +
                \logp{
                    \frac{1+e^{b}}{1+e^{a}}
                    }
                ,
            $
        for $a,b\in \R$.
        By Taylor's Theorem, with the Lagrange form of the remainder for some $\lambda\in(0,1)$
            \begin{equation*}
            \begin{aligned}
                \ftn(b)
            &=
                \ftn(a)
                +
                (b-a)
                \left[
                    \frac{d}{d x}\,
                    \ftn(x)
                \right]_{x=a}
                +
                \frac{(b-a)^2}{2}
                \left[
                    \frac{d^2}{d x^2}\,
                    \ftn(x)
                \right]_{x=\lambda a + (1-\lambda)b}
            .
            \end{aligned}
            \end{equation*}
        Note that
            $$
                \frac{d}{d x}\,
                \ftn(x)
            =
                \frac{{e}^{x}}{1+{e}^{x}} - \frac{{e}^{a}}{1+{e}^{a}}
            ,
            \quad
            \text{ and }
            \quad
                \frac{d^2}{d x^2}\,
                \ftn(x)
            =
                \frac{{e}^{x}}{\left(1+{e}^{x}\right)^{2}}
            =
                \frac{1}{e^x +e^{-x}+2}
            ,
            $$
        in addition,
            $\ftn(a)=0$,
        and
            $\left[\frac{d}{d x}\,
                \ftn(x)\right]_{x=a}=0$.
        Then choosing $a=\fO(\Xt)$ and $b=f(\Xt)$,
            $$
                \E\big[\crit(\Zt,f)\big] - \E\big[\crit(\Zt,\fO)\big]
            =
                \frac{1}{2}
                \E\left[
                |f(\Xt)-\fO(\Xt)|^2
                \left[
                    \frac{1}{e^x +e^{-x}+2}
                \right]_{x=\lambda \fO(\Xt) + (1-\lambda)f(\Xt)}
                \right]
            .
            $$
        To obtain $\Cpcq$ and $\CpcqB$ note that $\DNNBoundfix\geq \Clog$ and
            $$
                \min_{x\in[-\DNNBoundfix,\DNNBoundfix]}
                \left\{
                    \frac{1}{e^x +e^{-x}+2}
                \right\}
            =
                \frac{1}{e^{\DNNBoundfix} +e^{-\DNNBoundfix}+2}
            ,
            \quad
            \text{ and }
            \quad
                \max_{x\in\R}
                \left\{
                    \frac{1}{e^x +e^{-x}+2}
                \right\}
            =
                1/4
            .
            $$
        With Assumption \ref{as:data_logistic_fbound},
        combining the previous two displays implies the desired result. 
\end{proof}
\vx

\begin{proof}[Proof of Lemma \ref{lem:logistic_conditions}(ii)]
    For any $f,f':\Xspace\to\R$, and $\z=(y,\x)\in\{0,1\}\times\Xspace$
        \begin{equation*}
        \begin{aligned}
             \bigg|
                \logp{
                    \frac{1+e^{ f(\x)}}{1+e^{ f'(\x)}}
                    }
            \bigg|
        &
        =
                \logp{
                    \frac{1+e^{ f(\x)}}{1+e^{ f'(\x)}}
                    }
                \mathbbm{1}_{\{f(\x)>f'(\x)\}}
                +
                \logp{
                    \frac{1+e^{ f'(\x)}}{1+e^{ f(\x)}}
                    }
                \mathbbm{1}_{\{f(\x)<f'(\x)\}}
        \\&
        \leq
                \logp{
                    \frac{e^{ f(\x)}}{e^{ f'(\x)}}
                    }
                \mathbbm{1}_{\{f(\x)>f'(\x)\}}
                +
                \logp{
                    \frac{e^{ f'(\x)}}{e^{ f(\x)}}
                    }
                \mathbbm{1}_{\{f(\x)<f'(\x)\}}
        \\&
        =
            \bigg|
            \logp{
                \frac{e^{ f'(\x)}}{e^{ f(\x)}}
                }
            \bigg|
        =
            \big|
                f'(\x)
                -
                f(\x)
            \big|
        .
        \end{aligned}
        \end{equation*}
    Thus, since $y\in\{0,1\}$,
        \begin{equation*}
        \begin{aligned}[b]
            |\crit(\z,f)-\crit(\z,f')|
        &= 
            \bigg|
                y
                \Big(
                    f'(\x)
                    -
                    f(\x)
                \Big)
                +
                \logp{
                    \frac{1+e^{ f(\x)}}{1+e^{ f'(\x)}}
                    }
            \bigg|
        \\&
        \leq
            \Big|
                y
                \Big(
                    f'(\x)
                    -
                    f(\x)
                \Big)
            \Big|
                +
            \bigg|
                \logp{
                    \frac{1+e^{ f(\x)}}{1+e^{ f'(\x)}}
                    }
            \bigg|
        \leq
            2 
            \big|
                f'(\x)
                -
                f(\x)
            \big|
        .
        \end{aligned}
        \end{equation*}
\end{proof}
\vx

\begin{proof}[Proof of Lemma \ref{lem:logistic_conditions}(iii)]
    Consider arbitrary $t\in \{1,\ldots,n\}$, and $f,f'\in\left\{\FMLP\cup\{\fO\}\right\}$.
    Since $\Yt\in\{0,1\}$ and $f(\Xt)\leq\DNNBoundfix$,
        \begin{equation*}
        \begin{aligned}
                \big|\critt\big(f\big)\big|
        &
        =
                \left|
                    \logp{1+e^{f(\Xt)}}
                    -\Yt f(\Xt)
                \right|
        \leq
                \logp{1+e^{\DNNBoundfix}}
                +\DNNBoundfix
        .
        \end{aligned}
        \end{equation*}
    Hence,
        \begin{equation*}
        \begin{aligned}
            \E\Big[
                \big|\critt\big(f\big)\big|
                \mathbbm{1}_{\{\Zt\notin\Zbound\}}
            \Big]
        &
        \leq
            \left[
                    \logp{1+e^{\DNNBoundfix}}
                    +\DNNBoundfix
            \right]
            \P(\Zt\notin\Zbound)
        ,
        \end{aligned}
        \end{equation*}
    and the desired result follows from Assumption \ref{as:data_subg} and \eqref{eq:xbound_ybound_choice}, since $t$ is arbitrary.
\end{proof}
\vx

    \begin{proof}[Proof of Corollary \ref{cor:FCDNN_log}]
    The proof will follow by applying Theorem \ref{thrm:FCDNN_ROC}(i) with Lemma \ref{lem:logistic_conditions}.
\end{proof}
\vx

\subsection{Proofs for FCDNN results in nonparametric conditional quantile settings}\label{sec:PROOF_FCDNN_quant}
{
\renewcommand{\check}{\rho_\tau}%
    
    \begin{lemma}\label{lem:quantile_conditions}
    Consider the setting of Example \ref{ex:quant}. 
    Let $\FMLPg=\FMLP$ and 
    $\DNNBound=\DNNBoundfix\coloneqq2\Cquant$.
    \begin{enumerate}
        \item 
            If Assumption \ref{as:data_quantile} holds, 
            then Condition \ref{as:curv} holds with
            $\Cpcq={\cdensi}/{2}$ and $\CpcqB={\cdensii}/{2}$.
        \item 
            Condition \ref{as:lip} holds with
                $\Mbound = 1.$
        \item 
            Let $\Zbound=\Ybound\times\Xbound=[-\Yboundval,\Yboundval]\times[-\Xboundval,\Xboundval]^d$ 
            where
            $\Yboundval$
            and
            $\Xboundval$ are chosen as in 
            \eqref{eq:xbound_ybound_choice}.
            If 
            Assumptions \ref{as:data_holder}, \ref{as:data_subg} and \ref{as:data_quantile} hold,
            then there exists 
            $\trunc$
            satisfying \eqref{eq:trunc_term_def} such that 
            $\trunc\lesssim n^{-1}$.
    \end{enumerate}
    \end{lemma}
\begin{proof}[Proof of Lemma \ref{lem:quantile_conditions}(i)]
\newcommand{\Rx}{R}%
    For brevity, the following omits the qualifier ``almost surely'' from statements where it applies.
    Fix $\tau \in (0,1)$, 
    and define
        $$
            \check(u)\coloneqq u\cdot\big(\tau-\mathbbm{1}_{\{u\leq0\}}\big) 
        ,
        \qquad
            \Rx(u) \coloneqq \E\big[\check(\Yt - u)\mid\! \Xt\big]
        .
        $$ 
    Then,
        \begin{equation}\label{eq:RXcriteq}
            \crit(\Zt,f)
            =
            \check(\Yt-f(\Xt))
        \quad
        \text{ and }
        \quad
            \E\big[\critt(f)-\crit(\fO) \mid  \Xt\big]
        =
            \Rx(f(\Xt))
            -
            \Rx(\fO(\Xt))
        .
        \end{equation}
    For almost every $u$ (i.e., $u\neq \Yt$)
        $$
            \frac{\partial }{\partial u}
            \Rx(u)
        =
            \E\Big[
                -(\tau-\mathbbm{1}_{\{\Yt\leq u\}})
            \mid
                \Xt
            \Big]
        =
            \FXY(u)
            -
            \tau
        .
        $$
    By the fundamental theorem of calculus 
        $$
            \Rx(f(\Xt))
            -
            \Rx(\fO(\Xt))
        =
            \int_{\fO(\Xt)}^{f(\Xt)}
                \Big(
                \FXY(u)
                -
                \tau
                \Big)
                du
        =
            \int_{0}^{f(\Xt)-\fO(\Xt)}
                \Big(
                \FXY(\fO(\Xt)+s)
                -
                \tau
                \Big)
                ds
        .
        $$
    Thus, with \eqref{eq:RXcriteq}
    and
        $
        \FXY(\fO(\Xt))
        =
        \tau
        $
    by \eqref{eq:ex3},
        \begin{equation*}
        \begin{aligned}
            \E\big[\critt(f)-\crit(\fO) \mid  \Xt\big]
        &=
            \int_{0}^{f(\Xt)-\fO(\Xt)}
                \Big(
                \FXY(\fO(\Xt)+s)
                -
                \FXY(\fO(\Xt))
                \Big)
            ds
        ,
        \end{aligned}
        \end{equation*}
    and by the assumed density $\densXY$ from Assumption \ref{as:data_quantile}
        \begin{equation}\label{eq:densdif}
        \begin{aligned}
            \E\big[\critt(f)-\crit(\fO) \mid  \Xt\big]
        &=
            \int_{0}^{f(\Xt)-\fO(\Xt)\!}\!
            \int_{\fO(\Xt)}^{\fO(\Xt)+s}
            \densXY(y)
            dy
            ds
        \\&
        =
            \int_{0}^{f(\Xt)-\fO(\Xt)\!}\!
                \!\int_{0}^{s} \densXY(\fO(\Xt) + u) du\,
            ds
        .
        \end{aligned}
        \end{equation}
    By assumption, 
        $\DNNBound$ satisfies \eqref{eq:width_bound_order}
    so
        $\DNNBound=\DNNBoundfix=2\Cquant$.
    With this,
        $
        |f(\Xt)-\fO(\Xt)|\leq 3\Cquant
        $
    since
    $\snorm{f}\leq2\Cquant$ for any $f\in\FMLP$,
    and $\snorm{\fO}\leq \Cquant$ by Assumption \ref{as:data_quantile}.
    Then, 
        $$
        -\big|f(\Xt)-\fO(\Xt)\big|
        \leq
        u
        \leq
        \big|f(\Xt)-\fO(\Xt)\big|
        \quad
        \implies 
        \quad
        \Big|\big(\fO(\Xt)+u\big)-\fO(\Xt)\Big|
        =
        |u|
        \leq 
        3\Cquant
        .
        $$
    Thus, the bounds on $\densXY$ from Assumption \ref{as:data_quantile} apply, and we have
        \begin{equation*}
        \begin{aligned}
        \text{if } 
        0 \leq s \leq f(\Xt)-\fO(\Xt)
        &\quad \implies \quad&
            s\cdensi
        \leq
            &\int_{0}^{s} \densXY(\fO(\Xt) + u) du
        \leq
            s\cdensii
        ,
        \\
        \text{if } 
        0 \geq s \geq f(\Xt)-\fO(\Xt)
        & \quad \implies  &
            (-s)\cdensi
        \leq
            &\int_{s}^{0} \densXY(\fO(\Xt) + u) du 
        \leq
            (-s)\cdensii
        .
        \end{aligned}
        \end{equation*}
    This implies
        \begin{equation}\label{eq:distdifbnd}
        \begin{aligned}
            \frac{\cdensi|f(\Xt)-\fO(\Xt)|^2}{2}
        \leq
            \int_{0}^{f(\Xt)-\fO(\Xt)\!}\!
                \!\int_{0}^{s} \densXY(\fO(\Xt) + u) du\,
            ds
        \leq
            \frac{\cdensii|f(\Xt)-\fO(\Xt)|^2}{2}
        .
        \end{aligned}
        \end{equation}
    To see this, if $f(\Xt)-\fO(\Xt) \geq 0$, then \eqref{eq:distdifbnd} follows from the $s\geq 0 $ bound;
    and if $f(\Xt)-\fO(\Xt) < 0$, then \eqref{eq:distdifbnd} follows from the $s\leq 0 $ bound by using
        \begin{equation*}
        \begin{aligned}
            \int_{0}^{f(\Xt)-\fO(\Xt)\!}\!
                \!\int_{0}^{s} \densXY(\fO(\Xt) + u) du\,
            ds
        &
        =
            \int_{f(\Xt)-\fO(\Xt)}^{0}
                \!\int_{s}^{0} \densXY(\fO(\Xt) + u) du\,
            ds
        .
        \end{aligned}
        \end{equation*}
    Thus, combining \eqref{eq:densdif}, and \eqref{eq:distdifbnd}
        \begin{equation*}
        \begin{aligned}
            \frac{\cdensi|f(\Xt)-\fO(\Xt)|^2}{2}
        \leq
            \E\big[\critt(f)-\crit(\fO) \mid  \Xt\big]
        \leq
            \frac{\cdensii|f(\Xt)-\fO(\Xt)|^2}{2}
        ,
        \end{aligned}
        \end{equation*}
    and the desired result follows by taking the expectation. 
\end{proof}
\vx

\begin{proof}[Proof of Lemma \ref{lem:quantile_conditions}(ii)]
    For
    $\tau\in(0,1),
    $
    let
        $\rho_\tau(u)\coloneqq u\cdot\big(\tau-\mathbbm{1}_{\{u\leq0\}}\big) $.
    so 
        $\rho_\tau(\Yt-f(\Xt))=\crit(\Zt,f)$
    
    Consider the case where $u,v\geq0$. Then, 
        $\rho_\tau(u)=\tau u,$ and $ \rho_\tau(v)=\tau v,$
    so
        $$
        |\rho_\tau(u)-\rho_\tau(v)|
        =|\tau||u-v|
        =\tau |u-v|
        \leq |u-v|,
        $$
    since $0<\tau<1$.
    The case where $u,v<0$ follows by a similar argument.
    
    Consider the case where $u\leq0< v$, so 
    $\rho_\tau(u)=(\tau-1)u, $ 
    and
    $ \rho_\tau(v)=\tau v.$ 
    Then,
        $$
        |\rho_\tau(u)-\rho_\tau(v)|
        =
        |(\tau-1)u-\tau v|
        =
        |(1-\tau)(-u)-\tau v|
        \leq
            |(1-\tau)(-u)|+|\tau v|
        ,
        $$
    by the triangle inequality. 
    Thus, since $0<\tau<1$ and $(-u),v\geq 0$,
        $$
        |\rho_\tau(u)-\rho_\tau(v)|
        \leq
            (1-\tau)(-u)+\tau v
        \leq 
            v-u
        =
            |u-v|
        .
        $$
    This completes the proof since $v\leq0< u$ follows by a similar argument.
\end{proof}
\vx

\begin{proof}[Proof of Lemma \ref{lem:quantile_conditions}(iii)]
\renewcommand{\DNNBound}{\DNNBoundfix}%
    Consider arbitrary $t\in \{1,\ldots,n\}$, and $f,f'\in\left\{\FMLP\cup\{\fO\}\right\}$. 
    Note that 
        $
        \big|
            \tau
            -
            \mathbbm{1}_{\{\Yt \leq f(\Xt)\}}
        \big|
        < 1
        ,
        $
    and $\sup_{f\in\left\{\FMLP\cup\{\fO\}\right\}}\snormDNNBound{f}\leq \DNNBound$,
    so we have
            \begin{equation}\label{eq:trunc_0_quant}
            \begin{aligned}[b]
                \E\Big[
                    \big|\critt\big(f\big)\big|
                    \mathbbm{1}_{\{\Zt\notin\Zbound\}}
                \Big]
            &
            =
                \E\Big[
                    \big|\big(
                        \tau
                        -
                        \mathbbm{1}_{\{\Yt \leq f(\Xt)\}}
                    \big)
                    \big(
                        \Yt - f(\Xt)
                    \big)
                    \big|
                    \mathbbm{1}_{\{\Zt\notin\Zbound\}}
                \Big]
            \\&
            \leq
                \E\Big[
                    \big|
                        \Yt - f(\Xt)
                    \big|
                    \mathbbm{1}_{\{\Zt\notin\Zbound\}}
                \Big]
            \\&
            \leq
                \E\big[
                    | \Yt |
                    \mathbbm{1}_{\{\Zt\notin\Zbound\}}
                \big]
                +
                \DNNBound
                \P\big(\{\Zt\notin\Zbound\}\big)
            .
            \end{aligned}
            \end{equation}
    For the first term in the last line of \eqref{eq:trunc_0_quant} 
        \begin{equation*}
        \begin{aligned}
            \E\big[
                | \Yt |
                \mathbbm{1}_{\{\Zt\notin\Zbound\}}
            \big]
        &=
            \E\big[
                | \Yt |
                \mathbbm{1}_{\{|\Yt|\leq \Yboundval\}}
                \mathbbm{1}_{\{\Zt\notin\Zbound\}}
            \big]
            +
            \E\big[
                | \Yt |
                \mathbbm{1}_{\{|\Yt|> \Yboundval\}}
                \mathbbm{1}_{\{\Zt\notin\Zbound\}}
            \big]
        \\&
        \leq
            \Yboundval
            \P\big(\{\Zt\notin\Zbound\}\big)
            +
            \E\big[
                | \Yt |
                \mathbbm{1}_{\{|\Yt|> \Yboundval\}}
            \big]
        .
        \end{aligned}
        \end{equation*}
    Combining the previous two displays,
        \begin{equation*}
            \begin{aligned}[b]
                \E\big[
                    \big|\critt\big(f\big)\big|
                    \mathbbm{1}_{\{\Zt\notin\Zbound\}}
                \big]
            &
            \leq
                \E\big[
                    | \Yt |
                    \mathbbm{1}_{\{|\Yt|> \Yboundval\}}
                \big]
                +
                (\DNNBound+\Yboundval)
                \P\big(\{\Zt\notin\Zbound\}\big)
            .
            \end{aligned}
            \end{equation*}
    Thus, the desired result follows from Assumption \ref{as:data_subg} with \eqref{eq:xbound_ybound_choice} and \eqref{eq:Yunif_int}, since $t$ is arbitrary.
\end{proof}
\vx


    \begin{proof}[Proof of Corollary \ref{cor:FCDNN_quant}]
    The proof will follow by applying Theorem \ref{thrm:FCDNN_ROC}(i) with Lemma \ref{lem:quantile_conditions}.
\end{proof}
\vx

}

\subsection{Supporting Results}\label{sec:FCDNN_sup_res}

\begin{lemma}\label{lem:subgausexp}
Let $X$ be a real-valued random variable. Assume that there exists $C>0$ such that
\[
\P(|X|\geq t)\leq 2e^{-t^2C}
\qquad\text{for all } t\geq 0.
\]
Then,
for any $\alpha>0$,
        \begin{equation*}
        \begin{aligned}
            \E \left[e^{\alpha |X|}\right]
        \leq
            1
            +
            2\alpha
            e^{\frac{\alpha^2}{4C}}
            \sqrt{\frac{\pi}{C}}
        .
        \end{aligned}
        \end{equation*}
\end{lemma}
\begin{proof}
\renewcommand{\X}{X}
    For any $\alpha>0$,
        \begin{equation*}
        \begin{aligned}
            \E \left[e^{\alpha |\X|}\right]
        &= 
            \int_0^\infty\! 
            \P\!\left(
                e^{\alpha |\X|}
            \geq 
                u
            \right)
            du 
        =
            1
            +
            \int_1^\infty\! 
            \P\!\left(
                e^{\alpha |\X|}
            \geq 
                u
            \right)
            du
        =
            1
            +
            \int_1^\infty\! 
            \P\!\left(
                |\X|
            \geq 
                \frac{\log(u)}{\alpha}
            \right)
            du
        .
        \end{aligned}
        \end{equation*}
    Substituting $s= \log(u)/\alpha$ and using the assumed tail bound,
        \begin{equation*}
        \begin{aligned}
            \int_1^\infty\!
            \P\!\left(
                |\X|
            \geq 
                \frac{\log(u)}{\alpha}
            \right)
            du
        &
        =
            \alpha
            \int_0^\infty\!
                e^{s\alpha}
                \P\!\left(
                    |\X|
                \geq 
                    s
                \right)
            ds
        \leq
            2\alpha
            \int_0^\infty\!
                e^{s\alpha-s^2C}
            ds
        .
        \end{aligned}
        \end{equation*}
    Completing the square
        $
        s\alpha-s^2C
        =
        -C\left[s-{\alpha}/({2C})\right]^2+{\alpha^2}/({4C})
        .
        $
    Thus,
        \begin{equation*}
        \begin{aligned}
            \int_0^\infty\!
                e^{s\alpha-s^2C}
            ds
        &
        =
            e^{\frac{\alpha^2}{4C}}
            \int_0^\infty\!
                e^{-C\left(s-\frac{\alpha}{2C}\right)^2}
            ds
        \leq
            e^{\frac{\alpha^2}{4C}}
            \int_{-\infty}^\infty\!
                e^{-Ct^2}
            dt
        =
            e^{\frac{\alpha^2}{4C}}
            \sqrt{\frac{\pi}{C}}
        ,
        \end{aligned}
        \end{equation*}
    where the last equality uses the well-known result 
        $\textstyle
        \int_{-\infty}^\infty e^{-x^2}dx=\sqrt{\pi}
        .
        $
    The desired result follows from combining the previous three displays.
\end{proof}
    \vx

The following is a restatement of \citet[Theorem 2]{kohler_rate_2021}. For $\Xboundval>0$ and $f \in \HolderCust{\smooth}
    ([-\Xboundval,\Xboundval]^d)$
    define the norm 
        $$
        \norm*{f}_{\HolderCust{\holD}{ }([-\Xboundval,\Xboundval]^d)}
        \coloneqq
        \max\left\{\norm{D^{\kb}f}_{\infty|[-\Xboundval,\Xboundval]^d}
        :\,
        \norm{\kb}_{\lp{1}}
            \leq \holD
        ,
        \,
        \kb \in 
            \N_0^d
        \right\}
        .
        $$

\begin{lemma}[\citealp{kohler_rate_2021}]\label{lem:approximationNEW}
\newcommand{\temperr}{\varepsilon}%
\newcommand{\Cyar}{\const{Cyar}}%
\renewcommand{\L}{L}%
\renewcommand{\H}{H}%
\renewcommand{\Hb}{\boldsymbol{H}}%
\renewcommand{\Hi}[1]{H_{#1}}%
\renewcommand{\DNNBound}{B}%
\renewcommand{\fpr}{f}
\renewcommand{\Xboundval}{\overline{\mathrm{x}}}%
\renewcommand{\growth}{\growthnon}%

    Suppose 
    $\fO \in \HolderCust{\smooth}
    ([-\Xboundval,\Xboundval]^d)$ for some 
    $\Xboundval\geq 1$ and 
    $\smooth=\holD+\holE$ 
        where
            $\holD \in \N_0$, 
            $\holE\in(0,1]$.
    Let 
        $\FMLPnon(\L,\Hb,\DNNBound)$
    be defined as in \eqref{eq:sievespace}.
    For some $C>0$ sufficiently large let
        $$
        \growthnon^{{2p}/{d}}
        \geq
        C
        \Big[\max\left\{\Xboundval,\, \norm*{\fO}_{\HolderCust{\holD}{ }([-\Xboundval,\Xboundval]^d)}\right\}
            \Big]^{4(\holD+1)}
        .
        $$
    There exists $\Cyar\geq 1$ such that 
    if 
        $$
            \DNNBound
        \geq 
            \snormDNNBoundval{\fO} + 
            \Cyar
            \Big[\max\left\{\Xboundval,\, \norm*{\fO}_{\HolderCust{\holD}{ }([-\Xboundval,\Xboundval]^d)}\right\}
            \Big]^{4(\holD+1)}
                \cdot
                \growthnon^{-{2p}/{d}}
        ,
        $$ 
    and $\L$ and $\Hb$ are either: deep and wide
        \begin{equation}
        \begin{aligned}
                \L 
        &
            \geq
                5
                +
                \left\lceil 
                    \log_4(\growthnon)
                \right\rceil
                \cdot
                \left(\frac{2\,\smooth}{d}\right)
                \left(
                    \big\lceil 
                        \log_2(\max\{\holD,d\}+1)
                    \big\rceil
                    +1
                \right)
        ,
        \\
                \min_{l\in\{1,2,\ldots,\L\}}\Hi{l}
        &
            \geq
                \left\lceil 
                    \growthnon
                \right\rceil
                \cdot
                2^d
                \,
                (8d)^2
                \,
                \binom{d+\holD}{d}
                \,
                (\holD+1)
        ,
        \end{aligned}
        \end{equation}
    or, very deep with fixed width 
        \begin{equation}
        \begin{aligned}
                \L 
        &
            \geq
                5
                \left\lceil 
                    \growthnon
                \right\rceil
                +
                \left\lceil 
                    (\growthnon-1)
                    \log_4\left(
                    e^{
                        4(\holD+1)
                    }
                    \right)
                +
                    \log_4\left(
                    \growthnon
                    \right)
                    \cdot
                    {\frac{1}{d}}
                        (2\smooth + 4d(\holD+1))
                \right\rceil
            \\&
            \quad
                \cdot
                \left\lceil
                \log_2(\max\{\holD,d\}+1)
                \right\rceil
                +
                \left\lceil
                \log_4(\growthnon)
                \left(\frac{2\,\smooth}{d}\right)
                \right\rceil
        ,
        \\
                \min_{l\in\{1,2,\ldots,\L\}}\Hi{l}
        &
            \geq
                132
                \cdot
                2^d
                \cdot
                \lceil e^d \rceil
                \cdot
                \binom{d+\holD}{d}
                \cdot
                \max\{\holD+1,d^2\}
        ,
        \end{aligned}
        \end{equation}
    then, 
    there is some 
    $\fpr\in\FMLPnon(\L,\Hb,\DNNBound)$
    where
        $$
        \snormDNNBoundval{\fpr-\fO} \leq 
        \Cyar
        \Big[\max\left\{\Xboundval,\, \norm*{\fO}_{\HolderCust{\holD}{ }([-\Xboundval,\Xboundval]^d)}\right\}
        \Big]^{4(\holD+1)}
        \cdot
        \growthnon^{-{2p}/{d}}
        .
        $$
\end{lemma}
    \vx

\begin{lemma}\label{lem:WHLBound}
    Let 
        $\FMLP= \FMLPnon(\L,\Hb,\DNNBound)$
    be defined as in \eqref{eq:sievespace}
    where 
    the sequences 
        $\{\L\}_{n\in\N}$,
        $\{\Hi{l}\}_{n\in\N}$ 
    for each
        $l\in\N$,
    are non-decreasing, and
        $\Hi{l}\asymp \H$ 
    for all 
        $l\in\N.$
    Then 
        $\W\asymp\H^2\L$.
\end{lemma}
\begin{proof}
\newcommand{\Hmin}{H^{(min)}_n}%
    First, consider an MLP architecture with $W$ weights and $L$ hidden layers that each have $H$ nodes, 
    then
        $$W = H^2(L-1)+H(d+L+1) +1.$$
    Let $\Hmax \coloneqq \max_{l\in\{1,2,\ldots,\L\}}\Hi{l}$,
    and
        $\Hmin \coloneqq \min_{l\in\{1,2,\ldots,\L\}}\Hi{l}$.
    Then, for the architecture $\FMLP$ where the number of nodes may vary between layers, since $\L\geq1$, it follows that 
        $$
            \big(\Hmin\big)^2(\L-1)+\Hmin(d+\L+1) +1
        \;\leq \;
            \W 
        \;\leq \;
            \big(\Hmax\big)^2(\L-1)+\Hmax(d+\L+1) +1
        .
        $$
    By assumption 
        $\H \asymp \Hmax\asymp \Hmin$,
    so we have 
        \begin{equation*}
        \begin{aligned}
            \W &
        \;\lesssim\;
                \H^2(\L-1)+\H(d+\L+1) +1
        \;\asymp\;
            \H^2\L+\H\L
        \;\asymp\; 
            \H^2\L
        .
        \end{aligned}
        \end{equation*}
\end{proof}
    \vx

\begin{lemma}\label{lem:Pdim_boundNEW}%
    Let 
        $\FMLP= \FMLPnon(\L,\Hb,\DNNBound)$
    be defined as in \eqref{eq:sievespace}
    where 
    the sequences 
        $\{\L\}_{n\in\N}$,
        $\{\Hi{l}\}_{n\in\N}$ 
    for each
        $l\in\N$,
    are non-decreasing,
        $\Hi{l}\asymp \H$ 
    for all 
        $l\in\N.$
    Then,
        $$
            \H^2\L^2\log(\H)
        \;
        \lesssim
        \;
            \Pdim(\FMLP) 
        \;\lesssim\;
            \H^2\L^2\log(\H^2\L)
        .
        $$
\end{lemma}
\begin{proof}
    By \citet[Theorems 3 and 7]{bartlett_nearly_tight_2019},%
\footnote{These bounds are written explicitly in display (2) of \citet{bartlett_nearly_tight_2019}. 
Display (2) uses the Vapnik-Chervonenkis dimension instead of Pseudo-Dimension, however, these are equivalent for function classes generated by a neural network with fixed architecture and fixed activation functions. For details see the discussion following \citet[Definition 2]{bartlett_nearly_tight_2019}, and \citet[Theorem 14.1]{anthony_bartlett_neural_1999}.
}
    there exist constants $c,C>0$ such that, for all $n\in\N$,
        $$
        c\, \W \L \log(\W / \L)
        \leq 
        \Pdim(\FMLP) 
        \leq 
        C\,\W\L\log(\W)
        .
        $$
    Using this, with
    $\W\asymp\H^2\L$
    by Lemma \ref{lem:WHLBound}, 
    implies the result.
    \end{proof}
    \vx

\section{Proofs for Section \texorpdfstring{\ref{sec:CDNN}}{4}}\label{sec:PROOF_SEC_CDNN}

\begin{proof}[Proof of Lemma \ref{lem:CDNN_pdim}]
\newcommand{\U}{U_n}%
\renewcommand{\Hmax}{H_{\text{max}}}%
    Denote the max width of all layers in $\FCONVnon(\L,\filtlen,\DNNBoundfix)$ 
    as 
        $\Hmax\coloneqq \max_{1\leq l \leq \L}\Hicnv{l}$,
    and note that 
        $\Hmax\leq \Hreduc+\LpoolII\filtlen<3(2d+10)^2$
    which does not depend on $n$.
    Let $\FMLP\coloneqq\FMLPnon(\L,\boldsymbol{H}, \DNNBoundfix)$ be a FCDNN with fixed widths $\boldsymbol{H}\coloneqq[\Hmax,\ldots,\Hmax]^{\tr}\in\N^{\L}$.
    Clearly, 
        $\FCONV\coloneqq\FCONVnon(\L,\filtlen,\DNNBoundfix)\subseteq\FMLPnon(\L,\boldsymbol{H}, \DNNBoundfix)$.
    Then, by \citet[Theorem 7]{bartlett_nearly_tight_2019} (see display (2) therein), 
        $$
        \Pdim(\FCONV)\leq \Pdim(\FMLP) \lesssim (\L \Hmax) \L \log (\L\Hmax) \lesssim \L^2 \log(\L)
        .
        $$
\end{proof}
    \vx

{%
\newcommand{\Wmatrix}{\boldsymbol{W}}%
\newcommand{\uvec}{\boldsymbol{u}}%
\newcommand{\drow}{d'}%
\newcommand{\dcol}{r}%
\newcommand{\Ltemp}{L}%
\newcommand{\Ltempa}{L'}%
\newcommand{\Htemp}{\overline{H}}%
\renewcommand{\drow}{\Htemp}%
\newcommand{\onevec}[1]{\boldsymbol{1}_{\dcol+#1 \filtlen}}%
\newcommand{\Bl}[1]{B^{(#1)}}%
\newcommand{\bl}[1]{b^{(#1)}}%
\renewcommand{\Bbar}{\overline{\boldsymbol{b}}}%
\newcommand{\FFDNNbiash}{\kappa}%
\newcommand{\FFDNNbias}{\boldsymbol{\FFDNNbiash}}%
\newcommand{\biasvC}{\biasv^{(\Ltemp+1)}}%
\newcommand{\Htempcap}{H^*}%
\newcommand{\GcvFINAL}{\Gcv^{\delta}}%
\newcommand{\tempbnd}{M}%
\newcommand{\compset}{[-\tempbnd,\tempbnd]^{\dcol}}%
%
%
\begin{lemma}\label{lem:CDNN_FCDNN_singlelayer}
    Suppose 
    $\Htemp,\dcol, \filtlen \in\N$ with $2\leq\filtlen\leq \dcol$. Define
        $$
            \Ltemp
        \coloneqq
            \left\lceil\frac{\drow\dcol}{s-1}\right\rceil
        .
        $$
    Then, given any
        $\tempbnd\geq 1$,
        $\FFDNNbias=[\FFDNNbiash_{1},\ldots,\FFDNNbiash_{\Htemp}]^{\tr}\in\R^{\Htemp}$
    and
        $\gv_{h}\in \R^{\dcol}$ for $h=1,\ldots,\Htemp$,
    there exist
    convolution filters
        $
        \{\gcv^{(l)}\}_{l=1}^{\Ltemp}
        $ 
    of length $\filtlen$,
    intercept vectors $\biasv^{(l)}\in \R^{\dcol+l\filtlen}$ for $l=1,\ldots,\Ltemp$, 
    and an \LCRM\ $\Av_{\dcol}$ defined as in \eqref{eq:LCRMdef},
    such that 
        $$
            \nv^{(\Ltemp+1)}\!\left(\x \right)
        =
            \left[\relu\left(
            \gv_h^{\tr}\,\x
            +
            \FFDNNbiash_{h}
            \right)
            \right]_{h=1}^{\Htemp}
        ,
        \quad
        \forall\, \x\in \compset
        ,
        $$
    where
        $\nv^{(\Ltemp+1)}\!\left(\x \right)$ 
    is defined recursively by  
    \begin{equation*}
            \nv^{(l)}\!\left(\x \right) 
    \coloneqq 
    \begin{cases}
        \relu\!\left(
            \Gcv^{(l)}
            \x + \biasv^{(l)}
        \right)
    ,
    &
        l=1
    \\
        \relu\!\left(
            \Gcv^{(l)}
            \nv^{(l-1)}(\x) + \biasv^{(l)}
        \right)
    ,
    & 
        l\in\{2,\ldots,\Ltemp\} 
    \\
        \relu\!\left(
            \Av_{\dcol} 
            \nv^{(\Ltemp)}(\x)
        \right)
    ,
    &
        l=\Ltemp+1
    ,
    \end{cases}
    \end{equation*}
and $\Gcv^{(l)}\in\R^{\dcol+l\filtlen}\times\R^{\dcol+(l-1)\filtlen}$
with
    $[\Gcv^{(l)}]_{i,j}=[\gcv^{(l)}]_{i-j}$,
for each $l\in\{1,\ldots,\Ltemp\}$.
\end{lemma}
\begin{proof}
In the first step, we show that linear combinations of vectors can be replicated with repeated convolutions.
Let $\gcv=\{\gc_k\}_{k=-\infty}^\infty$ 
denote a filter 
where 
$\gc_k=0$ whenever $k\notin\{0,1,\ldots,\Htemp\dcol-1\}$, and
    $   
    \gc_{h\dcol-j}
    \coloneqq 
    [\,\gv_h\,]_{j}
    ,
    $
for $h=1,\dots,\drow,$ $j=1,\dots,\dcol$. Then we have
$$
\{\gc_{\Htemp\dcol-1},\ldots,\gc_0\}
=
\{\gv_{\Htemp}^{\tr},\ldots,\gv_{1}^{\tr}\}
.
$$
By 
\citet[Theorem 3]{zhou_universality_2020}
for any $s\geq 2$ there exist 
$\textstyle\Ltempa<\frac{\drow\dcol-1}{s-1}+1$ and 
convolution filters $\{\gcv_{l}\}_{l=1}^{\Ltempa}$, each of length $s$, such that 
$
\gcv = \gcv_{\Ltempa}*\cdots*\gcv_{1}
.
$
Define
$$
    \Ltemp
\coloneqq
    \left\lceil\frac{\drow\dcol}{s-1}\right\rceil
=
    \left\lfloor\frac{\drow\dcol-1}{s-1}\right\rfloor+1
\geq
    \max\left\{
    l\in\N : l<\frac{\drow\dcol-1}{s-1}+1
    \right\}
,
$$
so $\Ltempa\leq \Ltemp$.
Then, 
    $$\gcv = \gcv_{\Ltemp}*\cdots*\gcv_{1}$$
by letting
    $\gcv_{\Ltempa+1}=\ldots=\gcv_{\Ltemp}$
be the delta sequence.
Define the matrix 
$
\Gcv^{\gcv}
\in\R^{(\dcol+\Ltemp\filtlen)\times \dcol}
$
where $[\,\Gcv^{\gcv}\,]_{i,j}=\gc_{i-j}$.
For $l=1,\ldots,\Ltemp$, define
    $
    \Gcv^{\gcv_l}
    \in\R^{(\dcol+l\filtlen)\times (\dcol+(l-1)\filtlen)}
    $
    where $[\,\Gcv^{\gcv_l}\,]_{i,j}=[\,\gcv_l\,]_{i-j}$.
Then,
(see, e.g.,
\citealp[Lemma 4.5]{lin_optimal_2023}; and
\citealp[proof of Theorem 2]{zhou_universality_2020})
we have 
    $$
    \Gcv^{\gcv}
    =
    \Gcv^{\gcv_\Ltemp}
    \Gcv^{\gcv_{\Ltemp-1}}
    \cdots
    \Gcv^{\gcv_{1}}
    ,
    $$
and, for any $\x\in\R^{\dcol}$
    $$
            \Gcv^{\gcv}\x
        \,=\,
            \Gcv^{\gcv_\Ltemp}
            \Gcv^{\gcv_{\Ltemp-1}}
            \cdots
            \Gcv^{\gcv_{1}}\x
        \,=\,
            \gcv_{\Ltemp}*\cdots*\gcv_{1} * \x
        \,=\,
            \gcv * \x
        .
    $$
Note that
    $$
    [\,\Gcv^{\gcv}\,]_{h\dcol,j}
    =
    \gc_{h\dcol-j}
    =
    [\,\gv_h\,]_{j}
    ,
    \quad
    h=1,\ldots,\Htemp,
    \;
    j=1,\dots,\dcol
    ,
    $$
so 
for $h=1,\ldots,\Htemp$,
the $h\dcol$-th row of $\Gcv^{\gcv}$ is $\gv_h^{\tr}$,
and
    \begin{equation}\label{eq:conv_lincom_indices}
        [\,\Gcv^{\gcv}\x\,]_{h\dcol}
    =
        \gv_h^{\tr}\x
    ,
    \quad
    h=1,\ldots,\Htemp
    .
    \end{equation}
    
In the second step, 
we show that linear combinations of vectors can be replicated with repeated convolutions with a ReLU activation in between each convolution.
Define
    $$
    \Bl{l}\coloneqq
    \prod_{k=1}^l
    \max\left\{
    1,
    \norm{\gcv_{k}}_{\lp{1}}
    \right\}
    ,
    \quad 
    \text{where}
    \quad
    \norm{\gcv_{l}}_{\lp{1}}=\sum_{j=-\infty}^\infty|[\,\gcv_{l}\,]_j|
    .
    $$
Then, it follows via induction, for every $\x\in\compset$,
    \begin{equation}\label{eq:conv_bounds}
        \norm{
            \Gcv^{\gcv_l}
            \Gcv^{\gcv_{l-1}}
            \cdots
            \Gcv^{\gcv_{1}}
            \x
        }_{\lp{\infty}}
        \leq
            \Bl{l}
            \tempbnd
        .
    \end{equation}
\newcommand{\cl}[1]{c^{(#1)}}%
\newcommand{\shiftl}[1]{S^{(#1)}}%
Define
$\onevec{l}\coloneqq[1,\ldots,1]^{\tr}\in\R^{\dcol+ls}$.
Choose constants $\{\cl{l}\}_{l=1}^{\Ltemp-1}$ and define $\{\shiftl{l}\}_{l=0}^{\Ltemp-1}$ recursively such that 
    $\shiftl{0} \coloneqq [0,\ldots,0]^\tr\in\R^{\dcol}$,
and for $l= 1,\ldots,\Ltemp-1$
    $$
        \cl{l} > \Bl{l}\tempbnd + \norm{\Gcv^{\gcv_l} \shiftl{l-1} }_{\lp{\infty}}
    ,
    \quad
    \text{and}
    \quad
            \shiftl{l} 
        \coloneqq 
            \Gcv^{\gcv_l}
            \shiftl{l-1}
            +
            \cl{l}\onevec{l}
    .
    $$
For every $\x\in\compset$ and $l\in\{1,\ldots,\Ltemp-1\}$,
note that for each coordinate $j$
    $$
        \left[
            \Gcv^{\gcv_{l}}
                \cdots
                \Gcv^{\gcv_{1}}
                \x
            +
            \Gcv^{\gcv_{l}}
            \shiftl{l-1}
        \right]_j
    \geq
        -(
        \Bl{l}\tempbnd
        +
        \norm{\Gcv^{\gcv_l} \shiftl{l-1} }_{\lp{\infty}}
        )
    >
        -\cl{l}
    $$
Then, for any $\x\in\compset$,
    \begin{equation*}
    \begin{aligned}
        \relu\left(
            \Gcv^{\gcv_{1}}\x
            +
            \cl{1}\onevec{ }
        \right)
    &=
        \Gcv^{\gcv_{1}}\x
        +
        \cl{1}\onevec{ }
    =
        \Gcv^{\gcv_{1}}\x
        +
        \shiftl{1}
    ,
    \end{aligned}
    \end{equation*}
and
it follows via induction, for $l=1,\ldots,\Ltemp-1$,
    \begin{equation*}
    \begin{aligned}
    &
        \relu\left(
            \Gcv^{\gcv_{l}}
            \relu\left(
    \ldots
                    \relu\left(
                        \Gcv^{\gcv_{1}}\x
                        +
                        \cl{1}\onevec{ }
                    \right)
    \ldots
            \right)
        +
            \cl{l}\onevec{l}
        \right)
    =
        \Gcv^{\gcv_{l}}
        \cdots
        \Gcv^{\gcv_{1}}
        \x
        +
        \shiftl{l}
    .
    \end{aligned}
    \end{equation*}
Now, set 
    $$
    \biasv^{(l)}  
    \coloneqq
    \cl{l}\onevec{l}
    ,
    \qquad
    l\in\{1,\ldots,\Ltemp-1\}
    ,
    $$
so we have 
    $$
    \nv^{(\Ltemp-1)}\!\left(\x \right) 
    =
        \Gcv^{\gcv_{\Ltemp-1}}
        \cdots
        \Gcv^{\gcv_{1}}
        \x
        +
        \shiftl{\Ltemp-1}
    .
    $$
For the $l=\Ltemp$ layer,
given $\FFDNNbias=[\FFDNNbiash_{1},\ldots,\FFDNNbiash_{\Htemp}]^{\tr}\in\R^{\Htemp}$,
choose    
    $\biasv^{(\Ltemp)}\in\R^{\dcol+\Ltemp\filtlen}$
such that
    \begin{equation*}
    \begin{aligned}
        [\biasv^{(\Ltemp)}]_j
    =
        \begin{cases}
            \FFDNNbiash_{j/\dcol}
            -
            \left[
            \Gcv^{\gcv_{\Ltemp}}
            \shiftl{\Ltemp-1}
            \right]_j
        &
            j\in\big\{\dcol,\,2\dcol,\ldots,\,(\Htemp-1)\dcol,\,\Htemp\dcol\big\}
        \\
            -
            \left[
            \Gcv^{\gcv_{\Ltemp}}
            \shiftl{\Ltemp-1}
            \right]_j
            -
            \Bl{\Ltemp}\tempbnd
        ,
        &
            \text{otherwise}.
        \end{cases}
    \end{aligned}
    \end{equation*}
Then, for any $\x\in\compset$
    $$
    \nv^{(\Ltemp)}\!\left(\x \right) 
    =
        \relu\left(
        \Gcv^{\gcv}
        \x
        +
        \Gcv^{\gcv_{\Ltemp}}
        \shiftl{\Ltemp-1}
        +\bl{\Ltemp}
        \right)
    ,
    $$
and, by \eqref{eq:conv_bounds} and \eqref{eq:conv_lincom_indices},
    \begin{gather*}
        \sup_{\x\in\compset}
        \left[
        \Gcv^{\gcv}\x
        +\Gcv^{\gcv_{\Ltemp}}\shiftl{\Ltemp-1}
        +
        \biasv^{(\Ltemp)}
        \right]_j
    \leq
        0
    ,
    \qquad
        \forall j\notin\big\{\dcol,\,2\dcol,\ldots,\,(\Htemp-1)\dcol,\,\Htemp\dcol\big\}
    ,
    \qquad
    \text{ and }
    \\
        [\Gcv^{\gcv}\x
        +\Gcv^{\gcv_{\Ltemp}}\shiftl{\Ltemp-1}
        +
        \biasv^{(\Ltemp)}
        ]_j
        =
        \gv_{j/\dcol}^{\tr}\,\x
        +
        \FFDNNbiash_{j/\dcol}
    ,
    \quad
        \forall j\in\big\{\dcol,\,2\dcol,\ldots,\,(\Htemp-1)\dcol,\,\Htemp\dcol\big\}
    .
    \end{gather*}
Thus, for each $\x\in\compset$
    \begin{equation}\label{eq:RELU_conv_lincom_indices}
    \begin{aligned}
    &
        \left[\nv^{(\Ltemp)}\!\left(\x \right)\right]_j=0,
        &\quad 
        &j\notin\big\{\dcol,\,2\dcol,\ldots,\,(\Htemp-1)\dcol,\,\Htemp\dcol\big\}
    \\&
        \left[\nv^{(\Ltemp)}\!\left(\x \right)\right]_{h\dcol}
    =
        \relu\left(
        \gv_h^{\tr}\,\x
        +
        \FFDNNbiash_{h}
        \right)
    ,
    &
    &h=1,\ldots,\Htemp
    .
    \end{aligned}
    \end{equation}

In the last step, we show that the \LCRM\ dimension reduction layer can output the correct nodes to represent the FCDNN layer. 
To do this, we show $[\Av_{\dcol}]_{ij}\neq 0$ when $j=i\dcol$ is compatible with \eqref{eq:LCRMdef}.
For any $i=1,\ldots,\Htemp$ the lower bound follows immediately from
    $j=i\dcol>(i-1)\dcol$,
and the upper bound
    $j=i\dcol<(i-1)\dcol + w_{\dcol}$
follows from the definitions of $w_{\dcol}$ and $m_{\dcol}$ since
    \begin{equation*}
    \begin{aligned}
        (i-1)\dcol + w_{\dcol}
    &=
        i\dcol
        + 
        \filtlen\left\lceil\frac{\drow\dcol}{\filtlen-1}\right\rceil
        - (\Htemp-1)\dcol
    \geq
        i\dcol
        + 
        \drow\dcol
        \left(
        \frac{\filtlen}{\filtlen-1}
        \right)
        - (\Htemp-1)\dcol
    \geq
        i\dcol
        +
        \dcol
    > i\dcol
    .
    \end{aligned}
    \end{equation*}
Then, we may choose
$\Av_{\dcol} \in \R^{\Htemp\times(\dcol+\Ltemp\filtlen)}$ such that
$$
    [\Av_\dcol]_{ij}
    =
    \begin{cases}
    1, & j=i\dcol 
    \\
    0, & \text{otherwise}.
    \end{cases}
$$
Thus, by \eqref{eq:RELU_conv_lincom_indices},
\[
[\nv^{(\Ltemp+1)}(\x)]_{h}=\relu(\gv_h^{\tr} \x +\FFDNNbiash_h), \qquad h=1,\ldots,\Htemp
.
\]
\end{proof}
\vx

{
\newcommand{\Hfix}{H}%
\newcommand{\HtempB}{\Htemp'}%
\renewcommand{\onevec}[1]{\boldsymbol{1}_{#1}}%
\renewcommand{\Ltemp}{L_{\text{FC}}}%
\renewcommand{\Ltemp}{L'}%
\renewcommand{\Ltempa}{L}%
\newcommand{\LtempI}{\overline{L}_1}%
\newcommand{\LtempII}{\overline{L}_2}%
\newcommand{\nvtemp}{\tilde{\nv}}%
\newcommand{\CconvapproxB}{\const{CconvapproxB}}%

\begin{proof}[Proof of Proposition \ref{prop:CDNN_approx}]
\renewcommand{\Hfix}{\Hreduc}%
    Consider an arbitrary FCDNN with $\Ltemp$ hidden layers of width $\Hfix=2d+10$ denoted as 
        \begin{equation*}
                \nvff^{(l)}(\x)
        \coloneqq 
        \begin{cases}
            \left[\relu\left(
            (\gv_h^{(1)})^{\tr}\,\x
            +
            \FFDNNbiash_{h}^{(1)}
            \right)
            \right]_{h=1}^{\Hfix},
        &
            l=1,
        \\
            \left[\relu\left(
            (\gv_h^{(l)})^{\tr}\,\nvff^{(l-1)}(\x)
            +
            \FFDNNbiash_{h}^{(l)}
            \right)
            \right]_{h=1}^{\Hfix}
        & 
            l\in\{2,\ldots,\Ltemp\} 
        ,
        \end{cases}
        \end{equation*}
    and output given by 
        $$
            g(\x)
        \coloneqq
            (\gv^{(\Ltemp+1)})^{\tr}\,\nvff^{(\Ltemp)}(\x)
            +
            \FFDNNbiash^{(\Ltemp+1)}
        .
        $$
    Let $\nv^{(l)}$ be defined as in \eqref{eq:nodes_conv_banded} where $\L=\Ltempa \coloneqq \LtempI +(\Ltemp-1)\LtempII$.
    Then, by Lemma \ref{lem:CDNN_FCDNN_singlelayer} there exist
    convolution filters
        $
        \{\gcv^{(l)}\}_{l=1}^{\LtempI-1}
        $
    of length $\filtlen$,
    an \LCRM\
        $\Av^{(\LtempI)} \in \R^{\Htemp\times(\dcol + \LtempI\filtlen)}$
    and
    intercept vectors 
        $\biasv^{(l)}\in \R^{\Hicnv{l}}$, $l=1,\ldots,\LtempI$,
    such that   
        $$
            \nv^{(\LtempI)}\!\left(\x \mid \gV\right)
        =
            \nvff^{(1)}(\x)
        ,
        \quad
        \forall\, \x\in \XspaceC
        .
        $$
    It follows via induction using Lemma \ref{lem:CDNN_FCDNN_singlelayer} that there exists $\gV\in \R^{\W}$ such that 
    for each $j\in\{1,\ldots,\Ltemp\}$
        $$
            \nv^{(\LtempI+(j-1)\LtempII)}\!\left(\x \mid \gV\right)
        =
            \nvff^{(j)}(\x)
        ,
        \quad
        \forall\, \x\in \XspaceC
        ,
        $$
    and choosing $\gV$ to have the same output weights as $g$ we obtain
        $$
        \nv^{(\Ltempa)}\!\left(\x \mid \gV\right)
        =
        g(\x)
        \quad
        \forall\, \x\in \XspaceC
        .
        $$

    Let $\boldsymbol{H}=[\Hfix,\ldots,\Hfix]^{\tr}\in\N^{\Ltemp}$. 
    Since the above $g$ was arbitrary, it follows that for any 
    $g\in\FMLPnon(\Ltemp,\boldsymbol{H}, \DNNBoundfix)$  
    there exists 
    $f\in\FCONVnon(\Ltempa,\filtlen,\DNNBoundfix)$
    such that 
        $f(\x)=g(\x)$ for all $\x\in\XspaceC$.
    Then, by \citet[Theorem 4.1]{yarotsky_phase_2020}, for a constant $\CconvapproxB$ depending only on $\smooth,d$,
        $$
            \sup_{\fO\in \HolderCust{\smooth}(\XspaceC)}\!
            \left\{
                \inf_{f\in\FCONVnon(\Ltempa,\filtlen,\DNNBoundfix)}
                \snorm{\fO-f}\!
            \right\}
        \leq
            \sup_{\fO\in \HolderCust{\smooth}(\XspaceC)}\!
            \left\{
                \inf_{g\in\FMLPnon(\Ltemp,\boldsymbol{H}, \DNNBoundfix)}
                \snorm{\fO-g}\!
            \right\}
        \leq
            \CconvapproxB
            \Ltemp^{-\frac{2\smooth}{d}}
            \log(\Ltemp)^{\frac{2\smooth}{d}}
        .
        $$
\end{proof}
    \vx

}

    \begin{proof}[Proof of Theorem \ref{thrm:CDNN_ROC}]
        Follows by similar reasoning as the proof of Theorem \ref{thrm:FCDNN_ROC} using Proposition \ref{prop:CDNN_approx}, Lemma \ref{lem:CDNN_pdim}, Lemma \ref{lem:Ptrunc_truncf_env_verify} and choosing $\delta \asymp \L^2\log(n)^2$.
    \end{proof}
    \vx

    \begin{proof}[Proof of Corollary \ref{cor:CDNN}]
        Follows from Theorem \ref{thrm:CDNN_ROC} using Lemmas \ref{lem:regression_conditions}, \ref{lem:logistic_conditions}, and \ref{lem:quantile_conditions}.
    \end{proof}
    \vx

}

\section{Considerations for alternative DNN architectures}\label{sec:DNN_Alt_Arch}
\newcommand{\reluALT}{\varphi}%
\newcommand{\FMLPalt}{\mathcal{N}_{n,\reluALT}}%
\newcommand{\FMLPnonalt}{\mathcal{N}_{\reluALT}}%
\newcommand{\anB}{b}%
This section considers extensions of this paper's results to alternative feedforward DNN architectures.
The key to this will be obtaining results for the complexity--either covering number or pseudo-dimension--and approximation power of the DNN sieve spaces under consideration.
\begin{remark}\label{rem:Pdim}
    In this section 
    I will use bounds on  
    the Vapnik Chervonenkis dimension (see \citealp[Definition 1]{bartlett_nearly_tight_2019}) of DNNs that take values in $\{0,1\}$
    from
    \citet{anthony_bartlett_neural_1999}
    and 
    \citet{bartlett_nearly_tight_2019}.
    This is without loss of generality since these results can be directly applied to the pseudo-dimension of real valued DNNs using \citet[Theorem 14.1]{anthony_bartlett_neural_1999} (also see discussion following \citealp[Definition 2]{bartlett_nearly_tight_2019}).
\end{remark}

The first result follows immediately from \citet[Proposition 1]{yarotsky_error_2017}. This shows that 
Theorems \ref{thrm:FCDNN_ROC} and \ref{thrm:CDNN_ROC}, and their subsequent corollaries, 
also apply (with possibly larger constants in the bounds) when $\FMLP$ and $\FCONV$ are equipped with any continuous piecewise-linear activation function, $\reluALT$, that has a fixed number of breakpoints, $\anB\in\N$. One important example of this is the ``leaky'' ReLU (LReLU) activation function 
    \begin{equation*}
        \reluALT^{\textrm{LReLU}}(x)
    =
    \left\{
    \begin{aligned}
        &
            cx,
        &\quad&
            x<0,
        \\&
            x,
        &\quad&
            x\geq0,
    \end{aligned}
    \right.
    \end{equation*}
for some small constant $c>0$ (often set to $c=0.01$) that is predetermined before optimizing over the weights $\gV$. 
The LReLU is often used in practice to address the vanishing gradient problem that arises with the ReLU where certain computation units may never have non-zero outputs. 

    \begin{corollary}\label{prop:nonReLU}
        Let $\anB\in\N$ be a constant and $\reluALT:\R\to\R$ be any continuous piece-wise linear function with $\anB$ breakpoints.
        If
            $\FMLPalt$ 
        and 
            $\FCONV$
        are defined as in \eqref{eq:sievespace} and \eqref{eq:sievespaceCNN}, respectively, except with the ReLU activation function
        replaced by $\reluALT$,
        then Theorems \ref{thrm:FCDNN_ROC} and \ref{thrm:CDNN_ROC} apply with a possibly larger constant $C$.
    \end{corollary}

\newcommand{\FMLPFNN}{\mathcal{N}_{n,\reluALT}^{\mathrm{FFN}}}%
\newcommand{\units}{U_n}%

{
\newcommand{\lin}{0}
\newcommand{\lI}{3}
\newcommand{\lII}{7}
\newcommand{\figlabLI}{5}
\newcommand{\lout}{10}

\newcommand{\shft}{3pt}

\begin{figure}[t]    
    \centering

    \definecolor{input_node_fg}{HTML}{E90F0E}
    \definecolor{input_node_bg}{HTML}{F5C5C1}
    \definecolor{hidden_node_fg}{HTML}{3972FE}
    \definecolor{hidden_node_bg}{HTML}{D9E8FB}
    \definecolor{reduce_node_fg}{HTML}{8A6D3B}
    \definecolor{reduce_node_bg}{HTML}{FCF8E3}
    \definecolor{output_node_fg}{HTML}{9DBD88}
    \definecolor{output_node_bg}{HTML}{D6E7D4}

    
    \begin{tikzpicture}[
        node distance=0.55cm,
        input/.style={circle, draw=input_node_fg, thick, minimum size=0.8cm, inner sep=0pt, fill=input_node_bg},
        hidden/.style={rectangle, draw=hidden_node_fg, thick, minimum size=0.8cm, inner sep=0pt, fill=hidden_node_bg},
        output/.style={diamond, draw=output_node_fg, thick, minimum size=0.1cm, inner sep=1.0pt, fill=output_node_bg},
        conn/.style={->, shorten >= 1.8pt}
    ]
    

        \node[input] (x1) at (\lin, 0.75) {$x_{1}$};
        \node[input] (x2) at (\lin, -0.75) {$x_{2}$};

        \node[hidden] (u11) at (\lI, {1.4}) {$\!\nff^{(1)}_{1}\!$};
        \node[hidden] (u12) at (\lI, 0.0) {$\!\nff^{(1)}_{2}\!$};
        \node[hidden] (u13) at (\lI, -1.4) {$\!\nff^{(1)}_{3}\!$};

        \node[hidden] (u21) at (\lII, 0.7) {$\!\nff^{(2)}_{1}\!$};
        \node[hidden] (u22) at (\lII, -0.7) {$\!\nff^{(2)}_{2}\!$};


        \node[output] (y) at (\lout, 0.0) {$\!\n^{(3)}_{1}\!$};

        \draw[conn] (x1.east) -- (u11);
        \draw[conn] (x1.east) -- (u21);
        \draw[conn] (x1.east) -- (u12);
        \draw[conn] (x2.east) -- (u12);
        \draw[conn] (x2.east) -- (u13);

        \draw[conn] (u11.east) -- (u21);
        \draw[conn] (u12.east) -- (y);
        \draw[conn] (u12.east) -- (u22);
        \draw[conn] (u13.east) -- (u22);

        \draw[conn] (u21.east) -- (y);
        \draw[conn] (u22.east) -- (y);

        \newcommand{\labfontsz}{10pt}
        \newcommand{\labheight}{2.1}
        \node at (\lin,\labheight) {\fontsize{\labfontsz}{2pt}\textbf{Inputs}};
        \node at (\figlabLI,{\labheight+0.2}) {\fontsize{\labfontsz}{2pt}\textbf{Hidden Layers}};
        \node at (\lout,{\labheight-0.0}) {\fontsize{\labfontsz}{2pt}\textbf{Output}};

    \end{tikzpicture}

    \caption{
    Example of $\FMLPFNN$ architecture graph structure where 
        $\L=2$, 
        $\W=17$, 
        and $d=2$.
    }
    \label{fig:DNN}
\end{figure}
}

    The next result provides complexity bounds for applying Theorem \ref{thrm:ROC_V2} to
    DNN estimators with a wide variety of activation functions and any feedforward graph structure, provided an approximation result like Lemma \ref{lem:approximationNEW} or Proposition \ref{prop:CDNN_approx} exists for the DNN under consideration. 
    These DNNs will be denoted as $\FMLPFNN$, which allow for any graph structure where 
    units in layer $l\in\{1,\ldots,\L+1\}$
    takes inputs from any of the units in layers $l'\in\{0,\ldots,l-1\}$.
    See Figure \ref{fig:DNN} for an example of $\FMLPFNN$'s graph structure.
    
    \begin{corollary}%
    \label{prop:FFN}
    \newcommand{\temp}{\eta}
        Let $\FMLPFNN$ be any feedforward neural network with $\L$ layers, $\W$ weights, $\units$ computation units and some continuous activation function $\reluALT:\R\to\R$, such that 
            $\sup_{f\in\FMLPFNN}\snorm{f}\leq \DNNBound$. 
        Define the complexity bound $\temp_{n,\reluALT}$ for the following three classes of activation functions:
            \begin{enumerate}
                \item 
                    if $\reluALT$ is piecewise-linear with $\anB\in\N$ breakpoints let 
                    $
                        \temp_{n,\reluALT}
                    \coloneqq
                        \W\L \log(\W);
                    $
                \item 
                    if $\reluALT$ is piecewise-polynomial with $\anB\in\N$ breakpoints where each piece is a polynomial with degree $\leq p \in\N$,
                    let
                    $
                        \temp_{n,\reluALT}
                    \coloneqq
                        \W\units\log\big((p+1)\anB\big)
                    ;
                    $
                \item 
                    if $\reluALT$ is the sigmoid function, 
                    $\textstyle\reluALT(x)=\frac{1}{1+e^{-x}}$,
                    let
                    $$
                        \temp_{n,\reluALT}
                    \coloneqq
                        \big((\W+2)\units\big)^2
                        +
                        (\W+2)\units\log_2\big(18(\W+2)\units^2\big)
                    .
                    $$
            \end{enumerate}
        Suppose $\reluALT$ is described by one of the three above classes,
        then there exists $C\geq 1$ such that 
        $\Pdim(\FMLPFNN)\;\leq\; C\,\temp_{n,\reluALT}$ 
        for any $n\in\N$.
    \end{corollary}

    The pseudo-dimension bounds follow from \citet[pp. 5,6]{bartlett_nearly_tight_2019} for the piecewise-linear case, 
    \citet[Theorem 10]{bartlett_nearly_tight_2019} 
    for the
    piecewise-polynomial case, and \citet[Theorem 8.13]{anthony_bartlett_neural_1999} for the sigmoid case.

\section*{Declaration of generative AI and AI-assisted technologies in the writing process.}

During the preparation of this work, the author used ChatGPT (OpenAI) and Refine.ink to help identify potential issues in exposition, grammar, clarity of presentation, and consistency of mathematical notation. After using these tools/services, the author reviewed and edited the content as needed and takes full responsibility for the content of the published article.

\singlespacing
\newpage
\bibliography{References.bib}


                            \end{document}